\documentclass{article}

\usepackage[accepted]{neutral2col}

\usepackage[T1]{fontenc}
\usepackage{microtype}
\usepackage{graphicx}
\usepackage{booktabs}
\usepackage{amsmath}
\usepackage{amssymb}
\usepackage{mathtools}
\usepackage{xcolor}
\usepackage{tikz}
\usepackage{float}        % provides the [H] float placement used in figures
\usepackage{subcaption}   % provides appendix subfigure panels and \captionsetup

\definecolor{light_orange}{RGB}{255,229,191}
\definecolor{light_red}{RGB}{253,195,183}
\definecolor{light_gray}{RGB}{198,230,233}
\definecolor{light_blue}{RGB}{230, 230, 255}
\definecolor{accent_orange}{RGB}{255, 128, 2}
\definecolor{accent_blue}{RGB}{178, 178, 255}
\definecolor{accent_blue_dark}{RGB}{102, 102, 255}
\definecolor{light_grey}{gray}{0.95}
\definecolor{phviolet}{RGB}{124,58,237}

\providecommand{\val}[1]{%
  \ifcsname valstore/#1\endcsname\csname valstore/#1\endcsname
  \else\errmessage{unknown value key: #1}\fi}
\expandafter\def\csname valstore/yearbook/models\endcsname{21}
\expandafter\def\csname valstore/yearbook/cutoffs\endcsname{104}
\expandafter\def\csname valstore/yearbook/seeds\endcsname{5}
\expandafter\def\csname valstore/yearbook/cutoff-first\endcsname{1905}
\expandafter\def\csname valstore/yearbook/cutoff-last\endcsname{2013}
\expandafter\def\csname valstore/yearbook/scratch-models\endcsname{12}
\expandafter\def\csname valstore/yearbook/frozen-models\endcsname{9}
\expandafter\def\csname valstore/yearbook/model/clip_b32_frozen/indist\endcsname{73.2}
\expandafter\def\csname valstore/yearbook/model/clip_b32_frozen/future\endcsname{64.1}
\expandafter\def\csname valstore/yearbook/model/clip_b32_frozen/decay\endcsname{9.1}
\expandafter\def\csname valstore/yearbook/model/cnn_l/indist\endcsname{92.8}
\expandafter\def\csname valstore/yearbook/model/cnn_l/future\endcsname{79.0}
\expandafter\def\csname valstore/yearbook/model/cnn_l/decay\endcsname{13.7}
\expandafter\def\csname valstore/yearbook/model/cnn_m/indist\endcsname{92.3}
\expandafter\def\csname valstore/yearbook/model/cnn_m/future\endcsname{78.4}
\expandafter\def\csname valstore/yearbook/model/cnn_m/decay\endcsname{13.9}
\expandafter\def\csname valstore/yearbook/model/cnn_s/indist\endcsname{87.0}
\expandafter\def\csname valstore/yearbook/model/cnn_s/future\endcsname{74.7}
\expandafter\def\csname valstore/yearbook/model/cnn_s/decay\endcsname{12.3}
\expandafter\def\csname valstore/yearbook/model/convnext_s_frozen/indist\endcsname{79.9}
\expandafter\def\csname valstore/yearbook/model/convnext_s_frozen/future\endcsname{70.6}
\expandafter\def\csname valstore/yearbook/model/convnext_s_frozen/decay\endcsname{9.2}
\expandafter\def\csname valstore/yearbook/model/dinov2_s_frozen/indist\endcsname{72.5}
\expandafter\def\csname valstore/yearbook/model/dinov2_s_frozen/future\endcsname{62.7}
\expandafter\def\csname valstore/yearbook/model/dinov2_s_frozen/decay\endcsname{9.8}
\expandafter\def\csname valstore/yearbook/model/dinov3_s_frozen/indist\endcsname{65.8}
\expandafter\def\csname valstore/yearbook/model/dinov3_s_frozen/future\endcsname{57.9}
\expandafter\def\csname valstore/yearbook/model/dinov3_s_frozen/decay\endcsname{7.9}
\expandafter\def\csname valstore/yearbook/model/eva02_b_frozen/indist\endcsname{72.7}
\expandafter\def\csname valstore/yearbook/model/eva02_b_frozen/future\endcsname{63.4}
\expandafter\def\csname valstore/yearbook/model/eva02_b_frozen/decay\endcsname{9.3}
\expandafter\def\csname valstore/yearbook/model/mae_b_frozen/indist\endcsname{67.4}
\expandafter\def\csname valstore/yearbook/model/mae_b_frozen/future\endcsname{59.0}
\expandafter\def\csname valstore/yearbook/model/mae_b_frozen/decay\endcsname{8.4}
\expandafter\def\csname valstore/yearbook/model/mlp_l/indist\endcsname{84.9}
\expandafter\def\csname valstore/yearbook/model/mlp_l/future\endcsname{72.7}
\expandafter\def\csname valstore/yearbook/model/mlp_l/decay\endcsname{12.2}
\expandafter\def\csname valstore/yearbook/model/mlp_m/indist\endcsname{85.7}
\expandafter\def\csname valstore/yearbook/model/mlp_m/future\endcsname{73.0}
\expandafter\def\csname valstore/yearbook/model/mlp_m/decay\endcsname{12.7}
\expandafter\def\csname valstore/yearbook/model/mlp_s/indist\endcsname{69.3}
\expandafter\def\csname valstore/yearbook/model/mlp_s/future\endcsname{62.3}
\expandafter\def\csname valstore/yearbook/model/mlp_s/decay\endcsname{7.1}
\expandafter\def\csname valstore/yearbook/model/resnet50_in_frozen/indist\endcsname{70.7}
\expandafter\def\csname valstore/yearbook/model/resnet50_in_frozen/future\endcsname{61.9}
\expandafter\def\csname valstore/yearbook/model/resnet50_in_frozen/decay\endcsname{8.7}
\expandafter\def\csname valstore/yearbook/model/resnet_l/indist\endcsname{90.0}
\expandafter\def\csname valstore/yearbook/model/resnet_l/future\endcsname{76.1}
\expandafter\def\csname valstore/yearbook/model/resnet_l/decay\endcsname{13.9}
\expandafter\def\csname valstore/yearbook/model/resnet_m/indist\endcsname{88.9}
\expandafter\def\csname valstore/yearbook/model/resnet_m/future\endcsname{75.8}
\expandafter\def\csname valstore/yearbook/model/resnet_m/decay\endcsname{13.1}
\expandafter\def\csname valstore/yearbook/model/resnet_s/indist\endcsname{87.3}
\expandafter\def\csname valstore/yearbook/model/resnet_s/future\endcsname{75.0}
\expandafter\def\csname valstore/yearbook/model/resnet_s/decay\endcsname{12.3}
\expandafter\def\csname valstore/yearbook/model/siglip_b_frozen/indist\endcsname{71.6}
\expandafter\def\csname valstore/yearbook/model/siglip_b_frozen/future\endcsname{62.4}
\expandafter\def\csname valstore/yearbook/model/siglip_b_frozen/decay\endcsname{9.2}
\expandafter\def\csname valstore/yearbook/model/vit_l/indist\endcsname{68.3}
\expandafter\def\csname valstore/yearbook/model/vit_l/future\endcsname{60.3}
\expandafter\def\csname valstore/yearbook/model/vit_l/decay\endcsname{8.0}
\expandafter\def\csname valstore/yearbook/model/vit_m/indist\endcsname{76.3}
\expandafter\def\csname valstore/yearbook/model/vit_m/future\endcsname{66.1}
\expandafter\def\csname valstore/yearbook/model/vit_m/decay\endcsname{10.1}
\expandafter\def\csname valstore/yearbook/model/vit_s/indist\endcsname{74.2}
\expandafter\def\csname valstore/yearbook/model/vit_s/future\endcsname{64.6}
\expandafter\def\csname valstore/yearbook/model/vit_s/decay\endcsname{9.6}
\expandafter\def\csname valstore/yearbook/model/vit_s16_in21k_frozen/indist\endcsname{72.2}
\expandafter\def\csname valstore/yearbook/model/vit_s16_in21k_frozen/future\endcsname{61.8}
\expandafter\def\csname valstore/yearbook/model/vit_s16_in21k_frozen/decay\endcsname{10.4}
\expandafter\def\csname valstore/yearbook/scratch/indist-min\endcsname{68.3}
\expandafter\def\csname valstore/yearbook/scratch/indist-max\endcsname{92.8}
\expandafter\def\csname valstore/yearbook/scratch/future-min\endcsname{60.3}
\expandafter\def\csname valstore/yearbook/scratch/future-max\endcsname{79.0}
\expandafter\def\csname valstore/yearbook/scratch/decay-min\endcsname{7.1}
\expandafter\def\csname valstore/yearbook/scratch/decay-max\endcsname{13.9}
\expandafter\def\csname valstore/yearbook/frozen/indist-min\endcsname{65.8}
\expandafter\def\csname valstore/yearbook/frozen/indist-max\endcsname{79.9}
\expandafter\def\csname valstore/yearbook/frozen/future-min\endcsname{57.9}
\expandafter\def\csname valstore/yearbook/frozen/future-max\endcsname{70.6}
\expandafter\def\csname valstore/yearbook/frozen/decay-min\endcsname{7.9}
\expandafter\def\csname valstore/yearbook/frozen/decay-max\endcsname{10.4}
\expandafter\def\csname valstore/yearbook/family/cnn/1905/future\endcsname{45.0}
\expandafter\def\csname valstore/yearbook/family/cnn/1905/decay\endcsname{8.3}
\expandafter\def\csname valstore/yearbook/family/cnn/1944/future\endcsname{80.3}
\expandafter\def\csname valstore/yearbook/family/cnn/1944/decay\endcsname{16.7}
\expandafter\def\csname valstore/yearbook/family/cnn/1978/future\endcsname{87.1}
\expandafter\def\csname valstore/yearbook/family/cnn/1978/decay\endcsname{-5.3}
\expandafter\def\csname valstore/yearbook/family/cnn/2012/future\endcsname{96.4}
\expandafter\def\csname valstore/yearbook/family/cnn/2012/decay\endcsname{-6.4}
\expandafter\def\csname valstore/yearbook/family/mlp/1905/future\endcsname{47.0}
\expandafter\def\csname valstore/yearbook/family/mlp/1905/decay\endcsname{-3.7}
\expandafter\def\csname valstore/yearbook/family/mlp/1944/future\endcsname{70.0}
\expandafter\def\csname valstore/yearbook/family/mlp/1944/decay\endcsname{12.4}
\expandafter\def\csname valstore/yearbook/family/mlp/1978/future\endcsname{79.0}
\expandafter\def\csname valstore/yearbook/family/mlp/1978/decay\endcsname{-4.8}
\expandafter\def\csname valstore/yearbook/family/mlp/2012/future\endcsname{86.6}
\expandafter\def\csname valstore/yearbook/family/mlp/2012/decay\endcsname{-5.3}
\expandafter\def\csname valstore/yearbook/family/resnet/1905/future\endcsname{49.9}
\expandafter\def\csname valstore/yearbook/family/resnet/1905/decay\endcsname{16.8}
\expandafter\def\csname valstore/yearbook/family/resnet/1944/future\endcsname{82.0}
\expandafter\def\csname valstore/yearbook/family/resnet/1944/decay\endcsname{16.2}
\expandafter\def\csname valstore/yearbook/family/resnet/1978/future\endcsname{88.0}
\expandafter\def\csname valstore/yearbook/family/resnet/1978/decay\endcsname{-1.4}
\expandafter\def\csname valstore/yearbook/family/resnet/2012/future\endcsname{96.6}
\expandafter\def\csname valstore/yearbook/family/resnet/2012/decay\endcsname{-7.1}
\expandafter\def\csname valstore/yearbook/family/transfer/1905/future\endcsname{46.9}
\expandafter\def\csname valstore/yearbook/family/transfer/1905/decay\endcsname{1.9}
\expandafter\def\csname valstore/yearbook/family/transfer/1944/future\endcsname{67.2}
\expandafter\def\csname valstore/yearbook/family/transfer/1944/decay\endcsname{21.4}
\expandafter\def\csname valstore/yearbook/family/transfer/1978/future\endcsname{70.6}
\expandafter\def\csname valstore/yearbook/family/transfer/1978/decay\endcsname{-4.4}
\expandafter\def\csname valstore/yearbook/family/transfer/2012/future\endcsname{85.4}
\expandafter\def\csname valstore/yearbook/family/transfer/2012/decay\endcsname{-6.1}
\expandafter\def\csname valstore/yearbook/family/vit/1905/future\endcsname{45.9}
\expandafter\def\csname valstore/yearbook/family/vit/1905/decay\endcsname{4.1}
\expandafter\def\csname valstore/yearbook/family/vit/1944/future\endcsname{70.7}
\expandafter\def\csname valstore/yearbook/family/vit/1944/decay\endcsname{20.3}
\expandafter\def\csname valstore/yearbook/family/vit/1978/future\endcsname{73.7}
\expandafter\def\csname valstore/yearbook/family/vit/1978/decay\endcsname{-5.9}
\expandafter\def\csname valstore/yearbook/family/vit/2012/future\endcsname{83.5}
\expandafter\def\csname valstore/yearbook/family/vit/2012/decay\endcsname{-8.4}
\expandafter\def\csname valstore/yearbook/family/transfer/indist\endcsname{71.8}
\expandafter\def\csname valstore/yearbook/family/transfer/future\endcsname{62.7}
\expandafter\def\csname valstore/yearbook/family/transfer/decay\endcsname{9.1}
\expandafter\def\csname valstore/yearbook/family/cnn/indist\endcsname{90.7}
\expandafter\def\csname valstore/yearbook/family/cnn/future\endcsname{77.4}
\expandafter\def\csname valstore/yearbook/family/cnn/decay\endcsname{13.3}
\expandafter\def\csname valstore/yearbook/family/mlp/indist\endcsname{80.0}
\expandafter\def\csname valstore/yearbook/family/mlp/future\endcsname{69.3}
\expandafter\def\csname valstore/yearbook/family/mlp/decay\endcsname{10.7}
\expandafter\def\csname valstore/yearbook/family/resnet/indist\endcsname{88.7}
\expandafter\def\csname valstore/yearbook/family/resnet/future\endcsname{75.6}
\expandafter\def\csname valstore/yearbook/family/resnet/decay\endcsname{13.1}
\expandafter\def\csname valstore/yearbook/family/vit/indist\endcsname{72.9}
\expandafter\def\csname valstore/yearbook/family/vit/future\endcsname{63.7}
\expandafter\def\csname valstore/yearbook/family/vit/decay\endcsname{9.2}
\expandafter\def\csname valstore/yearbook/param/mlp_s/trainable\endcsname{98k}
\expandafter\def\csname valstore/yearbook/param/mlp_s/total\endcsname{98k}
\expandafter\def\csname valstore/yearbook/param/mlp_m/trainable\endcsname{410k}
\expandafter\def\csname valstore/yearbook/param/mlp_m/total\endcsname{410k}
\expandafter\def\csname valstore/yearbook/param/mlp_l/trainable\endcsname{2.1M}
\expandafter\def\csname valstore/yearbook/param/mlp_l/total\endcsname{2.1M}
\expandafter\def\csname valstore/yearbook/param/cnn_s/trainable\endcsname{94k}
\expandafter\def\csname valstore/yearbook/param/cnn_s/total\endcsname{94k}
\expandafter\def\csname valstore/yearbook/param/cnn_m/trainable\endcsname{542k}
\expandafter\def\csname valstore/yearbook/param/cnn_m/total\endcsname{542k}
\expandafter\def\csname valstore/yearbook/param/cnn_l/trainable\endcsname{2.1M}
\expandafter\def\csname valstore/yearbook/param/cnn_l/total\endcsname{2.1M}
\expandafter\def\csname valstore/yearbook/param/resnet_s/trainable\endcsname{98k}
\expandafter\def\csname valstore/yearbook/param/resnet_s/total\endcsname{98k}
\expandafter\def\csname valstore/yearbook/param/resnet_m/trainable\endcsname{400k}
\expandafter\def\csname valstore/yearbook/param/resnet_m/total\endcsname{400k}
\expandafter\def\csname valstore/yearbook/param/resnet_l/trainable\endcsname{2.1M}
\expandafter\def\csname valstore/yearbook/param/resnet_l/total\endcsname{2.1M}
\expandafter\def\csname valstore/yearbook/param/vit_s/trainable\endcsname{75k}
\expandafter\def\csname valstore/yearbook/param/vit_s/total\endcsname{75k}
\expandafter\def\csname valstore/yearbook/param/vit_m/trainable\endcsname{545k}
\expandafter\def\csname valstore/yearbook/param/vit_m/total\endcsname{545k}
\expandafter\def\csname valstore/yearbook/param/vit_l/trainable\endcsname{2.2M}
\expandafter\def\csname valstore/yearbook/param/vit_l/total\endcsname{2.2M}
\expandafter\def\csname valstore/yearbook/param/resnet50_in_frozen/trainable\endcsname{4k}
\expandafter\def\csname valstore/yearbook/param/resnet50_in_frozen/total\endcsname{23.5M}
\expandafter\def\csname valstore/yearbook/param/vit_s16_in21k_frozen/trainable\endcsname{770}
\expandafter\def\csname valstore/yearbook/param/vit_s16_in21k_frozen/total\endcsname{21.7M}
\expandafter\def\csname valstore/yearbook/param/dinov2_s_frozen/trainable\endcsname{770}
\expandafter\def\csname valstore/yearbook/param/dinov2_s_frozen/total\endcsname{22.1M}
\expandafter\def\csname valstore/yearbook/param/dinov3_s_frozen/trainable\endcsname{770}
\expandafter\def\csname valstore/yearbook/param/dinov3_s_frozen/total\endcsname{21.6M}
\expandafter\def\csname valstore/yearbook/param/convnext_s_frozen/trainable\endcsname{2k}
\expandafter\def\csname valstore/yearbook/param/convnext_s_frozen/total\endcsname{49.5M}
\expandafter\def\csname valstore/yearbook/param/mae_b_frozen/trainable\endcsname{2k}
\expandafter\def\csname valstore/yearbook/param/mae_b_frozen/total\endcsname{85.8M}
\expandafter\def\csname valstore/yearbook/param/eva02_b_frozen/trainable\endcsname{2k}
\expandafter\def\csname valstore/yearbook/param/eva02_b_frozen/total\endcsname{85.8M}
\expandafter\def\csname valstore/yearbook/param/clip_b32_frozen/trainable\endcsname{2k}
\expandafter\def\csname valstore/yearbook/param/clip_b32_frozen/total\endcsname{87.5M}
\expandafter\def\csname valstore/yearbook/param/siglip_b_frozen/trainable\endcsname{2k}
\expandafter\def\csname valstore/yearbook/param/siglip_b_frozen/total\endcsname{92.9M}
\expandafter\def\csname valstore/amazon_reviews/models\endcsname{19}
\expandafter\def\csname valstore/amazon_reviews/cutoffs\endcsname{20}
\expandafter\def\csname valstore/amazon_reviews/seeds\endcsname{3}
\expandafter\def\csname valstore/amazon_reviews/cutoff-first\endcsname{2014-H1}
\expandafter\def\csname valstore/amazon_reviews/cutoff-last\endcsname{2023-H2}
\expandafter\def\csname valstore/amazon_reviews/scratch-models\endcsname{12}
\expandafter\def\csname valstore/amazon_reviews/frozen-models\endcsname{7}
\expandafter\def\csname valstore/amazon_reviews/model/bert_base_frozen/indist\endcsname{0.924}
\expandafter\def\csname valstore/amazon_reviews/model/bert_base_frozen/future\endcsname{0.999}
\expandafter\def\csname valstore/amazon_reviews/model/bert_base_frozen/decay\endcsname{0.075}
\expandafter\def\csname valstore/amazon_reviews/model/bigru_s/indist\endcsname{0.680}
\expandafter\def\csname valstore/amazon_reviews/model/bigru_s/future\endcsname{0.797}
\expandafter\def\csname valstore/amazon_reviews/model/bigru_s/decay\endcsname{0.116}
\expandafter\def\csname valstore/amazon_reviews/model/bilstm_attn_l/indist\endcsname{0.713}
\expandafter\def\csname valstore/amazon_reviews/model/bilstm_attn_l/future\endcsname{0.819}
\expandafter\def\csname valstore/amazon_reviews/model/bilstm_attn_l/decay\endcsname{0.106}
\expandafter\def\csname valstore/amazon_reviews/model/bilstm_m/indist\endcsname{0.696}
\expandafter\def\csname valstore/amazon_reviews/model/bilstm_m/future\endcsname{0.824}
\expandafter\def\csname valstore/amazon_reviews/model/bilstm_m/decay\endcsname{0.128}
\expandafter\def\csname valstore/amazon_reviews/model/deberta_v3_base_frozen/indist\endcsname{0.806}
\expandafter\def\csname valstore/amazon_reviews/model/deberta_v3_base_frozen/future\endcsname{0.849}
\expandafter\def\csname valstore/amazon_reviews/model/deberta_v3_base_frozen/decay\endcsname{0.043}
\expandafter\def\csname valstore/amazon_reviews/model/distilbert_base_frozen/indist\endcsname{0.965}
\expandafter\def\csname valstore/amazon_reviews/model/distilbert_base_frozen/future\endcsname{1.047}
\expandafter\def\csname valstore/amazon_reviews/model/distilbert_base_frozen/decay\endcsname{0.082}
\expandafter\def\csname valstore/amazon_reviews/model/electra_base_frozen/indist\endcsname{0.924}
\expandafter\def\csname valstore/amazon_reviews/model/electra_base_frozen/future\endcsname{1.000}
\expandafter\def\csname valstore/amazon_reviews/model/electra_base_frozen/decay\endcsname{0.076}
\expandafter\def\csname valstore/amazon_reviews/model/ffn_l/indist\endcsname{0.759}
\expandafter\def\csname valstore/amazon_reviews/model/ffn_l/future\endcsname{0.848}
\expandafter\def\csname valstore/amazon_reviews/model/ffn_l/decay\endcsname{0.089}
\expandafter\def\csname valstore/amazon_reviews/model/ffn_m/indist\endcsname{0.757}
\expandafter\def\csname valstore/amazon_reviews/model/ffn_m/future\endcsname{0.834}
\expandafter\def\csname valstore/amazon_reviews/model/ffn_m/decay\endcsname{0.077}
\expandafter\def\csname valstore/amazon_reviews/model/ffn_s/indist\endcsname{0.766}
\expandafter\def\csname valstore/amazon_reviews/model/ffn_s/future\endcsname{0.841}
\expandafter\def\csname valstore/amazon_reviews/model/ffn_s/decay\endcsname{0.075}
\expandafter\def\csname valstore/amazon_reviews/model/modernbert_base_frozen/indist\endcsname{1.022}
\expandafter\def\csname valstore/amazon_reviews/model/modernbert_base_frozen/future\endcsname{1.086}
\expandafter\def\csname valstore/amazon_reviews/model/modernbert_base_frozen/decay\endcsname{0.064}
\expandafter\def\csname valstore/amazon_reviews/model/mpnet_base_frozen/indist\endcsname{0.948}
\expandafter\def\csname valstore/amazon_reviews/model/mpnet_base_frozen/future\endcsname{1.047}
\expandafter\def\csname valstore/amazon_reviews/model/mpnet_base_frozen/decay\endcsname{0.099}
\expandafter\def\csname valstore/amazon_reviews/model/roberta_base_frozen/indist\endcsname{0.856}
\expandafter\def\csname valstore/amazon_reviews/model/roberta_base_frozen/future\endcsname{0.944}
\expandafter\def\csname valstore/amazon_reviews/model/roberta_base_frozen/decay\endcsname{0.088}
\expandafter\def\csname valstore/amazon_reviews/model/textcnn_l/indist\endcsname{0.793}
\expandafter\def\csname valstore/amazon_reviews/model/textcnn_l/future\endcsname{0.871}
\expandafter\def\csname valstore/amazon_reviews/model/textcnn_l/decay\endcsname{0.078}
\expandafter\def\csname valstore/amazon_reviews/model/textcnn_m/indist\endcsname{0.792}
\expandafter\def\csname valstore/amazon_reviews/model/textcnn_m/future\endcsname{0.885}
\expandafter\def\csname valstore/amazon_reviews/model/textcnn_m/decay\endcsname{0.093}
\expandafter\def\csname valstore/amazon_reviews/model/textcnn_s/indist\endcsname{0.746}
\expandafter\def\csname valstore/amazon_reviews/model/textcnn_s/future\endcsname{0.860}
\expandafter\def\csname valstore/amazon_reviews/model/textcnn_s/decay\endcsname{0.114}
\expandafter\def\csname valstore/amazon_reviews/model/tx_l/indist\endcsname{0.735}
\expandafter\def\csname valstore/amazon_reviews/model/tx_l/future\endcsname{0.829}
\expandafter\def\csname valstore/amazon_reviews/model/tx_l/decay\endcsname{0.094}
\expandafter\def\csname valstore/amazon_reviews/model/tx_m/indist\endcsname{0.708}
\expandafter\def\csname valstore/amazon_reviews/model/tx_m/future\endcsname{0.801}
\expandafter\def\csname valstore/amazon_reviews/model/tx_m/decay\endcsname{0.092}
\expandafter\def\csname valstore/amazon_reviews/model/tx_s/indist\endcsname{0.686}
\expandafter\def\csname valstore/amazon_reviews/model/tx_s/future\endcsname{0.776}
\expandafter\def\csname valstore/amazon_reviews/model/tx_s/decay\endcsname{0.090}
\expandafter\def\csname valstore/amazon_reviews/scratch/indist-min\endcsname{0.680}
\expandafter\def\csname valstore/amazon_reviews/scratch/indist-max\endcsname{0.793}
\expandafter\def\csname valstore/amazon_reviews/scratch/future-min\endcsname{0.776}
\expandafter\def\csname valstore/amazon_reviews/scratch/future-max\endcsname{0.885}
\expandafter\def\csname valstore/amazon_reviews/scratch/decay-min\endcsname{0.075}
\expandafter\def\csname valstore/amazon_reviews/scratch/decay-max\endcsname{0.128}
\expandafter\def\csname valstore/amazon_reviews/frozen/indist-min\endcsname{0.806}
\expandafter\def\csname valstore/amazon_reviews/frozen/indist-max\endcsname{1.022}
\expandafter\def\csname valstore/amazon_reviews/frozen/future-min\endcsname{0.849}
\expandafter\def\csname valstore/amazon_reviews/frozen/future-max\endcsname{1.086}
\expandafter\def\csname valstore/amazon_reviews/frozen/decay-min\endcsname{0.043}
\expandafter\def\csname valstore/amazon_reviews/frozen/decay-max\endcsname{0.099}
\expandafter\def\csname valstore/amazon_reviews/family/ffn/28/future\endcsname{0.993}
\expandafter\def\csname valstore/amazon_reviews/family/ffn/28/decay\endcsname{0.117}
\expandafter\def\csname valstore/amazon_reviews/family/ffn/34/future\endcsname{0.817}
\expandafter\def\csname valstore/amazon_reviews/family/ffn/34/decay\endcsname{0.087}
\expandafter\def\csname valstore/amazon_reviews/family/ffn/40/future\endcsname{0.802}
\expandafter\def\csname valstore/amazon_reviews/family/ffn/40/decay\endcsname{0.073}
\expandafter\def\csname valstore/amazon_reviews/family/ffn/46/future\endcsname{0.789}
\expandafter\def\csname valstore/amazon_reviews/family/ffn/46/decay\endcsname{0.056}
\expandafter\def\csname valstore/amazon_reviews/family/frozen/28/future\endcsname{1.351}
\expandafter\def\csname valstore/amazon_reviews/family/frozen/28/decay\endcsname{0.131}
\expandafter\def\csname valstore/amazon_reviews/family/frozen/34/future\endcsname{0.943}
\expandafter\def\csname valstore/amazon_reviews/family/frozen/34/decay\endcsname{0.068}
\expandafter\def\csname valstore/amazon_reviews/family/frozen/40/future\endcsname{0.934}
\expandafter\def\csname valstore/amazon_reviews/family/frozen/40/decay\endcsname{0.073}
\expandafter\def\csname valstore/amazon_reviews/family/frozen/46/future\endcsname{0.921}
\expandafter\def\csname valstore/amazon_reviews/family/frozen/46/decay\endcsname{0.039}
\expandafter\def\csname valstore/amazon_reviews/family/recurrent/28/future\endcsname{1.101}
\expandafter\def\csname valstore/amazon_reviews/family/recurrent/28/decay\endcsname{0.242}
\expandafter\def\csname valstore/amazon_reviews/family/recurrent/34/future\endcsname{0.794}
\expandafter\def\csname valstore/amazon_reviews/family/recurrent/34/decay\endcsname{0.134}
\expandafter\def\csname valstore/amazon_reviews/family/recurrent/40/future\endcsname{0.723}
\expandafter\def\csname valstore/amazon_reviews/family/recurrent/40/decay\endcsname{0.094}
\expandafter\def\csname valstore/amazon_reviews/family/recurrent/46/future\endcsname{0.710}
\expandafter\def\csname valstore/amazon_reviews/family/recurrent/46/decay\endcsname{0.059}
\expandafter\def\csname valstore/amazon_reviews/family/textcnn/28/future\endcsname{1.002}
\expandafter\def\csname valstore/amazon_reviews/family/textcnn/28/decay\endcsname{0.227}
\expandafter\def\csname valstore/amazon_reviews/family/textcnn/34/future\endcsname{0.859}
\expandafter\def\csname valstore/amazon_reviews/family/textcnn/34/decay\endcsname{0.102}
\expandafter\def\csname valstore/amazon_reviews/family/textcnn/40/future\endcsname{0.811}
\expandafter\def\csname valstore/amazon_reviews/family/textcnn/40/decay\endcsname{0.090}
\expandafter\def\csname valstore/amazon_reviews/family/textcnn/46/future\endcsname{0.818}
\expandafter\def\csname valstore/amazon_reviews/family/textcnn/46/decay\endcsname{0.076}
\expandafter\def\csname valstore/amazon_reviews/family/transformer/28/future\endcsname{0.967}
\expandafter\def\csname valstore/amazon_reviews/family/transformer/28/decay\endcsname{0.173}
\expandafter\def\csname valstore/amazon_reviews/family/transformer/34/future\endcsname{0.784}
\expandafter\def\csname valstore/amazon_reviews/family/transformer/34/decay\endcsname{0.101}
\expandafter\def\csname valstore/amazon_reviews/family/transformer/40/future\endcsname{0.748}
\expandafter\def\csname valstore/amazon_reviews/family/transformer/40/decay\endcsname{0.080}
\expandafter\def\csname valstore/amazon_reviews/family/transformer/46/future\endcsname{0.721}
\expandafter\def\csname valstore/amazon_reviews/family/transformer/46/decay\endcsname{0.056}
\expandafter\def\csname valstore/amazon_reviews/family/frozen/indist\endcsname{0.921}
\expandafter\def\csname valstore/amazon_reviews/family/frozen/future\endcsname{0.996}
\expandafter\def\csname valstore/amazon_reviews/family/frozen/decay\endcsname{0.075}
\expandafter\def\csname valstore/amazon_reviews/family/recurrent/indist\endcsname{0.696}
\expandafter\def\csname valstore/amazon_reviews/family/recurrent/future\endcsname{0.813}
\expandafter\def\csname valstore/amazon_reviews/family/recurrent/decay\endcsname{0.117}
\expandafter\def\csname valstore/amazon_reviews/family/ffn/indist\endcsname{0.761}
\expandafter\def\csname valstore/amazon_reviews/family/ffn/future\endcsname{0.841}
\expandafter\def\csname valstore/amazon_reviews/family/ffn/decay\endcsname{0.080}
\expandafter\def\csname valstore/amazon_reviews/family/textcnn/indist\endcsname{0.777}
\expandafter\def\csname valstore/amazon_reviews/family/textcnn/future\endcsname{0.872}
\expandafter\def\csname valstore/amazon_reviews/family/textcnn/decay\endcsname{0.095}
\expandafter\def\csname valstore/amazon_reviews/family/transformer/indist\endcsname{0.710}
\expandafter\def\csname valstore/amazon_reviews/family/transformer/future\endcsname{0.802}
\expandafter\def\csname valstore/amazon_reviews/family/transformer/decay\endcsname{0.092}
\expandafter\def\csname valstore/amazon_reviews/param/ffn_s/trainable\endcsname{99k}
\expandafter\def\csname valstore/amazon_reviews/param/ffn_s/total\endcsname{124.7M}
\expandafter\def\csname valstore/amazon_reviews/param/ffn_m/trainable\endcsname{394k}
\expandafter\def\csname valstore/amazon_reviews/param/ffn_m/total\endcsname{125.0M}
\expandafter\def\csname valstore/amazon_reviews/param/ffn_l/trainable\endcsname{1.6M}
\expandafter\def\csname valstore/amazon_reviews/param/ffn_l/total\endcsname{126.2M}
\expandafter\def\csname valstore/amazon_reviews/param/textcnn_s/trainable\endcsname{88k}
\expandafter\def\csname valstore/amazon_reviews/param/textcnn_s/total\endcsname{124.7M}
\expandafter\def\csname valstore/amazon_reviews/param/textcnn_m/trainable\endcsname{461k}
\expandafter\def\csname valstore/amazon_reviews/param/textcnn_m/total\endcsname{125.1M}
\expandafter\def\csname valstore/amazon_reviews/param/textcnn_l/trainable\endcsname{1.9M}
\expandafter\def\csname valstore/amazon_reviews/param/textcnn_l/total\endcsname{126.6M}
\expandafter\def\csname valstore/amazon_reviews/param/bigru_s/trainable\endcsname{99k}
\expandafter\def\csname valstore/amazon_reviews/param/bigru_s/total\endcsname{124.7M}
\expandafter\def\csname valstore/amazon_reviews/param/bilstm_m/trainable\endcsname{536k}
\expandafter\def\csname valstore/amazon_reviews/param/bilstm_m/total\endcsname{125.2M}
\expandafter\def\csname valstore/amazon_reviews/param/bilstm_attn_l/trainable\endcsname{2.2M}
\expandafter\def\csname valstore/amazon_reviews/param/bilstm_attn_l/total\endcsname{126.8M}
\expandafter\def\csname valstore/amazon_reviews/param/tx_s/trainable\endcsname{83k}
\expandafter\def\csname valstore/amazon_reviews/param/tx_s/total\endcsname{124.7M}
\expandafter\def\csname valstore/amazon_reviews/param/tx_m/trainable\endcsname{493k}
\expandafter\def\csname valstore/amazon_reviews/param/tx_m/total\endcsname{125.1M}
\expandafter\def\csname valstore/amazon_reviews/param/tx_l/trainable\endcsname{1.9M}
\expandafter\def\csname valstore/amazon_reviews/param/tx_l/total\endcsname{126.5M}
\expandafter\def\csname valstore/amazon_reviews/param/distilbert_base_frozen/trainable\endcsname{769}
\expandafter\def\csname valstore/amazon_reviews/param/distilbert_base_frozen/total\endcsname{66.4M}
\expandafter\def\csname valstore/amazon_reviews/param/bert_base_frozen/trainable\endcsname{769}
\expandafter\def\csname valstore/amazon_reviews/param/bert_base_frozen/total\endcsname{109.5M}
\expandafter\def\csname valstore/amazon_reviews/param/roberta_base_frozen/trainable\endcsname{769}
\expandafter\def\csname valstore/amazon_reviews/param/roberta_base_frozen/total\endcsname{124.6M}
\expandafter\def\csname valstore/amazon_reviews/param/deberta_v3_base_frozen/trainable\endcsname{769}
\expandafter\def\csname valstore/amazon_reviews/param/deberta_v3_base_frozen/total\endcsname{183.8M}
\expandafter\def\csname valstore/amazon_reviews/param/electra_base_frozen/trainable\endcsname{769}
\expandafter\def\csname valstore/amazon_reviews/param/electra_base_frozen/total\endcsname{108.9M}
\expandafter\def\csname valstore/amazon_reviews/param/mpnet_base_frozen/trainable\endcsname{769}
\expandafter\def\csname valstore/amazon_reviews/param/mpnet_base_frozen/total\endcsname{109.5M}
\expandafter\def\csname valstore/amazon_reviews/param/modernbert_base_frozen/trainable\endcsname{769}
\expandafter\def\csname valstore/amazon_reviews/param/modernbert_base_frozen/total\endcsname{149.0M}
\expandafter\def\csname valstore/arxiv/models\endcsname{20}
\expandafter\def\csname valstore/arxiv/cutoffs\endcsname{26}
\expandafter\def\csname valstore/arxiv/seeds\endcsname{3}
\expandafter\def\csname valstore/arxiv/cutoff-first\endcsname{2000}
\expandafter\def\csname valstore/arxiv/cutoff-last\endcsname{2025}
\expandafter\def\csname valstore/arxiv/scratch-models\endcsname{12}
\expandafter\def\csname valstore/arxiv/frozen-models\endcsname{8}
\expandafter\def\csname valstore/arxiv/model/bert_base_frozen/indist\endcsname{96.7}
\expandafter\def\csname valstore/arxiv/model/bert_base_frozen/future\endcsname{93.6}
\expandafter\def\csname valstore/arxiv/model/bert_base_frozen/decay\endcsname{3.1}
\expandafter\def\csname valstore/arxiv/model/bigru_s/indist\endcsname{97.9}
\expandafter\def\csname valstore/arxiv/model/bigru_s/future\endcsname{94.7}
\expandafter\def\csname valstore/arxiv/model/bigru_s/decay\endcsname{3.2}
\expandafter\def\csname valstore/arxiv/model/bilstm_attn_l/indist\endcsname{98.0}
\expandafter\def\csname valstore/arxiv/model/bilstm_attn_l/future\endcsname{95.1}
\expandafter\def\csname valstore/arxiv/model/bilstm_attn_l/decay\endcsname{2.9}
\expandafter\def\csname valstore/arxiv/model/bilstm_m/indist\endcsname{98.1}
\expandafter\def\csname valstore/arxiv/model/bilstm_m/future\endcsname{95.2}
\expandafter\def\csname valstore/arxiv/model/bilstm_m/decay\endcsname{2.9}
\expandafter\def\csname valstore/arxiv/model/deberta_v3_base_frozen/indist\endcsname{92.7}
\expandafter\def\csname valstore/arxiv/model/deberta_v3_base_frozen/future\endcsname{86.1}
\expandafter\def\csname valstore/arxiv/model/deberta_v3_base_frozen/decay\endcsname{6.5}
\expandafter\def\csname valstore/arxiv/model/distilbert_base_frozen/indist\endcsname{96.4}
\expandafter\def\csname valstore/arxiv/model/distilbert_base_frozen/future\endcsname{93.1}
\expandafter\def\csname valstore/arxiv/model/distilbert_base_frozen/decay\endcsname{3.3}
\expandafter\def\csname valstore/arxiv/model/electra_base_frozen/indist\endcsname{91.8}
\expandafter\def\csname valstore/arxiv/model/electra_base_frozen/future\endcsname{84.4}
\expandafter\def\csname valstore/arxiv/model/electra_base_frozen/decay\endcsname{7.3}
\expandafter\def\csname valstore/arxiv/model/ffn_l/indist\endcsname{97.6}
\expandafter\def\csname valstore/arxiv/model/ffn_l/future\endcsname{94.9}
\expandafter\def\csname valstore/arxiv/model/ffn_l/decay\endcsname{2.7}
\expandafter\def\csname valstore/arxiv/model/ffn_m/indist\endcsname{97.5}
\expandafter\def\csname valstore/arxiv/model/ffn_m/future\endcsname{94.6}
\expandafter\def\csname valstore/arxiv/model/ffn_m/decay\endcsname{2.9}
\expandafter\def\csname valstore/arxiv/model/ffn_s/indist\endcsname{97.2}
\expandafter\def\csname valstore/arxiv/model/ffn_s/future\endcsname{93.9}
\expandafter\def\csname valstore/arxiv/model/ffn_s/decay\endcsname{3.4}
\expandafter\def\csname valstore/arxiv/model/minilm_l6_frozen/indist\endcsname{97.7}
\expandafter\def\csname valstore/arxiv/model/minilm_l6_frozen/future\endcsname{94.8}
\expandafter\def\csname valstore/arxiv/model/minilm_l6_frozen/decay\endcsname{2.8}
\expandafter\def\csname valstore/arxiv/model/modernbert_base_frozen/indist\endcsname{97.4}
\expandafter\def\csname valstore/arxiv/model/modernbert_base_frozen/future\endcsname{94.9}
\expandafter\def\csname valstore/arxiv/model/modernbert_base_frozen/decay\endcsname{2.5}
\expandafter\def\csname valstore/arxiv/model/mpnet_base_frozen/indist\endcsname{95.7}
\expandafter\def\csname valstore/arxiv/model/mpnet_base_frozen/future\endcsname{92.0}
\expandafter\def\csname valstore/arxiv/model/mpnet_base_frozen/decay\endcsname{3.7}
\expandafter\def\csname valstore/arxiv/model/roberta_base_frozen/indist\endcsname{95.9}
\expandafter\def\csname valstore/arxiv/model/roberta_base_frozen/future\endcsname{92.3}
\expandafter\def\csname valstore/arxiv/model/roberta_base_frozen/decay\endcsname{3.6}
\expandafter\def\csname valstore/arxiv/model/textcnn_l/indist\endcsname{98.0}
\expandafter\def\csname valstore/arxiv/model/textcnn_l/future\endcsname{95.2}
\expandafter\def\csname valstore/arxiv/model/textcnn_l/decay\endcsname{2.7}
\expandafter\def\csname valstore/arxiv/model/textcnn_m/indist\endcsname{97.9}
\expandafter\def\csname valstore/arxiv/model/textcnn_m/future\endcsname{94.8}
\expandafter\def\csname valstore/arxiv/model/textcnn_m/decay\endcsname{3.1}
\expandafter\def\csname valstore/arxiv/model/textcnn_s/indist\endcsname{97.9}
\expandafter\def\csname valstore/arxiv/model/textcnn_s/future\endcsname{94.5}
\expandafter\def\csname valstore/arxiv/model/textcnn_s/decay\endcsname{3.4}
\expandafter\def\csname valstore/arxiv/model/tx_l/indist\endcsname{97.9}
\expandafter\def\csname valstore/arxiv/model/tx_l/future\endcsname{94.7}
\expandafter\def\csname valstore/arxiv/model/tx_l/decay\endcsname{3.1}
\expandafter\def\csname valstore/arxiv/model/tx_m/indist\endcsname{97.9}
\expandafter\def\csname valstore/arxiv/model/tx_m/future\endcsname{95.1}
\expandafter\def\csname valstore/arxiv/model/tx_m/decay\endcsname{2.8}
\expandafter\def\csname valstore/arxiv/model/tx_s/indist\endcsname{98.0}
\expandafter\def\csname valstore/arxiv/model/tx_s/future\endcsname{95.1}
\expandafter\def\csname valstore/arxiv/model/tx_s/decay\endcsname{2.9}
\expandafter\def\csname valstore/arxiv/scratch/indist-min\endcsname{97.2}
\expandafter\def\csname valstore/arxiv/scratch/indist-max\endcsname{98.1}
\expandafter\def\csname valstore/arxiv/scratch/future-min\endcsname{93.9}
\expandafter\def\csname valstore/arxiv/scratch/future-max\endcsname{95.2}
\expandafter\def\csname valstore/arxiv/scratch/decay-min\endcsname{2.7}
\expandafter\def\csname valstore/arxiv/scratch/decay-max\endcsname{3.4}
\expandafter\def\csname valstore/arxiv/frozen/indist-min\endcsname{91.8}
\expandafter\def\csname valstore/arxiv/frozen/indist-max\endcsname{97.7}
\expandafter\def\csname valstore/arxiv/frozen/future-min\endcsname{84.4}
\expandafter\def\csname valstore/arxiv/frozen/future-max\endcsname{94.9}
\expandafter\def\csname valstore/arxiv/frozen/decay-min\endcsname{2.5}
\expandafter\def\csname valstore/arxiv/frozen/decay-max\endcsname{7.3}
\expandafter\def\csname valstore/arxiv/family/ffn/2000/future\endcsname{92.2}
\expandafter\def\csname valstore/arxiv/family/ffn/2000/decay\endcsname{4.3}
\expandafter\def\csname valstore/arxiv/family/ffn/2008/future\endcsname{94.6}
\expandafter\def\csname valstore/arxiv/family/ffn/2008/decay\endcsname{3.3}
\expandafter\def\csname valstore/arxiv/family/ffn/2016/future\endcsname{96.1}
\expandafter\def\csname valstore/arxiv/family/ffn/2016/decay\endcsname{2.2}
\expandafter\def\csname valstore/arxiv/family/ffn/2024/future\endcsname{95.8}
\expandafter\def\csname valstore/arxiv/family/ffn/2024/decay\endcsname{0.2}
\expandafter\def\csname valstore/arxiv/family/frozen/2000/future\endcsname{87.8}
\expandafter\def\csname valstore/arxiv/family/frozen/2000/decay\endcsname{5.4}
\expandafter\def\csname valstore/arxiv/family/frozen/2008/future\endcsname{91.6}
\expandafter\def\csname valstore/arxiv/family/frozen/2008/decay\endcsname{4.6}
\expandafter\def\csname valstore/arxiv/family/frozen/2016/future\endcsname{93.6}
\expandafter\def\csname valstore/arxiv/family/frozen/2016/decay\endcsname{2.8}
\expandafter\def\csname valstore/arxiv/family/frozen/2024/future\endcsname{93.8}
\expandafter\def\csname valstore/arxiv/family/frozen/2024/decay\endcsname{0.5}
\expandafter\def\csname valstore/arxiv/family/recurrent/2000/future\endcsname{92.7}
\expandafter\def\csname valstore/arxiv/family/recurrent/2000/decay\endcsname{3.4}
\expandafter\def\csname valstore/arxiv/family/recurrent/2008/future\endcsname{95.2}
\expandafter\def\csname valstore/arxiv/family/recurrent/2008/decay\endcsname{3.4}
\expandafter\def\csname valstore/arxiv/family/recurrent/2016/future\endcsname{96.5}
\expandafter\def\csname valstore/arxiv/family/recurrent/2016/decay\endcsname{2.2}
\expandafter\def\csname valstore/arxiv/family/recurrent/2024/future\endcsname{96.3}
\expandafter\def\csname valstore/arxiv/family/recurrent/2024/decay\endcsname{0.2}
\expandafter\def\csname valstore/arxiv/family/textcnn/2000/future\endcsname{93.4}
\expandafter\def\csname valstore/arxiv/family/textcnn/2000/decay\endcsname{4.3}
\expandafter\def\csname valstore/arxiv/family/textcnn/2008/future\endcsname{94.7}
\expandafter\def\csname valstore/arxiv/family/textcnn/2008/decay\endcsname{3.9}
\expandafter\def\csname valstore/arxiv/family/textcnn/2016/future\endcsname{96.2}
\expandafter\def\csname valstore/arxiv/family/textcnn/2016/decay\endcsname{2.3}
\expandafter\def\csname valstore/arxiv/family/textcnn/2024/future\endcsname{95.8}
\expandafter\def\csname valstore/arxiv/family/textcnn/2024/decay\endcsname{0.3}
\expandafter\def\csname valstore/arxiv/family/transformer/2000/future\endcsname{93.5}
\expandafter\def\csname valstore/arxiv/family/transformer/2000/decay\endcsname{2.9}
\expandafter\def\csname valstore/arxiv/family/transformer/2008/future\endcsname{95.4}
\expandafter\def\csname valstore/arxiv/family/transformer/2008/decay\endcsname{3.0}
\expandafter\def\csname valstore/arxiv/family/transformer/2016/future\endcsname{96.5}
\expandafter\def\csname valstore/arxiv/family/transformer/2016/decay\endcsname{2.2}
\expandafter\def\csname valstore/arxiv/family/transformer/2024/future\endcsname{96.3}
\expandafter\def\csname valstore/arxiv/family/transformer/2024/decay\endcsname{0.2}
\expandafter\def\csname valstore/arxiv/family/frozen/indist\endcsname{95.5}
\expandafter\def\csname valstore/arxiv/family/frozen/future\endcsname{91.4}
\expandafter\def\csname valstore/arxiv/family/frozen/decay\endcsname{4.1}
\expandafter\def\csname valstore/arxiv/family/recurrent/indist\endcsname{98.0}
\expandafter\def\csname valstore/arxiv/family/recurrent/future\endcsname{95.0}
\expandafter\def\csname valstore/arxiv/family/recurrent/decay\endcsname{3.0}
\expandafter\def\csname valstore/arxiv/family/ffn/indist\endcsname{97.4}
\expandafter\def\csname valstore/arxiv/family/ffn/future\endcsname{94.4}
\expandafter\def\csname valstore/arxiv/family/ffn/decay\endcsname{3.0}
\expandafter\def\csname valstore/arxiv/family/textcnn/indist\endcsname{97.9}
\expandafter\def\csname valstore/arxiv/family/textcnn/future\endcsname{94.8}
\expandafter\def\csname valstore/arxiv/family/textcnn/decay\endcsname{3.1}
\expandafter\def\csname valstore/arxiv/family/transformer/indist\endcsname{97.9}
\expandafter\def\csname valstore/arxiv/family/transformer/future\endcsname{95.0}
\expandafter\def\csname valstore/arxiv/family/transformer/decay\endcsname{2.9}
\expandafter\def\csname valstore/arxiv/param/ffn_s/trainable\endcsname{99k}
\expandafter\def\csname valstore/arxiv/param/ffn_s/total\endcsname{124.7M}
\expandafter\def\csname valstore/arxiv/param/ffn_m/trainable\endcsname{397k}
\expandafter\def\csname valstore/arxiv/param/ffn_m/total\endcsname{125.0M}
\expandafter\def\csname valstore/arxiv/param/ffn_l/trainable\endcsname{1.6M}
\expandafter\def\csname valstore/arxiv/param/ffn_l/total\endcsname{126.2M}
\expandafter\def\csname valstore/arxiv/param/textcnn_s/trainable\endcsname{89k}
\expandafter\def\csname valstore/arxiv/param/textcnn_s/total\endcsname{124.7M}
\expandafter\def\csname valstore/arxiv/param/textcnn_m/trainable\endcsname{462k}
\expandafter\def\csname valstore/arxiv/param/textcnn_m/total\endcsname{125.1M}
\expandafter\def\csname valstore/arxiv/param/textcnn_l/trainable\endcsname{1.9M}
\expandafter\def\csname valstore/arxiv/param/textcnn_l/total\endcsname{126.6M}
\expandafter\def\csname valstore/arxiv/param/bigru_s/trainable\endcsname{100k}
\expandafter\def\csname valstore/arxiv/param/bigru_s/total\endcsname{124.7M}
\expandafter\def\csname valstore/arxiv/param/bilstm_m/trainable\endcsname{537k}
\expandafter\def\csname valstore/arxiv/param/bilstm_m/total\endcsname{125.2M}
\expandafter\def\csname valstore/arxiv/param/bilstm_attn_l/trainable\endcsname{2.2M}
\expandafter\def\csname valstore/arxiv/param/bilstm_attn_l/total\endcsname{126.8M}
\expandafter\def\csname valstore/arxiv/param/tx_s/trainable\endcsname{83k}
\expandafter\def\csname valstore/arxiv/param/tx_s/total\endcsname{124.7M}
\expandafter\def\csname valstore/arxiv/param/tx_m/trainable\endcsname{493k}
\expandafter\def\csname valstore/arxiv/param/tx_m/total\endcsname{125.1M}
\expandafter\def\csname valstore/arxiv/param/tx_l/trainable\endcsname{1.9M}
\expandafter\def\csname valstore/arxiv/param/tx_l/total\endcsname{126.5M}
\expandafter\def\csname valstore/arxiv/param/minilm_l6_frozen/trainable\endcsname{3k}
\expandafter\def\csname valstore/arxiv/param/minilm_l6_frozen/total\endcsname{22.7M}
\expandafter\def\csname valstore/arxiv/param/distilbert_base_frozen/trainable\endcsname{5k}
\expandafter\def\csname valstore/arxiv/param/distilbert_base_frozen/total\endcsname{66.4M}
\expandafter\def\csname valstore/arxiv/param/bert_base_frozen/trainable\endcsname{5k}
\expandafter\def\csname valstore/arxiv/param/bert_base_frozen/total\endcsname{109.5M}
\expandafter\def\csname valstore/arxiv/param/roberta_base_frozen/trainable\endcsname{5k}
\expandafter\def\csname valstore/arxiv/param/roberta_base_frozen/total\endcsname{124.7M}
\expandafter\def\csname valstore/arxiv/param/deberta_v3_base_frozen/trainable\endcsname{5k}
\expandafter\def\csname valstore/arxiv/param/deberta_v3_base_frozen/total\endcsname{183.8M}
\expandafter\def\csname valstore/arxiv/param/electra_base_frozen/trainable\endcsname{5k}
\expandafter\def\csname valstore/arxiv/param/electra_base_frozen/total\endcsname{108.9M}
\expandafter\def\csname valstore/arxiv/param/mpnet_base_frozen/trainable\endcsname{5k}
\expandafter\def\csname valstore/arxiv/param/mpnet_base_frozen/total\endcsname{109.5M}
\expandafter\def\csname valstore/arxiv/param/modernbert_base_frozen/trainable\endcsname{5k}
\expandafter\def\csname valstore/arxiv/param/modernbert_base_frozen/total\endcsname{149.0M}

\usepackage{hyperref}
\usepackage[capitalize,noabbrev]{cleveref}
\definecolor{linknavy}{rgb}{0,0.08,0.45}
\hypersetup{colorlinks=true,linkcolor=linknavy,citecolor=linknavy,urlcolor=linknavy,pdfborder={0 0 0}}

\newif\iflncs

\providecommand{\tightvspace}[1]{}

\newif\ifverbosemath
\verbosemathfalse

\newcommand{\extendedpreprintlink}{\href{https://drift-happens.org/drift-happens.pdf}{extended preprint}}
\newcommand{\resultssitelink}{\href{https://drift-happens.org}{drift-happens.org}}

\providecommand{\forgettingfloat}{\begin{figure}[H]}
\providecommand{\forgettinggraphic}{%
  \includegraphics[width=0.45\textwidth]{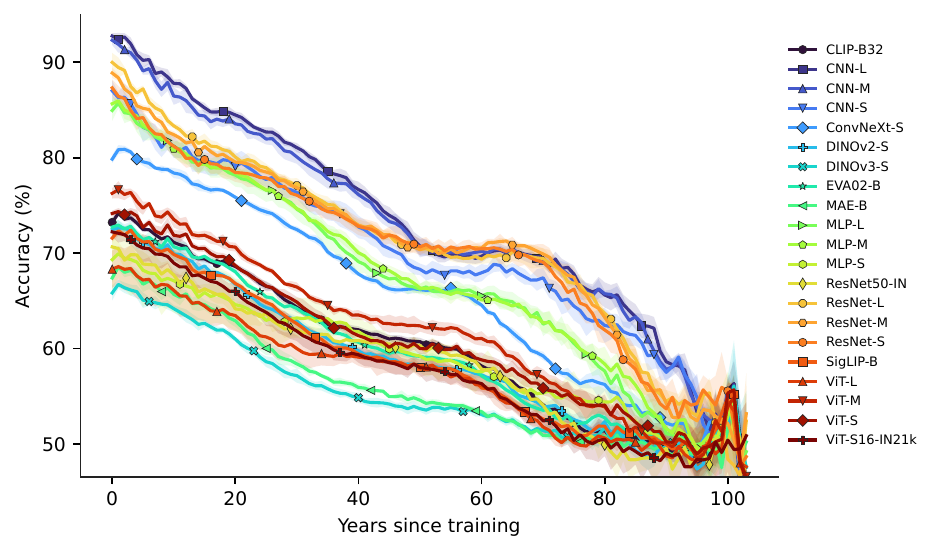}}

\lncsfalse  % extended build: keep the full reference list and author names

\renewcommand{\tightvspace}[1]{\vspace{#1}}

\verbosemathtrue

\setcitestyle{numbers,square}

\titlerunning{Drift Happens: Neural Architecture Robustness to Temporal Distribution Shift}

\begin{document}

\hypersetup{pdfauthor={Robin Holzinger, Riccardo Colletti}}

% The equal-contribution note is a symbol-marked footnote hung off the authors;
% switch footnote marks to symbols for it, then restore numerals for the body.
\renewcommand{\thefootnote}{\fnsymbol{footnote}}

\twocolumn[
\papertitle{Drift Happens: An Empirical Study of Neural Architecture \\ Robustness to Temporal Distribution Shift}
\paperkeywords{Temporal distribution shift, Concept drift, Inductive bias, Neural architecture robustness, Pretrained encoders}

{\centering
{\large\bfseries Robin Holzinger\footnotemark[1] \qquad Riccardo Colletti\footnotemark[1]\par}
\vskip 0.9em
{\normalsize Department of Electrical Engineering and Computer Sciences,\\
University of California, Berkeley, USA\par}
}

\vskip 1.4em
\venuebanner{Extended version. Accepted at QCDS 2026; the proceedings version will appear in Springer LNCS.}
\vskip 0.15in
]

% Footnote text for the author markers; lands in the footnote area of column one.
\footnotetext[1]{Both authors contributed equally.}
\renewcommand{\thefootnote}{\arabic{footnote}}
\setcounter{footnote}{0}

\begin{abstract}
% Abstract body only. The surrounding \begin{abstract}...\end{abstract}
% (and keyword handling) lives in each build root, since LNCS and the
% two-column preprint wrap the abstract differently.
Real-world data distributions evolve over time, inducing temporal distribution shift that can substantially degrade the reliability of deployed machine learning systems. However, the extent to which architectural choices and their associated inductive biases affect temporal robustness remains insufficiently understood.

We present a systematic empirical comparison of temporal robustness across three heterogeneous, time-indexed domains encompassing image classification, multi-label text classification, and text regression tasks. Using a unified evaluation framework based on temporal drift matrices, we train models on cumulative historical data and evaluate their performance on both earlier and later time periods, thereby quantifying cross-temporal generalization. Our study spans model families ranging from simple multilayer perceptrons and convolutional networks to recurrent networks and pretrained Transformer-based encoders.

Collectively, the results show that architectural inductive biases systematically shape temporal robustness: models whose inductive biases lead them to exploit localized, highly discriminative features attain the highest in-distribution accuracy, yet those features are often the ones that change most over time, so these models degrade fastest, while pretrained encoders that draw on coarser, more stable representations drift more gradually. These observations offer practical guidance for selecting architectures for real-world systems subject to temporal drift.

\end{abstract}

\tightvspace{-0.55cm}
\section{Introduction}
\label{section_introduction}

Machine learning models are typically trained under the assumption that training and test data are drawn from the same distribution. In practice, this assumption rarely holds: real-world data evolves over time, exhibiting \textit{temporal distribution shift} that can degrade model performance in ways that standard held-out evaluation fails to capture. A model that achieves high accuracy on held-out data from the same time period may fail when deployed on future inputs.

While distribution shift is well-documented, less understood is how architectural choices influence a model's robustness to temporal drift. \begin{quote}
\emph{Do different inductive biases (the translation invariance of convolutions, the sequential modeling of recurrent nets, the attention of Transformers) lead to different rates of temporal degradation?}

\emph{Do frozen, pretrained encoders resist temporal drift better than models trained end to end?}
\end{quote} These questions have practical implications for model selection, yet systematic comparisons across architectures and domains remain scarce.

This work investigates the temporal robustness of neural classifiers across three domains: image classification (\texttt{Yearbook}), text regression (\texttt{Amazon Reviews}), and multi-label text classification (\texttt{arXiv}). For each domain, we evaluate diverse architectures, from simple baselines to pretrained transformers, using a unified framework based on temporal drift matrices and provide qualitative explanations for model degradation via gradient saliency maps. Our contributions are threefold:

\tightvspace{-7pt}
\begin{itemize}
    \item We provide a unified empirical assessment of temporal robustness across three long-range, time-indexed domains, enabling direct comparison of how neural architectures behave under temporal distribution shift.
    \item We organize time-indexed evaluation in the spirit of Wild-Time \cite{yao2022wildtime} into temporal drift matrices, a compact representation that quantifies cross-temporal generalization by measuring performance when training on cumulative historical data and testing on both earlier and later time periods.
    \item We systematically compare a broad spectrum of model families, from multilayer perceptrons, convolutional and residual networks, and recurrent networks to transformers trained on the data and frozen pretrained encoders, highlighting how architectural assumptions relate to degradation patterns across modalities and tasks.
\end{itemize}
\tightvspace{-7pt}

\noindent Taken together, our results characterize how performance deteriorates as the temporal gap between training and evaluation widens across domains and model classes. \textbf{The features an inductive bias extracts within the training period are what lift in-distribution performance, yet they are the most tied to it and the first to degrade as the data drifts, so the very features that make a model accurate in distribution are the least robust over time.} Frozen pretrained encoders, relying on coarser and more transferable representations, trade in-distribution accuracy for steadier degradation. These findings guide architecture selection in non-stationary real-world environments.

\section{Background and Related Work}
\label{section_background}

\subsection{Temporal Distribution Shift}
A central challenge in real-world machine learning systems is that the data-generating process is rarely stationary. As models are deployed over months, years, or decades, both inputs and label semantics may evolve, causing systematic discrepancies between distributions encountered during training and those observed at inference. Such temporal evolution, commonly referred to as \emph{temporal distribution shift}, is pervasive across domains. Understanding the structure of this shift is essential for assessing and improving temporal robustness. However, prior work has largely focused on static or domain-level distribution shifts \cite{ben2010theory,gulrajani2021in_search_of_lost,koh2021wilds}, with limited attention to drift that unfolds sequentially over time, particularly in non-generated, in-the-wild settings \cite{yao2022wildtime}.

Following established terminology in concept drift research \cite{gama2014a_survey_on_concept_drift}, temporal distribution shift can be characterized along three complementary axes (Figure \ref{fig:distribution_shift}):

\textbf{Covariate shift} occurs when the input distribution $P(X)$ changes while the conditional $P(Y \mid X)$ remains fixed. This type of shift commonly arises in natural data streams where observational setups, sociocultural conventions, or user behavior gradually evolve.

\textbf{Label shift} refers to changes in the marginal label distribution $P(Y)$. Long-term textual or behavioral datasets frequently exhibit such imbalance drift as the prevalence of topics, categories, or rating patterns changes over time.

\textbf{Concept drift} denotes changes in the conditional distribution $P(Y \mid X)$, implying that identical inputs may correspond to different labels at different time points. This form of drift is particularly pronounced in domains where semantics or visual attributes evolve, such as historical portraits \cite{ginosar2015century} or online platforms.

In practice, these forms of drift rarely occur in isolation and often manifest in combination \cite{aguiar2023comprehensiveanalysisconceptdrift}. As a result, the temporal robustness of a model depends not only on the magnitude of drift but also on how its architectural assumptions and inductive biases interact with the evolving data distribution.

\subsection{Model Robustness to Distribution Shift}
% Shared by paper/main-lncs.tex and paper/main-preprint.tex.
% Define a command to draw stars
\newcommand{\drawstar}[2]{
    \draw[fill=#1] (#2) -- ++(0.1,0.03) -- ++(0.03,0.1) -- ++(0.03,-0.1) -- ++(0.1,-0.03) -- ++(-0.1,-0.03) -- ++(-0.03,-0.1) -- ++(-0.03,0.1) -- cycle;
}
\begin{figure}[t]
    \centering
    \resizebox{\linewidth}{!}{
        \begin{tikzpicture}
            % Axes
            \draw[->] (-0.2,0) -- (1.5,0) node[below] {};
            \draw[->] (0,-0.2) -- (0,1.5) node[left] {};

            % Decision boundary (black line)
            \draw[thick] (0.1, 0.25) .. controls (0.5, 0.2) and (0.9, 1)  .. (1.4, 1);

            % Blue dots (class 1) with black border
            \filldraw[black, fill=accent_blue] (0.3, 0.5) circle (0.1);
            \filldraw[black, fill=accent_blue] (0.65, 1) circle (0.1);
            \filldraw[black, fill=accent_blue] (1.2, 1.2) circle (0.1);

            % Red stars (class 2)
            \drawstar{accent_orange}{0.45, 0.25};
            \drawstar{accent_orange}{0.85, 0.5};
            \drawstar{accent_orange}{1.05, 0.25};
            \drawstar{accent_orange}{1.25, 0.75};

            % Label
            \node at (0.75, 1.7) {\scriptsize Original};
        \end{tikzpicture}

        \hspace{0.15cm}

        \begin{tikzpicture}
            % Concept Drift
            % Axes
            \draw[->] (-0.2,0) -- (1.5,0) node[below] {};
            \draw[->] (0,-0.2) -- (0,1.5) node[left] {};

            % Decision boundary (black line)
            \draw[thick] (0.1, 0.7) .. controls (0.9, 1) and (1.3, 0.5)  .. (1.4, 0.42);

            % Blue dots (class 1) with black border
            \filldraw[black, fill=accent_blue] (0.65, 1) circle (0.1);
            \filldraw[black, fill=accent_blue] (1.2, 1.2) circle (0.1);
            \filldraw[black, fill=accent_blue] (1.25, 0.75) circle (0.1);

            % Red stars (class 2)
            \drawstar{accent_orange}{0.3, 0.5};
            \drawstar{accent_orange}{0.45, 0.25};
            \drawstar{accent_orange}{0.85, 0.5};
            \drawstar{accent_orange}{1.05, 0.25};

            % Label
            \node at (0.75, 1.7) {\scriptsize Concept Drift};
        \end{tikzpicture}

        \hspace{0.15cm}

        \begin{tikzpicture}
            % Virtual Drift
            % Axes
            \draw[->] (-0.2,0) -- (1.5,0) node[below] {};
            \draw[->] (0,-0.2) -- (0,1.5) node[left] {};

            % Decision boundary (black line)
            \draw[thick] (0.1, 0.25) .. controls (0.5, 0.2) and (0.9, 1)  .. (1.4, 1);

            % slightly shifted
            % \draw[thick] (0.1, 0.15) .. controls (0.8, 0.2) and (0.9, 1)  .. (1.4, 1);

            % Blue dots (class 1) with black border
            \filldraw[black, fill=accent_blue] (0.3, 0.55) circle (0.1);
            \filldraw[black, fill=accent_blue] (0.25, 0.75) circle (0.1);
            \filldraw[black, fill=accent_blue] (0.5, 0.85) circle (0.1);

            % Red stars (class 2)
            \drawstar{accent_orange}{1, 0.5};
            \drawstar{accent_orange}{1.25, 0.35};
            \drawstar{accent_orange}{1.3, 0.7};

            % Label
            \node at (0.75, 1.7) {\scriptsize Covariate Drift};
            \node at (0.75, 1.4) {\tiny (Virtual Drift)};
        \end{tikzpicture}

        \hspace{0.15cm}

        \begin{tikzpicture}
            % Prior Probability Shift
            % Axes
            \draw[->] (-0.2,0) -- (1.5,0) node[below] {};
            \draw[->] (0,-0.2) -- (0,1.5) node[left] {};

            % Decision boundary (black line)
            \draw[thick] (0.1, 0.25) .. controls (0.5, 0.2) and (0.9, 1)  .. (1.4, 1);

            % slightly shifted
            % \draw[thick] (0.2, 0.15) .. controls (0.4, 0.1) and (0.8, 1)  .. (1.4, 0.9);

            % Blue dots (class 1) with black border
            \filldraw[black, fill=accent_blue] (0.3, 0.5) circle (0.1);
            \filldraw[black, fill=accent_blue] (0.65, 1) circle (0.1);
            \filldraw[black, fill=accent_blue] (1.2, 1.2) circle (0.1);
            \filldraw[black, fill=accent_blue] (0.35, 1.1) circle (0.1);
            \filldraw[black, fill=accent_blue] (0.4, 0.8) circle (0.1);
            \filldraw[black, fill=accent_blue] (0.6, 1.2) circle (0.1);
            \filldraw[black, fill=accent_blue] (0.15, 0.7) circle (0.1);
            \filldraw[black, fill=accent_blue] (0.25, 0.85) circle (0.1);

            % Red stars (class 2)
            \drawstar{accent_orange}{0.85, 0.5};
            \drawstar{accent_orange}{1.05, 0.25};

            % Label
            \node at (0.75, 1.7) {\scriptsize Label Drift};
        \end{tikzpicture}

    }
    \caption{Drift types in a binary classification setting. Circles and stars indicate the label classes; the curve represents the decision boundary. \textit{Concept drift} induces a change in the decision boundary, \textit{Covariate drift} (virtual drift) changes the input distribution, and \textit{Label drift} alters the relative class frequency \iflncs\cite{gama2014a_survey_on_concept_drift}\else\cite{holzinger2024analysis}\fi.}
    \tightvspace{-5pt}
    \label{fig:distribution_shift}
    \tightvspace{-0.2cm}
\end{figure}

Understanding how learning algorithms behave under distribution shift has become an important research direction. Early work approached robustness through domain adaptation \cite{ben2010theory} and out-of-distribution generalization \cite{gulrajani2021in_search_of_lost}, formalizing shift as a transition between a small number of discrete source and target domains. The introduction of large-scale benchmarks such as \texttt{WILDS} \cite{koh2021wilds}, which focuses on naturally occurring distribution shifts without explicit temporal indexing, and more recently \texttt{Wild-Time} \cite{yao2022wildtime}, which explicitly models time-indexed data and temporal distribution shifts, has broadened this perspective by providing evaluation protocols that more closely reflect real-world deployment scenarios.

A complementary line of research has examined robustness through model diagnostics and monitoring. Methods for detecting drift onset by examining changes in latent representations or predictive uncertainty have shown promise in deep learning settings \iflncs\cite{rabanser_failing_2019}\else\cite{rabanser_failing_2019,ackerman2021automaticallydetectingdatadrift}\fi. Recent empirical analyses have characterized how concept drift manifests in practice, including its locality and temporal progression across large-scale data streams \cite{aguiar2023comprehensiveanalysisconceptdrift}. At the systems level, work on data-centric and continual-learning infrastructures has explored how to maintain model quality over time through cost-aware retraining and pipeline orchestration \iflncs\cite{mahadevan_cost_aware_2024,boether_modyn_2025}\else\cite{mahadevan_cost_effective_2023,mahadevan_cost_aware_2024,tian_continuum_2018,boether_modyn_2025,holzinger2024analysis}\fi.

Despite these advances, robustness studies commonly focus either on a single architecture evaluated across multiple datasets or on a single dataset used to compare a narrow set of architectures. As a result, comparatively little is known about how architectural design choices interact with long-range temporal drift across heterogeneous modalities and tasks. This gap is particularly relevant given the diversity of inductive biases exhibited by modern neural models: convolutional networks encode locality and translation equivariance, recurrent networks capture sequential structure \cite{hochreiter1997long,cho2014learning}, and Transformer-based encoders rely on self-attention with minimal structural priors \cite{vaswani2017attention}. Likewise, large-scale pretraining in vision and language \cite{devlin2019bert,liu2019roberta} introduces representations shaped by broad historical data, yet their behavior under extended temporal drift remains poorly characterized.

The present study addresses this open question by comparing these architectural families under a shared temporal evaluation protocol and across multiple modalities, enabling a controlled analysis of how model design influences robustness under real-world temporal distribution shift.

\subsection{Robustness Evaluation vs.\ Temporal Adaptation}

This work evaluates the inherent temporal robustness of neural architectures under distribution shift, measuring how different model families, trained only on cumulative historical data, perform as the temporal gap between training and evaluation widens. This isolates the robustness arising from architectural design and pretraining alone, prior to any adaptive intervention.

A complementary line of research instead investigates how models can \emph{adapt} once drift is detected, \iflncs through scheduled or cost-aware retraining \cite{mahadevan_cost_aware_2024} and continual-learning pipelines \cite{boether_modyn_2025}\else through scheduled or cost-aware retraining \cite{mahadevan_cost_effective_2023,mahadevan_cost_aware_2024}, continual-learning pipelines \cite{tian_continuum_2018,boether_modyn_2025,holzinger2024analysis}, or accuracy-aware data maintenance \cite{ralf2023wooders}\fi, addressing when to retrain, how much data to incorporate, and how to trade performance against cost. Our study is orthogonal to that literature and provides a foundation on which such adaptation strategies can be designed, evaluated, and compared.

\section{Methods and Experimental Approach}
\label{section_methods}

Temporal distribution shift manifests differently across modalities, tasks, and time scales, yet existing empirical studies typically vary a single axis at a time, leaving open how \emph{architectural design}, \emph{label structure}, and \emph{drift mechanism} jointly shape temporal robustness (Section~\ref{section_background}). Our setup targets exactly this cross-cutting comparison: we combine three long-range, time-indexed datasets with a diverse suite of neural architectures spanning multiple inductive biases, all evaluated under the unified temporal protocol of Section~\ref{section:evaluation_framework} (temporal drift matrices), so that architectural differences, rather than differences in splitting or evaluation, drive the observed robustness patterns.

\subsection{Datasets}
Our empirical analysis spans three qualitatively distinct temporal scenarios, chosen to cover different modalities, tasks, and sources of distribution shift. Each dataset captures multiple decades of real-world temporal evolution, making it suitable for cross-temporal evaluation.

\subsubsection{Yearbook: A Century of Portraits}
The \texttt{Yearbook} dataset \iflncs\cite{ginosar2015century}\else\cite{yearbook_ginosar_2019}\fi{} contains 37{,}921 frontal portraits of American high-school seniors from \val{yearbook/cutoff-first} to \val{yearbook/cutoff-last}, which we align and process as 3-channel $32\times32$ tensors. Although acquisition is largely standardized, stylistic attributes (e.g., hairstyles, clothing, accessories) vary substantially across decades. The dataset provides binary sex labels that are approximately balanced over time. It is a canonical benchmark for temporal shift: Wild-Time \cite{yao2022wildtime} uses a pre-/post-1970 split and reports marked out-of-distribution degradation, and Modyn \cite{boether_modyn_2025} observes accuracy decay as the train--test time gap grows. We reproduce this trend (Fig.~\ref{fig:mean_matrices}), consistent with covariate and concept drift.

\subsubsection{Amazon Reviews 2023: E-commerce}
The \texttt{Amazon Reviews} dataset \cite{hou2024bridging} comprises 571.54 million reviews across 33 product categories, spanning May 1996 to September 2023. Each review has a timestamp, star rating, and free-text content. We cast a review-level sentiment regression task, predicting the 1--5 rating from the text, which exhibits strong covariate and concept drift due to evolving language, consumer behavior, and platform usage. We focus on seven categories, restrict to 2014--2023, and draw a stratified sample of 300{,}000 reviews.

\subsubsection{arXiv: Scientific Discourse}
The \texttt{arXiv} dataset \cite{arxiv2024} defines a multi-label task over 2{,}866{,}787 title--abstract records annotated with 176 subject categories. We concatenate titles and abstracts, keep seven leaf categories (\cref{app:arxiv}), and use \val{arxiv/cutoff-first}--\val{arxiv/cutoff-last} submissions whose categories fall within them (papers may carry several). Category mix and terminology shift, inducing label and covariate drift\iflncs\else\ \cite{holzinger2024analysis}\fi. Wild-Time \cite{yao2022wildtime} reports roughly 20\% degradation under temporal splits for a related arXiv task, and the gap is not substantially closed by domain generalization or continual learning methods, motivating our architectural comparison.

\subsection{Model Implementation}

We evaluate architectures spanning different inductive biases to understand how model design affects temporal robustness. Rather than covering the full architecture landscape, we select families that span the spectrum of inductive-bias strength: from simple baselines without structural assumptions that establish lower bounds on performance, to modern architectures that incorporate structural priors, to pre-trained models that leverage large-scale external data. This spread is what lets us attribute robustness differences to the priors themselves.

\subsubsection{Image Classification}

For the \texttt{Yearbook} dataset, we evaluate four model families trained on the data, each at three sizes (small, medium, large), for \val{yearbook/scratch-models} architectures, complemented by \val{yearbook/frozen-models} pretrained vision encoders used as frozen backbones.

The four families differ in the spatial prior they encode. At one extreme, the \emph{multilayer perceptron} (\texttt{MLP}) flattens the image and applies only fully connected layers, imposing no spatial structure. The \emph{convolutional network} (\texttt{CNN}) builds in locality and translation equivariance through convolutional filters with batch normalization and pooling, and the \emph{residual network} (\texttt{ResNet}) extends it with skip connections \cite{he2016resnet} that ease optimization at greater depth. At the other extreme, the \emph{vision Transformer} (\texttt{ViT}) drops convolution for self-attention over patches \cite{dosovitskiy2021vit}, a far weaker spatial prior, with a small patch size suited to the $32\times32$ inputs. Across all four, the small, medium, and large variants scale depth and width.

Finally, we assess \emph{transfer learning} using pretrained vision encoders trained on large-scale curated or web-scale datasets. The underlying hypothesis is that representations learned from diverse data may exhibit stronger temporal robustness than features learned solely from historical portraits. We consider \val{yearbook/frozen-models} pretrained models. The self-supervised \texttt{DINOv2-S} \cite{oquab2024dinov2} and \texttt{DINOv3-S} \cite{simeoni2025dinov3} are pretrained on large curated image collections; \texttt{CLIP-B32} \cite{radford2021clip} and \texttt{SigLIP-B} \cite{zhai2023siglip} use contrastive image-text pretraining, the latter with a sigmoid loss; \texttt{ConvNeXt-S} \cite{liu2022convnext}, \texttt{ResNet50-IN} \cite{he2016resnet}, and \texttt{ViT-S16-IN21k} \cite{dosovitskiy2021vit} are trained with supervised \texttt{ImageNet} labels, the last on \texttt{ImageNet-21k}; and \texttt{MAE-B} \cite{he2022mae} and \texttt{EVA02-B} \cite{fang2024eva02} are pretrained with masked image modeling. In all cases, we freeze the pretrained backbone and train only a linear classification head, isolating the contribution of pretrained representations from the effects of fine-tuning dynamics.

\subsubsection{Text Models}

For the \texttt{Amazon Reviews} and \texttt{arXiv} datasets, we evaluate model families that differ in their inductive biases for representing textual structure. The same architectures serve both datasets, differing only in output layer and loss, with \texttt{Amazon Reviews} using regression under a weighted mean squared error and arXiv multi-label classification under a weighted binary cross-entropy.

All four families operate on cached \texttt{RoBERTa} token embeddings. The \emph{feed-forward baseline} (\texttt{FFN}) averages them into a single 768-dimensional vector and passes it through a small MLP head that discards order entirely, with variants scaling the hidden width from 128 to 2048. The \emph{convolutional model} (\texttt{TextCNN}) \cite{kim2014convolutional} applies one-dimensional convolutions to capture local n-gram patterns, scaling the number and width of its filters. The \emph{recurrent} models read the sequence token by token, as a single-layer bidirectional \texttt{GRU} \cite{cho2014learning}, a single-layer bidirectional \texttt{LSTM} \cite{hochreiter1997long}, and a two-layer bidirectional \texttt{LSTM} with attention. The \emph{Transformer} encoders replace recurrence with self-attention \cite{vaswani2017attention} and learnable positional embeddings, ranging from 1 to 5 layers and 4 to 6 heads.

\iflncs\noindent\fi Finally, we assess transfer learning from pretrained language encoders. We consider frozen encoders with a light head trained on the pooled output: \texttt{BERT} \cite{devlin2019bert}, \texttt{RoBERTa} \cite{liu2019roberta}, \texttt{DeBERTa-v3} \cite{he2023debertav3}, \texttt{ELECTRA} \cite{clark2020electra}, \texttt{MPNet} \cite{song2020mpnet}, \texttt{ModernBERT} \cite{warner2024modernbert}, and \texttt{DistilBERT} \cite{sanh2019distilbert} on both text tasks, plus \texttt{MiniLM-L6} \cite{wang2020minilm} on \texttt{arXiv}. As with the vision encoders, freezing isolates pretrained representations from fine-tuning.

\section{Evaluation Framework}
\label{section:evaluation_framework}

Standard held-out evaluation measures performance on data from the training distribution, which under temporal shift can look strong even as the model fails on future data. We therefore measure cross-temporal generalization explicitly.

\subsection{Temporal Drift Matrices}

Let $\mathcal{D} = \{(x_i, y_i, t_i)\}_{i=1}^{N}$ denote a dataset where each example is associated with a timestamp $t_i$. We divide the timeline into $K$ disjoint intervals $\mathcal{T} = \{T_1, \ldots, T_K\}$, where $T_k = [t_k^{\text{start}}, t_k^{\text{end}})$. Let $\mathcal{D}_k = \{(x, y, t) \in \mathcal{D} : t \in T_k\}$ denote the subset of data from interval $k$.

For training, we construct cumulative datasets $\mathcal{D}_{\leq k} = \bigcup_{j=1}^{k} \mathcal{D}_j$ containing all data up to and including interval $k$. A model $f_k$ trained on $\mathcal{D}_{\leq k}$ has access to historical data over time $t_k^{\text{end}}$ but no knowledge of future periods. This cumulative strategy reflects realistic deployment scenarios in which models are periodically retrained on all historical data.

The \textit{temporal drift matrix} $M \in \mathbb{R}^{K \times K}$ captures cross-temporal generalization:
\begin{equation}
    M_{ij} = \text{perf}(f_i, \mathcal{D}_j)
\end{equation}
where $\text{perf}(\cdot, \cdot)$ denotes a performance metric (accuracy, macro AUC, or balanced MSE depending on the task). Entry $M_{ij}$ measures how well a model trained on data through period $i$ performs on data from period $j$.

\iflncs\noindent\fi The structure of $M$ reveals aspects of temporal robustness. Following our plotting convention, rows are the training cutoff (``trained up to'') and columns the evaluation period (``evaluated on''), with time increasing upward and to the right from a lower-left origin. The diagonal $M_{ii}$ is in-distribution performance on data held out from the training period; the upper-left ($j < i$) is held-out performance on earlier in-training periods; and the lower-right ($j > i$) is forward generalization to data unseen during training. This lets us read how performance degrades (or improves) as the temporal gap between training and evaluation widens.

\subsection{Temporal Splitting Strategy}

Each slice $\mathcal{D}_k$ is partitioned once into a stratified training split $\mathcal{D}_k^{\text{train}}$ (70\%) and a held-out test split $\mathcal{D}_k^{\text{test}}$ (30\%), shared across all models via a fixed split seed. For in-distribution evaluation (when $j \leq i$), we use only the held-out test split of $\mathcal{D}_j$ to prevent data leakage:

\begin{equation}
    \nonumber
    \mathcal{D}_{\text{eval}}^{(i,j)} =
    \begin{cases}
        \mathcal{D}_{j}^{\text{test}} & \text{if } j \leq i \text{ (in-distribution)} \\
        \mathcal{D}_{j}^{\text{train}} \cup \mathcal{D}_{j}^{\text{test}} & \text{if } j > i \text{ (out-of-distrib.)}
    \end{cases}
\end{equation}

For out-of-distribution evaluation on future time slices, we use all available samples from that period to maximize statistical power, since by definition none of this data was seen during training. Evaluating on periods that overlap the training data ($j \leq i$) is deliberate: these held-out entries verify that a model retains performance across the historical periods it was trained on and provide the in-distribution reference from which forward decay is measured.

\subsection{Training Protocol}

The image models are trained with Adam \cite{kingma2015adam} and the text models with its decoupled-weight-decay variant AdamW, a standard choice that behaves robustly across heterogeneous architectures; fixing one optimizer per modality avoids per-model optimizer tuning as a confound. Learning rates are task-specific, with exact hyperparameters pinned in versioned experiment presets. Training proceeds for a fixed number of epochs, and model selection is based on the final checkpoint. Every configuration is trained under multiple random seeds, five on \texttt{Yearbook} and three on the text tasks (\cref{app:reproducibility}), and all results average over seeds.

We adopt a cumulative temporal training strategy: for each slice $T_k$ we train a model $f_k$ on $\mathcal{D}_{\leq k}$, all examples observed up to $T_k$, yielding a sequence $\{f_1, \dots, f_K\}$ that each represent the best model obtainable from the data available at that point in time. All checkpoints are stored and evaluated on every slice to construct the full temporal drift matrix $M$.

The loss follows the task structure and is tailored to address label imbalance.

For \emph{binary classification} on \texttt{Yearbook}, we minimize standard two-class cross-entropy on logits and labels, the maximum-likelihood objective for a two-class softmax model.

\ifverbosemath
For \emph{multi-label classification} on \texttt{arXiv}, each paper may belong to multiple subject categories. We keep only seven leaf categories as the label space, retaining papers that carry at least one of them. The label distribution remains skewed, with negative examples substantially outnumbering positives for each category. We therefore use a weighted binary cross-entropy with logits, applied independently to each of the $C = 7$ categories:
\begin{equation}
\begin{aligned}
\mathcal{L}_{\text{BCE}}
    &= -\frac{1}{n}\sum_{i=1}^{n} \sum_{c=1}^{C}
       \big[ w_c \, y_{ic} \log(\sigma(z_{ic})) \\
    &\qquad + (1-y_{ic}) \log(1-\sigma(z_{ic})) \big],
\end{aligned}
\end{equation}
where $z_{ic}$ denotes the logit for class $c$, $\sigma(\cdot)$ is the sigmoid function, and $y_{ic} \in \{0,1\}$ is the binary label. The class weights $w_c$ are defined as
\begin{equation}
    w_c = \frac{|\{i : y_{ic} = 0\}|}{|\{i : y_{ic} = 1\}|},
\end{equation}
i.e., the ratio of negative to positive examples for each category. This reweighting compensates for label imbalance by amplifying the contribution of rare positive labels, preventing the model from achieving deceptively high accuracy by predicting the all-zero vector.

For \emph{regression} on \texttt{Amazon Reviews}, we train models to predict the star rating using a weighted mean squared error:
\begin{equation}
    \mathcal{L}_{\text{WMSE}} = \frac{1}{n}\sum_{i=1}^{n} w_{y_i} (y_i - \hat{y}_i)^2,
\end{equation}
where $y_i \in \{1,2,3,4,5\}$ is the true rating and $\hat{y}_i$ is the prediction. The rating-specific weights $w_r$ are defined as
\begin{equation}
    w_r = \frac{n}{n_r},
\end{equation}
with $n$ the total number of samples and $n_r$ the number of samples with rating $r$. Because the empirical rating distribution is heavily skewed toward high scores, an unweighted \texttt{MSE} would be dominated by the majority class and largely ignore rare low-rating events. The inverse-frequency weighting counteracts this imbalance, ensuring that errors on minority ratings remain visible in the objective and that temporal degradation in performance cannot be explained solely by changes in the prevalence of positive reviews.
\else
For \emph{multi-label classification} on \texttt{arXiv}, where each paper's categories all lie within the seven leaf categories and negatives greatly outnumber positives, we use a class-weighted binary cross-entropy with logits over the $C = 7$ categories. Each class $c$ is weighted by its negative-to-positive ratio $w_c = |\{i : y_{ic} = 0\}| / |\{i : y_{ic} = 1\}|$, which amplifies rare positives so the all-zero vector cannot score well.

For \emph{regression} on \texttt{Amazon Reviews}, we predict the star rating $y_i \in \{1,\dots,5\}$ with an inverse-frequency-weighted mean squared error, weighting each rating $r$ by $w_r = n/n_r$ ($n$ total samples, $n_r$ with rating $r$). Because the rating distribution is skewed toward high scores, this keeps errors on rare low ratings visible, so temporal degradation cannot be explained by rating prevalence alone.
\fi

\subsection{Evaluation Metrics}

To populate each entry of the drift matrix $M$, we require a scalar performance metric per train-test time pair. We match each metric to the task's label structure: accuracy for \texttt{Yearbook}, whose binary labels are approximately balanced (\cref{app:metrics_yearbook}); balanced MSE ($\text{MSE}_{\text{bal}}$) for \texttt{Amazon Reviews}, which reweights its skewed rating distribution (\cref{app:metrics_amazon}); and macro AUC ($\overline{\text{AUC}}$) for \texttt{arXiv}, whose imbalanced multi-label categories require a threshold-free, per-class average (\cref{app:metrics_arxiv}). The two classification tasks thus receive different metrics because their label structures differ; the drift protocol itself is identical.

\subsection{Summary Statistics of the Drift Matrix}
\label{subsec:drift_summary}

In the drift matrix $M$, the row index $i$ is the training cutoff and the column index $j$ is the evaluation period, so the entry $M_{ij}$ is the performance of a model trained on all data up to period $i$ when it is tested on period $j$.
\ifverbosemath
Three numbers summarize the matrix, each averaged only over the cells that are filled, since a train-test pair with no completed run leaves its cell empty. The \emph{in-distribution} score is the average of the diagonal,
\begin{equation}
    \mathrm{ID}(M) = \operatorname{mean}_{i} M_{ii},
    \label{eq:in_distribution}
\end{equation}
a model's performance on the same period it was trained through. The \emph{future} score averages the cells with $j > i$, where the evaluation period falls after the training cutoff,
\begin{equation}
    \mathrm{Fut}(M) = \operatorname{mean}_{j > i} M_{ij}.
    \label{eq:future}
\end{equation}
The \emph{decay} is the difference between the two, oriented so a larger value always means worse temporal robustness, since accuracy and $\overline{\text{AUC}}$ are better when high while $\text{MSE}_{\text{bal}}$ is better when low,
\begin{equation}
    \mathrm{Dec}(M) =
    \begin{cases}
        \mathrm{ID}(M) - \mathrm{Fut}(M) & (\text{accuracy},\ \overline{\text{AUC}}),\\
        \mathrm{Fut}(M) - \mathrm{ID}(M) & (\text{MSE}_{\text{bal}}).
    \end{cases}
    \label{eq:decay}
\end{equation}
so a positive decay is a loss of performance on future periods and a negative one, which is uncommon, a gain. Together, future score and decay order every model from most to least temporally robust.
\else
Three numbers summarize the matrix, each averaged only over filled cells (a train-test pair with no completed run leaves its cell empty): the \emph{in-distribution} score $\mathrm{ID}(M) = \operatorname{mean}_{i} M_{ii}$, a model's performance on the period it was trained through; the \emph{future} score $\mathrm{Fut}(M) = \operatorname{mean}_{j > i} M_{ij}$, averaging the cells whose evaluation period falls after the training cutoff; and the \emph{decay} $\mathrm{Dec}(M) = \mathrm{ID}(M) - \mathrm{Fut}(M)$ (the two terms swapped for $\text{MSE}_{\text{bal}}$, which is better when low), oriented so a larger value always means worse temporal robustness. A positive decay is a loss on future periods and a negative one, uncommon, a gain; together, future score and decay order every model from most to least temporally robust.
\fi

Beyond these whole-matrix scores, we look at how robustness depends on the training time itself. Fixing one cutoff $i$, a single row of the matrix, we pair its diagonal cell $M_{ii}$ with the average over the later periods in that row ($\operatorname{mean}_{j>i} M_{ij}$) and the decay between them, reading these at a few cutoffs spread evenly across the timeline (the last one is left out, as it has no future). Averaging the same quantities within each architecture family lets us compare the families directly.

\ifverbosemath
To place each model against the others, we average the matrices over the cohort $\mathcal{C}$ of models into the \emph{cohort-mean matrix}
\begin{equation}
    \bar{M}_{ij} = \frac{1}{|\mathcal{C}|}\sum_{m \in \mathcal{C}} M^{(m)}_{ij},
    \label{eq:cohort_mean}
\end{equation}
defined at the cells every model fills; each model's \emph{deviation} is $\Delta^{(m)}_{ij} = M^{(m)}_{ij} - \bar{M}_{ij}$.
\else
To place each model against the others, we average the matrices over the cohort $\mathcal{C}$ into the \emph{cohort-mean matrix} $\bar{M}_{ij} = |\mathcal{C}|^{-1}\sum_{m \in \mathcal{C}} M^{(m)}_{ij}$, defined at the cells every model fills; each model's \emph{deviation} is $\Delta^{(m)}_{ij} = M^{(m)}_{ij} - \bar{M}_{ij}$.
\fi

\section{Results}

Each model is summarised through its drift matrix by the in-distribution, forward, and decay scores of \cref{subsec:drift_summary}, namely its performance on the period it was trained through, its mean performance on the later periods held out from training, and the difference between the two. How far a model degrades between them is governed by two factors that recur across the three domains, the strength of its inductive bias and whether it relies on a frozen pretrained encoder.

\begin{figure*}[t]
\centering
\includegraphics[width=\textwidth]{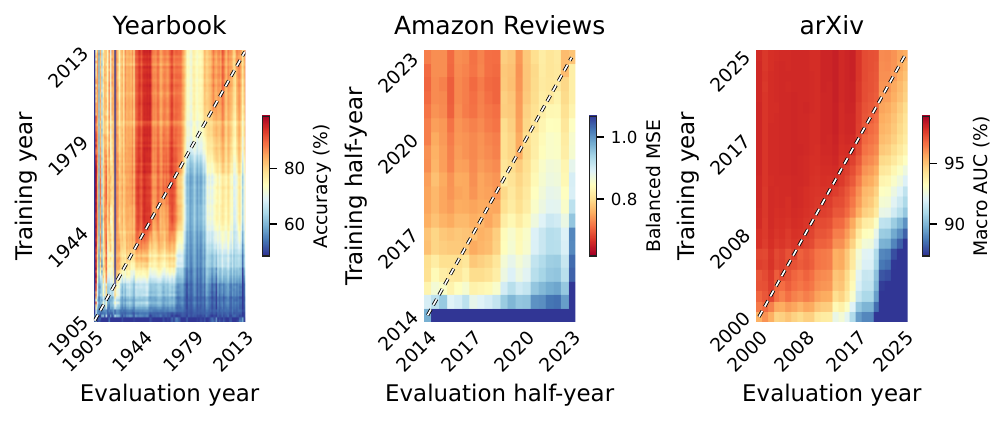}
\iflncs
\caption{Cohort-mean drift matrices, one panel per domain. Each cell $(i,j)$ is mean performance across models when trained through period $i$ (row) and evaluated on period $j$ (column): accuracy for \texttt{Yearbook}, balanced MSE for \texttt{Amazon Reviews} (lower is better), and macro AUC for \texttt{arXiv}. The dashed diagonal marks in-distribution evaluation; the lower-right region is forward generalization. Vertical white bands mark periods with insufficient samples.}
\else
\caption{Cohort-mean drift matrices, one panel per domain. Each cell $(i,j)$ is the mean performance across all models of that dataset when trained on data through period $i$ (row) and evaluated on period $j$ (column): accuracy for \texttt{Yearbook}, balanced MSE for \texttt{Amazon Reviews} (lower is better), and macro AUC for \texttt{arXiv}. The dashed diagonal marks in-distribution evaluation; the lower-right region of each panel is forward generalization to unseen future periods. Vertical white bands mark periods with insufficient samples.}
\fi
\label{fig:mean_matrices}
\end{figure*}

\subsection{Image Classification}

On \texttt{Yearbook}, the model families separate sharply in distribution, along the diagonal of the cohort-mean matrix in the \texttt{Yearbook} panel of \cref{fig:mean_matrices}. MLP-S, which flattens the image into a vector and imposes no spatial structure, sits at about $\val{yearbook/model/mlp_s/indist}\%$, while the CNNs and ResNets reach near $\val{yearbook/model/cnn_l/indist}\%$ for CNN-M and CNN-L (\cref{tab:robustness_yearbook}). What sets these apart is their inductive bias toward locality and translation equivariance, by which they build features from small local regions of the image and recognise a pattern wherever it appears, and this lets them exploit the cues that most sharply separate male from female portraits within a given period. The ViTs and frozen encoders fall between these two ends.

Temporal robustness reverses this ordering, and we read it from the decay, the accuracy a model loses once the evaluation year moves past its training cutoff (\cref{tab:robustness_yearbook}). The CNNs and ResNets, strongest in distribution, decay the most, by $\val{yearbook/model/cnn_l/decay}$ and $\val{yearbook/model/cnn_m/decay}$ points for CNN-L and CNN-M, so their accuracy on future years falls to about $\val{yearbook/model/cnn_l/future}\%$. MLP-S is the steadiest model of all, with only $\val{yearbook/model/mlp_s/decay}$ points of decay, though from its lower starting accuracy of $\val{yearbook/model/mlp_s/future}\%$. The frozen pretrained encoders form a third group, more stable than the CNNs but less accurate to begin with, decaying by $\val{yearbook/frozen/decay-min}$ to $\val{yearbook/frozen/decay-max}$ points, with the self-supervised DINOv3-S steadiest among them and the supervised ImageNet-21k backbone least.

% Single-column LNCS: a normal figure places inline (avoids the top-float
% whitespace a double-column figure* strands); the two-column preprint keeps
% figure* so the wide grid spans both columns.
\iflncs\begin{figure}[htbp]\else\begin{figure*}[htbp]\fi
\centering
\includegraphics[width=\linewidth]{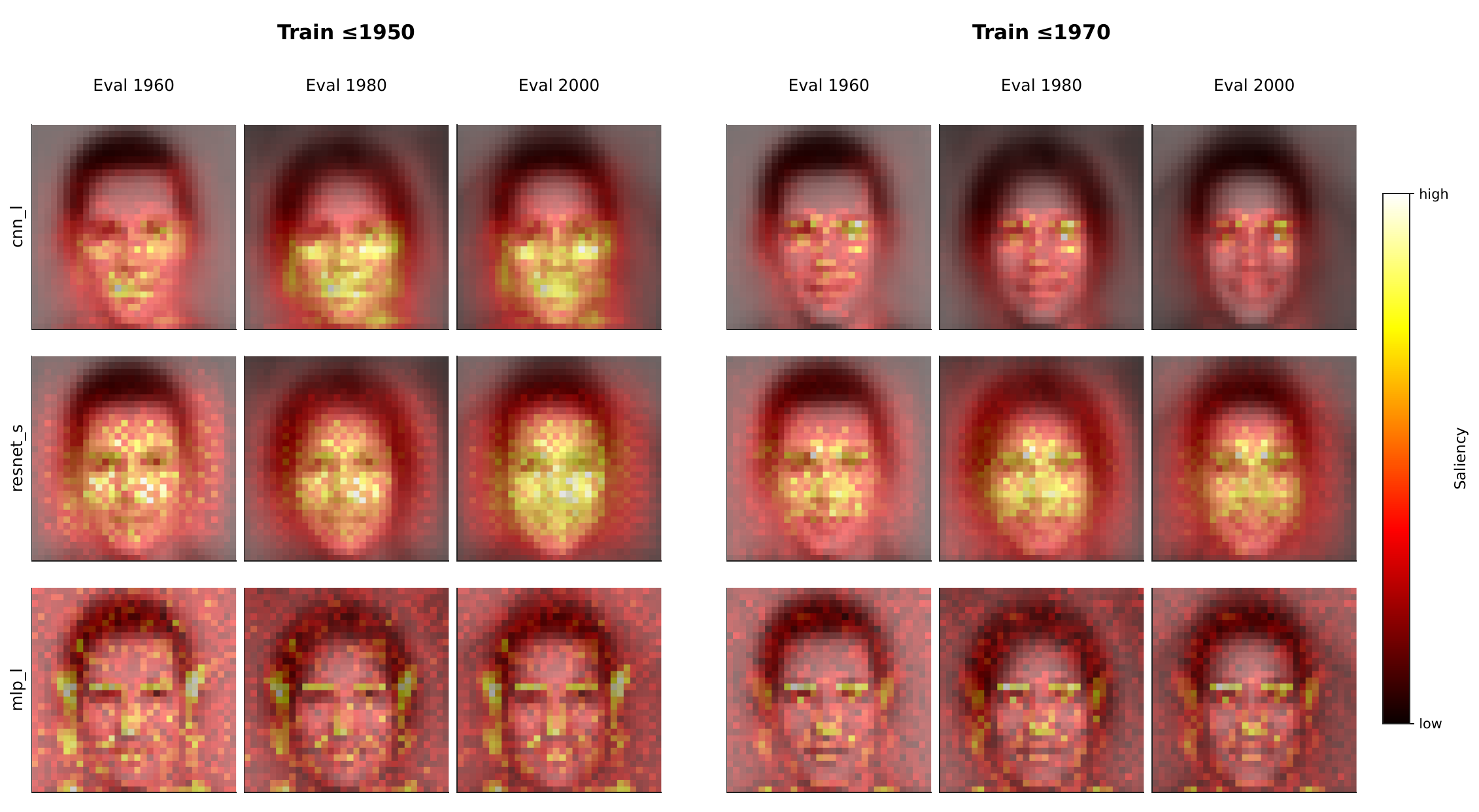}
\caption{Gradient saliency maps on \texttt{Yearbook} for CNN-L, ResNet-S, and MLP-L, each trained through 1950 (left) and 1970 (right) and evaluated on later years. Each panel averages gradient saliency over the selected portraits from the indicated evaluation year rather than a single example.}
\label{fig:saliency}
\iflncs\end{figure}\else\end{figure*}\fi

What lifts a model in distribution is also what undermines it over time. The features a strong inductive bias extracts to separate the classes \textbf{raise its in-distribution accuracy}, but they are \textbf{the most specific to the years it was trained on, and the first to lose their meaning} once hairstyles, clothing, and image quality drift. The sharper a model's in-distribution lead, the faster it erodes as the data ages. Robustness on \texttt{Yearbook} is therefore not a property an architecture optimises on its own, but the other side of fitting a single period too closely, closer to overfitting across time than across samples (\cref{app:yearbook}).

\forgettingfloat
\centering
\forgettinggraphic
\caption{Forgetting curves on \texttt{Yearbook}. Each curve is one model's mean accuracy as the gap between its training year and the evaluation year grows, averaged over training years and seeds. A zero gap is in-distribution, and the slope is the rate of forgetting (\cref{app:yearbook_forgetting}).}
\label{fig:forgetting_main}
\end{figure}

The forgetting curves in \cref{fig:forgetting_main} show this playing out year by year. Each curve plots a model's accuracy against the gap between its training cutoff and the evaluation year, so reading from left to right traces how quickly it forgets (\cref{app:yearbook_forgetting}). The strongly biased networks start far above the rest, near $90\%$ at a zero gap, and their curves descend the steepest. As the gap widens the curves draw together and then cross, so the families that led in distribution lose their edge on the most distant years, and every model settles near two-class chance.

\subsubsection{Saliency Maps}

The gradient saliency maps in \cref{fig:saliency} show where this fragility comes from. Extending the training window from 1950 to 1970 sharpens where the convolutional models look: the CNN tightens from a diffuse scatter over the cheeks and mouth to a compact, near-symmetric pair of bright spots at the eyes, the detail that most sharply tells the classes apart, and the ResNet shifts the same way while keeping a broader spread across the central face. The MLP, without that bias, spreads its attribution across the whole frame at either cutoff, the background included. The same precision that lets the convolutional models read this discriminative detail binds them to it: those localized features are the most specific to the training period and the most exposed to drift, while the MLP's blunter reading is less tied to any era.

\subsection{Text Models}

Each text model is a light head trained on cached, frozen \texttt{RoBERTa} embeddings (\cref{section_methods}), a shared representation over which the families differ only in inductive bias. \texttt{Amazon Reviews} is a rating regression scored by balanced MSE, where lower is better, and \texttt{arXiv} is a multi-label classification scored by macro AUC.

\subsubsection{Amazon Reviews}

On \texttt{Amazon Reviews} the recurrent and Transformer models fit the training reviews most closely, since both read the wording in context, the recurrent networks token by token and the Transformers through self-attention, and so capture how a review's words compose into its rating. That fit shows up as the lowest in-distribution error, down to $\val{amazon_reviews/model/bigru_s/indist}$ balanced MSE for BiGRU-S, with the Transformers alongside them at $\val{amazon_reviews/model/tx_s/indist}$ for TX-S (\cref{fig:mean_matrices}, \cref{tab:robustness_amazon_reviews}). The FFN, which averages the embeddings and discards word order, sits higher at about $\val{amazon_reviews/family/ffn/indist}$, and the frozen encoders higher still, between $\val{amazon_reviews/frozen/indist-min}$ and $\val{amazon_reviews/frozen/indist-max}$.

Temporal robustness runs the other way, and the models that fit the reviews most tightly are the ones whose error grows fastest. The recurrent networks decay the most, by up to $\val{amazon_reviews/model/bilstm_m/decay}$ balanced MSE, so their error on future reviews climbs to about $\val{amazon_reviews/model/bilstm_m/future}$, while the FFN is the steadiest of the trained models, gaining only $\val{amazon_reviews/model/ffn_s/decay}$ to $\val{amazon_reviews/model/ffn_l/decay}$. The decay is sharpest for models trained on the earliest reviews, where the recurrent family worsens by $\val{amazon_reviews/family/recurrent/28/decay}$, and it narrows toward the recent cutoffs as fewer years remain ahead (\cref{app:amazon_reviews}). The frozen encoders again sit at higher error but drift less than the trained models, and DeBERTa-v3 is the most stable model anywhere in the study, at $\val{amazon_reviews/model/deberta_v3_base_frozen/decay}$ of decay.

What makes these models fit the reviews so well is also what later undermines them. The way they compose the wording into a rating is the most specific to how reviews were written in the training years, and the first to lose its meaning as the language shifts. \textbf{Far enough forward the in-distribution ranking inverts, and the models that fit the reviews most tightly end among the least accurate}, while the order-free FFN, never sharp, stays the steadiest.

\subsubsection{arXiv}

On \texttt{arXiv} a stronger inductive bias yields no in-distribution advantage, and so costs no temporal robustness. Sorting a paper into its subject categories is close to recognising its topic, and the topic is already encoded in the frozen \texttt{RoBERTa} embeddings every model is built on, so no architecture finds structure the others miss. The feed-forward, convolutional, recurrent, and Transformer families therefore reach almost the same in-distribution macro AUC, between about $\val{arxiv/scratch/indist-min}\%$ and $\val{arxiv/scratch/indist-max}\%$, within a point of one another (\cref{fig:mean_matrices}, \cref{tab:robustness_arxiv}).

Robustness is just as uniform, and each family loses a similar small $\val{arxiv/scratch/decay-min}$ to $\val{arxiv/scratch/decay-max}$ points going forward, with the frozen encoders in the same band apart from DeBERTa-v3 and ELECTRA, which fall further at $\val{arxiv/model/deberta_v3_base_frozen/decay}$ and $\val{arxiv/model/electra_base_frozen/decay}$ points (\cref{app:arxiv}). \textbf{Where the bias gains nothing in distribution it forms no period-specific features to lose, so no family decays faster than the rest}. The later years stay close to the training distribution: the backbone learned this vocabulary before any model saw it, leaving temporal shift little to take away.

\tightvspace{-5pt}
\section{Conclusion and Future Work}

Across image classification, text regression, and multi-label text classification, in-distribution accuracy turns out to neither guarantee nor predict temporal robustness. What governs the rate of degradation is instead the strength of a model's inductive bias and whether it relies on a frozen pretrained encoder. A stronger bias extracts the most discriminative, period-specific features and leads in distribution, yet those same features are the first to lose their meaning as the data drifts, so the architectures that fit the training period most tightly are the ones that degrade fastest. Frozen pretrained encoders occupy a different regime, conceding in-distribution accuracy in return for steadier behaviour over time, and on \texttt{arXiv}, where the task is already solved by the pretrained representation and the bias buys no advantage, no family decays faster than the rest. The practical reading is that the model with the best held-out score at training time is often the least robust once deployed, so architecture selection should weigh the expected horizon between retraining cycles, not in-distribution accuracy alone.

\tightvspace{-5pt}
\subsection{Limitations}

\tightvspace{-5pt}
Our comparison covers the most common inductive biases, convolutional, recurrent, and attention-based, against simple baselines and frozen encoders, but many architectures remain untested and the study is best read as a first systematic pass rather than an exhaustive one. It spans three datasets across two modalities, and broadening it to further domains and other forms of distribution shift would show how widely the pattern holds. The text tasks are the most constrained, using a narrow label space, a five-point rating for \texttt{Amazon Reviews} and seven categories for \texttt{arXiv}, spanning only about a decade and a quarter of a century against a full century for \texttt{Yearbook}, and running on a stratified subsample rather than the complete corpora. Because too few examples fall within each time slice to train a text encoder from scratch, the text models also share a single frozen \texttt{RoBERTa} representation, so their absolute scores should be read in that light. We average over a small fixed seed set rather than conducting formal significance tests, and hyperparameter choices and seed variance may still affect fine-grained rankings; we therefore emphasize family-level patterns over individual placements. The families are also not capacity-matched. The small, medium, and large variants within each family expose scale effects, and on \texttt{Yearbook} the smallest \texttt{MLP}, \texttt{CNN}, and \texttt{ResNet} variants decay less than their larger siblings, so inductive bias and capacity remain partially confounded. The analysis is also descriptive rather than mechanistic. We measure how robustness varies with inductive bias and pretraining without isolating the precise cues responsible, and we evaluate inherent robustness under cumulative training rather than any adaptive policy, which leaves the question of retraining orchestration to systems such as Modyn \cite{boether_modyn_2025}.

\tightvspace{-5pt}
\subsection{Future Work}

\tightvspace{-5pt}
These limitations point to several extensions. The comparison can be widened to a broader range of inductive biases, to capacity-matched pairs that disentangle bias from scale, and to additional datasets, modalities, and drift mechanisms, and carried to longer temporal spans with the full corpora and richer label spaces, where language drift should be more pronounced and the gap between architectures wider. The drift matrices can also be paired with scheduled or drift-triggered retraining under explicit compute budgets, measuring not only how much a model decays but how cheaply that decay can be undone \cite{mahadevan_cost_aware_2024}. Most of all, moving from description to mechanism, by tying inductive bias and pretraining back to the specific features that drift, would turn the observed regularity into an account of why temporal robustness behaves as it does.

\tightvspace{-5pt}
\subsection{Availability}

\tightvspace{-5pt}
Code, experiment definitions, and paper-generation scripts are available at \href{https://github.com/learning-mechanisms/drift-happens/}{github.com/learning-mechanisms/drift-happens}; public run histories and matrix artifacts at \href{https://wandb.ai/drift-happens/drift-happens}{wandb.ai/drift-happens/drift-happens}. The \extendedpreprintlink{} and \resultssitelink{} retain the complete per-model drift-matrix galleries. Code is Apache-2.0; paper, figures, and documentation are CC BY 4.0 where we hold the rights, and external datasets and models keep their original licenses.

\bibliographystyle{plainnat}
\bibliography{bibliography}

\section*{Appendix}
\appendix

\section{Reproducibility}
\label{app:reproducibility}

Experiments are defined as versioned preset snapshots and run through the
repository command-line interface in a Pixi-pinned environment; run metadata
records the git state and \texttt{pixi.lock} hash, and regenerated paper and
website assets are checked against a committed checksum manifest. Each experiment
is a staged run for one dataset, architecture, and seed: the training stage fits
all cumulative temporal checkpoints, and the evaluation stage scores them on
every time slice to form the drift matrix. We use seeds $0$--$4$ for
\texttt{Yearbook} and $0$--$2$ for \texttt{Amazon Reviews} and \texttt{arXiv},
averaging over them. Public run histories and per-run matrix artifacts are
available in the Weights \& Biases project at
\url{https://wandb.ai/drift-happens/drift-happens}.

Table~\ref{tab:compute_estimate} summarizes the measured compute on NVIDIA A100
GPUs, summing the W\&B \texttt{\_runtime} of each finished train and evaluation
stage. GPU-hours are successful-stage wall times under one GPU per run; the
four-GPU node-hour column divides these by four for ideal packed execution,
excluding queueing, data preprocessing, failed attempts, and scheduling idle
time.

\begin{table}[htbp]
\caption{Approximate measured compute for the conference experiment campaign.}
\label{tab:compute_estimate}
\centering
\newcommand{\computeEstimateTabular}{%
\begin{tabular}{@{}lrrrr@{}}
\toprule
Dataset & Model-seed configs & Train GPU-h & Eval GPU-h & 4-GPU node-h \\
\midrule
\texttt{Yearbook} & 105 & 23.7 & 9.3 & 8.2 \\
\texttt{Amazon Reviews} & 57 & 116.6 & 136.7 & 63.3 \\
\texttt{arXiv} & 60 & 272.3 & 101.9 & 93.5 \\
\midrule
Total & 222 & 412.6 & 247.9 & 165.1 \\
\bottomrule
\end{tabular}
}
\iflncs
    \footnotesize
    \setlength{\tabcolsep}{3pt}
    \computeEstimateTabular
\else
    \resizebox{\columnwidth}{!}{%
        \footnotesize
        \setlength{\tabcolsep}{3pt}
        \computeEstimateTabular
    }
\fi
\end{table}

\section{Per-Dataset Protocol and Metrics}
\label{appendix:evaluation_metrics}
The drift-matrix protocol of \cref{section:evaluation_framework} is shared across the three datasets, but each instantiates it at its own time granularity and is scored with the metric matched to its label structure (\cref{tab:protocol_instantiation}). For each task we require a scalar performance measure that is both aligned with the task and robust to the class imbalance present in that dataset; a metric dominated by the label distribution would conflate shifts in the data with changes in the model.

\begin{table}[htbp]
\caption{Per-dataset instantiation of the drift-matrix protocol: time span, slice granularity, number of slices $K$, and primary metric.}
\label{tab:protocol_instantiation}
\centering
% Shrink to the column width only if the table is wider than it (as in the
% two-column preprint); never enlarge a narrow table, which plain
% \resizebox{\columnwidth}{!} would do in the roomy single-column LNCS build.
\resizebox{\ifdim\width>\columnwidth \columnwidth\else\width\fi}{!}{%
\begin{tabular}{@{}lllcl@{}}
\toprule
Dataset & Span & Slice & $K$ & Metric \\
\midrule
\texttt{Yearbook} & 1905--2013 & yearly & 104 & accuracy \\
\texttt{Amazon Reviews} & 2014--2023 & half-yearly & 20 & balanced MSE \\
\texttt{arXiv} & 2000--2025 & yearly & 26 & macro AUC \\
\bottomrule
\end{tabular}%
}
\end{table}

\subsection{Yearbook}
\label{app:metrics_yearbook}
We slice the 1905--2013 timeline into one-year intervals, keeping the $K = 104$ years with enough samples. The task is binary classification with roughly balanced classes, so we score it with standard accuracy, the fraction of portraits classified correctly. Balance is what makes accuracy trustworthy here: when neither class dominates, a model cannot earn a good score by always predicting the same label, so accuracy rises and falls only as the model genuinely classifies more or fewer cases correctly. Higher values are better.

\subsection{Amazon Reviews}
\label{app:metrics_amazon}
We slice the 2014--2023 timeline into half-year intervals, giving $K = 20$ slices. The task is regression: the model predicts a star rating from one to five and is scored by squared error. The difficulty is that the ratings are strongly skewed toward five-star reviews, so a single mean squared error taken over all reviews would mainly measure performance on the majority and would barely react to the rarer low ratings. To keep every rating level visible, we first compute the mean squared error separately within each rating and then average those per-rating values equally over the rating levels present,
\begin{equation}
    \nonumber
    \text{MSE}_{\text{bal}} = \frac{1}{|\mathcal{R}|}\sum_{r \in \mathcal{R}} \text{MSE}_r, \qquad
    \text{MSE}_r = \frac{1}{|\mathcal{S}_r|}\sum_{i \in \mathcal{S}_r}(y_i - \hat{y}_i)^2,
\end{equation}
where $\mathcal{S}_r = \{i : y_i = r\}$ is the set of reviews whose true rating is $r$ and $\mathcal{R} = \{r : |\mathcal{S}_r| > 0\}$ is the set of rating levels present in the slice (at most the five levels one to five). Because every present level contributes equally, a model that began to drift on the scarce one- and two-star reviews would reveal it here. Unlike the other two metrics, lower values are better.

\subsection{arXiv}
\label{app:metrics_arxiv}
We slice the 2000--2025 timeline into one-year intervals, giving $K = 26$ slices. The task is multi-label: a paper can belong to several of the $C = 7$ subject areas at once, and those areas are very unevenly populated. This raises two problems. First, fixing a single decision threshold is arbitrary and tends to favour the frequent subjects, so a threshold-based score such as accuracy is misleading. Second, an average that weights papers equally is again dominated by the common subjects. We address the first by scoring each subject with the Area Under the ROC Curve, which summarises performance over all thresholds at once: $\text{AUC}_c$ is the probability that a randomly chosen paper carrying subject $c$ is given a higher score for that subject than a randomly chosen paper that does not carry it. We address the second by averaging the per-subject values uniformly, so each subject counts the same regardless of how many papers it contains,
\begin{equation}
    \nonumber
    \overline{\text{AUC}} = \frac{1}{C}\sum_{c=1}^{C} \text{AUC}_c, \qquad
    \text{AUC}_c = \int_0^1 \text{TPR}_c(t)\, d\text{FPR}_c(t),
\end{equation}
where $\text{TPR}_c(t)$ and $\text{FPR}_c(t)$ denote the true- and false-positive rates for subject $c$ at threshold $t$. Higher values are better, and a model that ranked every paper correctly within every subject would reach $1$.

\tightvspace{\fill}

\pagebreak
\onecolumn

\captionsetup[figure]{font=footnotesize,labelfont=footnotesize}

\section{Yearbook -- Drift Matrices}
\label{app:yearbook}

\noindent The cohort-mean and per-model deviation matrices shown here, and the in-distribution, future, and decay quantities tabulated below, are defined in Section~\ref{subsec:drift_summary}.

\begin{figure}[H]
\centering
\includegraphics[width=0.5\textwidth]{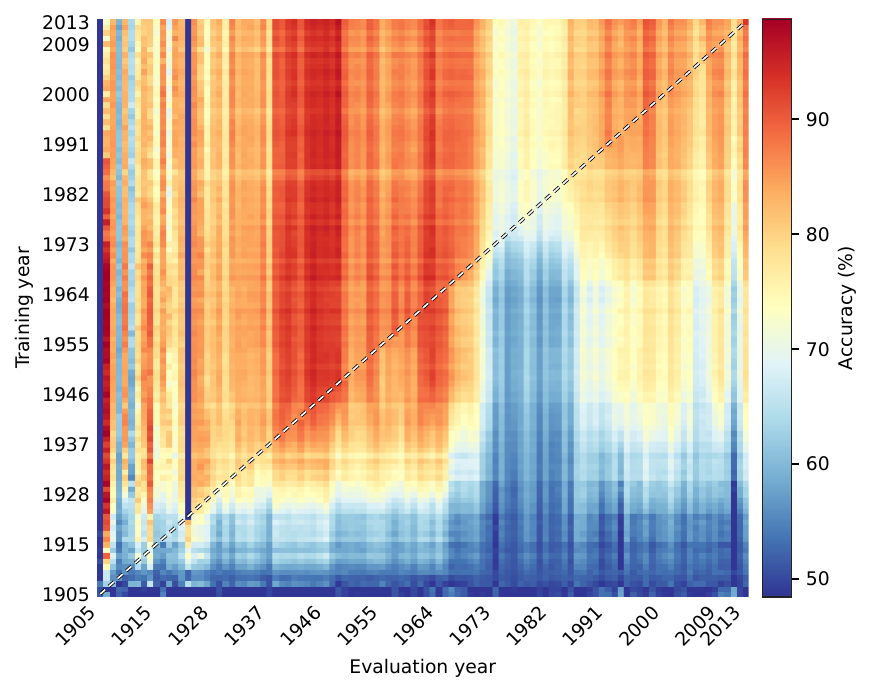}
\caption{Cohort-mean Accuracy matrix $\bar{M}$ over the Yearbook models. Cell $(i,j)$ is the mean across those models of the score from training through slice $i$ and evaluating on slice $j$.}
\label{fig:yearbook_mean_matrix}
\end{figure}

\subsection{Model Roster}

\noindent
\begin{minipage}[t]{0.49\linewidth}
\centering
\footnotesize
\captionof{table}{Yearbook: models trained from scratch.}
\label{tab:yearbook_roster}
\resizebox{\ifdim\width>\linewidth \linewidth\else\width\fi}{!}{%
\begin{tabular}{l l r}
\toprule
Model & Family & Parameters \\
\midrule
ViT-S & ViT & 75k \\
CNN-S & CNN & 94k \\
ResNet-S & ResNet & 98k \\
MLP-S & MLP & 98k \\
ResNet-M & ResNet & 400k \\
MLP-M & MLP & 410k \\
CNN-M & CNN & 542k \\
ViT-M & ViT & 545k \\
MLP-L & MLP & 2.1M \\
ResNet-L & ResNet & 2.1M \\
CNN-L & CNN & 2.1M \\
ViT-L & ViT & 2.2M \\
\bottomrule
\end{tabular}}
\end{minipage}
\hfill
\begin{minipage}[t]{0.49\linewidth}
\centering
\footnotesize
\captionof{table}{Yearbook: frozen pretrained encoders, with trainable head and total parameters.}
\label{tab:yearbook_roster_frozen}
\resizebox{\ifdim\width>\linewidth \linewidth\else\width\fi}{!}{%
\begin{tabular}{l l rr}
\toprule
Model & Family & Trainable & Total \\
\midrule
DINOv3-S & Transfer & 770 & 21.6M \\
ViT-S16-IN21k & Transfer & 770 & 21.7M \\
DINOv2-S & Transfer & 770 & 22.1M \\
ResNet50-IN & Transfer & 4k & 23.5M \\
ConvNeXt-S & Transfer & 2k & 49.5M \\
EVA02-B & Transfer & 2k & 85.8M \\
MAE-B & Transfer & 2k & 85.8M \\
CLIP-B32 & Transfer & 2k & 87.5M \\
SigLIP-B & Transfer & 2k & 92.9M \\
\bottomrule
\end{tabular}}
\end{minipage}

\subsection{MLP}

\begin{figure}[H]
\centering
\begin{subfigure}[t]{0.49\textwidth}\centering
    \includegraphics[width=0.49\linewidth]{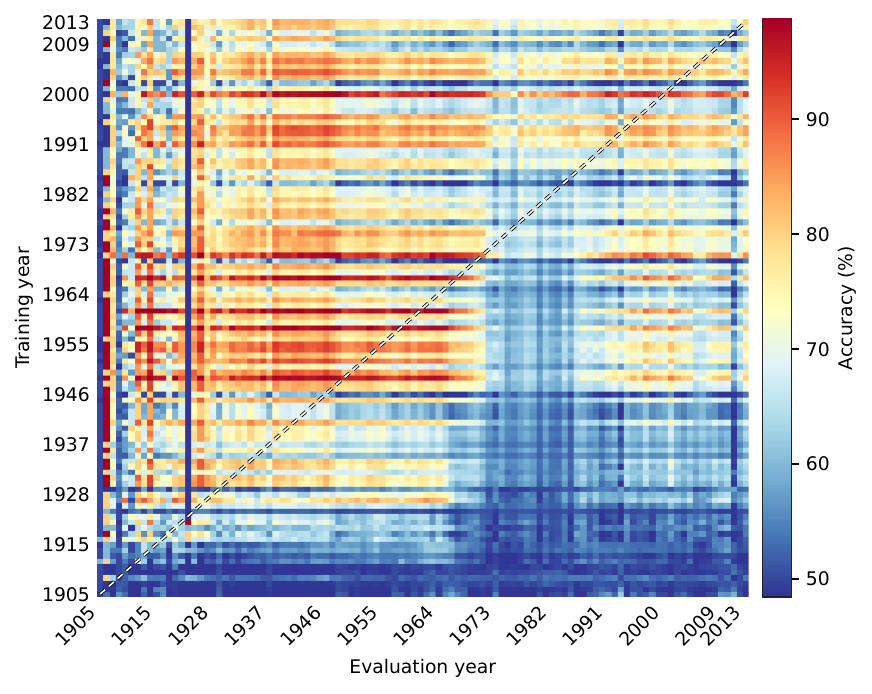}\hfill
    \includegraphics[width=0.49\linewidth]{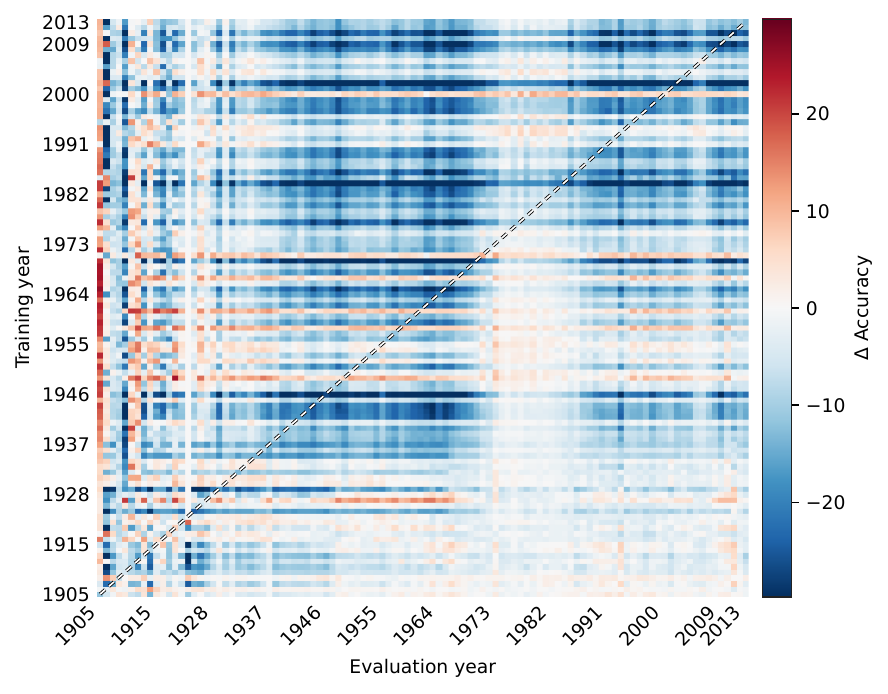}
    \caption{MLP-S}
    \label{fig:yearbook_MLP_S}
\end{subfigure}
\hfill
\begin{subfigure}[t]{0.49\textwidth}\centering
    \includegraphics[width=0.49\linewidth]{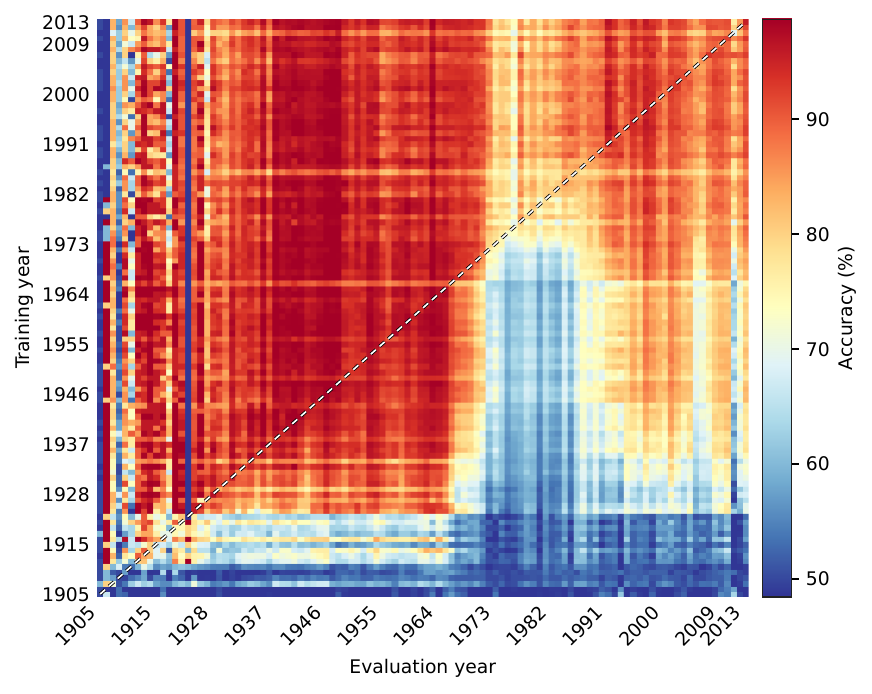}\hfill
    \includegraphics[width=0.49\linewidth]{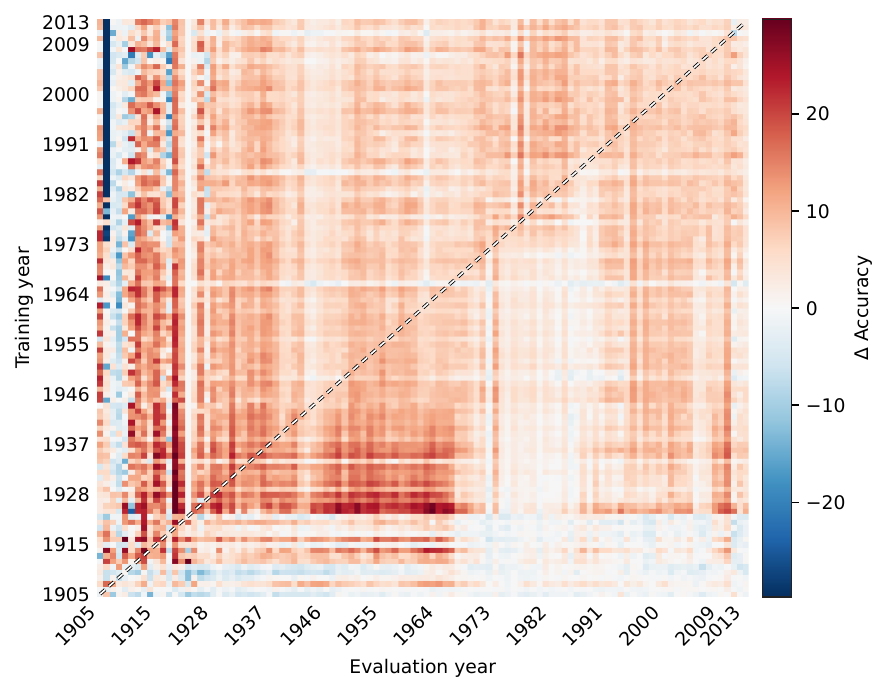}
    \caption{MLP-M}
    \label{fig:yearbook_MLP_M}
\end{subfigure}

\begin{subfigure}[t]{0.49\textwidth}\centering
    \includegraphics[width=0.49\linewidth]{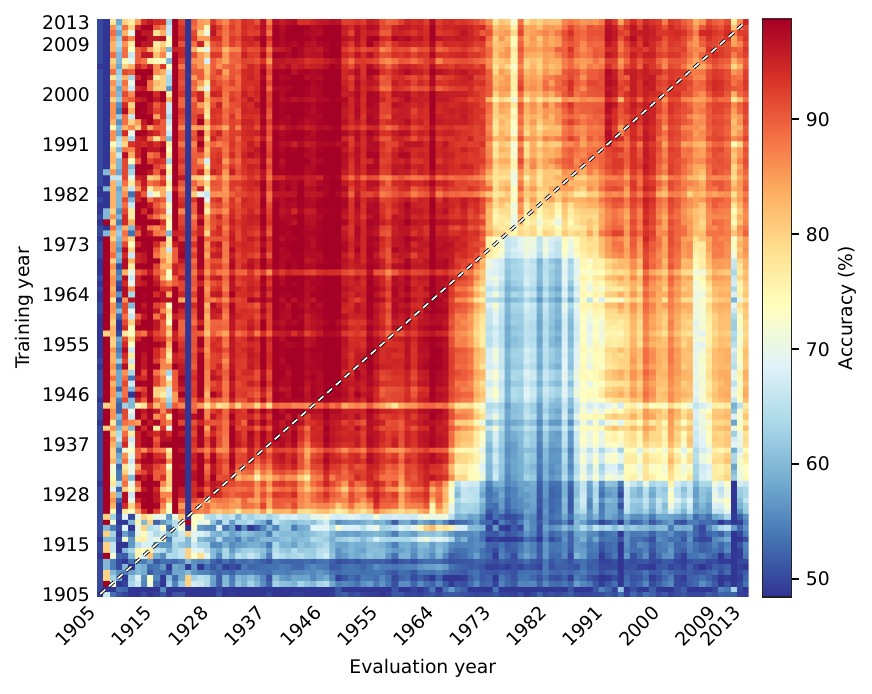}\hfill
    \includegraphics[width=0.49\linewidth]{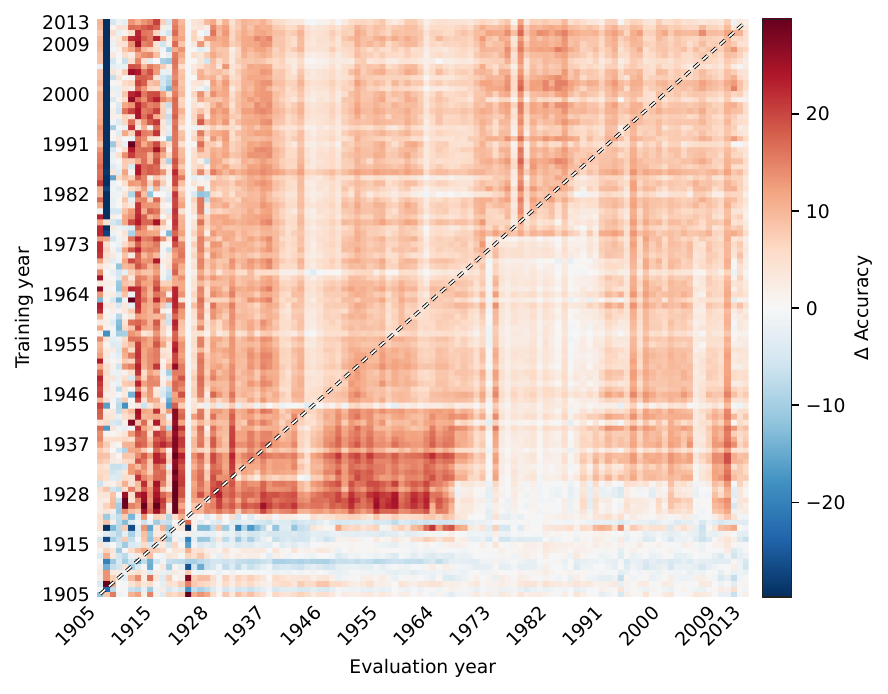}
    \caption{MLP-L}
    \label{fig:yearbook_MLP_L}
\end{subfigure}
\caption{MLP models: Accuracy drift matrix $M^{(m)}$ and deviation from the cohort mean $\Delta^{(m)} = M^{(m)} - \bar{M}$ for each model, shown on a sequential and a zero-centred diverging scale, respectively.}
\label{fig:yearbook_family_image_mlp}
\end{figure}

\subsection{CNN}

\begin{figure}[H]
\centering
\begin{subfigure}[t]{0.49\textwidth}\centering
    \includegraphics[width=0.49\linewidth]{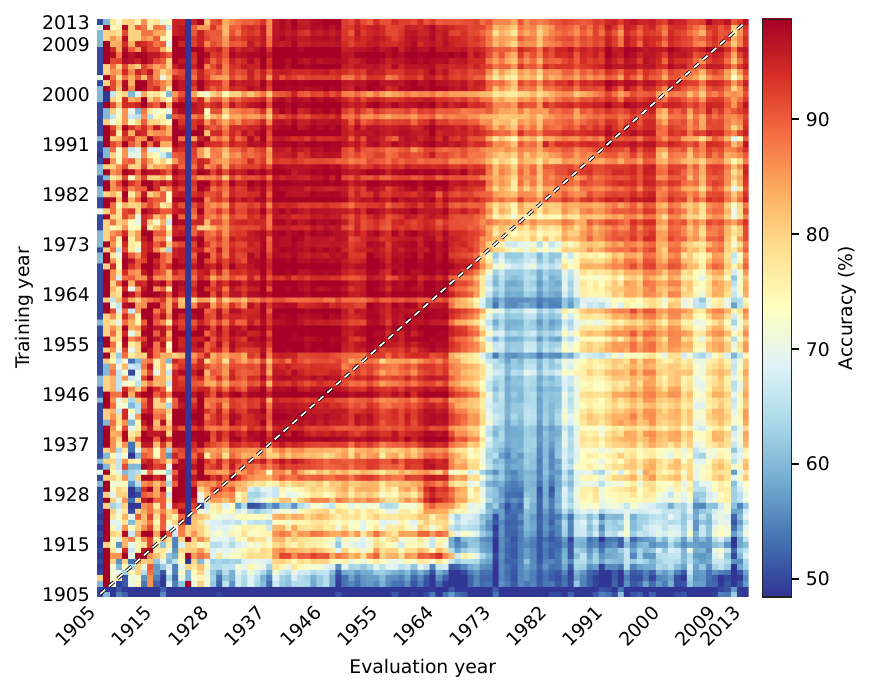}\hfill
    \includegraphics[width=0.49\linewidth]{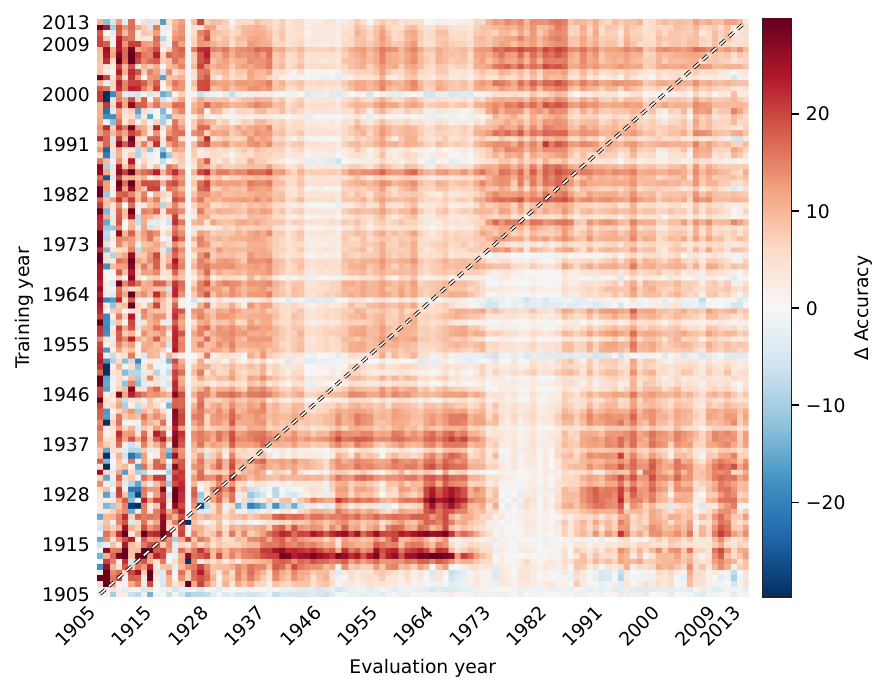}
    \caption{CNN-S}
    \label{fig:yearbook_CNN_S}
\end{subfigure}
\hfill
\begin{subfigure}[t]{0.49\textwidth}\centering
    \includegraphics[width=0.49\linewidth]{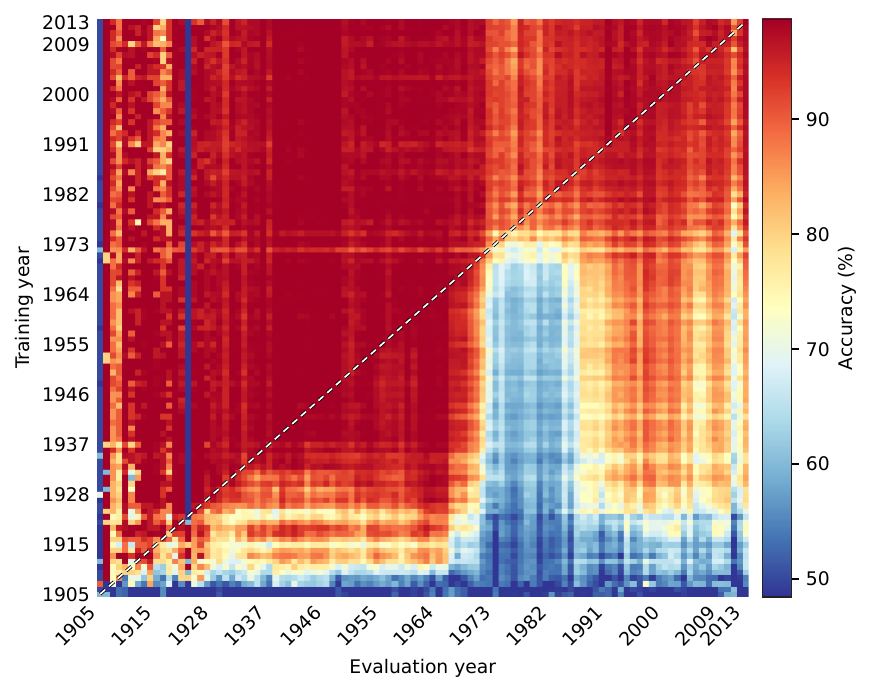}\hfill
    \includegraphics[width=0.49\linewidth]{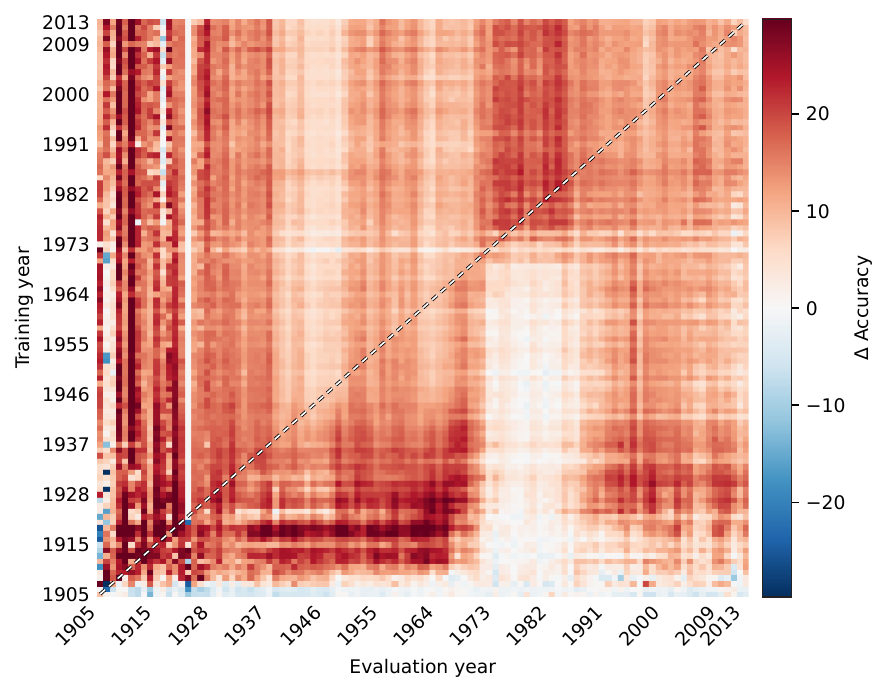}
    \caption{CNN-M}
    \label{fig:yearbook_CNN_M}
\end{subfigure}

\begin{subfigure}[t]{0.49\textwidth}\centering
    \includegraphics[width=0.49\linewidth]{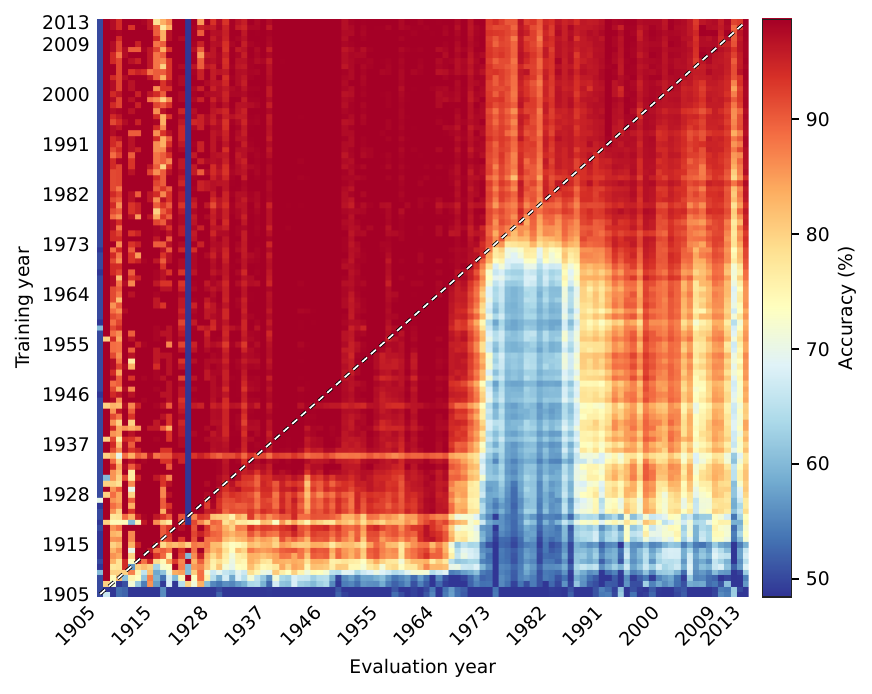}\hfill
    \includegraphics[width=0.49\linewidth]{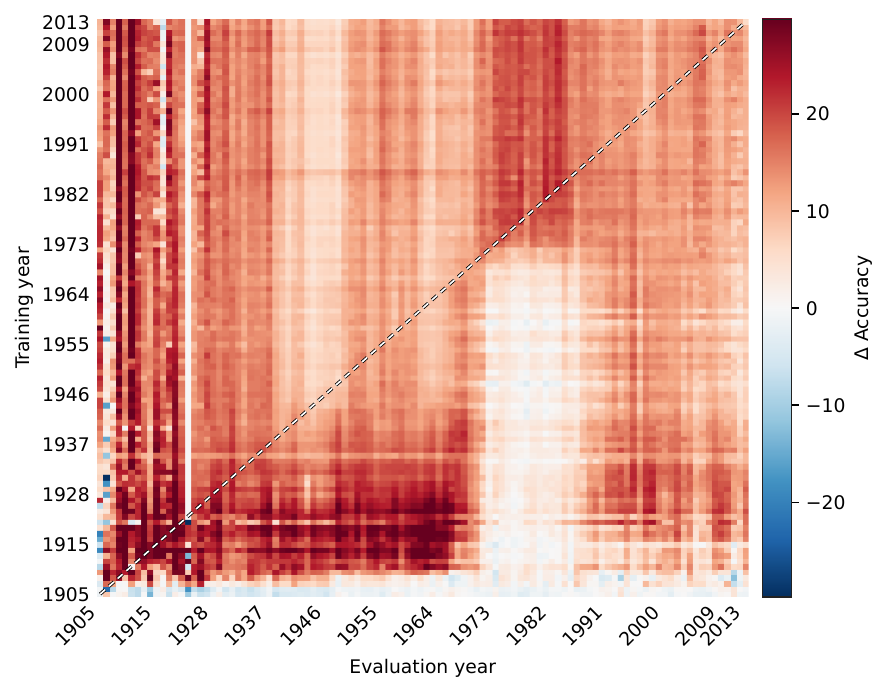}
    \caption{CNN-L}
    \label{fig:yearbook_CNN_L}
\end{subfigure}
\caption{CNN models: Accuracy drift matrix $M^{(m)}$ and deviation from the cohort mean $\Delta^{(m)} = M^{(m)} - \bar{M}$ for each model, shown on a sequential and a zero-centred diverging scale, respectively.}
\label{fig:yearbook_family_image_cnn}
\end{figure}

\subsection{ResNet}

\begin{figure}[H]
\centering
\begin{subfigure}[t]{0.49\textwidth}\centering
    \includegraphics[width=0.49\linewidth]{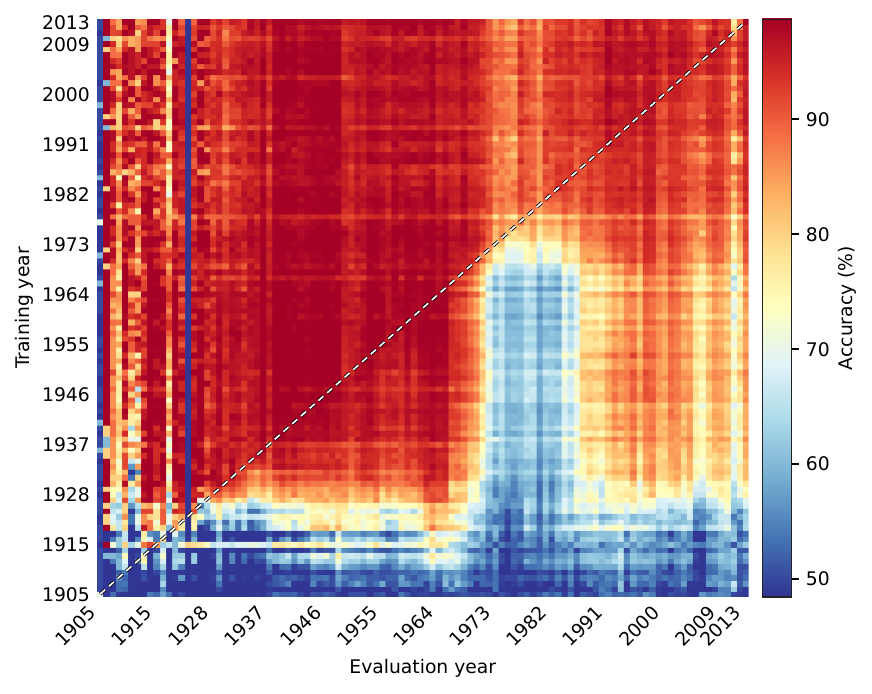}\hfill
    \includegraphics[width=0.49\linewidth]{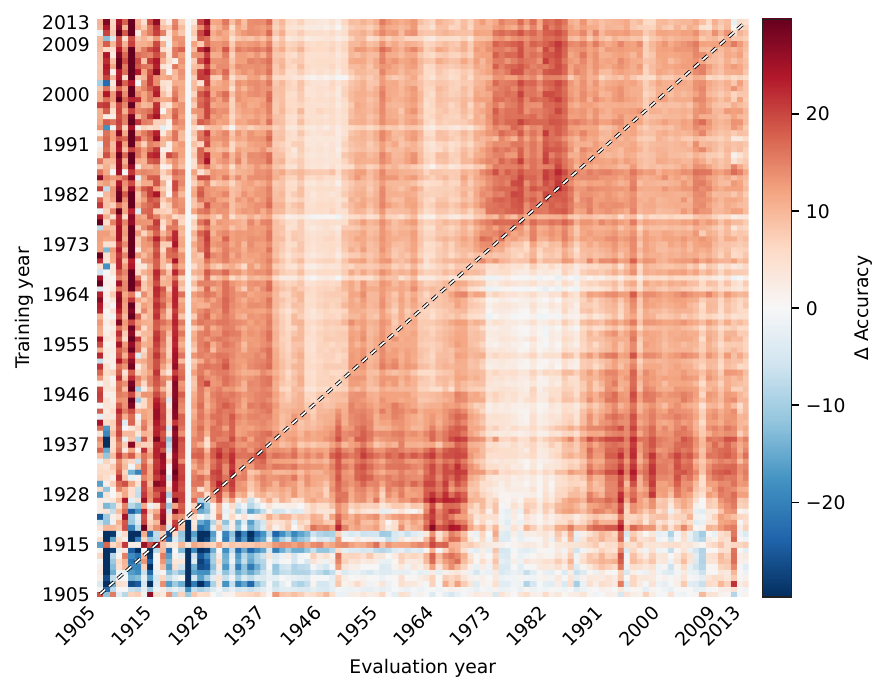}
    \caption{ResNet-S}
    \label{fig:yearbook_ResNet_S}
\end{subfigure}
\hfill
\begin{subfigure}[t]{0.49\textwidth}\centering
    \includegraphics[width=0.49\linewidth]{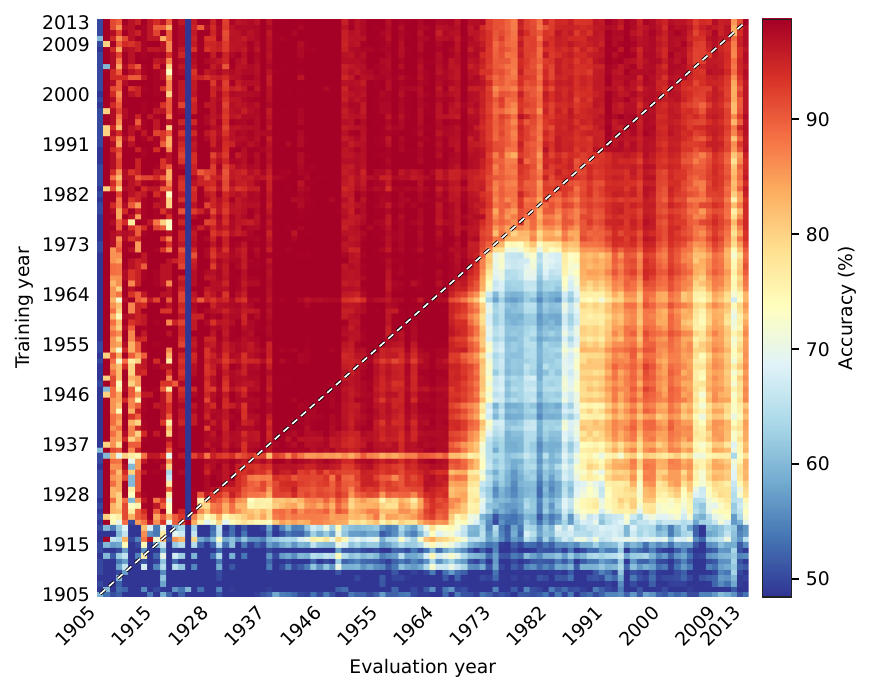}\hfill
    \includegraphics[width=0.49\linewidth]{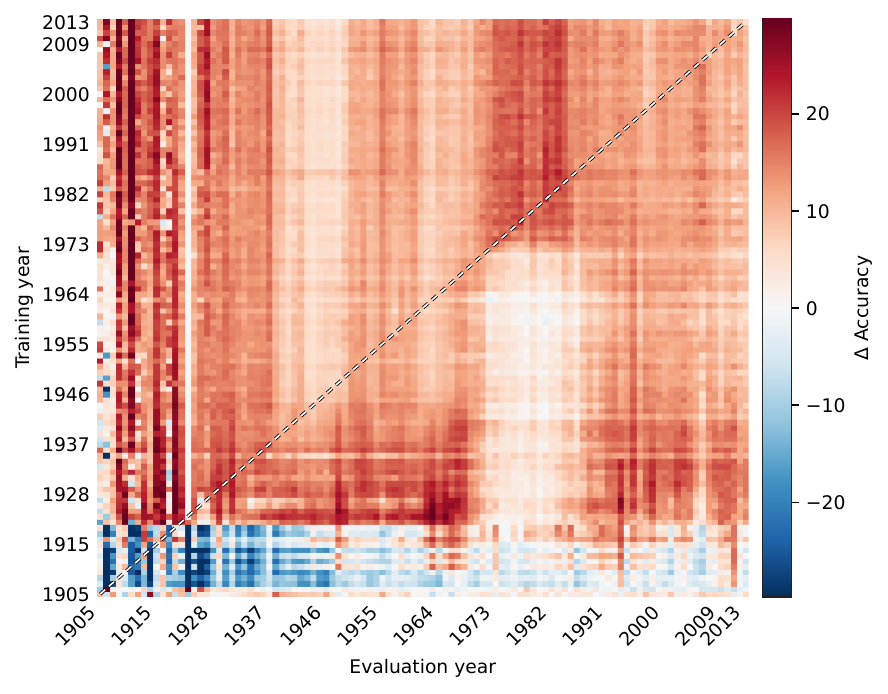}
    \caption{ResNet-M}
    \label{fig:yearbook_ResNet_M}
\end{subfigure}

\begin{subfigure}[t]{0.49\textwidth}\centering
    \includegraphics[width=0.49\linewidth]{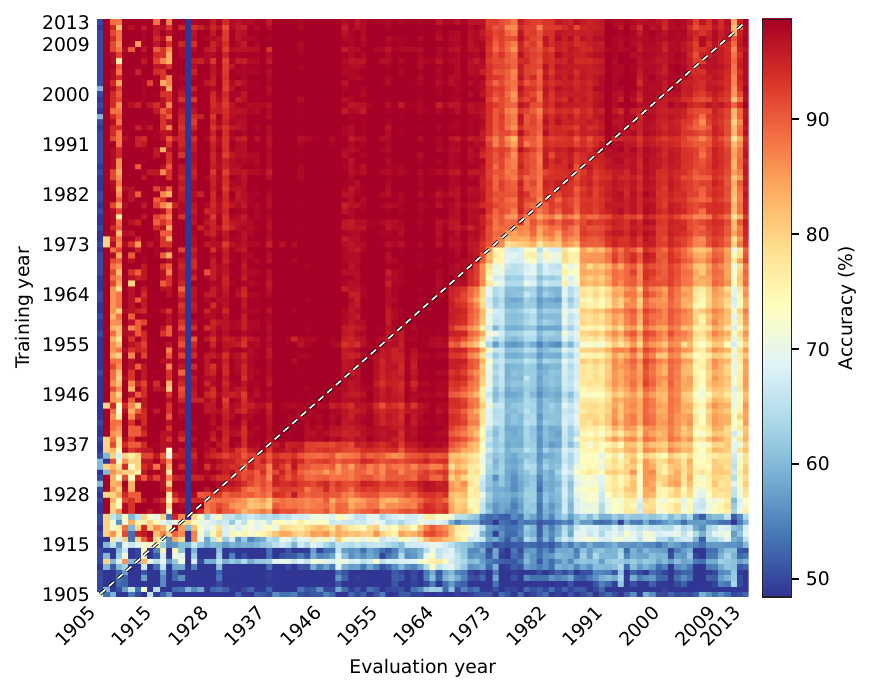}\hfill
    \includegraphics[width=0.49\linewidth]{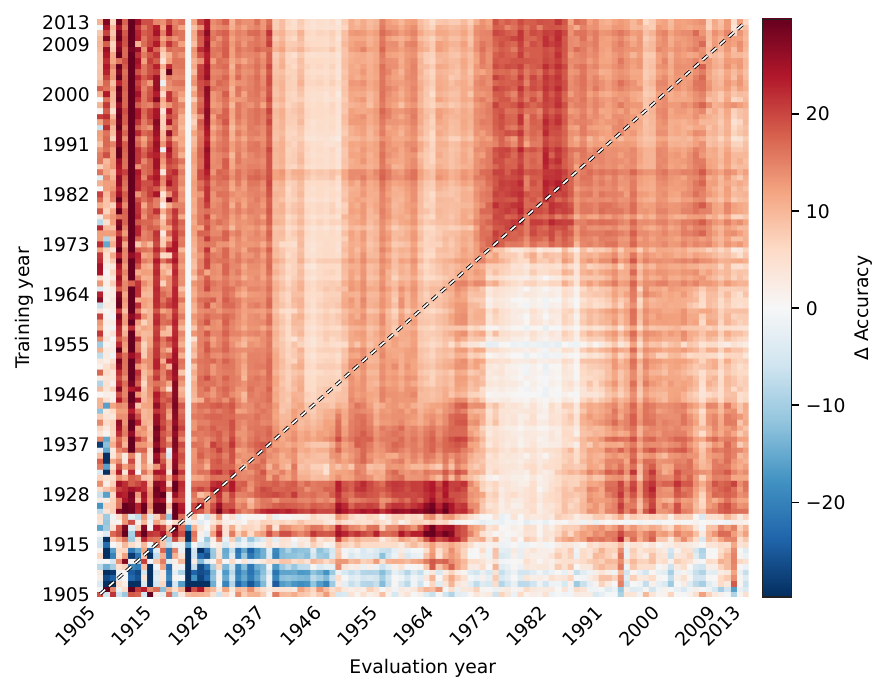}
    \caption{ResNet-L}
    \label{fig:yearbook_ResNet_L}
\end{subfigure}
\caption{ResNet models: Accuracy drift matrix $M^{(m)}$ and deviation from the cohort mean $\Delta^{(m)} = M^{(m)} - \bar{M}$ for each model, shown on a sequential and a zero-centred diverging scale, respectively.}
\label{fig:yearbook_family_image_resnet}
\end{figure}

\subsection{ViT}

\begin{figure}[H]
\centering
\begin{subfigure}[t]{0.49\textwidth}\centering
    \includegraphics[width=0.49\linewidth]{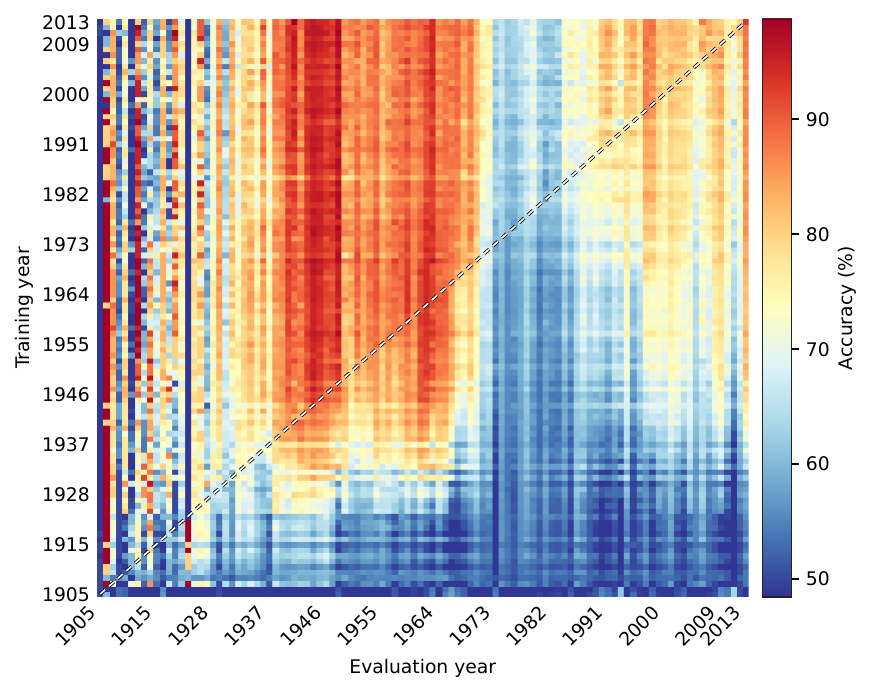}\hfill
    \includegraphics[width=0.49\linewidth]{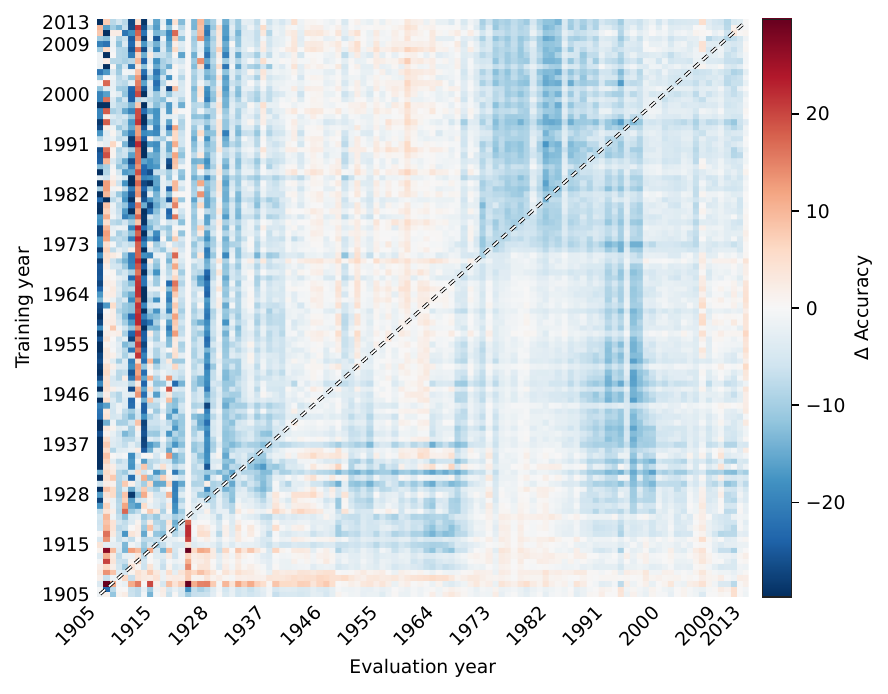}
    \caption{ViT-S}
    \label{fig:yearbook_ViT_S}
\end{subfigure}
\hfill
\begin{subfigure}[t]{0.49\textwidth}\centering
    \includegraphics[width=0.49\linewidth]{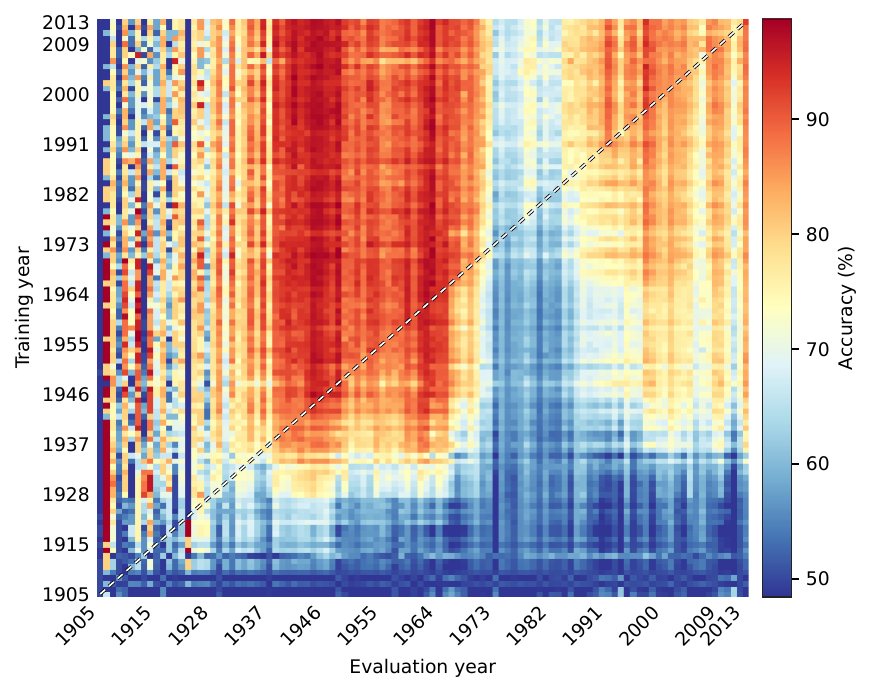}\hfill
    \includegraphics[width=0.49\linewidth]{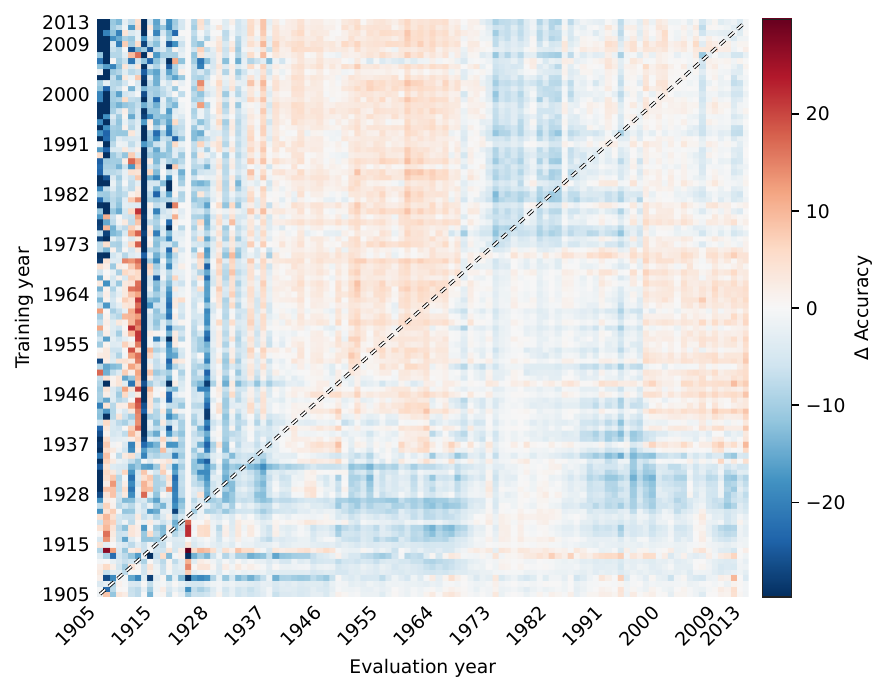}
    \caption{ViT-M}
    \label{fig:yearbook_ViT_M}
\end{subfigure}

\begin{subfigure}[t]{0.49\textwidth}\centering
    \includegraphics[width=0.49\linewidth]{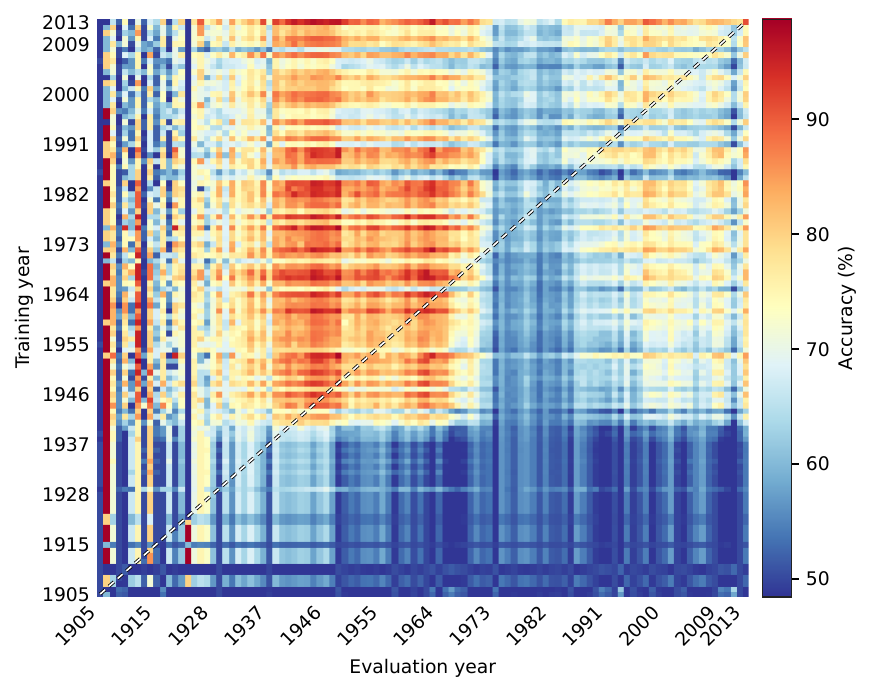}\hfill
    \includegraphics[width=0.49\linewidth]{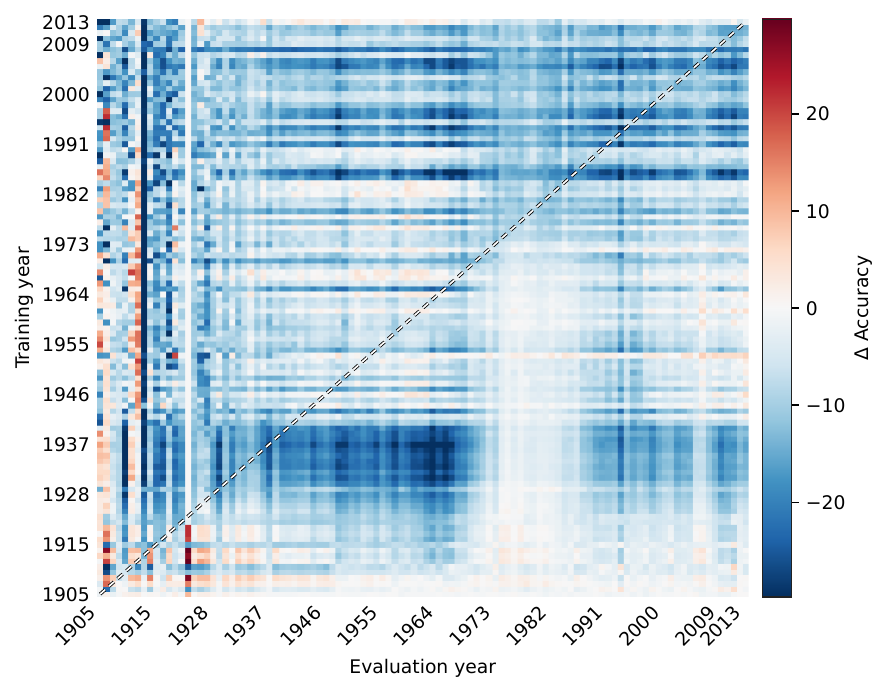}
    \caption{ViT-L}
    \label{fig:yearbook_ViT_L}
\end{subfigure}
\caption{ViT models: Accuracy drift matrix $M^{(m)}$ and deviation from the cohort mean $\Delta^{(m)} = M^{(m)} - \bar{M}$ for each model, shown on a sequential and a zero-centred diverging scale, respectively.}
\label{fig:yearbook_family_image_vit}
\end{figure}

\subsection{Transfer}

\begin{figure}[H]
\centering
\begin{subfigure}[t]{0.49\textwidth}\centering
    \includegraphics[width=0.49\linewidth]{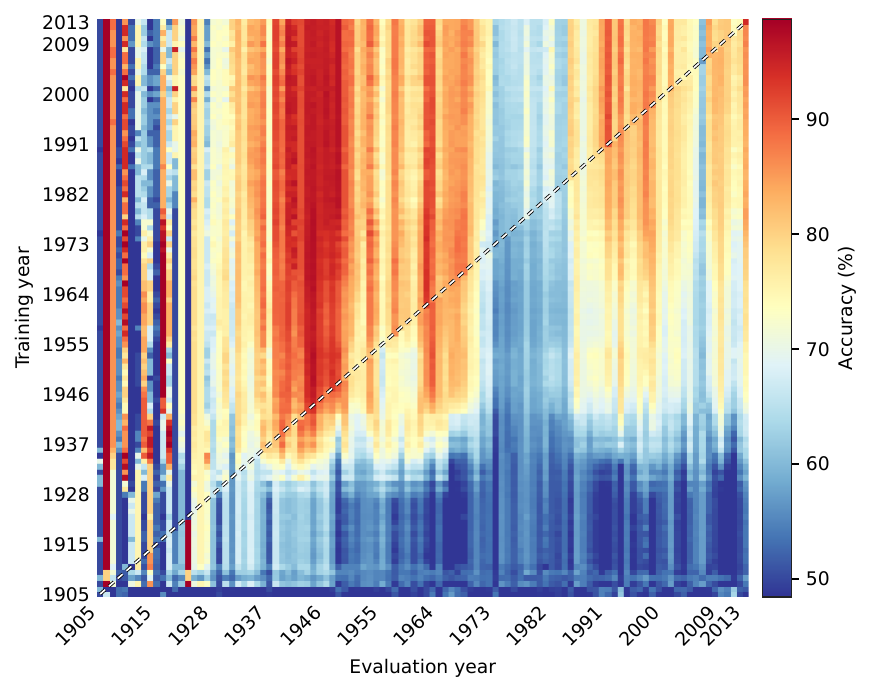}\hfill
    \includegraphics[width=0.49\linewidth]{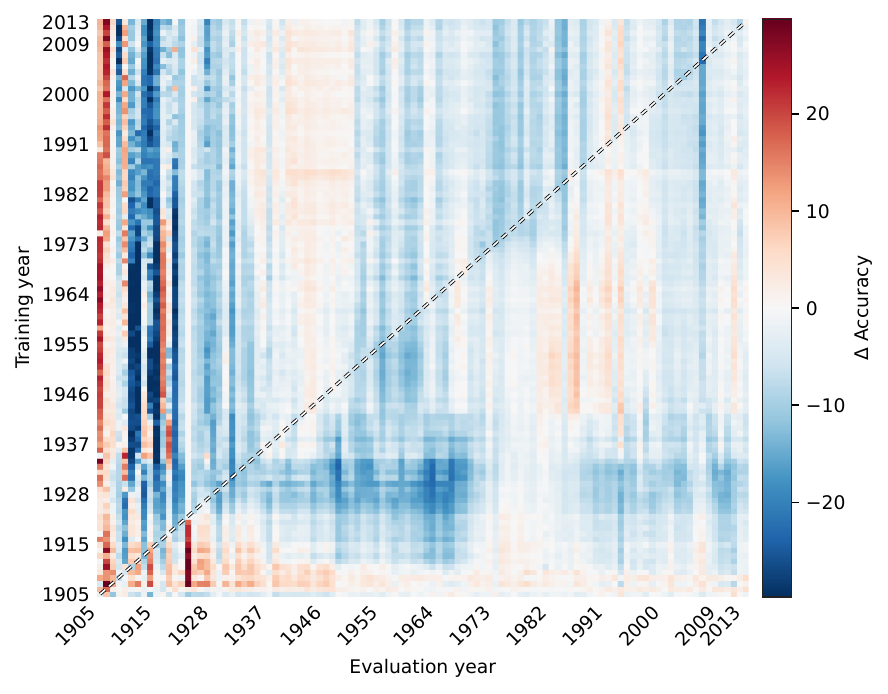}
    \caption{CLIP-B32}
    \label{fig:yearbook_CLIP_B32}
\end{subfigure}
\hfill
\begin{subfigure}[t]{0.49\textwidth}\centering
    \includegraphics[width=0.49\linewidth]{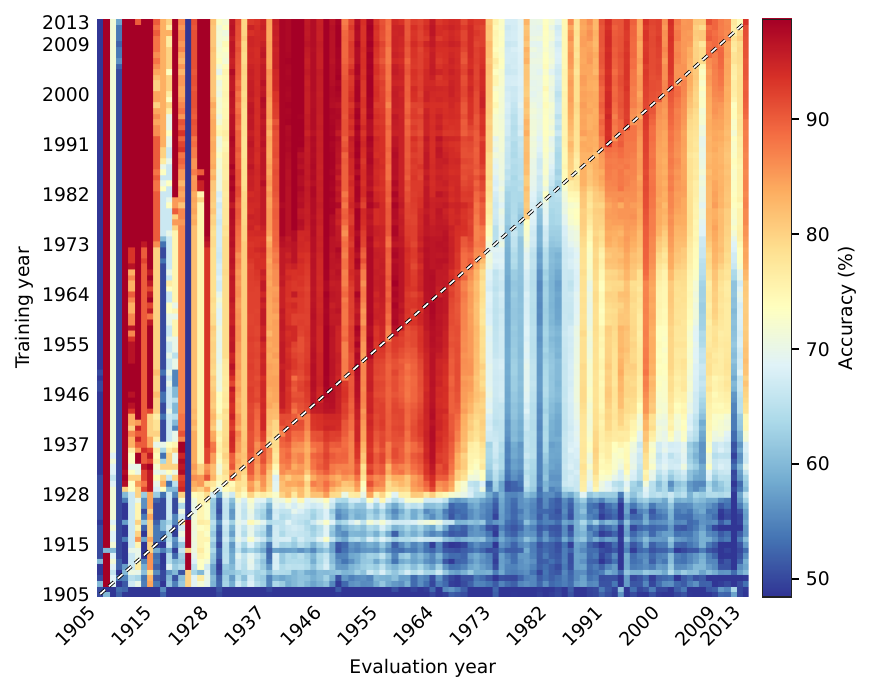}\hfill
    \includegraphics[width=0.49\linewidth]{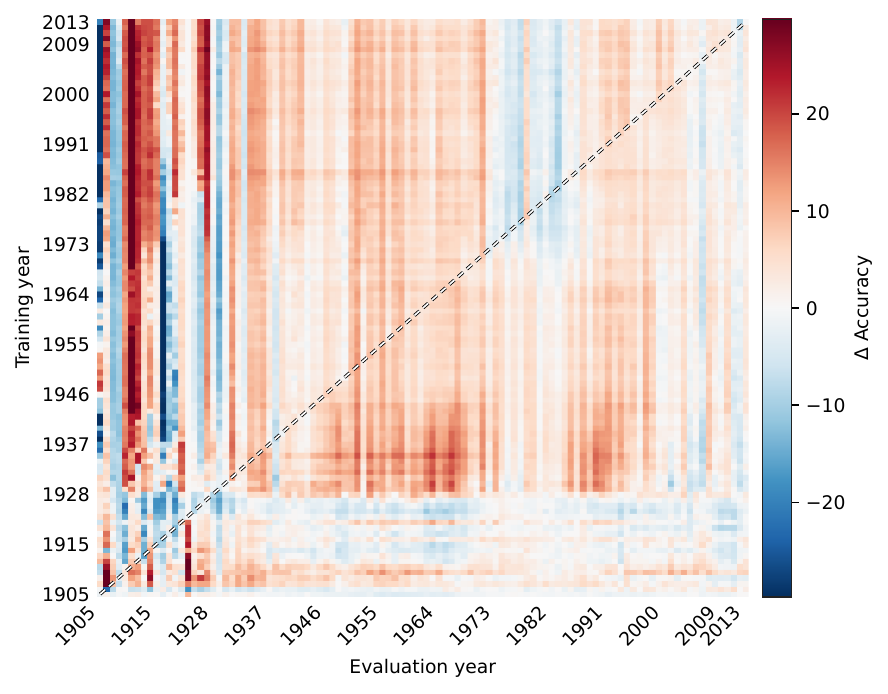}
    \caption{ConvNeXt-S}
    \label{fig:yearbook_ConvNeXt_S}
\end{subfigure}

\begin{subfigure}[t]{0.49\textwidth}\centering
    \includegraphics[width=0.49\linewidth]{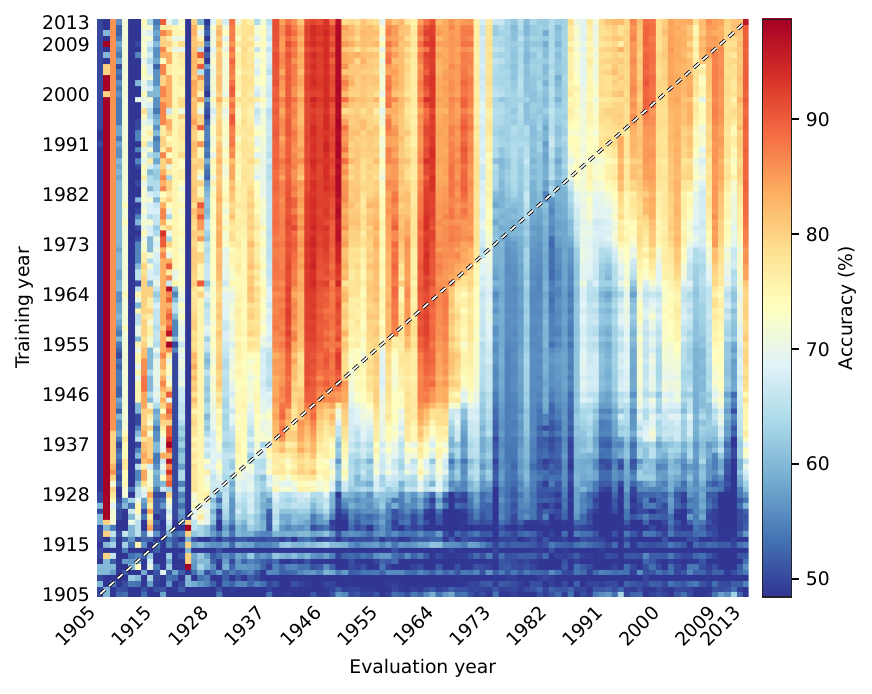}\hfill
    \includegraphics[width=0.49\linewidth]{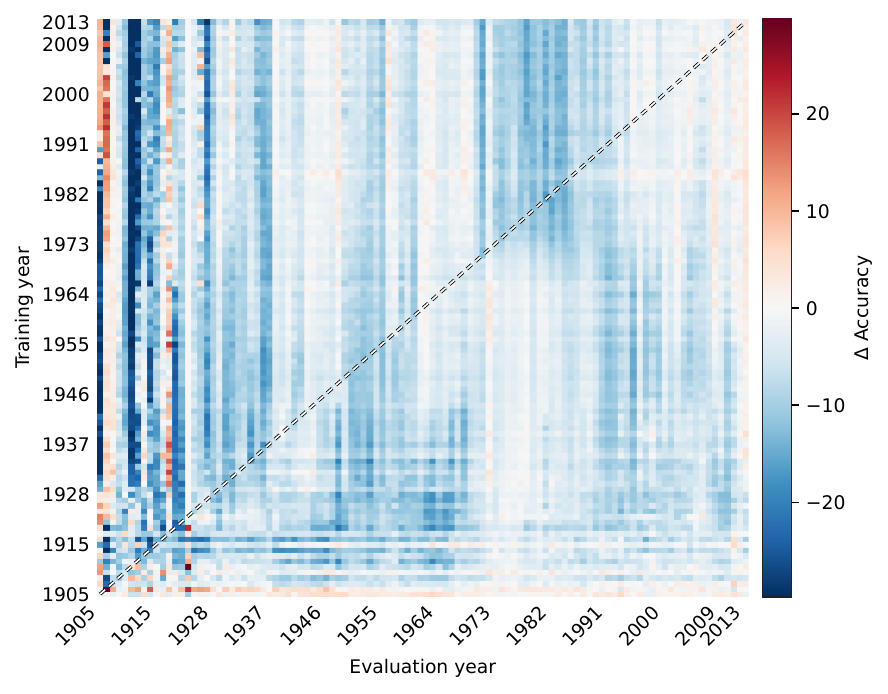}
    \caption{DINOv2-S}
    \label{fig:yearbook_DINOv2_S}
\end{subfigure}
\hfill
\begin{subfigure}[t]{0.49\textwidth}\centering
    \includegraphics[width=0.49\linewidth]{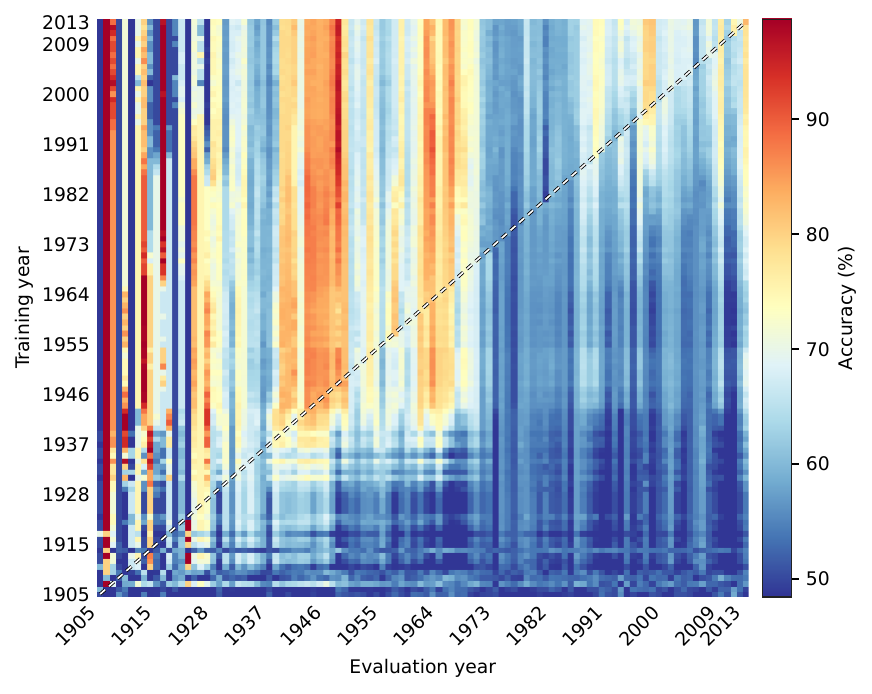}\hfill
    \includegraphics[width=0.49\linewidth]{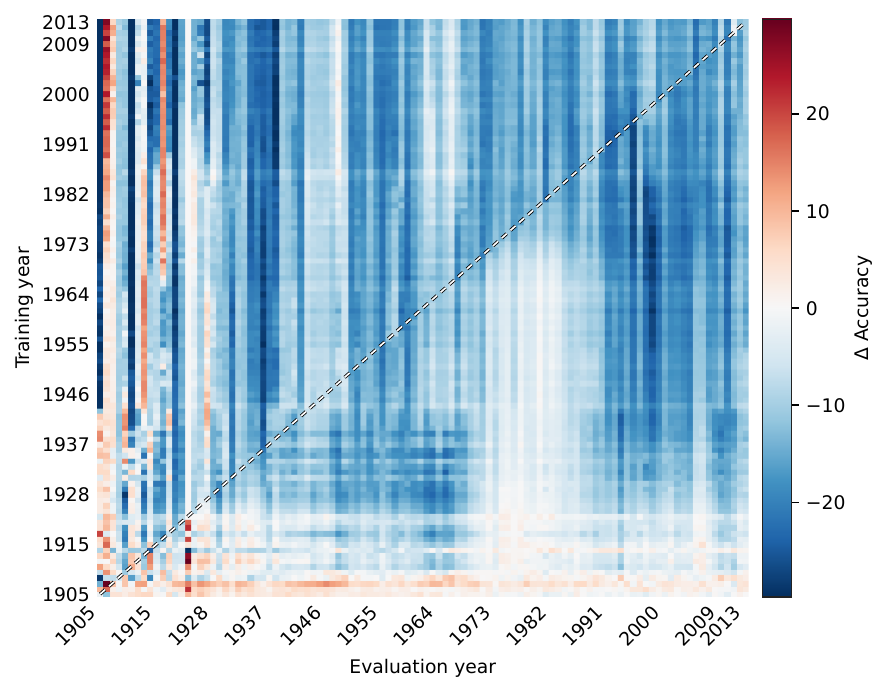}
    \caption{DINOv3-S}
    \label{fig:yearbook_DINOv3_S}
\end{subfigure}

\begin{subfigure}[t]{0.49\textwidth}\centering
    \includegraphics[width=0.49\linewidth]{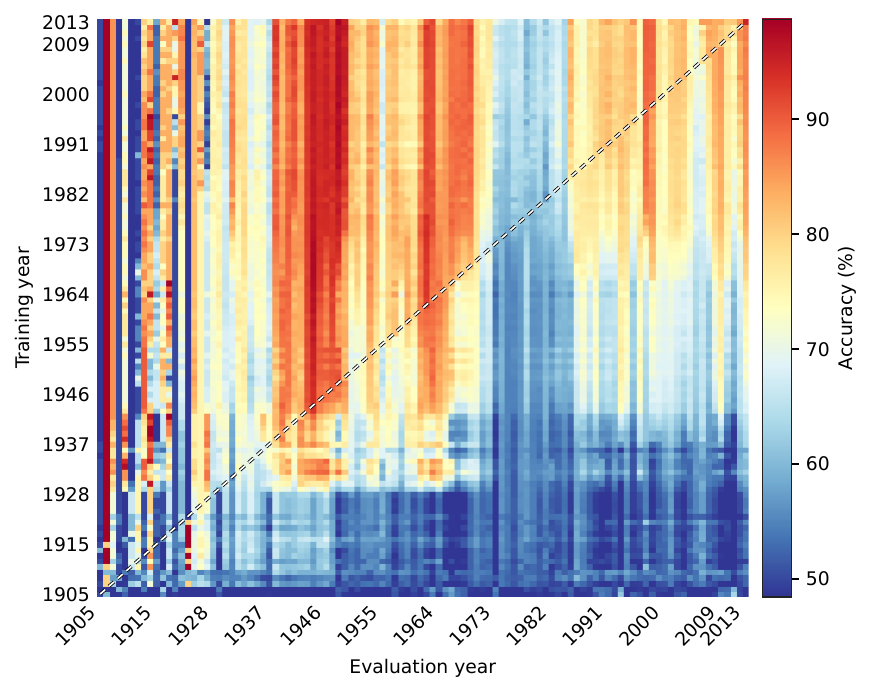}\hfill
    \includegraphics[width=0.49\linewidth]{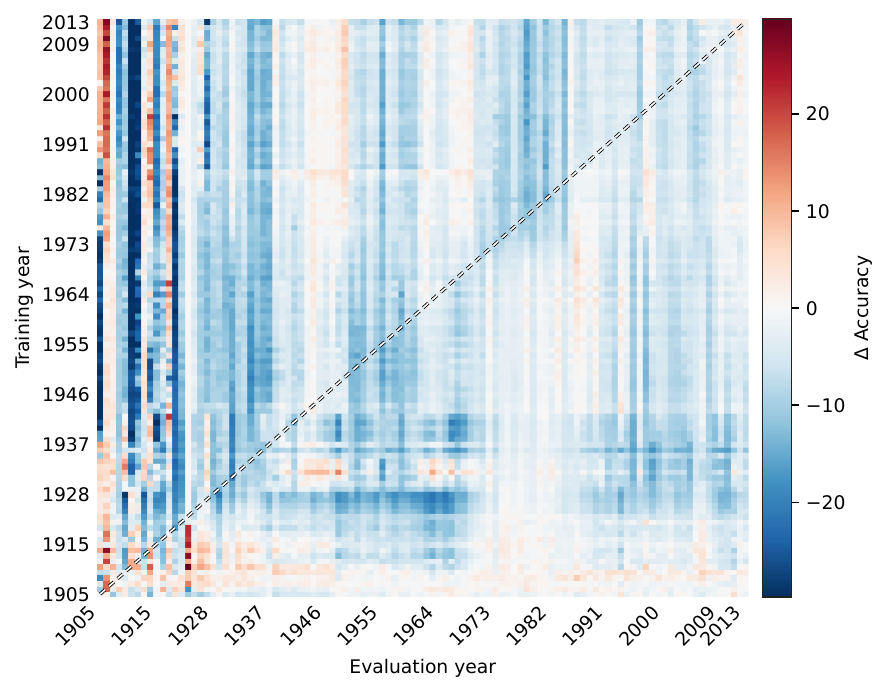}
    \caption{EVA02-B}
    \label{fig:yearbook_EVA02_B}
\end{subfigure}
\hfill
\begin{subfigure}[t]{0.49\textwidth}\centering
    \includegraphics[width=0.49\linewidth]{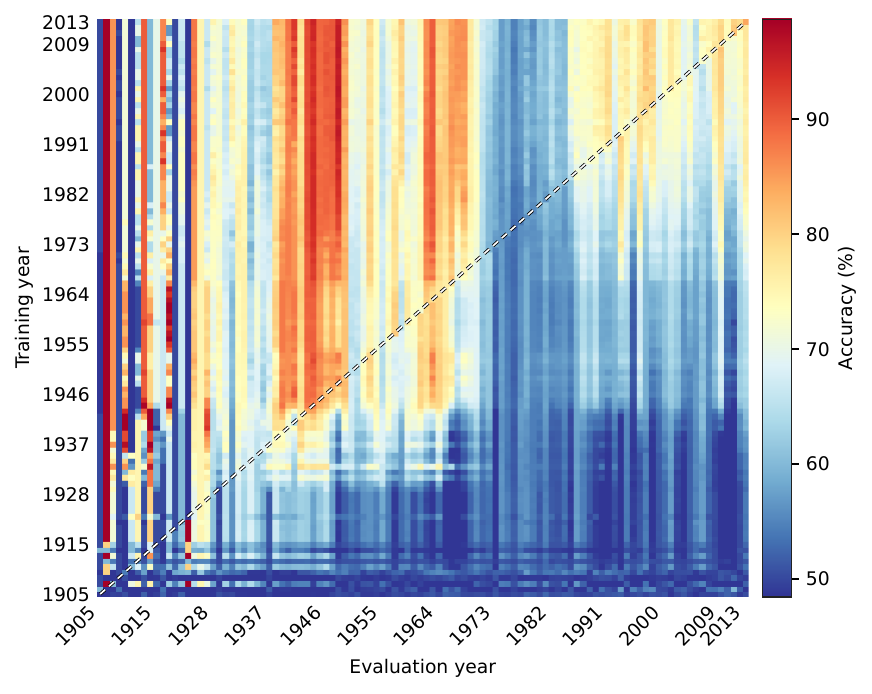}\hfill
    \includegraphics[width=0.49\linewidth]{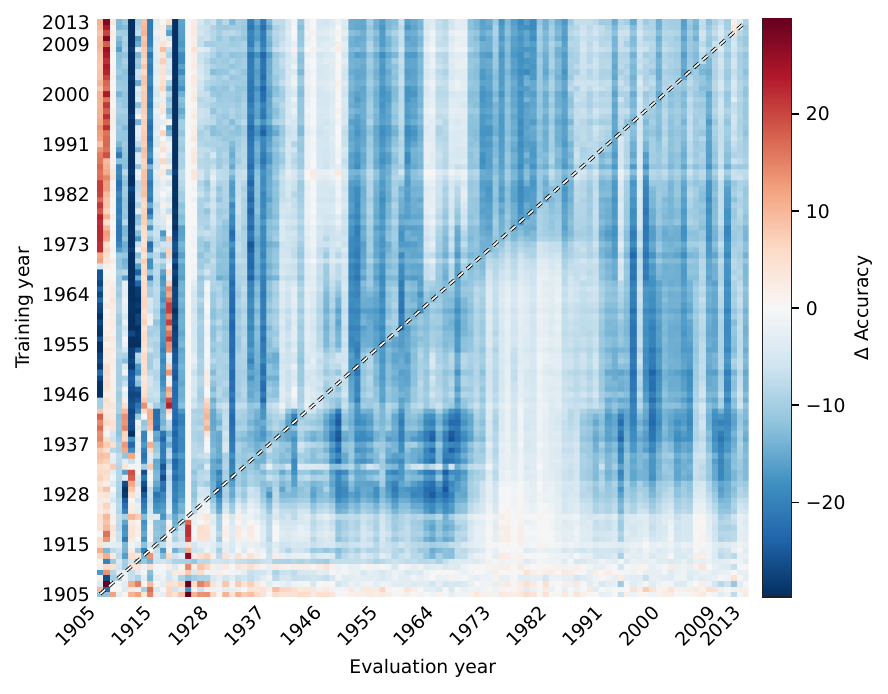}
    \caption{MAE-B}
    \label{fig:yearbook_MAE_B}
\end{subfigure}

\begin{subfigure}[t]{0.49\textwidth}\centering
    \includegraphics[width=0.49\linewidth]{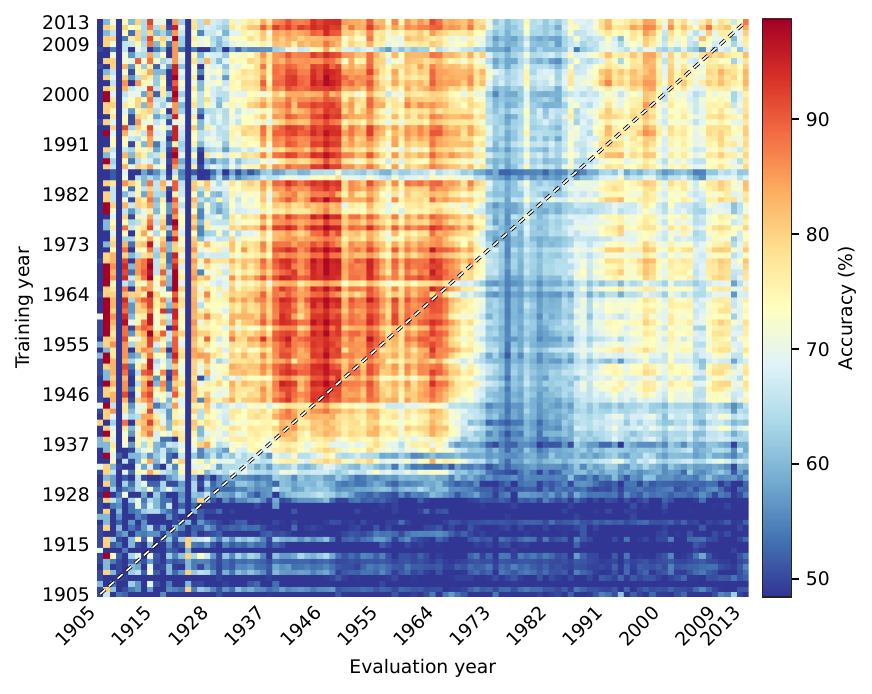}\hfill
    \includegraphics[width=0.49\linewidth]{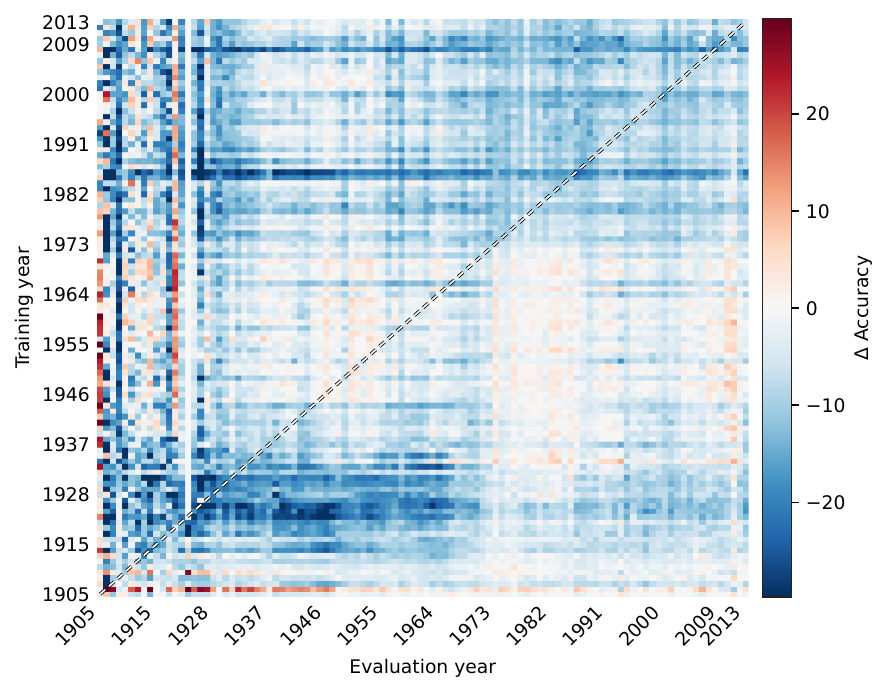}
    \caption{ResNet50-IN}
    \label{fig:yearbook_ResNet50_IN}
\end{subfigure}
\hfill
\begin{subfigure}[t]{0.49\textwidth}\centering
    \includegraphics[width=0.49\linewidth]{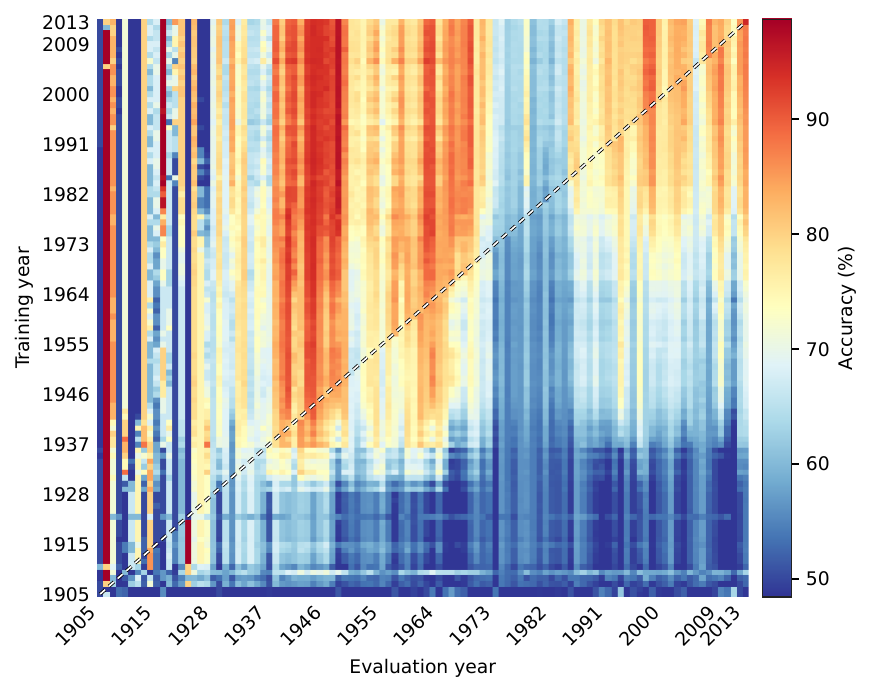}\hfill
    \includegraphics[width=0.49\linewidth]{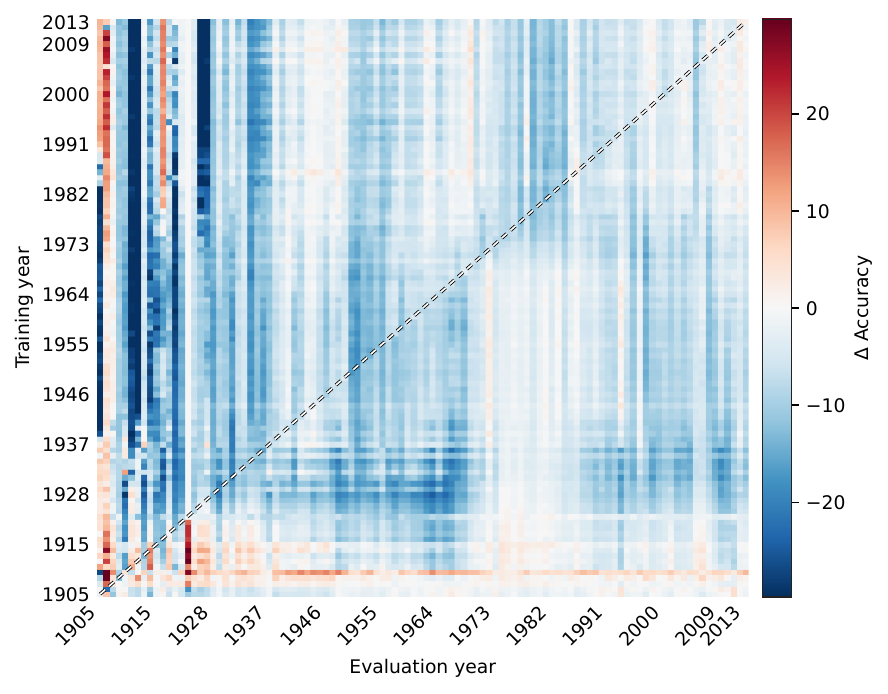}
    \caption{SigLIP-B}
    \label{fig:yearbook_SigLIP_B}
\end{subfigure}

\begin{subfigure}[t]{0.49\textwidth}\centering
    \includegraphics[width=0.49\linewidth]{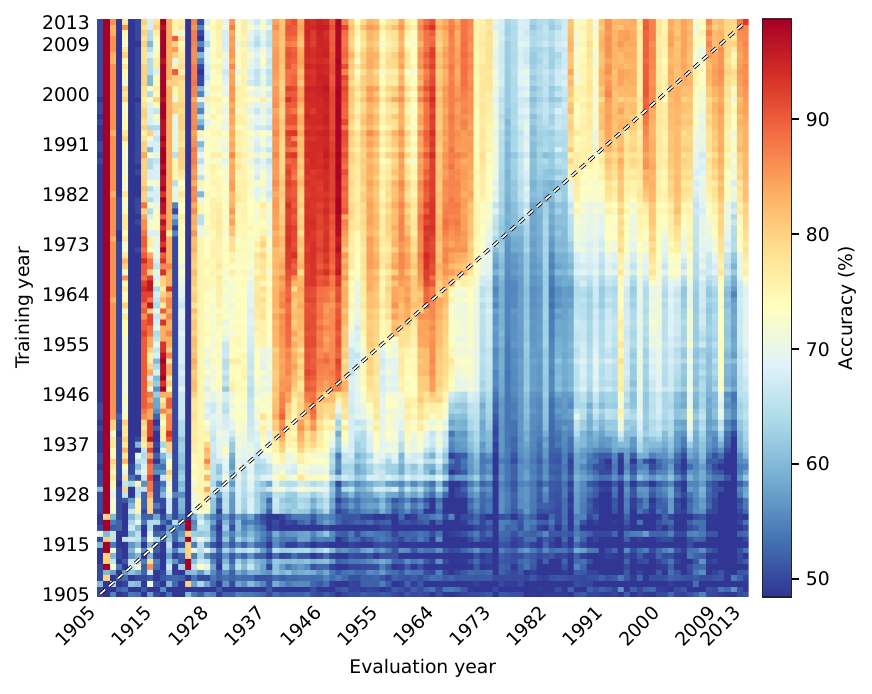}\hfill
    \includegraphics[width=0.49\linewidth]{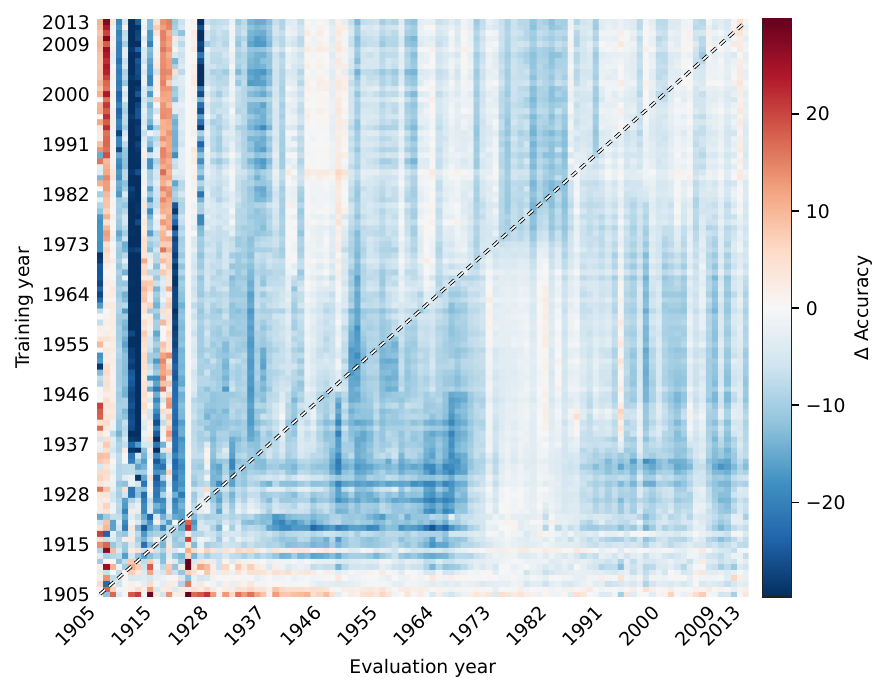}
    \caption{ViT-S16-IN21k}
    \label{fig:yearbook_ViT_S16_IN21k}
\end{subfigure}
\caption{Transfer models: Accuracy drift matrix $M^{(m)}$ and deviation from the cohort mean $\Delta^{(m)} = M^{(m)} - \bar{M}$ for each model, shown on a sequential and a zero-centred diverging scale, respectively.}
\label{fig:yearbook_family_image_transfer}
\end{figure}

\subsection{Forgetting and Rankings}
\label{app:yearbook_forgetting}

\noindent To see how quickly each model forgets, we summarize its drift matrix as a forgetting curve. The curve plots the Accuracy against the lag $\ell = j - i$, the number of slices between the training cutoff $i$ and the evaluation slice $j$. At each lag we average over all training cutoffs,
\[ F(\ell) = \operatorname{mean}_{i} M_{i,\,i+\ell}. \]
The result is the Accuracy at a fixed temporal distance, independent of which period a model was trained on. This separates the effect of temporal distance from the difficulty of any single slice.

\begin{figure}[H]
\centering
\begin{subfigure}[t]{0.49\textwidth}\centering
    \includegraphics[width=\linewidth]{plots_experiments/yearbook/forgetting.pdf}
    \caption{Per model}
    \label{fig:yearbook_forgetting}
\end{subfigure}\hfill
\begin{subfigure}[t]{0.49\textwidth}\centering
    \includegraphics[width=\linewidth]{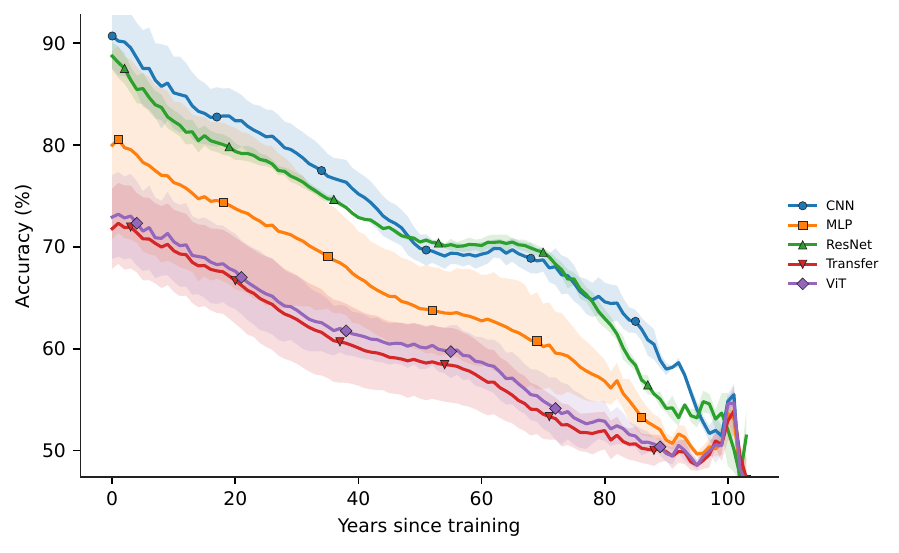}
    \caption{Per model family}
    \label{fig:yearbook_forgetting_family}
\end{subfigure}
\caption{Forgetting curves: each model (left) and averaged within each family (right).}
\label{fig:yearbook_forgetting_combined}
\end{figure}

\begin{figure}[H]
\centering
\begin{subfigure}[t]{0.49\textwidth}\centering
    \includegraphics[width=\linewidth]{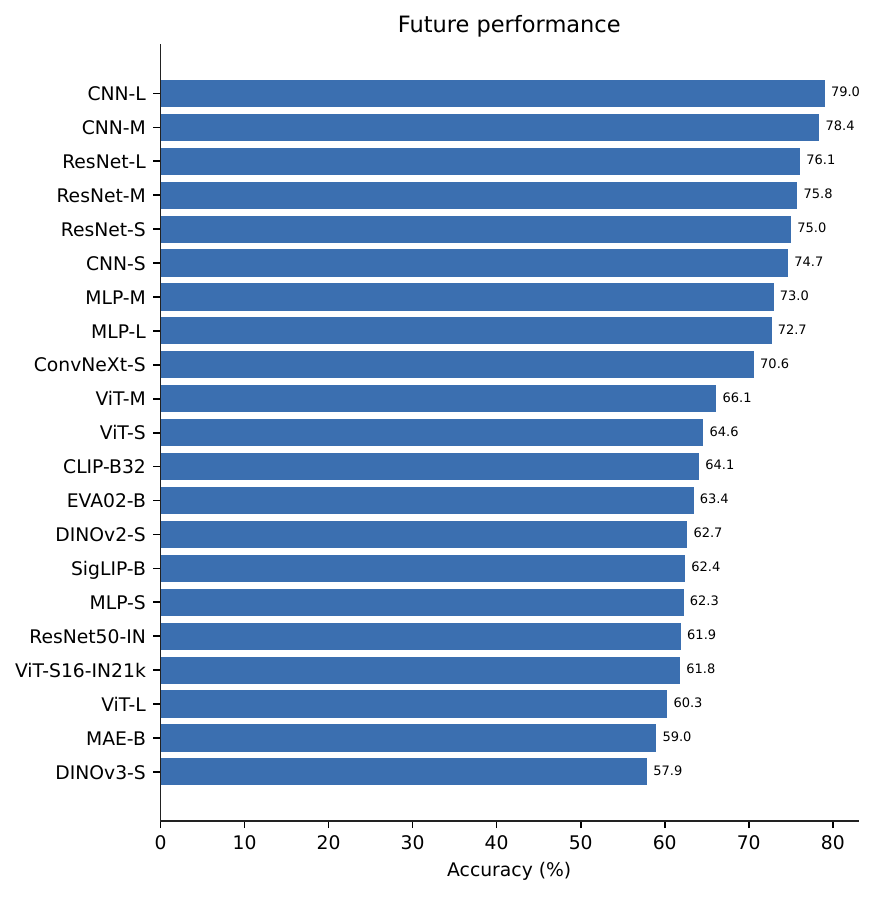}
    \caption{By future performance}
    \label{fig:yearbook_ranking_future}
\end{subfigure}\hfill
\begin{subfigure}[t]{0.49\textwidth}\centering
    \includegraphics[width=\linewidth]{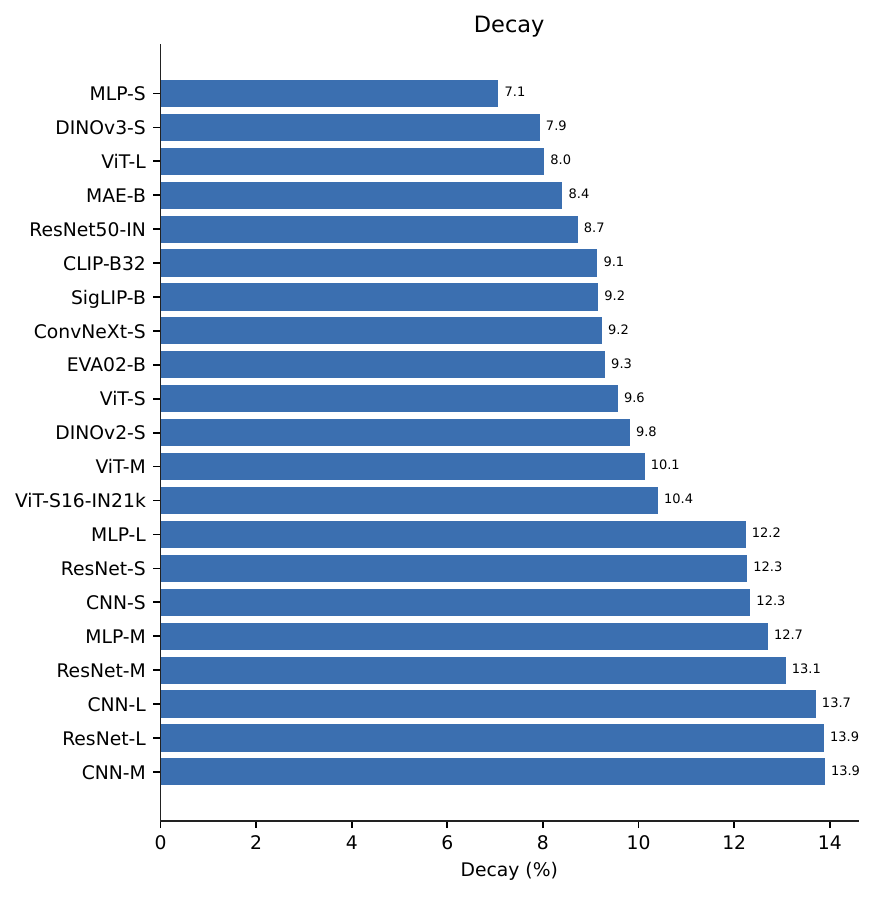}
    \caption{By decay}
    \label{fig:yearbook_ranking_decay}
\end{subfigure}
\caption{Models ranked by mean future performance and by temporal decay.}
\label{fig:yearbook_rankings}
\end{figure}

\subsection{Result Tables}

\begin{table}[H]
\centering
\footnotesize
\caption{Temporal robustness on Yearbook.}
\label{tab:robustness_yearbook}
\begin{minipage}[t]{0.49\linewidth}
\centering
\resizebox{\ifdim\width>\linewidth \linewidth\else\width\fi}{!}{%
\begin{tabular}{l rr}
\toprule
Model & Future (\%) & Decay (\%) \\
\midrule
MLP-S & 62.3 & 7.1 \\
DINOv3-S & 57.9 & 7.9 \\
ViT-L & 60.3 & 8.0 \\
MAE-B & 59.0 & 8.4 \\
ResNet50-IN & 61.9 & 8.7 \\
CLIP-B32 & 64.1 & 9.1 \\
SigLIP-B & 62.4 & 9.2 \\
ConvNeXt-S & 70.6 & 9.2 \\
EVA02-B & 63.4 & 9.3 \\
ViT-S & 64.6 & 9.6 \\
DINOv2-S & 62.7 & 9.8 \\
\bottomrule
\end{tabular}}
\end{minipage}\hfill
\begin{minipage}[t]{0.49\linewidth}
\centering
\resizebox{\ifdim\width>\linewidth \linewidth\else\width\fi}{!}{%
\begin{tabular}{l rr}
\toprule
Model & Future (\%) & Decay (\%) \\
\midrule
ViT-M & 66.1 & 10.1 \\
ViT-S16-IN21k & 61.8 & 10.4 \\
MLP-L & 72.7 & 12.2 \\
ResNet-S & 75.0 & 12.3 \\
CNN-S & 74.7 & 12.3 \\
MLP-M & 73.0 & 12.7 \\
ResNet-M & 75.8 & 13.1 \\
CNN-L & 79.0 & 13.7 \\
ResNet-L & 76.1 & 13.9 \\
CNN-M & 78.4 & 13.9 \\
\bottomrule
\end{tabular}}
\end{minipage}
\end{table}

\noindent
\begin{minipage}[t]{0.49\linewidth}
\centering
\scriptsize
\setlength{\tabcolsep}{4pt}
\captionof{table}{Yearbook: models trained up to 1905, ordered by future performance.}
\label{tab:yearbook_cutoff_1905}
\resizebox{\ifdim\width>\linewidth \linewidth\else\width\fi}{!}{%
\begin{tabular}{c l rrr}
\toprule
Rank & Model & Accuracy (\%) & Future (\%) & Decay (\%) \\
\midrule
1 & ViT-S16-IN21k & 50.0 & 50.7 & -0.7 \\
2 & ResNet-M & 60.0 & 50.3 & 9.7 \\
3 & ResNet-L & 70.0 & 50.1 & 19.9 \\
4 & MLP-L & 40.0 & 49.8 & -9.8 \\
5 & MAE-B & 40.0 & 49.5 & -9.5 \\
6 & ResNet-S & 70.0 & 49.3 & 20.7 \\
7 & DINOv2-S & 50.0 & 48.5 & 1.5 \\
8 & DINOv3-S & 50.0 & 48.3 & 1.7 \\
9 & ViT-L & 50.0 & 46.8 & 3.2 \\
10 & MLP-S & 40.0 & 46.3 & -6.3 \\
11 & ConvNeXt-S & 50.0 & 46.1 & 3.9 \\
12 & ViT-M & 50.0 & 45.7 & 4.3 \\
13 & CNN-L & 50.0 & 45.4 & 4.6 \\
14 & ViT-S & 50.0 & 45.1 & 4.9 \\
15 & ResNet50-IN & 50.0 & 45.1 & 4.9 \\
16 & EVA02-B & 50.0 & 44.9 & 5.1 \\
17 & MLP-M & 50.0 & 44.9 & 5.1 \\
18 & CNN-S & 50.0 & 44.9 & 5.1 \\
19 & CLIP-B32 & 50.0 & 44.7 & 5.3 \\
20 & CNN-M & 60.0 & 44.7 & 15.3 \\
21 & SigLIP-B & 50.0 & 44.7 & 5.3 \\
\bottomrule
\end{tabular}}
\end{minipage}
\hfill
\begin{minipage}[t]{0.49\linewidth}
\centering
\scriptsize
\setlength{\tabcolsep}{4pt}
\captionof{table}{Yearbook: models trained up to 1944, ordered by future performance.}
\label{tab:yearbook_cutoff_1944}
\resizebox{\ifdim\width>\linewidth \linewidth\else\width\fi}{!}{%
\begin{tabular}{c l rrr}
\toprule
Rank & Model & Accuracy (\%) & Future (\%) & Decay (\%) \\
\midrule
1 & ResNet-L & 98.4 & 82.9 & 15.5 \\
2 & ResNet-M & 98.3 & 81.9 & 16.4 \\
3 & CNN-M & 98.8 & 81.4 & 17.4 \\
4 & ResNet-S & 98.0 & 81.1 & 16.9 \\
5 & CNN-L & 97.1 & 80.2 & 16.8 \\
6 & CNN-S & 95.0 & 79.1 & 15.9 \\
7 & ConvNeXt-S & 96.3 & 78.5 & 17.8 \\
8 & MLP-M & 93.4 & 77.6 & 15.9 \\
9 & MLP-L & 87.9 & 74.6 & 13.3 \\
10 & ViT-M & 92.6 & 71.4 & 21.2 \\
11 & ViT-L & 91.0 & 70.7 & 20.3 \\
12 & ViT-S & 89.4 & 69.9 & 19.5 \\
13 & CLIP-B32 & 94.3 & 69.9 & 24.5 \\
14 & EVA02-B & 93.1 & 67.9 & 25.2 \\
15 & DINOv2-S & 88.5 & 67.1 & 21.5 \\
16 & SigLIP-B & 88.9 & 66.8 & 22.1 \\
17 & ViT-S16-IN21k & 86.4 & 65.0 & 21.4 \\
18 & ResNet50-IN & 80.2 & 64.8 & 15.3 \\
19 & MAE-B & 86.7 & 62.6 & 24.1 \\
20 & DINOv3-S & 82.7 & 62.4 & 20.3 \\
21 & MLP-S & 65.8 & 57.9 & 7.9 \\
\bottomrule
\end{tabular}}
\end{minipage}

\vspace{1.5ex}

\noindent
\begin{minipage}[t]{0.49\linewidth}
\centering
\scriptsize
\setlength{\tabcolsep}{4pt}
\captionof{table}{Yearbook: models trained up to 1978, ordered by future performance.}
\label{tab:yearbook_cutoff_1978}
\resizebox{\ifdim\width>\linewidth \linewidth\else\width\fi}{!}{%
\begin{tabular}{c l rrr}
\toprule
Rank & Model & Accuracy (\%) & Future (\%) & Decay (\%) \\
\midrule
1 & CNN-L & 85.9 & 91.4 & -5.4 \\
2 & ResNet-M & 88.7 & 90.4 & -1.6 \\
3 & ResNet-L & 87.6 & 89.3 & -1.7 \\
4 & CNN-M & 86.8 & 88.8 & -2.0 \\
5 & MLP-M & 79.4 & 84.5 & -5.1 \\
6 & ResNet-S & 83.4 & 84.4 & -1.0 \\
7 & MLP-L & 75.5 & 83.2 & -7.7 \\
8 & CNN-S & 72.7 & 81.1 & -8.5 \\
9 & ConvNeXt-S & 80.6 & 78.6 & 1.9 \\
10 & ViT-M & 72.4 & 76.8 & -4.4 \\
11 & CLIP-B32 & 69.9 & 74.8 & -5.0 \\
12 & EVA02-B & 60.6 & 73.3 & -12.8 \\
13 & ViT-L & 66.8 & 72.4 & -5.6 \\
14 & DINOv2-S & 64.5 & 72.3 & -7.8 \\
15 & ViT-S & 64.2 & 71.8 & -7.6 \\
16 & ViT-S16-IN21k & 66.5 & 70.7 & -4.3 \\
17 & ResNet50-IN & 72.7 & 70.7 & 2.0 \\
18 & SigLIP-B & 64.8 & 69.9 & -5.1 \\
19 & MLP-S & 67.9 & 69.4 & -1.5 \\
20 & MAE-B & 56.1 & 65.0 & -8.9 \\
21 & DINOv3-S & 60.3 & 59.9 & 0.4 \\
\bottomrule
\end{tabular}}
\end{minipage}
\hfill
\begin{minipage}[t]{0.49\linewidth}
\centering
\scriptsize
\setlength{\tabcolsep}{4pt}
\captionof{table}{Yearbook: models trained up to 2012, ordered by future performance.}
\label{tab:yearbook_cutoff_2012}
\resizebox{\ifdim\width>\linewidth \linewidth\else\width\fi}{!}{%
\begin{tabular}{c l rrr}
\toprule
Rank & Model & Accuracy (\%) & Future (\%) & Decay (\%) \\
\midrule
1 & CNN-L & 93.2 & 98.2 & -5.0 \\
2 & CNN-M & 93.2 & 97.7 & -4.6 \\
3 & ResNet-L & 88.9 & 97.5 & -8.5 \\
4 & ResNet-M & 92.6 & 97.0 & -4.3 \\
5 & ResNet-S & 86.8 & 95.4 & -8.5 \\
6 & CNN-S & 83.7 & 93.4 & -9.7 \\
7 & MLP-L & 87.4 & 91.9 & -4.6 \\
8 & MLP-M & 86.3 & 91.7 & -5.4 \\
9 & ConvNeXt-S & 73.7 & 91.2 & -17.5 \\
10 & DINOv2-S & 82.6 & 90.0 & -7.3 \\
11 & ViT-M & 77.9 & 88.1 & -10.2 \\
12 & ViT-S & 80.5 & 88.0 & -7.5 \\
13 & SigLIP-B & 85.8 & 87.5 & -1.7 \\
14 & EVA02-B & 82.1 & 87.4 & -5.3 \\
15 & CLIP-B32 & 74.7 & 86.6 & -11.9 \\
16 & ViT-S16-IN21k & 87.9 & 86.5 & 1.4 \\
17 & ResNet50-IN & 76.8 & 83.8 & -7.0 \\
18 & DINOv3-S & 71.1 & 77.9 & -6.9 \\
19 & MAE-B & 78.4 & 77.4 & 1.0 \\
20 & MLP-S & 70.0 & 76.1 & -6.1 \\
21 & ViT-L & 66.8 & 74.4 & -7.5 \\
\bottomrule
\end{tabular}}
\end{minipage}

\begin{table}[H]
\centering
\footnotesize
\caption{Yearbook: future performance and decay by model family.}
\label{tab:yearbook_by_family}
\begin{tabular}{l rrrrrrrr}
\toprule
 & \multicolumn{2}{c}{1905} & \multicolumn{2}{c}{1944} & \multicolumn{2}{c}{1978} & \multicolumn{2}{c}{2012} \\
Family & Future & Decay & Future & Decay & Future & Decay & Future & Decay \\
\midrule
CNN & 45.0 & 8.3 & 80.3 & 16.7 & 87.1 & -5.3 & 96.4 & -6.4 \\
MLP & 47.0 & -3.7 & 70.0 & 12.4 & 79.0 & -4.8 & 86.6 & -5.3 \\
ResNet & 49.9 & 16.8 & 82.0 & 16.2 & 88.0 & -1.4 & 96.6 & -7.1 \\
Transfer & 46.9 & 1.9 & 67.2 & 21.4 & 70.6 & -4.4 & 85.4 & -6.1 \\
ViT & 45.9 & 4.1 & 70.7 & 20.3 & 73.7 & -5.9 & 83.5 & -8.4 \\
\bottomrule
\end{tabular}
\end{table}

\pagebreak

\captionsetup[figure]{font=footnotesize,labelfont=footnotesize}

\section{Amazon Reviews -- Drift Matrices}
\label{app:amazon_reviews}

\noindent The cohort-mean and per-model deviation matrices shown here, and the in-distribution, future, and decay quantities tabulated below, are defined in Section~\ref{subsec:drift_summary}.

\begin{figure}[H]
\centering
\includegraphics[width=0.5\textwidth]{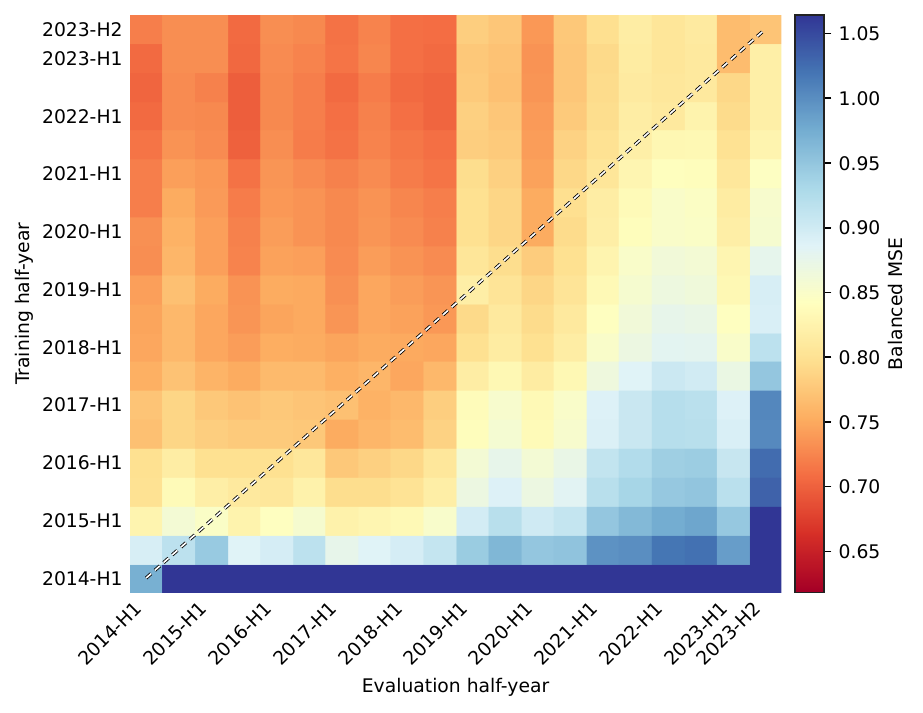}
\caption{Cohort-mean Balanced MSE matrix $\bar{M}$ over the Amazon Reviews models. Cell $(i,j)$ is the mean across those models of the score from training through slice $i$ and evaluating on slice $j$.}
\label{fig:amazon_reviews_mean_matrix}
\end{figure}

\subsection{Model Roster}

\noindent
\begin{minipage}[t]{0.49\linewidth}
\centering
\footnotesize
\captionof{table}{Amazon Reviews: models trained from scratch.}
\label{tab:amazon_reviews_roster}
\resizebox{\ifdim\width>\linewidth \linewidth\else\width\fi}{!}{%
\begin{tabular}{l l rr}
\toprule
Model & Family & Trainable & Total \\
\midrule
TX-S & Transformer & 83k & 124.7M \\
TextCNN-S & TextCNN & 88k & 124.7M \\
FFN-S & FFN & 99k & 124.7M \\
BiGRU-S & Recurrent & 99k & 124.7M \\
FFN-M & FFN & 394k & 125.0M \\
TextCNN-M & TextCNN & 461k & 125.1M \\
TX-M & Transformer & 493k & 125.1M \\
BiLSTM-M & Recurrent & 536k & 125.2M \\
FFN-L & FFN & 1.6M & 126.2M \\
TX-L & Transformer & 1.9M & 126.5M \\
TextCNN-L & TextCNN & 1.9M & 126.6M \\
BiLSTM-Attn-L & Recurrent & 2.2M & 126.8M \\
\bottomrule
\end{tabular}}
\end{minipage}
\hfill
\begin{minipage}[t]{0.49\linewidth}
\centering
\footnotesize
\captionof{table}{Amazon Reviews: frozen pretrained encoders, with trainable head and total parameters.}
\label{tab:amazon_reviews_roster_frozen}
\resizebox{\ifdim\width>\linewidth \linewidth\else\width\fi}{!}{%
\begin{tabular}{l l rr}
\toprule
Model & Family & Trainable & Total \\
\midrule
DistilBERT & Frozen & 769 & 66.4M \\
ELECTRA & Frozen & 769 & 108.9M \\
BERT & Frozen & 769 & 109.5M \\
MPNet & Frozen & 769 & 109.5M \\
RoBERTa & Frozen & 769 & 124.6M \\
ModernBERT & Frozen & 769 & 149.0M \\
DeBERTa-v3 & Frozen & 769 & 183.8M \\
\bottomrule
\end{tabular}}
\end{minipage}

\subsection{FFN}

\begin{figure}[H]
\centering
\begin{subfigure}[t]{0.49\textwidth}\centering
    \includegraphics[width=0.49\linewidth]{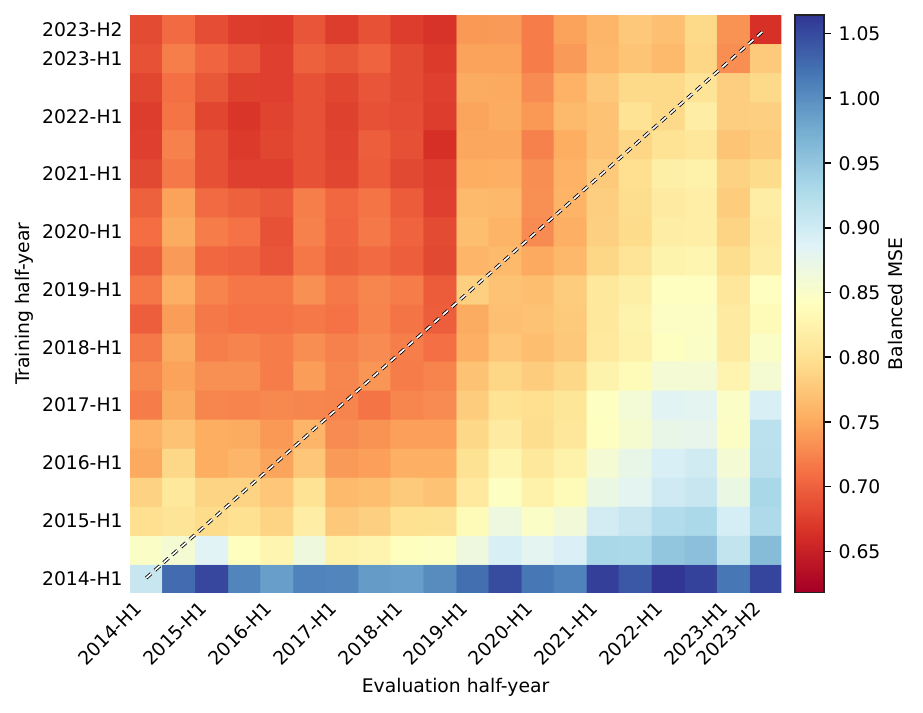}\hfill
    \includegraphics[width=0.49\linewidth]{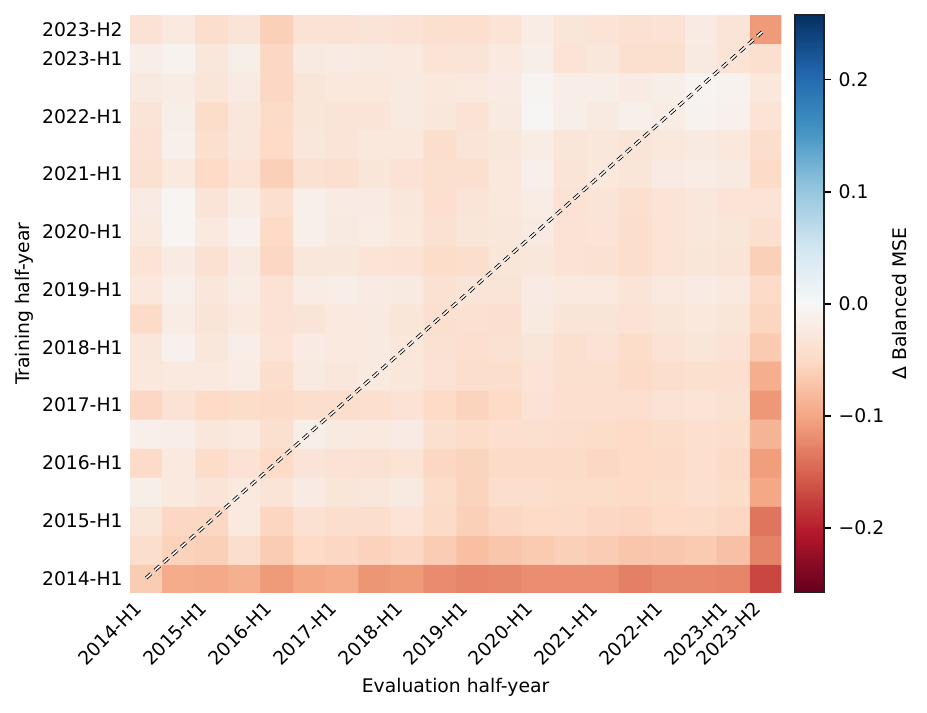}
    \caption{FFN-S}
    \label{fig:amazon_reviews_FFN_S}
\end{subfigure}
\hfill
\begin{subfigure}[t]{0.49\textwidth}\centering
    \includegraphics[width=0.49\linewidth]{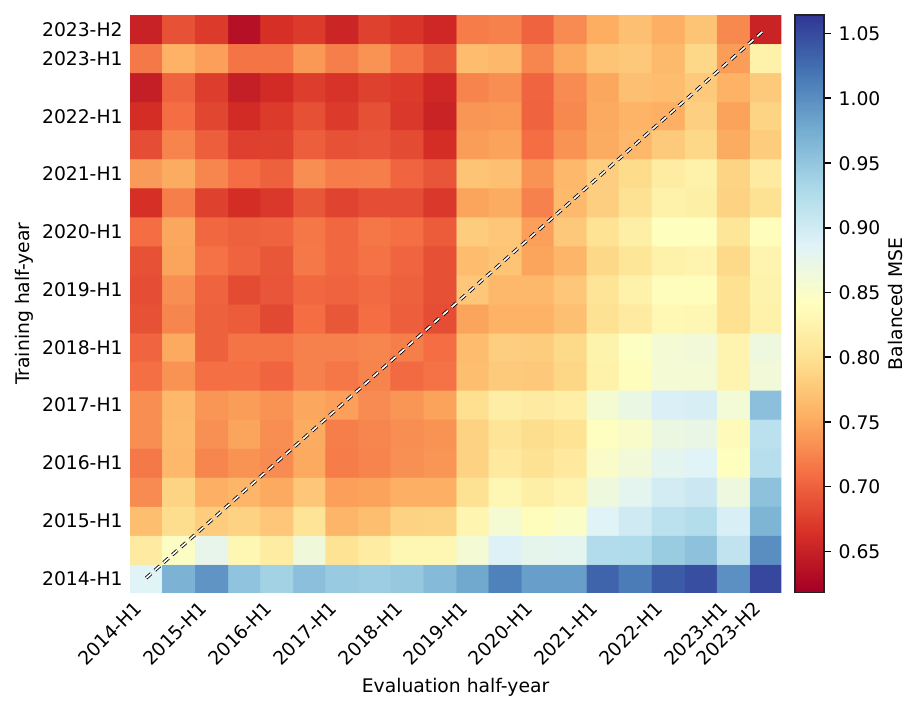}\hfill
    \includegraphics[width=0.49\linewidth]{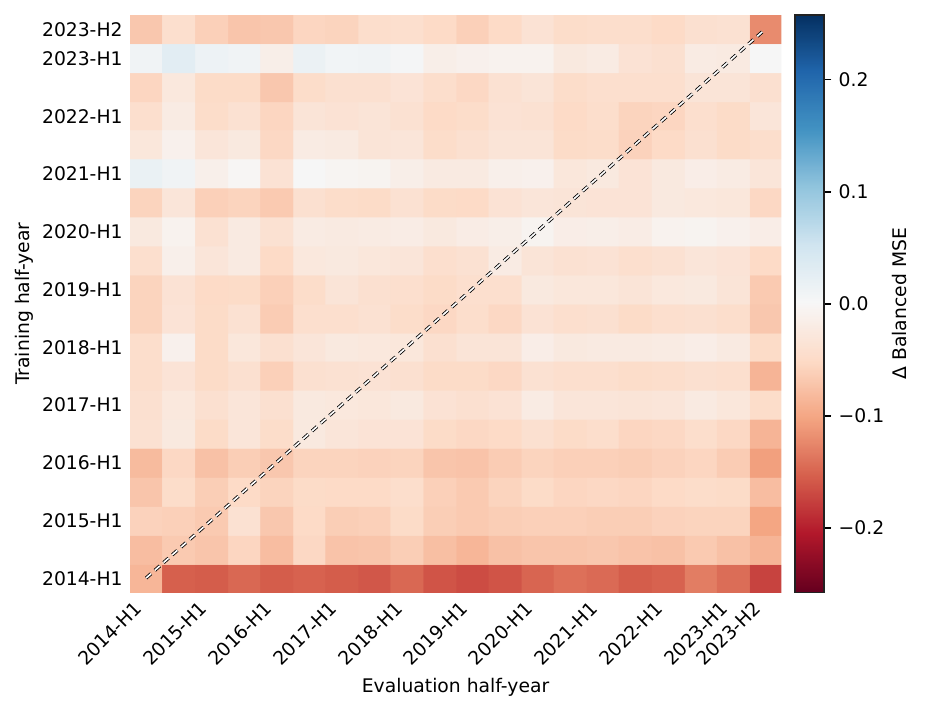}
    \caption{FFN-M}
    \label{fig:amazon_reviews_FFN_M}
\end{subfigure}

\begin{subfigure}[t]{0.49\textwidth}\centering
    \includegraphics[width=0.49\linewidth]{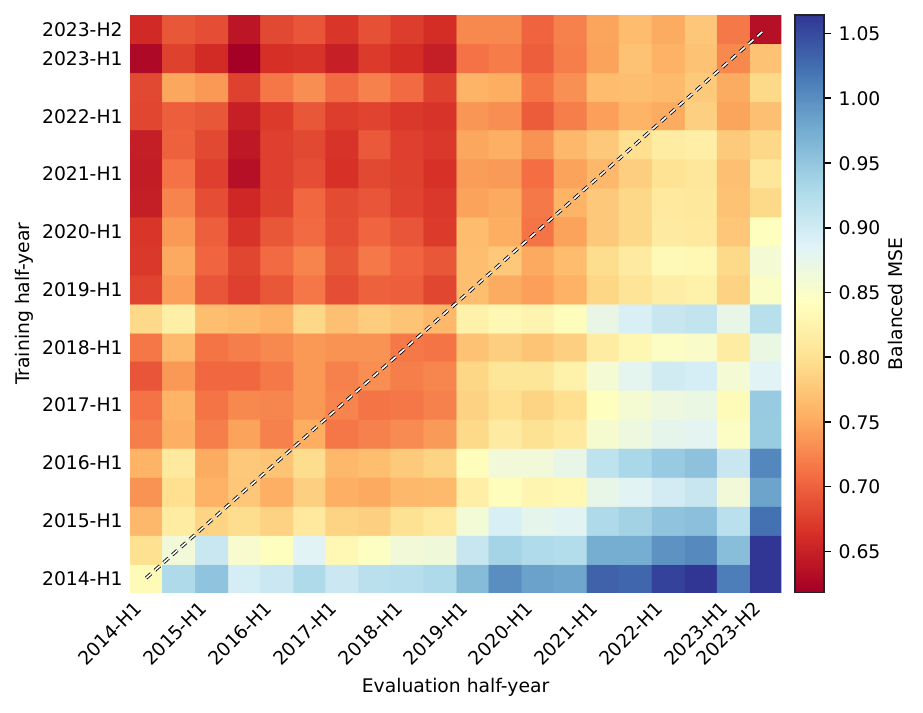}\hfill
    \includegraphics[width=0.49\linewidth]{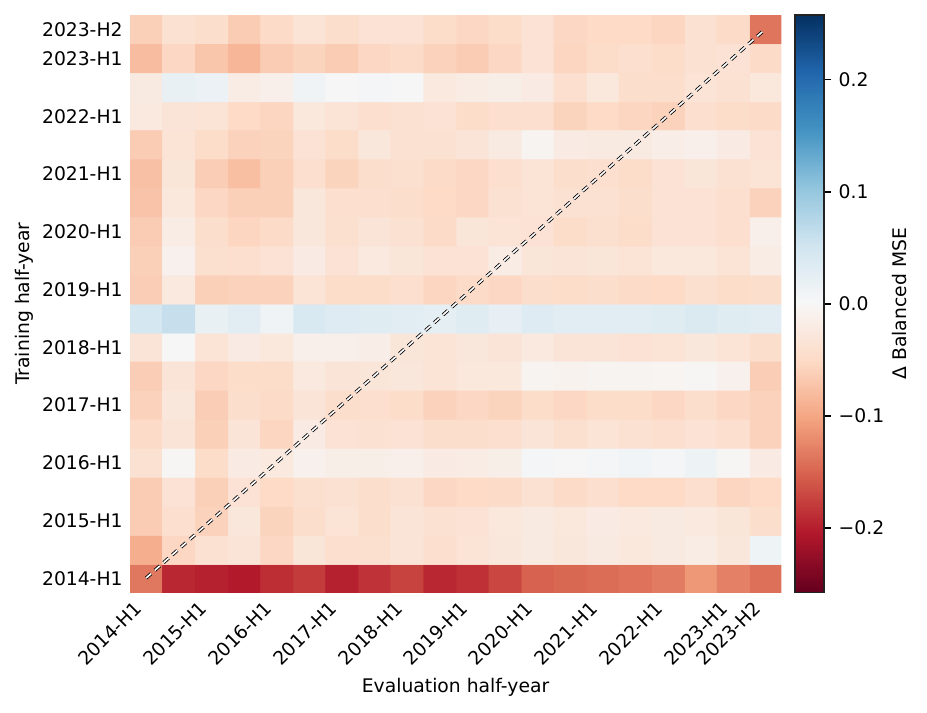}
    \caption{FFN-L}
    \label{fig:amazon_reviews_FFN_L}
\end{subfigure}
\caption{FFN models: Balanced MSE drift matrix $M^{(m)}$ and deviation from the cohort mean $\Delta^{(m)} = M^{(m)} - \bar{M}$ for each model, shown on a sequential and a zero-centred diverging scale, respectively.}
\label{fig:amazon_reviews_family_text_ffn_regression}
\end{figure}

\subsection{TextCNN}

\begin{figure}[H]
\centering
\begin{subfigure}[t]{0.49\textwidth}\centering
    \includegraphics[width=0.49\linewidth]{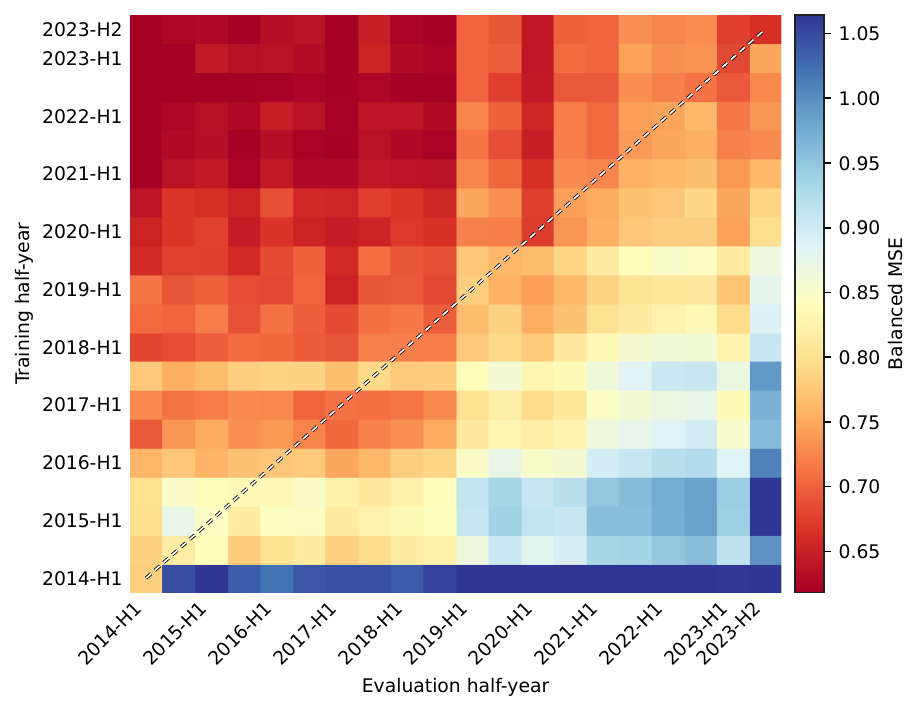}\hfill
    \includegraphics[width=0.49\linewidth]{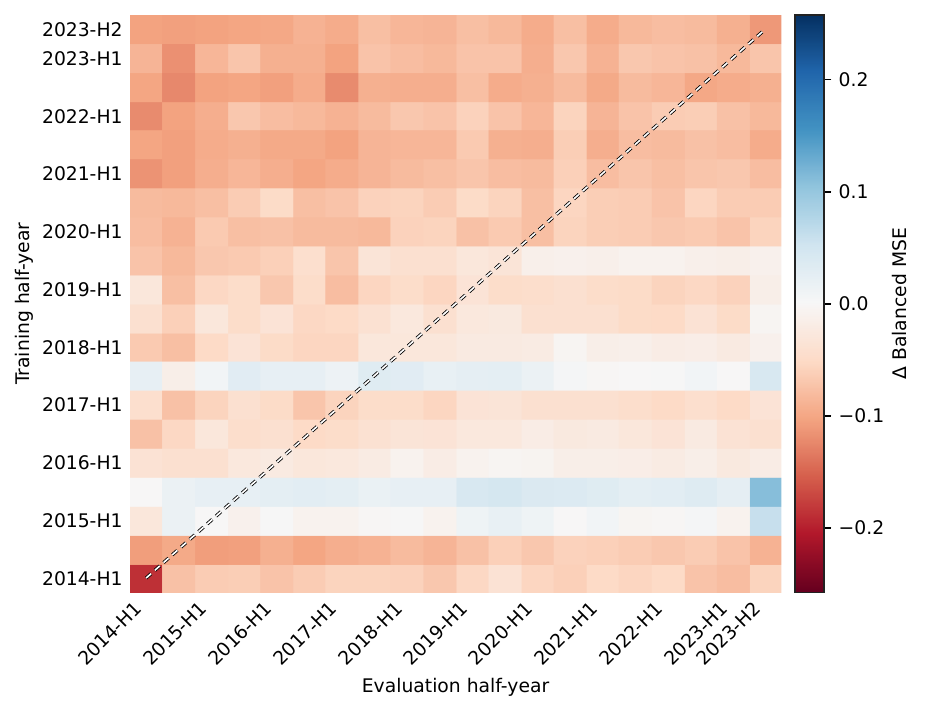}
    \caption{TextCNN-S}
    \label{fig:amazon_reviews_TextCNN_S}
\end{subfigure}
\hfill
\begin{subfigure}[t]{0.49\textwidth}\centering
    \includegraphics[width=0.49\linewidth]{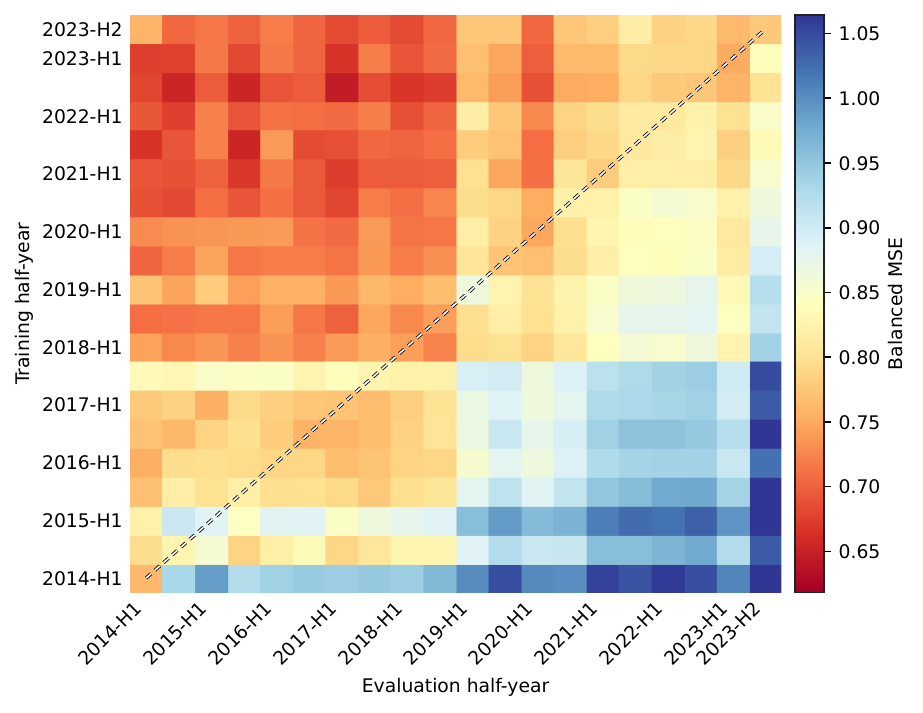}\hfill
    \includegraphics[width=0.49\linewidth]{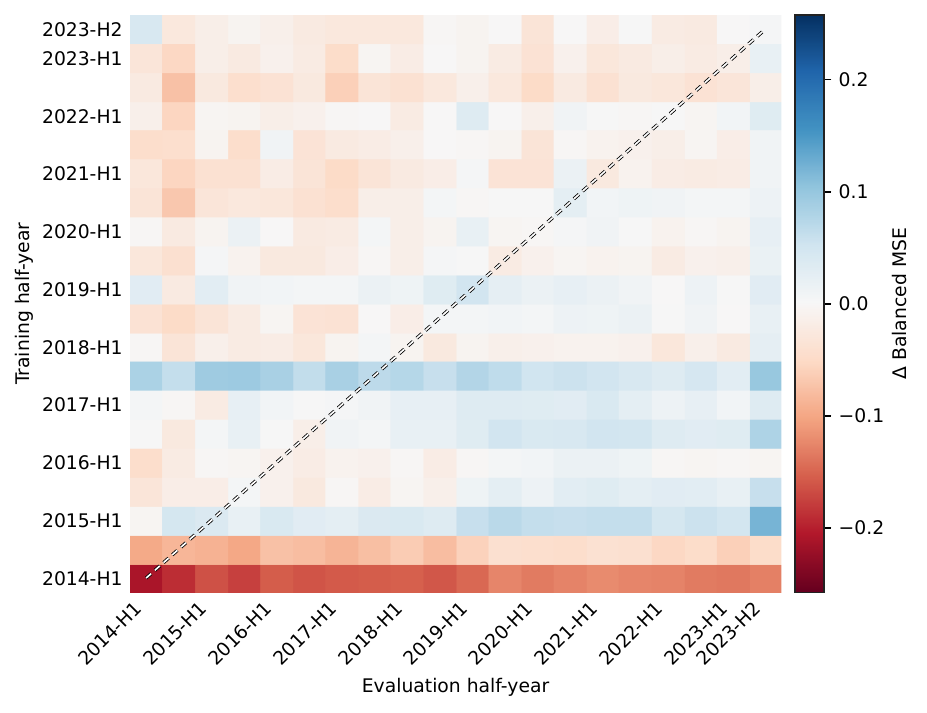}
    \caption{TextCNN-M}
    \label{fig:amazon_reviews_TextCNN_M}
\end{subfigure}

\begin{subfigure}[t]{0.49\textwidth}\centering
    \includegraphics[width=0.49\linewidth]{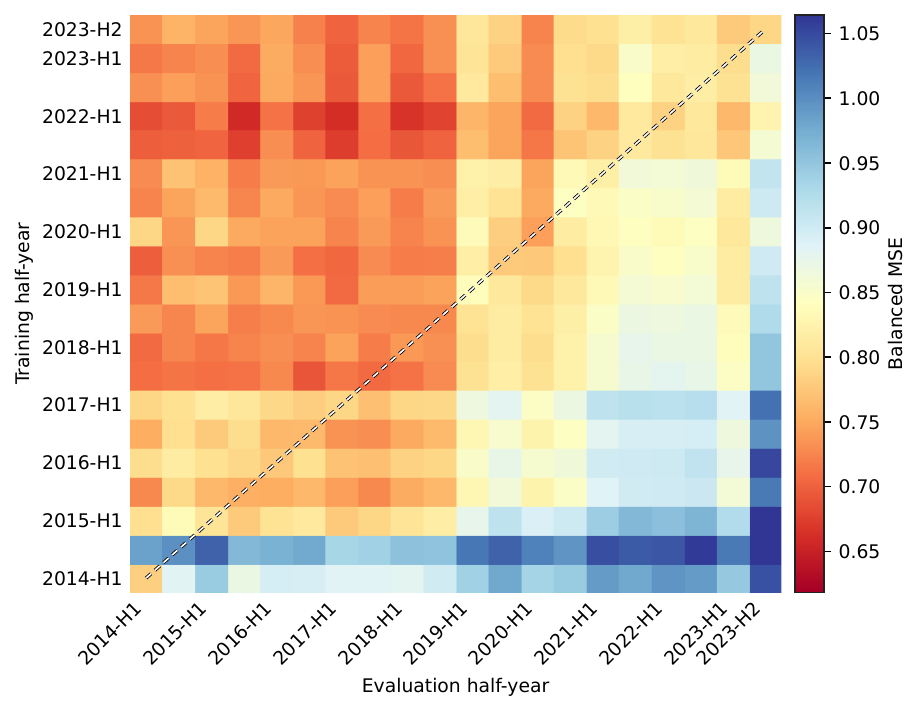}\hfill
    \includegraphics[width=0.49\linewidth]{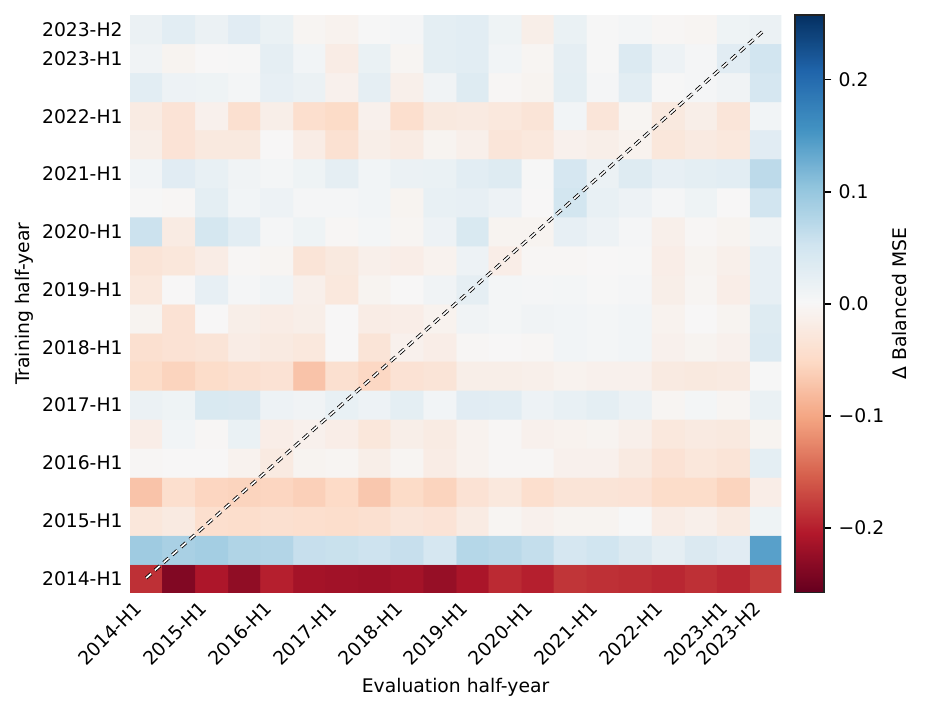}
    \caption{TextCNN-L}
    \label{fig:amazon_reviews_TextCNN_L}
\end{subfigure}
\caption{TextCNN models: Balanced MSE drift matrix $M^{(m)}$ and deviation from the cohort mean $\Delta^{(m)} = M^{(m)} - \bar{M}$ for each model, shown on a sequential and a zero-centred diverging scale, respectively.}
\label{fig:amazon_reviews_family_text_textcnn_regression}
\end{figure}

\subsection{Recurrent}

\begin{figure}[H]
\centering
\begin{subfigure}[t]{0.49\textwidth}\centering
    \includegraphics[width=0.49\linewidth]{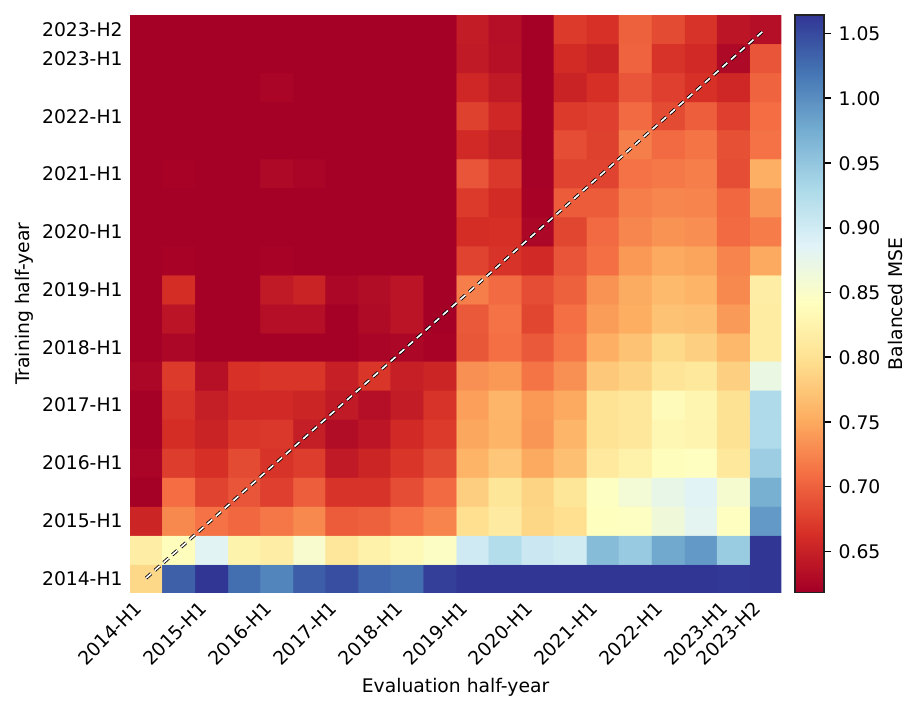}\hfill
    \includegraphics[width=0.49\linewidth]{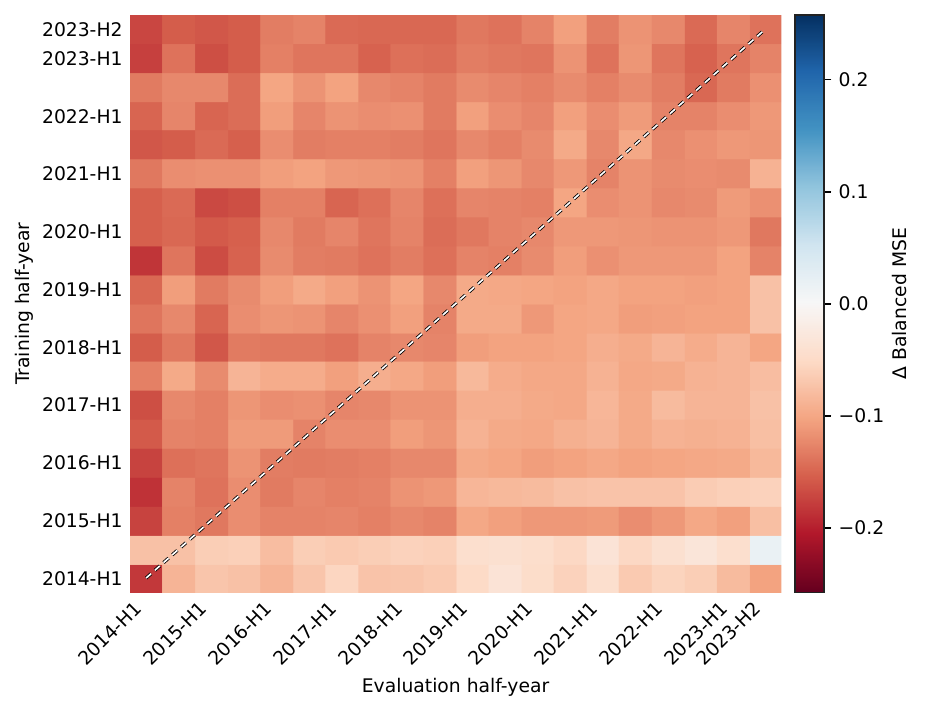}
    \caption{BiGRU-S}
    \label{fig:amazon_reviews_BiGRU_S}
\end{subfigure}
\hfill
\begin{subfigure}[t]{0.49\textwidth}\centering
    \includegraphics[width=0.49\linewidth]{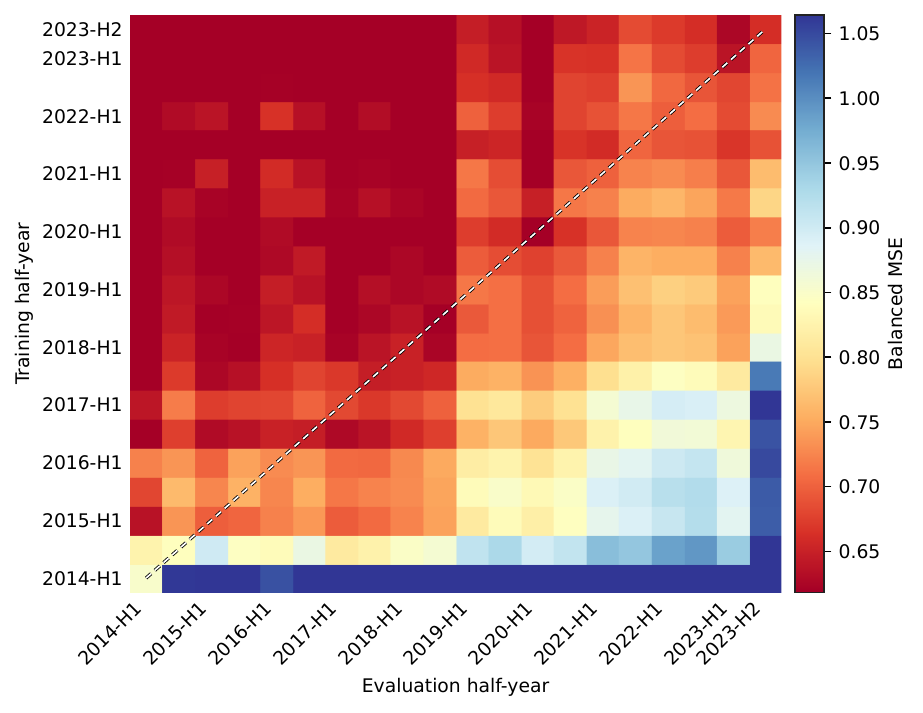}\hfill
    \includegraphics[width=0.49\linewidth]{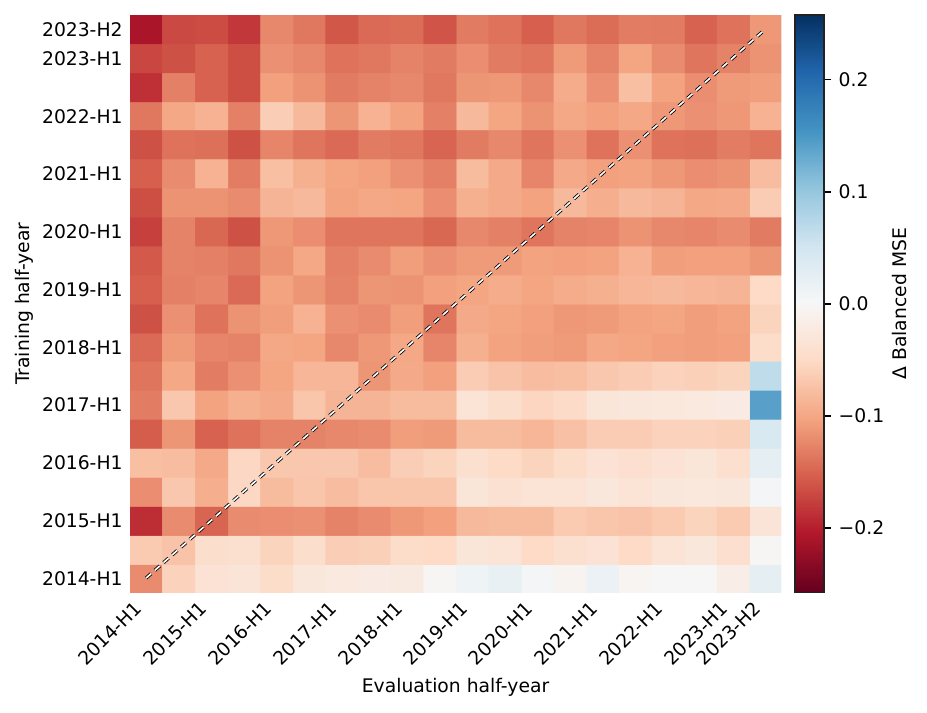}
    \caption{BiLSTM-M}
    \label{fig:amazon_reviews_BiLSTM_M}
\end{subfigure}

\begin{subfigure}[t]{0.49\textwidth}\centering
    \includegraphics[width=0.49\linewidth]{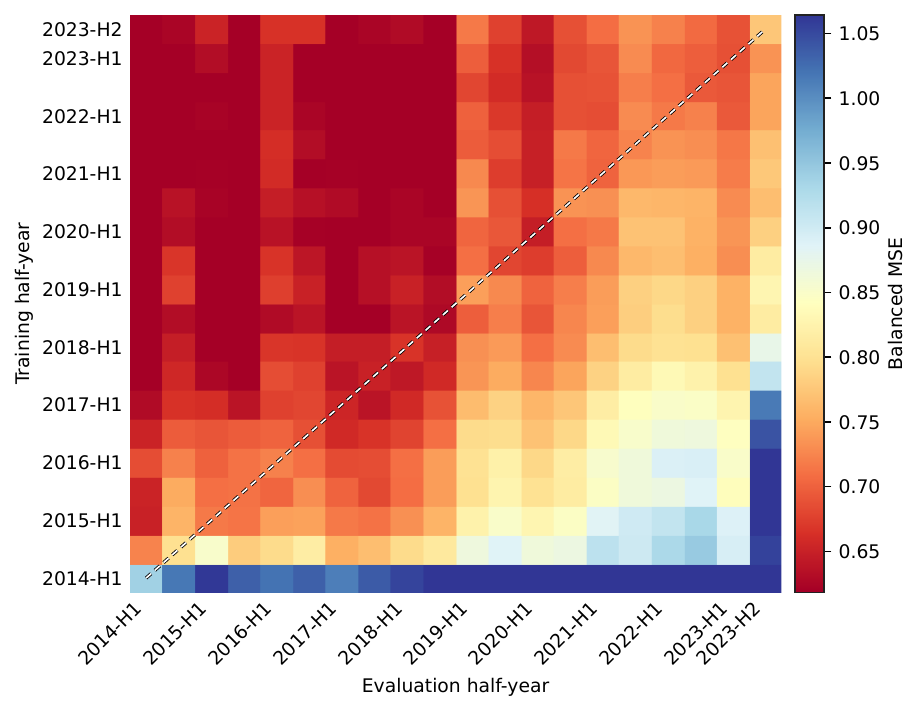}\hfill
    \includegraphics[width=0.49\linewidth]{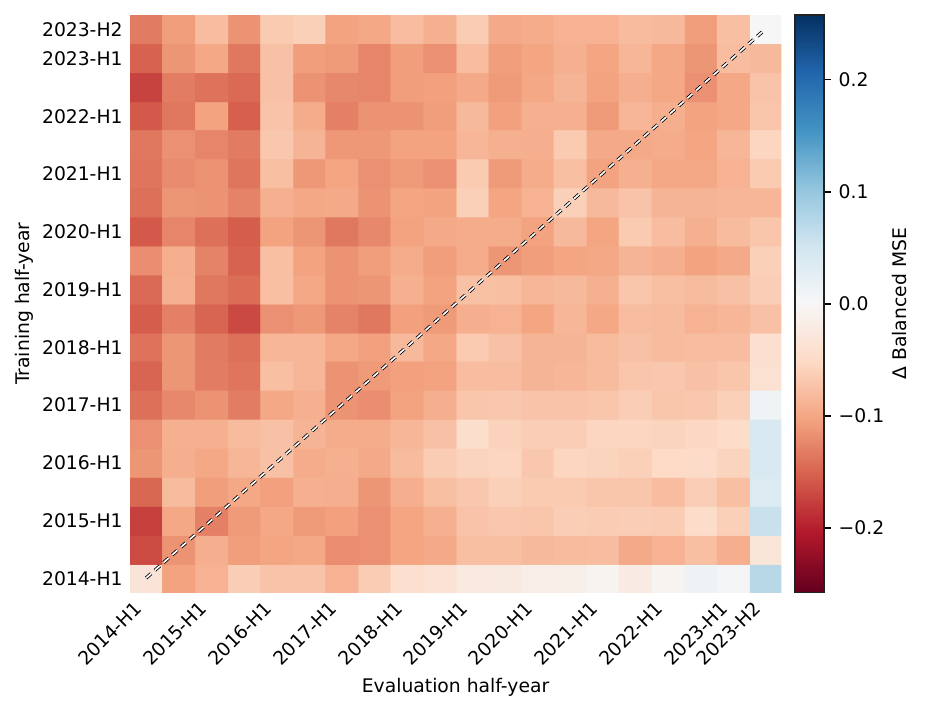}
    \caption{BiLSTM-Attn-L}
    \label{fig:amazon_reviews_BiLSTM_Attn_L}
\end{subfigure}
\caption{Recurrent models: Balanced MSE drift matrix $M^{(m)}$ and deviation from the cohort mean $\Delta^{(m)} = M^{(m)} - \bar{M}$ for each model, shown on a sequential and a zero-centred diverging scale, respectively.}
\label{fig:amazon_reviews_family_text_rnn_regression}
\end{figure}

\subsection{Transformer}

\begin{figure}[H]
\centering
\begin{subfigure}[t]{0.49\textwidth}\centering
    \includegraphics[width=0.49\linewidth]{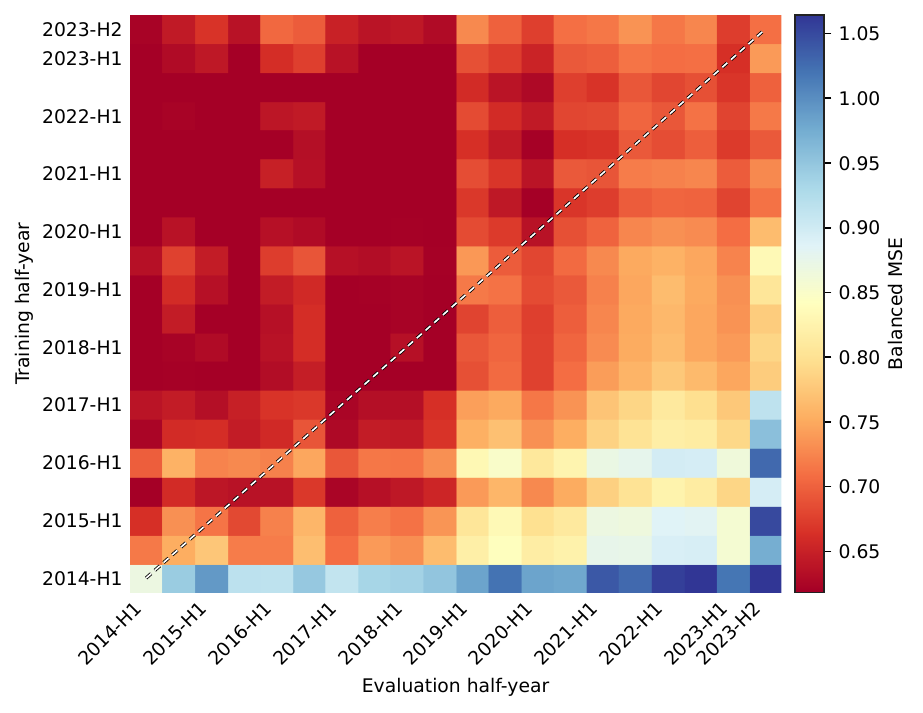}\hfill
    \includegraphics[width=0.49\linewidth]{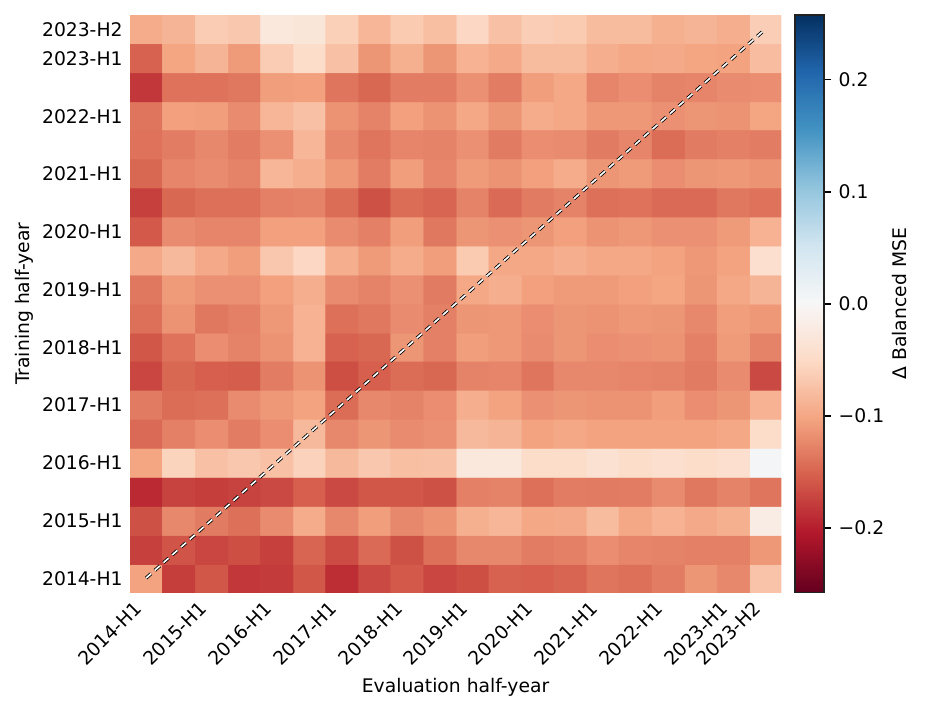}
    \caption{TX-S}
    \label{fig:amazon_reviews_TX_S}
\end{subfigure}
\hfill
\begin{subfigure}[t]{0.49\textwidth}\centering
    \includegraphics[width=0.49\linewidth]{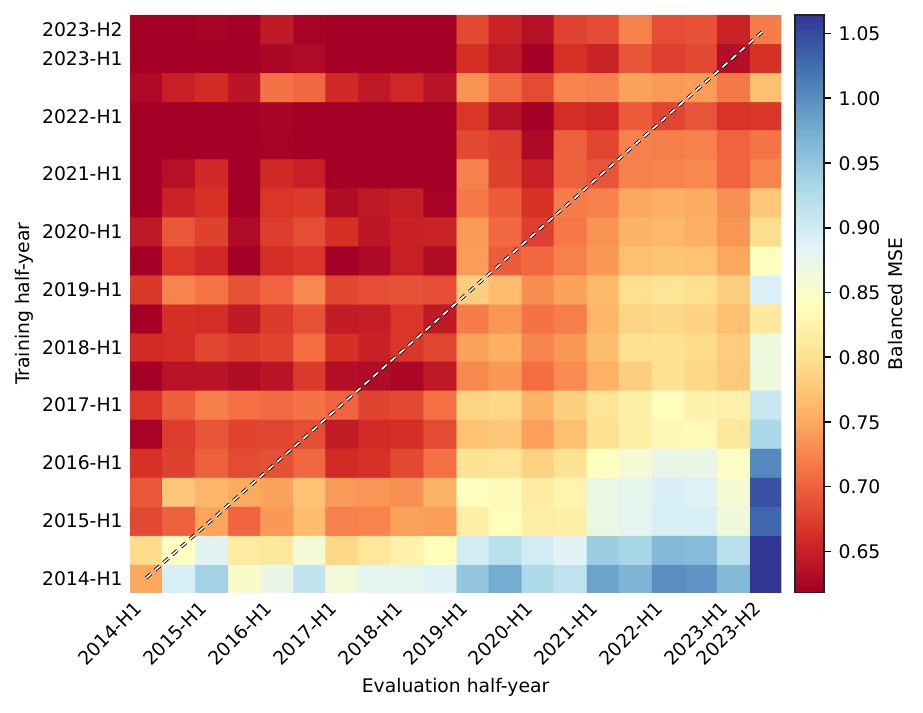}\hfill
    \includegraphics[width=0.49\linewidth]{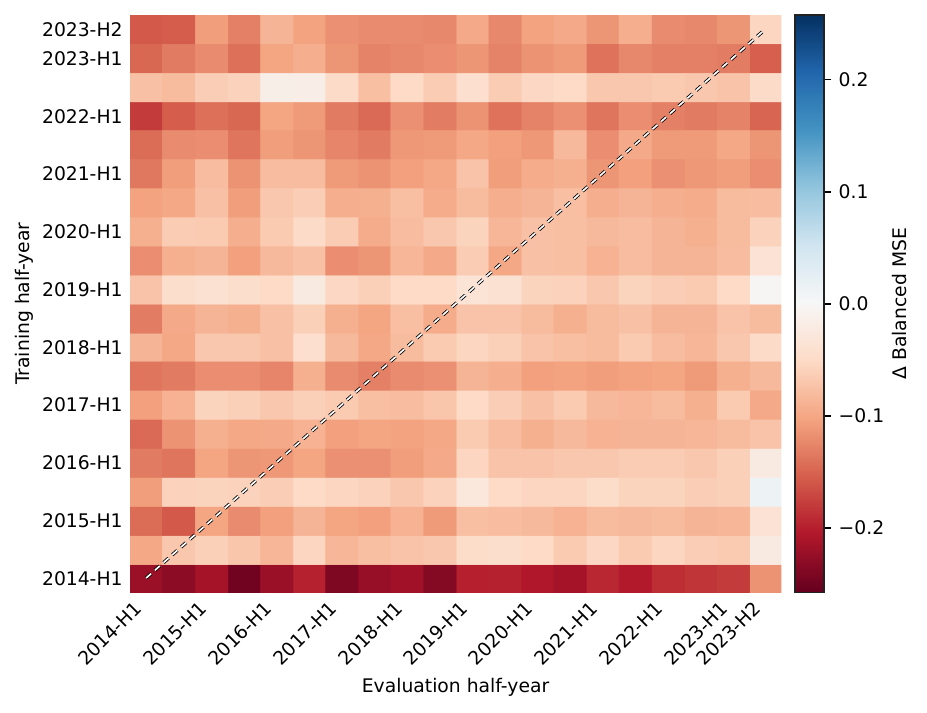}
    \caption{TX-M}
    \label{fig:amazon_reviews_TX_M}
\end{subfigure}

\begin{subfigure}[t]{0.49\textwidth}\centering
    \includegraphics[width=0.49\linewidth]{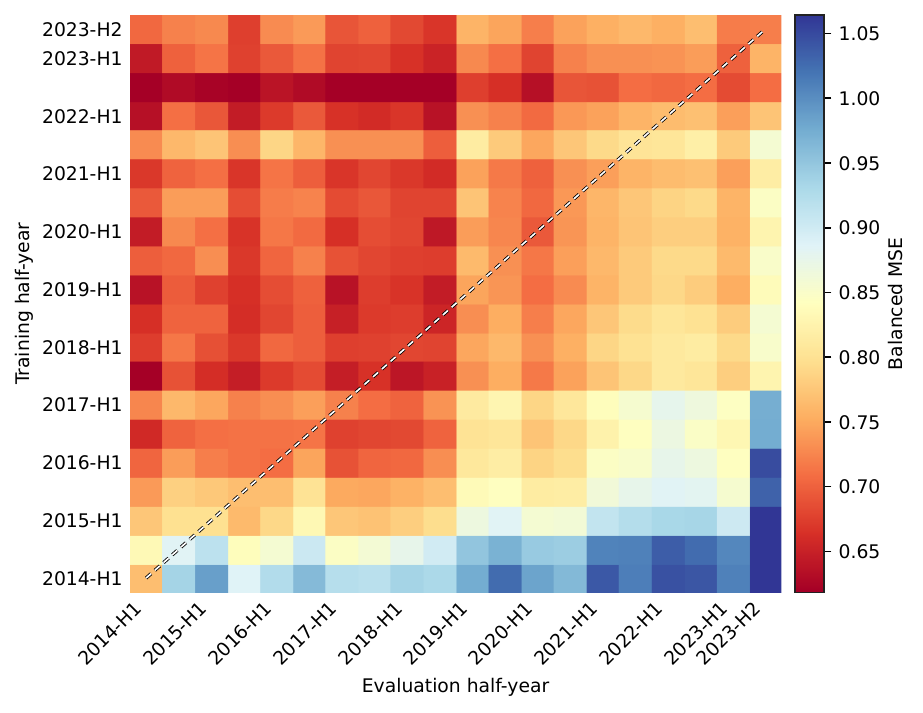}\hfill
    \includegraphics[width=0.49\linewidth]{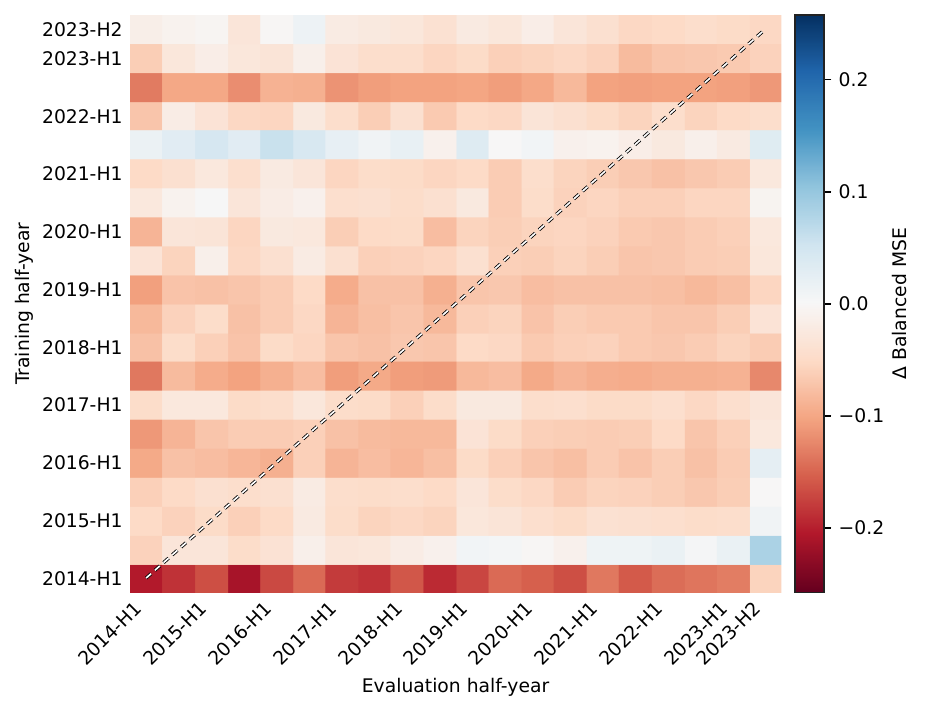}
    \caption{TX-L}
    \label{fig:amazon_reviews_TX_L}
\end{subfigure}
\caption{Transformer models: Balanced MSE drift matrix $M^{(m)}$ and deviation from the cohort mean $\Delta^{(m)} = M^{(m)} - \bar{M}$ for each model, shown on a sequential and a zero-centred diverging scale, respectively.}
\label{fig:amazon_reviews_family_text_tx_regression}
\end{figure}

\subsection{Frozen}

\begin{figure}[H]
\centering
\begin{subfigure}[t]{0.49\textwidth}\centering
    \includegraphics[width=0.49\linewidth]{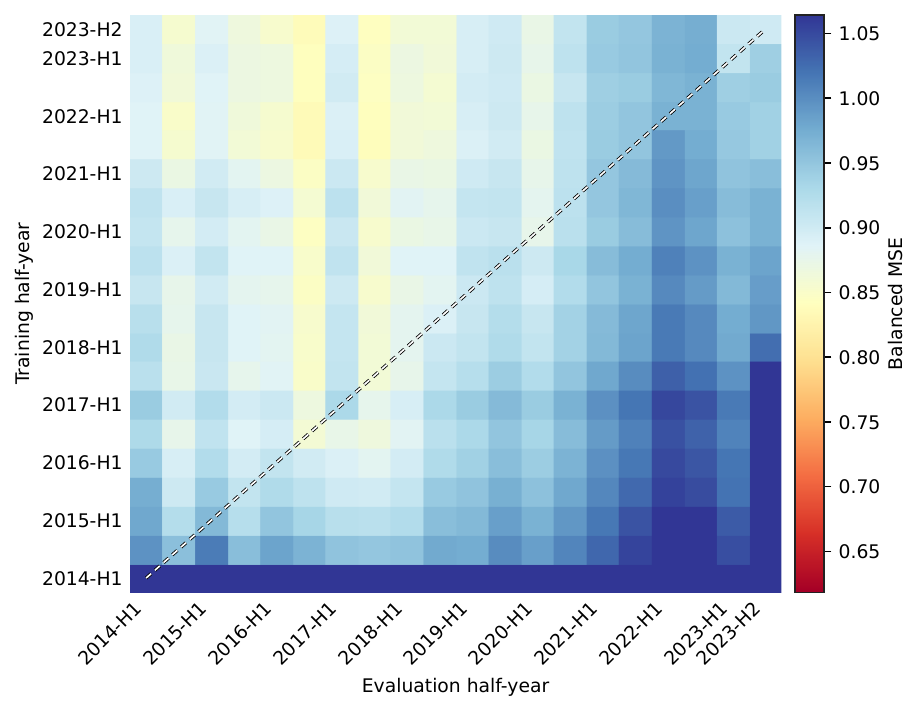}\hfill
    \includegraphics[width=0.49\linewidth]{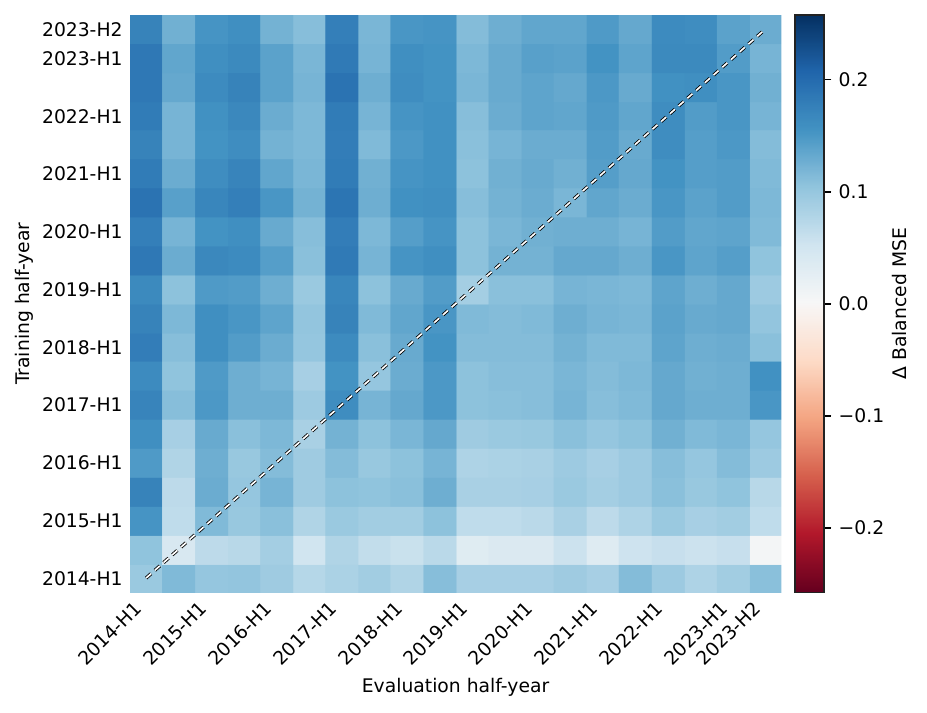}
    \caption{BERT}
    \label{fig:amazon_reviews_BERT}
\end{subfigure}
\hfill
\begin{subfigure}[t]{0.49\textwidth}\centering
    \includegraphics[width=0.49\linewidth]{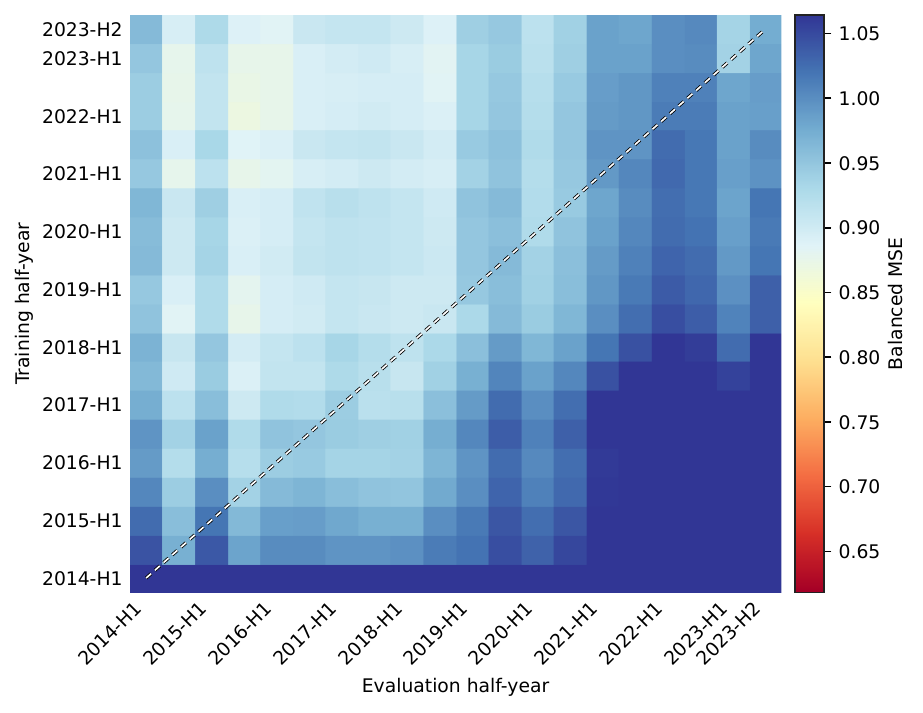}\hfill
    \includegraphics[width=0.49\linewidth]{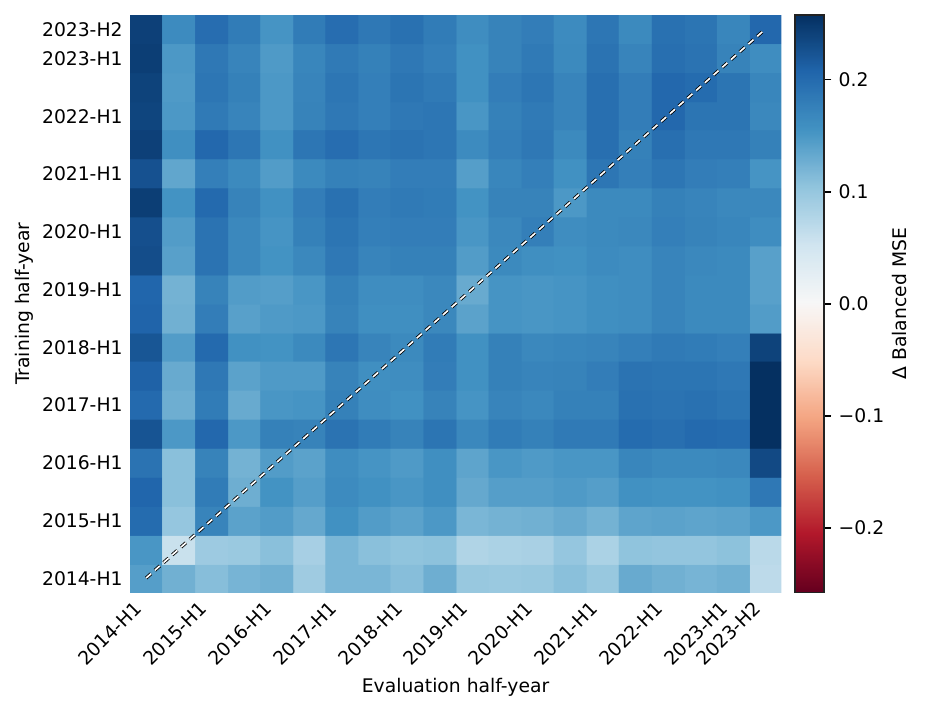}
    \caption{DistilBERT}
    \label{fig:amazon_reviews_DistilBERT}
\end{subfigure}

\begin{subfigure}[t]{0.49\textwidth}\centering
    \includegraphics[width=0.49\linewidth]{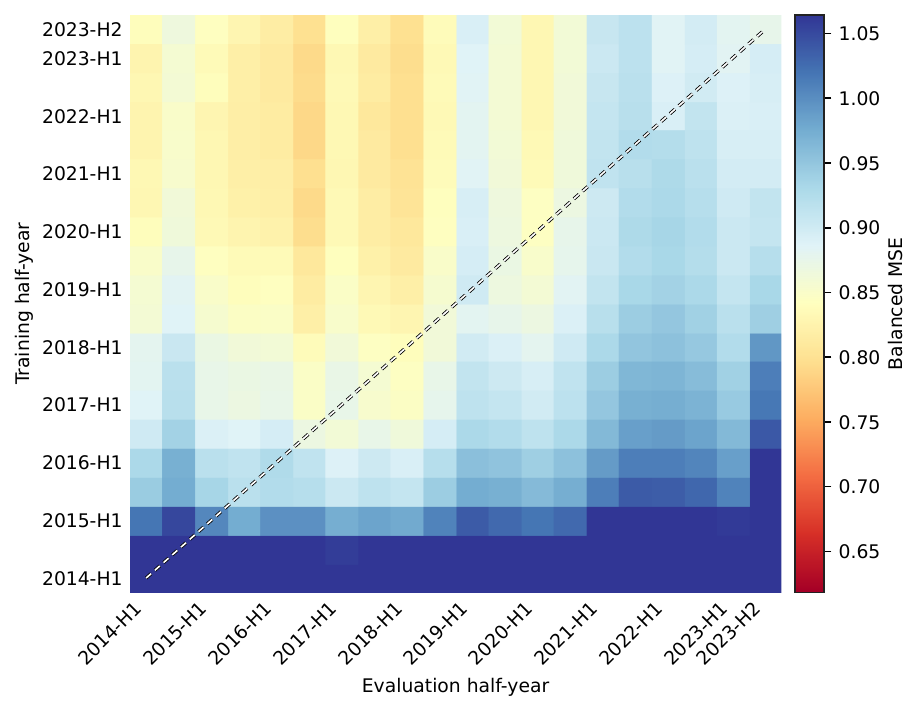}\hfill
    \includegraphics[width=0.49\linewidth]{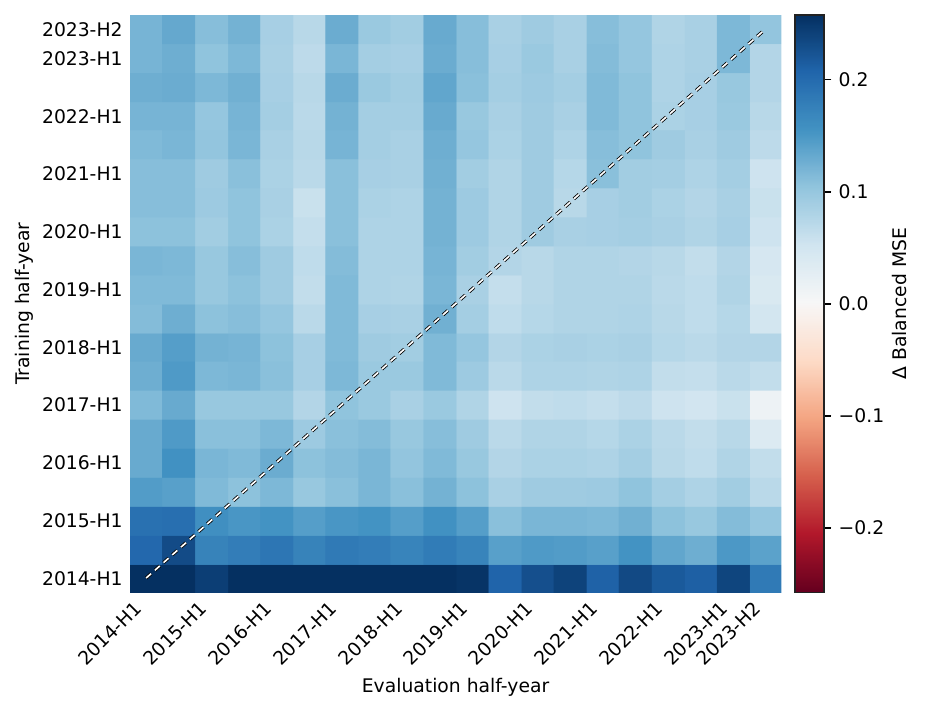}
    \caption{ELECTRA}
    \label{fig:amazon_reviews_ELECTRA}
\end{subfigure}
\hfill
\begin{subfigure}[t]{0.49\textwidth}\centering
    \includegraphics[width=0.49\linewidth]{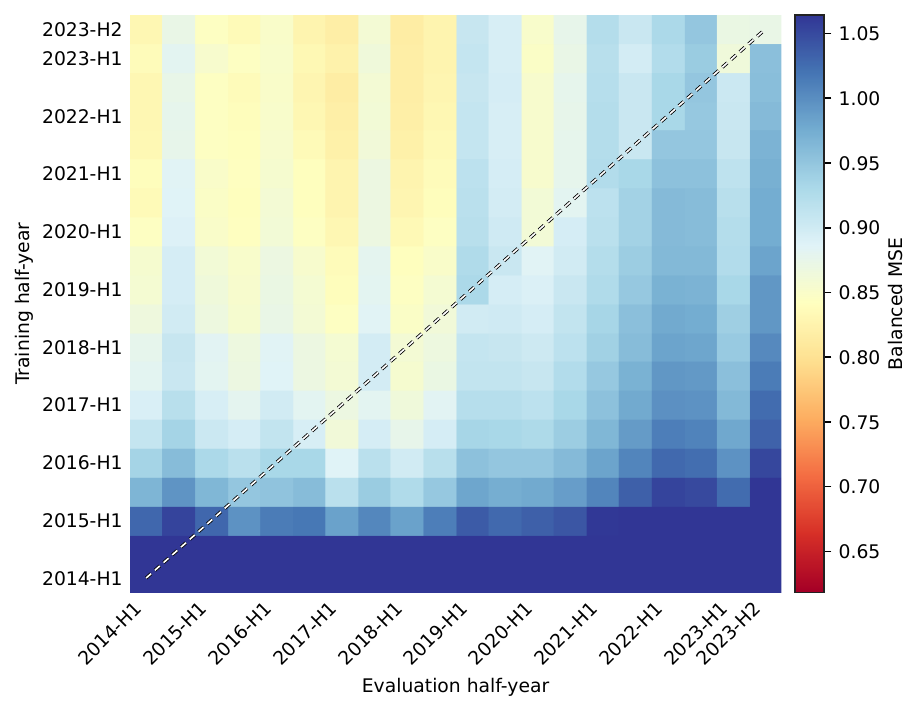}\hfill
    \includegraphics[width=0.49\linewidth]{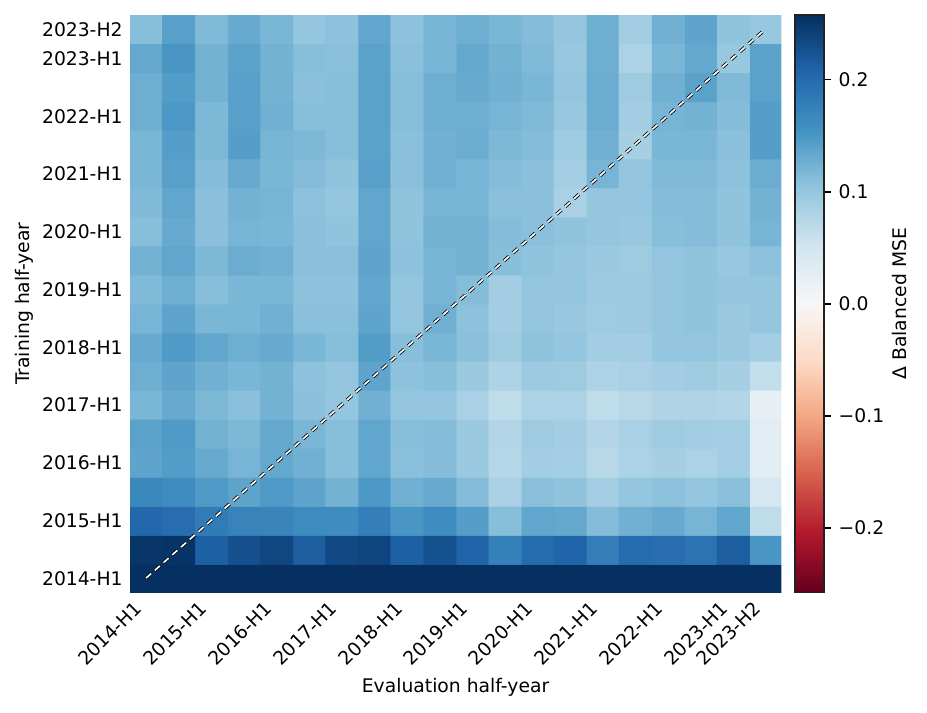}
    \caption{MPNet}
    \label{fig:amazon_reviews_MPNet}
\end{subfigure}

\begin{subfigure}[t]{0.49\textwidth}\centering
    \includegraphics[width=0.49\linewidth]{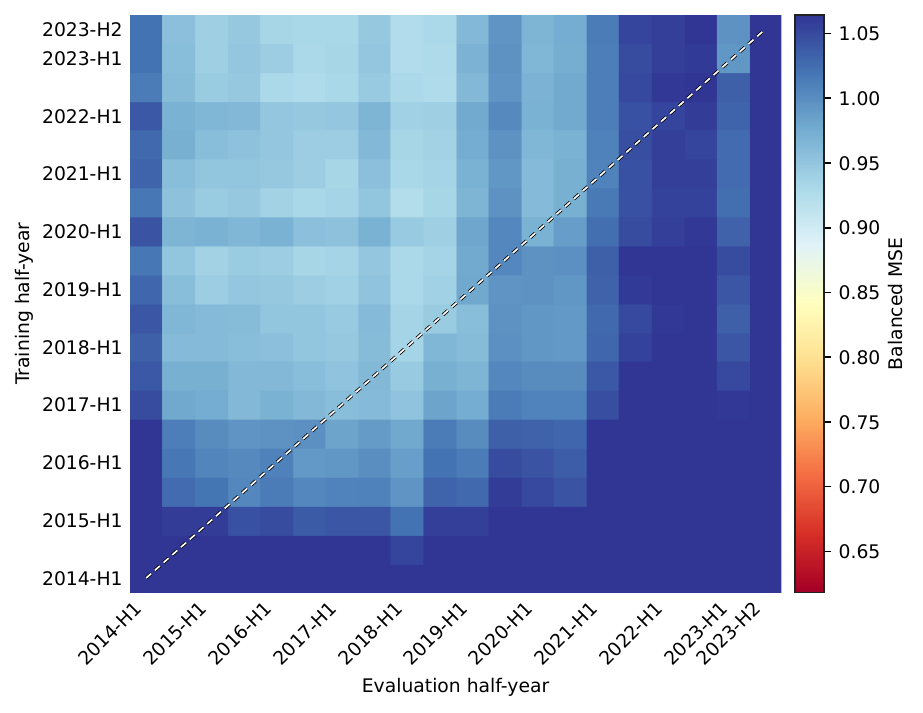}\hfill
    \includegraphics[width=0.49\linewidth]{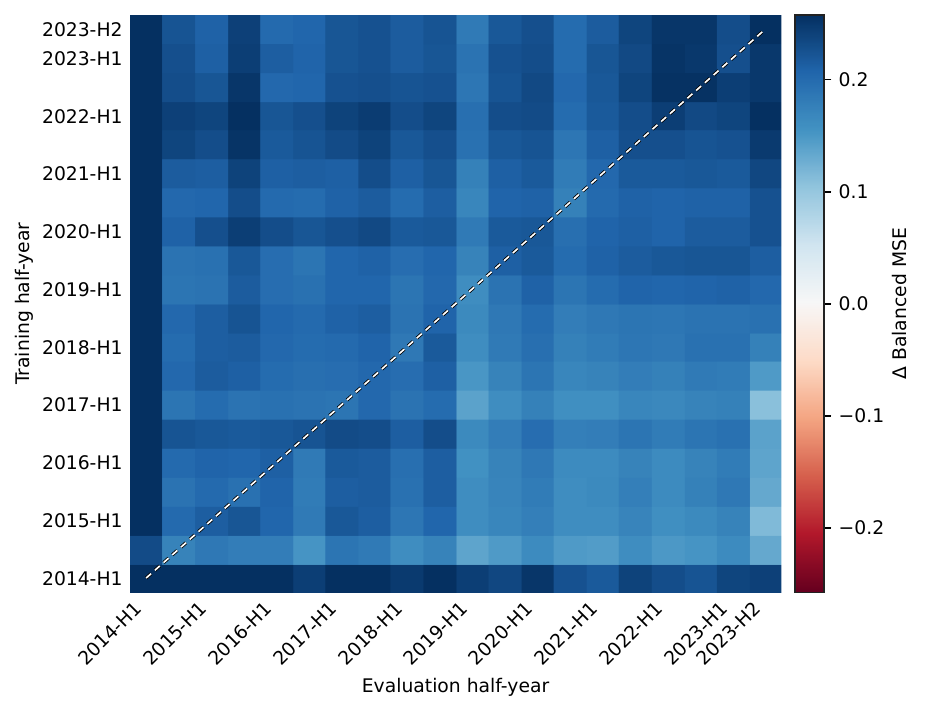}
    \caption{ModernBERT}
    \label{fig:amazon_reviews_ModernBERT}
\end{subfigure}
\hfill
\begin{subfigure}[t]{0.49\textwidth}\centering
    \includegraphics[width=0.49\linewidth]{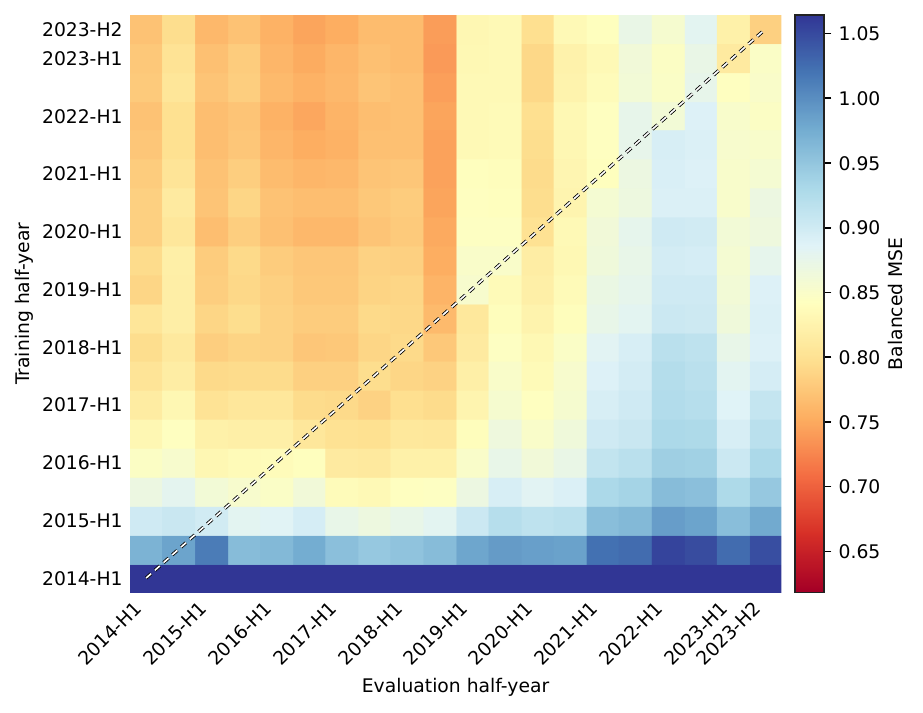}\hfill
    \includegraphics[width=0.49\linewidth]{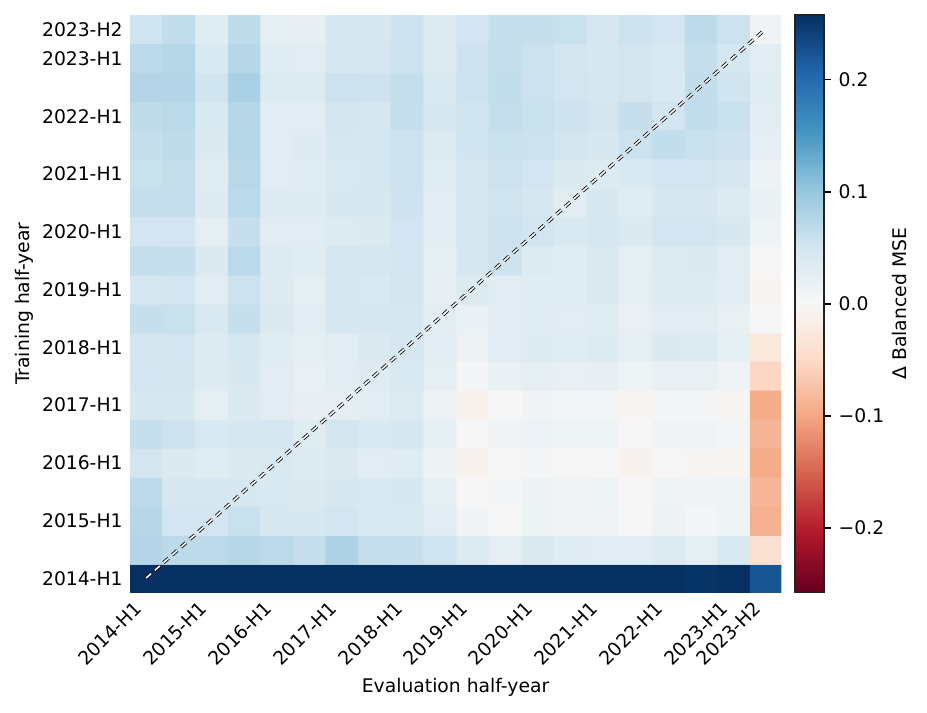}
    \caption{RoBERTa}
    \label{fig:amazon_reviews_RoBERTa}
\end{subfigure}

\begin{subfigure}[t]{0.49\textwidth}\centering
    \includegraphics[width=0.49\linewidth]{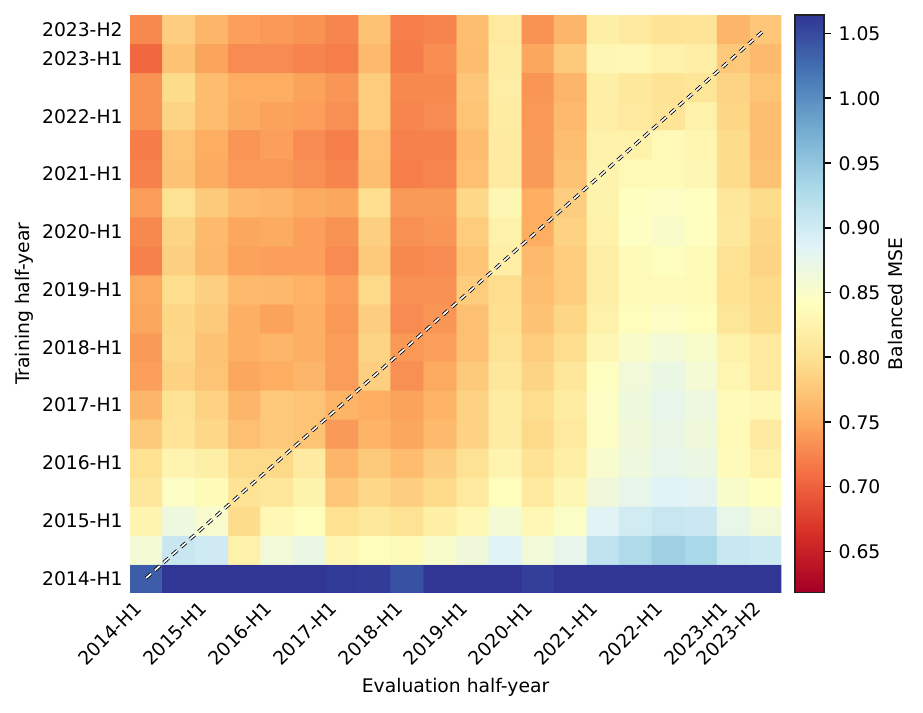}\hfill
    \includegraphics[width=0.49\linewidth]{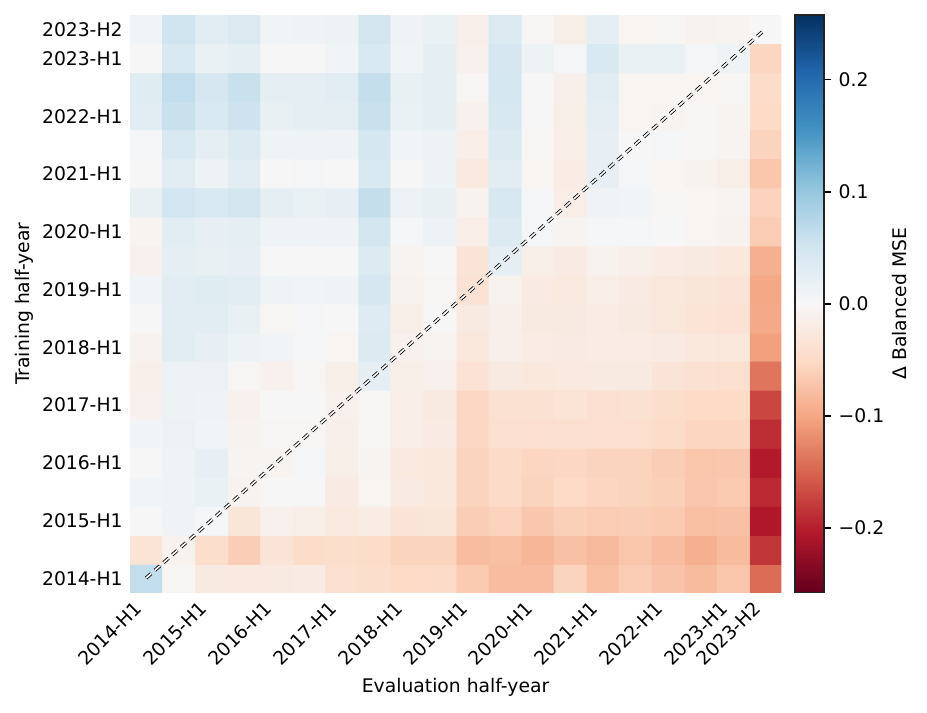}
    \caption{DeBERTa-v3}
    \label{fig:amazon_reviews_DeBERTa_v3}
\end{subfigure}
\caption{Frozen models: Balanced MSE drift matrix $M^{(m)}$ and deviation from the cohort mean $\Delta^{(m)} = M^{(m)} - \bar{M}$ for each model, shown on a sequential and a zero-centred diverging scale, respectively.}
\label{fig:amazon_reviews_family_text_frozen_head_regression}
\end{figure}

\subsection{Forgetting and Rankings}
\label{app:amazon_reviews_forgetting}

\noindent To see how quickly each model forgets, we summarize its drift matrix as a forgetting curve. The curve plots the Balanced MSE against the lag $\ell = j - i$, the number of slices between the training cutoff $i$ and the evaluation slice $j$. At each lag we average over all training cutoffs,
\[ F(\ell) = \operatorname{mean}_{i} M_{i,\,i+\ell}. \]
The result is the Balanced MSE at a fixed temporal distance, independent of which period a model was trained on. This separates the effect of temporal distance from the difficulty of any single slice.

\begin{figure}[H]
\centering
\begin{subfigure}[t]{0.49\textwidth}\centering
    \includegraphics[width=\linewidth]{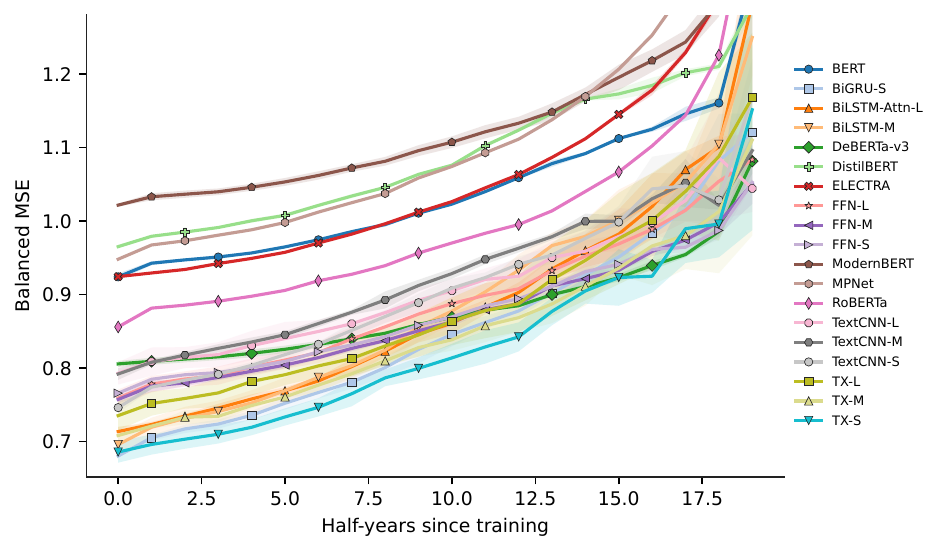}
    \caption{Per model}
    \label{fig:amazon_reviews_forgetting}
\end{subfigure}\hfill
\begin{subfigure}[t]{0.49\textwidth}\centering
    \includegraphics[width=\linewidth]{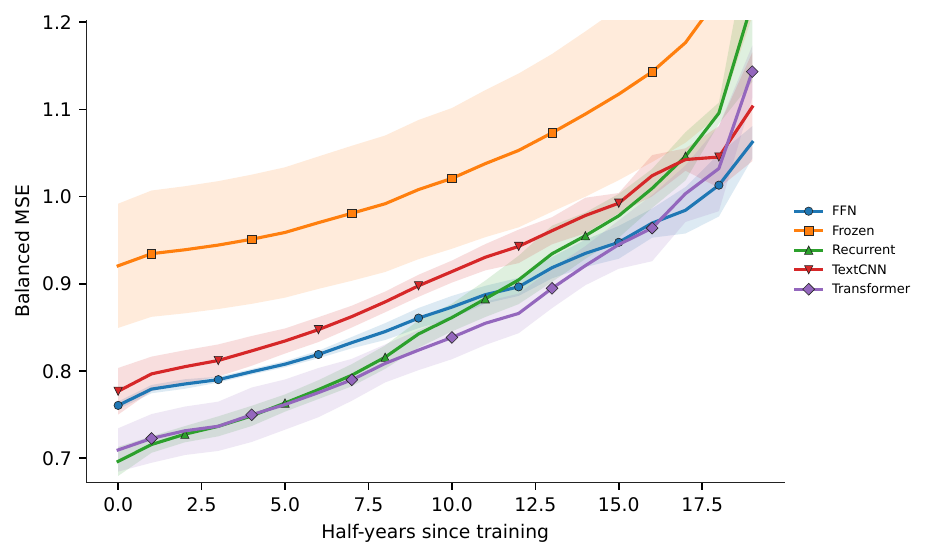}
    \caption{Per model family}
    \label{fig:amazon_reviews_forgetting_family}
\end{subfigure}
\caption{Forgetting curves: each model (left) and averaged within each family (right).}
\label{fig:amazon_reviews_forgetting_combined}
\end{figure}

\begin{figure}[H]
\centering
\begin{subfigure}[t]{0.49\textwidth}\centering
    \includegraphics[width=\linewidth]{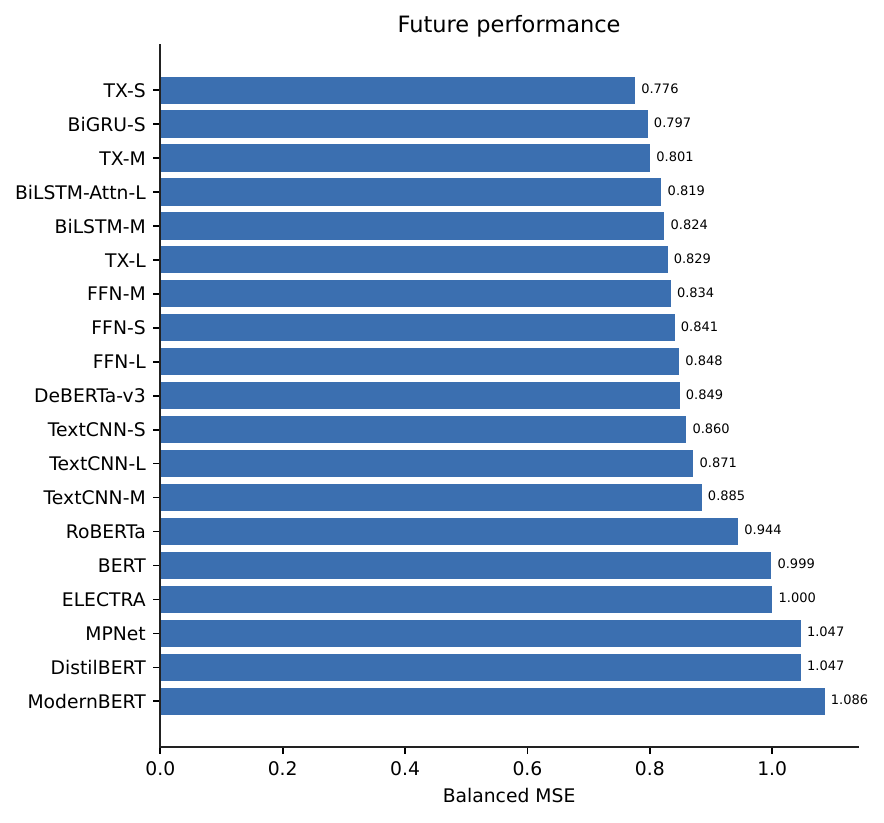}
    \caption{By future performance}
    \label{fig:amazon_reviews_ranking_future}
\end{subfigure}\hfill
\begin{subfigure}[t]{0.49\textwidth}\centering
    \includegraphics[width=\linewidth]{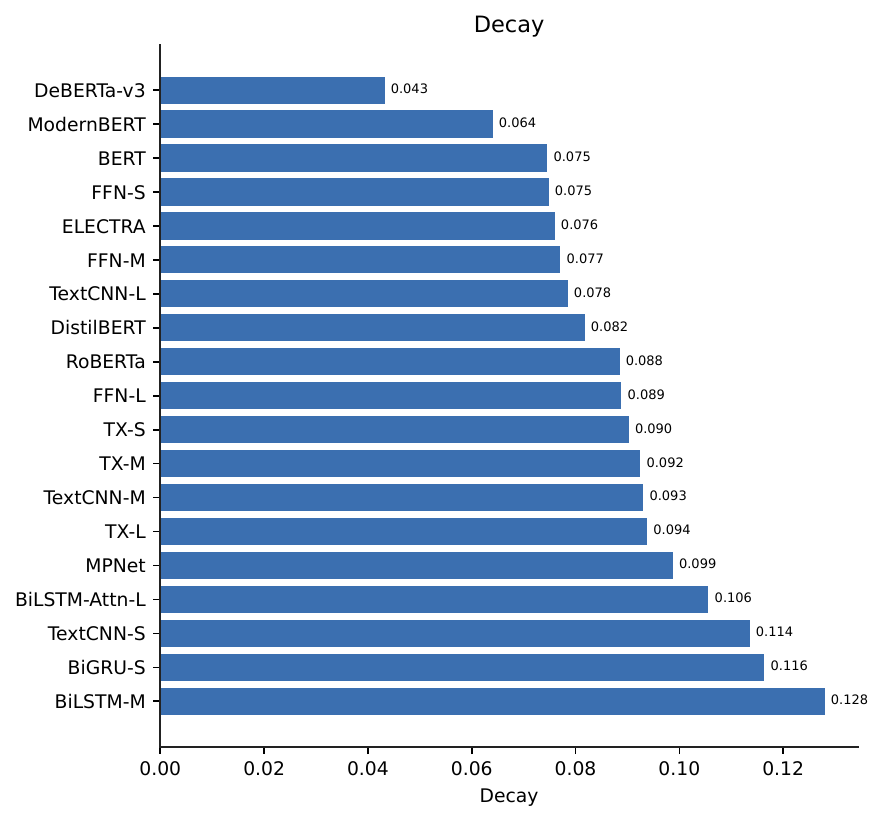}
    \caption{By decay}
    \label{fig:amazon_reviews_ranking_decay}
\end{subfigure}
\caption{Models ranked by mean future performance and by temporal decay.}
\label{fig:amazon_reviews_rankings}
\end{figure}

\subsection{Result Tables}

\begin{table}[H]
\centering
\footnotesize
\caption{Temporal robustness on Amazon Reviews.}
\label{tab:robustness_amazon_reviews}
\begin{minipage}[t]{0.49\linewidth}
\centering
\resizebox{\ifdim\width>\linewidth \linewidth\else\width\fi}{!}{%
\begin{tabular}{l rr}
\toprule
Model & Future & Decay \\
\midrule
DeBERTa-v3 & 0.849 & 0.043 \\
ModernBERT & 1.086 & 0.064 \\
BERT & 0.999 & 0.075 \\
FFN-S & 0.841 & 0.075 \\
ELECTRA & 1.000 & 0.076 \\
FFN-M & 0.834 & 0.077 \\
TextCNN-L & 0.871 & 0.078 \\
DistilBERT & 1.047 & 0.082 \\
RoBERTa & 0.944 & 0.088 \\
FFN-L & 0.848 & 0.089 \\
\bottomrule
\end{tabular}}
\end{minipage}\hfill
\begin{minipage}[t]{0.49\linewidth}
\centering
\resizebox{\ifdim\width>\linewidth \linewidth\else\width\fi}{!}{%
\begin{tabular}{l rr}
\toprule
Model & Future & Decay \\
\midrule
TX-S & 0.776 & 0.090 \\
TX-M & 0.801 & 0.092 \\
TextCNN-M & 0.885 & 0.093 \\
TX-L & 0.829 & 0.094 \\
MPNet & 1.047 & 0.099 \\
BiLSTM-Attn-L & 0.819 & 0.106 \\
TextCNN-S & 0.860 & 0.114 \\
BiGRU-S & 0.797 & 0.116 \\
BiLSTM-M & 0.824 & 0.128 \\
\bottomrule
\end{tabular}}
\end{minipage}
\end{table}

\noindent
\begin{minipage}[t]{0.49\linewidth}
\centering
\scriptsize
\setlength{\tabcolsep}{4pt}
\captionof{table}{Amazon Reviews: models trained up to 2014-H1, ordered by future performance.}
\label{tab:amazon_reviews_cutoff_28}
\resizebox{\ifdim\width>\linewidth \linewidth\else\width\fi}{!}{%
\begin{tabular}{c l rrr}
\toprule
Rank & Model & Balanced MSE & Future & Decay \\
\midrule
1 & TX-M & 0.748 & 0.933 & 0.185 \\
2 & TextCNN-L & 0.782 & 0.935 & 0.153 \\
3 & FFN-L & 0.836 & 0.972 & 0.136 \\
4 & TX-L & 0.767 & 0.981 & 0.215 \\
5 & FFN-M & 0.885 & 0.986 & 0.100 \\
6 & TX-S & 0.868 & 0.987 & 0.119 \\
7 & TextCNN-M & 0.761 & 0.993 & 0.231 \\
8 & FFN-S & 0.907 & 1.023 & 0.116 \\
9 & BiGRU-S & 0.790 & 1.073 & 0.283 \\
10 & TextCNN-S & 0.782 & 1.078 & 0.296 \\
11 & DeBERTa-v3 & 1.035 & 1.083 & 0.048 \\
12 & BiLSTM-Attn-L & 0.938 & 1.105 & 0.167 \\
13 & BiLSTM-M & 0.850 & 1.127 & 0.277 \\
14 & BERT & 1.067 & 1.232 & 0.165 \\
15 & DistilBERT & 1.113 & 1.250 & 0.137 \\
16 & ELECTRA & 1.332 & 1.382 & 0.050 \\
17 & ModernBERT & 1.273 & 1.386 & 0.113 \\
18 & RoBERTa & 1.227 & 1.468 & 0.241 \\
19 & MPNet & 1.492 & 1.657 & 0.165 \\
\bottomrule
\end{tabular}}
\end{minipage}
\hfill
\begin{minipage}[t]{0.49\linewidth}
\centering
\scriptsize
\setlength{\tabcolsep}{4pt}
\captionof{table}{Amazon Reviews: models trained up to 2017-H1, ordered by future performance.}
\label{tab:amazon_reviews_cutoff_34}
\resizebox{\ifdim\width>\linewidth \linewidth\else\width\fi}{!}{%
\begin{tabular}{c l rrr}
\toprule
Rank & Model & Balanced MSE & Future & Decay \\
\midrule
1 & TX-S & 0.625 & 0.748 & 0.123 \\
2 & BiGRU-S & 0.644 & 0.765 & 0.121 \\
3 & TX-M & 0.701 & 0.785 & 0.084 \\
4 & BiLSTM-Attn-L & 0.654 & 0.789 & 0.136 \\
5 & FFN-L & 0.724 & 0.810 & 0.086 \\
6 & FFN-S & 0.723 & 0.812 & 0.089 \\
7 & DeBERTa-v3 & 0.758 & 0.814 & 0.056 \\
8 & TextCNN-S & 0.711 & 0.817 & 0.106 \\
9 & TX-L & 0.722 & 0.818 & 0.096 \\
10 & BiLSTM-M & 0.681 & 0.828 & 0.147 \\
11 & FFN-M & 0.742 & 0.829 & 0.087 \\
12 & RoBERTa & 0.792 & 0.860 & 0.068 \\
13 & TextCNN-L & 0.789 & 0.876 & 0.087 \\
14 & TextCNN-M & 0.773 & 0.885 & 0.113 \\
15 & ELECTRA & 0.873 & 0.926 & 0.052 \\
16 & MPNet & 0.868 & 0.940 & 0.072 \\
17 & BERT & 0.929 & 0.984 & 0.055 \\
18 & ModernBERT & 0.958 & 1.028 & 0.070 \\
19 & DistilBERT & 0.942 & 1.047 & 0.104 \\
\bottomrule
\end{tabular}}
\end{minipage}

\vspace{1.5ex}

\noindent
\begin{minipage}[t]{0.49\linewidth}
\centering
\scriptsize
\setlength{\tabcolsep}{4pt}
\captionof{table}{Amazon Reviews: models trained up to 2020-H1, ordered by future performance.}
\label{tab:amazon_reviews_cutoff_40}
\resizebox{\ifdim\width>\linewidth \linewidth\else\width\fi}{!}{%
\begin{tabular}{c l rrr}
\toprule
Rank & Model & Balanced MSE & Future & Decay \\
\midrule
1 & BiLSTM-M & 0.614 & 0.706 & 0.092 \\
2 & BiGRU-S & 0.627 & 0.715 & 0.088 \\
3 & TX-S & 0.637 & 0.721 & 0.084 \\
4 & BiLSTM-Attn-L & 0.647 & 0.749 & 0.102 \\
5 & TX-M & 0.675 & 0.751 & 0.076 \\
6 & TextCNN-S & 0.672 & 0.767 & 0.094 \\
7 & TX-L & 0.693 & 0.773 & 0.080 \\
8 & FFN-L & 0.713 & 0.793 & 0.080 \\
9 & FFN-S & 0.729 & 0.795 & 0.066 \\
10 & FFN-M & 0.744 & 0.817 & 0.074 \\
11 & DeBERTa-v3 & 0.753 & 0.820 & 0.067 \\
12 & TextCNN-M & 0.748 & 0.833 & 0.085 \\
13 & TextCNN-L & 0.743 & 0.834 & 0.091 \\
14 & RoBERTa & 0.800 & 0.870 & 0.070 \\
15 & ELECTRA & 0.844 & 0.911 & 0.067 \\
16 & MPNet & 0.860 & 0.938 & 0.078 \\
17 & BERT & 0.876 & 0.959 & 0.084 \\
18 & DistilBERT & 0.926 & 0.998 & 0.072 \\
19 & ModernBERT & 0.969 & 1.041 & 0.072 \\
\bottomrule
\end{tabular}}
\end{minipage}
\hfill
\begin{minipage}[t]{0.49\linewidth}
\centering
\scriptsize
\setlength{\tabcolsep}{4pt}
\captionof{table}{Amazon Reviews: models trained up to 2023-H1, ordered by future performance.}
\label{tab:amazon_reviews_cutoff_46}
\resizebox{\ifdim\width>\linewidth \linewidth\else\width\fi}{!}{%
\begin{tabular}{c l rrr}
\toprule
Rank & Model & Balanced MSE & Future & Decay \\
\midrule
1 & TX-M & 0.634 & 0.665 & 0.031 \\
2 & BiGRU-S & 0.627 & 0.691 & 0.064 \\
3 & BiLSTM-M & 0.638 & 0.703 & 0.065 \\
4 & BiLSTM-Attn-L & 0.687 & 0.735 & 0.048 \\
5 & TX-S & 0.663 & 0.740 & 0.077 \\
6 & TextCNN-S & 0.682 & 0.747 & 0.066 \\
7 & TX-L & 0.699 & 0.759 & 0.060 \\
8 & DeBERTa-v3 & 0.776 & 0.764 & -0.013 \\
9 & FFN-L & 0.728 & 0.770 & 0.042 \\
10 & FFN-S & 0.731 & 0.777 & 0.046 \\
11 & FFN-M & 0.742 & 0.821 & 0.079 \\
12 & TextCNN-M & 0.751 & 0.838 & 0.087 \\
13 & RoBERTa & 0.812 & 0.847 & 0.035 \\
14 & TextCNN-L & 0.794 & 0.869 & 0.075 \\
15 & ELECTRA & 0.882 & 0.894 & 0.012 \\
16 & BERT & 0.911 & 0.939 & 0.029 \\
17 & MPNet & 0.863 & 0.956 & 0.093 \\
18 & DistilBERT & 0.937 & 0.980 & 0.043 \\
19 & ModernBERT & 0.992 & 1.069 & 0.076 \\
\bottomrule
\end{tabular}}
\end{minipage}

\begin{table}[H]
\centering
\footnotesize
\caption{Amazon Reviews: future performance and decay by model family.}
\label{tab:amazon_reviews_by_family}
\begin{tabular}{l rrrrrrrr}
\toprule
 & \multicolumn{2}{c}{2014-H1} & \multicolumn{2}{c}{2017-H1} & \multicolumn{2}{c}{2020-H1} & \multicolumn{2}{c}{2023-H1} \\
Family & Future & Decay & Future & Decay & Future & Decay & Future & Decay \\
\midrule
FFN & 0.993 & 0.117 & 0.817 & 0.087 & 0.802 & 0.073 & 0.789 & 0.056 \\
Frozen & 1.351 & 0.131 & 0.943 & 0.068 & 0.934 & 0.073 & 0.921 & 0.039 \\
Recurrent & 1.101 & 0.242 & 0.794 & 0.134 & 0.723 & 0.094 & 0.710 & 0.059 \\
TextCNN & 1.002 & 0.227 & 0.859 & 0.102 & 0.811 & 0.090 & 0.818 & 0.076 \\
Transformer & 0.967 & 0.173 & 0.784 & 0.101 & 0.748 & 0.080 & 0.721 & 0.056 \\
\bottomrule
\end{tabular}
\end{table}

\pagebreak

\captionsetup[figure]{font=footnotesize,labelfont=footnotesize}

\section{arXiv -- Drift Matrices}
\label{app:arxiv}

\noindent The cohort-mean and per-model deviation matrices shown here, and the in-distribution, future, and decay quantities tabulated below, are defined in Section~\ref{subsec:drift_summary}. The label space comprises the leaf categories \texttt{cs.LG}, \texttt{hep-ph}, \texttt{cs.CV}, \texttt{cs.AI}, \texttt{hep-th}, \texttt{quant-ph}, and \texttt{gr-qc}.

\begin{figure}[H]
\centering
\includegraphics[width=0.5\textwidth]{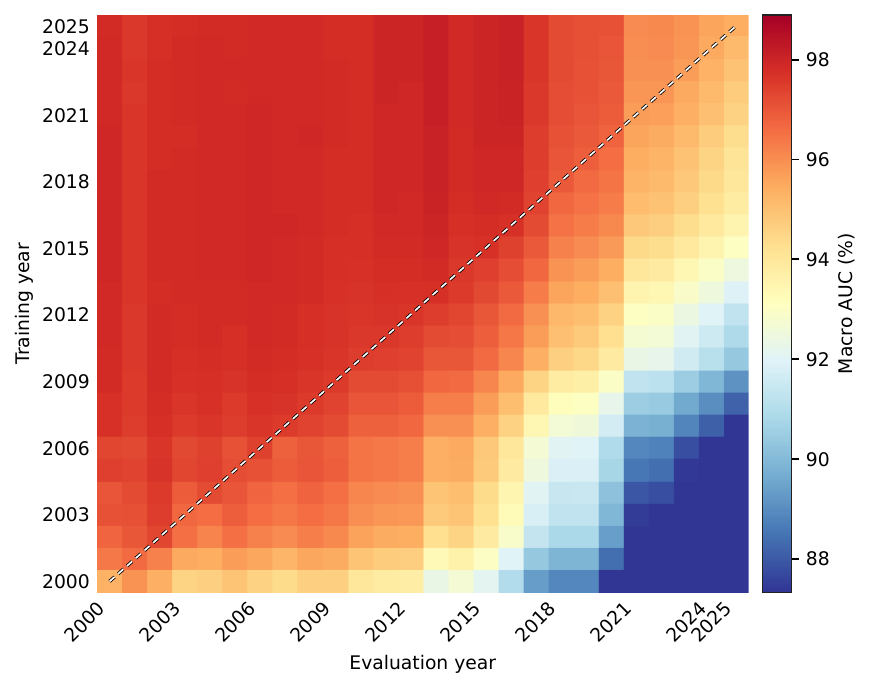}
\caption{Cohort-mean Macro AUC matrix $\bar{M}$ over the arXiv models. Cell $(i,j)$ is the mean across those models of the score from training through slice $i$ and evaluating on slice $j$.}
\label{fig:arxiv_mean_matrix}
\end{figure}

\subsection{Model Roster}

\noindent
\begin{minipage}[t]{0.49\linewidth}
\centering
\footnotesize
\captionof{table}{arXiv: models trained from scratch.}
\label{tab:arxiv_roster}
\resizebox{\ifdim\width>\linewidth \linewidth\else\width\fi}{!}{%
\begin{tabular}{l l rr}
\toprule
Model & Family & Trainable & Total \\
\midrule
TX-S & Transformer & 83k & 124.7M \\
TextCNN-S & TextCNN & 89k & 124.7M \\
FFN-S & FFN & 99k & 124.7M \\
BiGRU-S & Recurrent & 100k & 124.7M \\
FFN-M & FFN & 397k & 125.0M \\
TextCNN-M & TextCNN & 462k & 125.1M \\
TX-M & Transformer & 493k & 125.1M \\
BiLSTM-M & Recurrent & 537k & 125.2M \\
FFN-L & FFN & 1.6M & 126.2M \\
TX-L & Transformer & 1.9M & 126.5M \\
TextCNN-L & TextCNN & 1.9M & 126.6M \\
BiLSTM-Attn-L & Recurrent & 2.2M & 126.8M \\
\bottomrule
\end{tabular}}
\end{minipage}
\hfill
\begin{minipage}[t]{0.49\linewidth}
\centering
\footnotesize
\captionof{table}{arXiv: frozen pretrained encoders, with trainable head and total parameters.}
\label{tab:arxiv_roster_frozen}
\resizebox{\ifdim\width>\linewidth \linewidth\else\width\fi}{!}{%
\begin{tabular}{l l rr}
\toprule
Model & Family & Trainable & Total \\
\midrule
MiniLM-L6 & Frozen & 3k & 22.7M \\
DistilBERT & Frozen & 5k & 66.4M \\
ELECTRA & Frozen & 5k & 108.9M \\
BERT & Frozen & 5k & 109.5M \\
MPNet & Frozen & 5k & 109.5M \\
RoBERTa & Frozen & 5k & 124.7M \\
ModernBERT & Frozen & 5k & 149.0M \\
DeBERTa-v3 & Frozen & 5k & 183.8M \\
\bottomrule
\end{tabular}}
\end{minipage}

\subsection{FFN}

\begin{figure}[H]
\centering
\begin{subfigure}[t]{0.49\textwidth}\centering
    \includegraphics[width=0.49\linewidth]{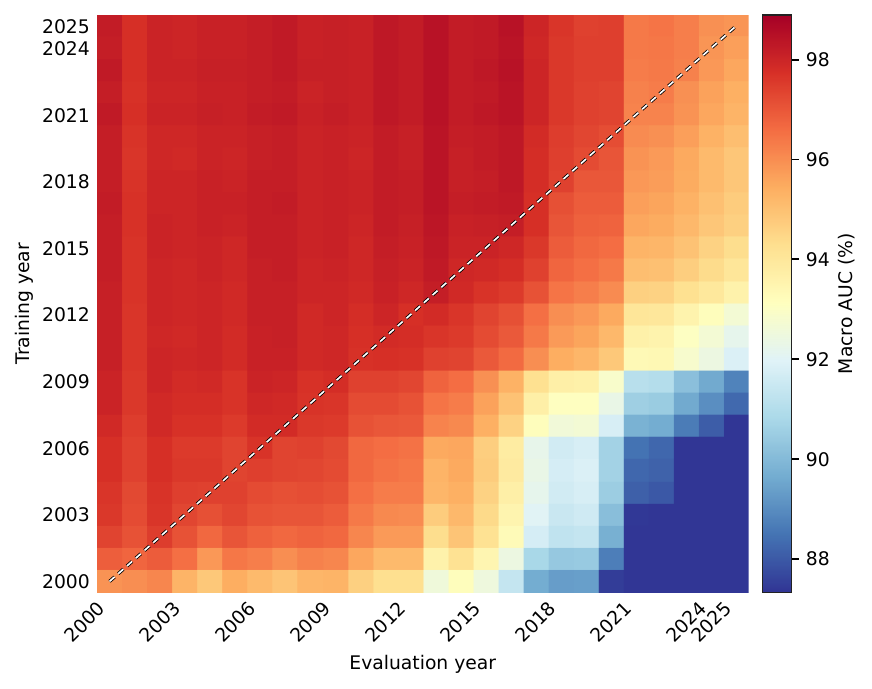}\hfill
    \includegraphics[width=0.49\linewidth]{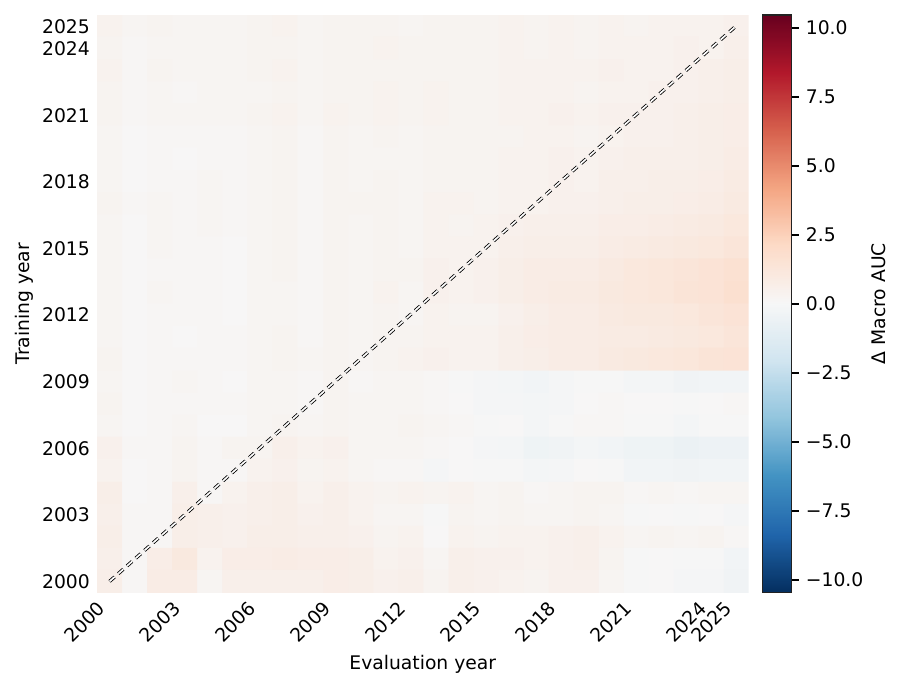}
    \caption{FFN-S}
    \label{fig:arxiv_FFN_S}
\end{subfigure}
\hfill
\begin{subfigure}[t]{0.49\textwidth}\centering
    \includegraphics[width=0.49\linewidth]{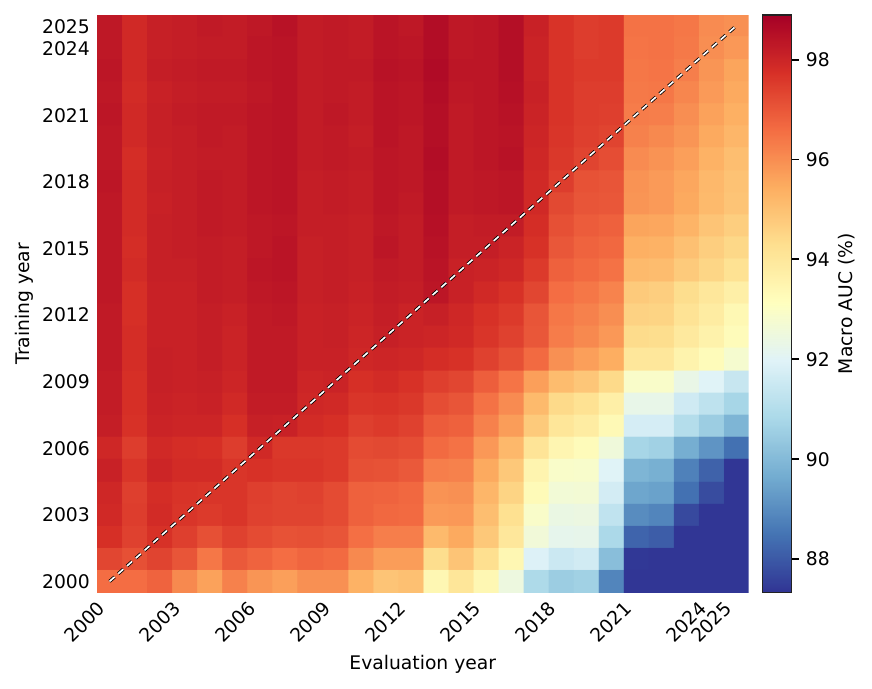}\hfill
    \includegraphics[width=0.49\linewidth]{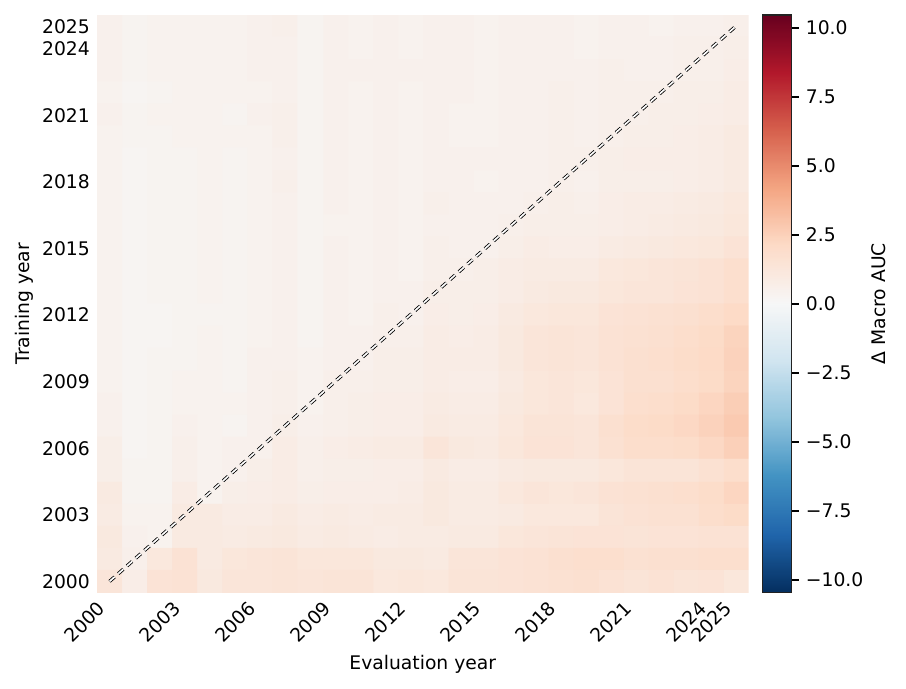}
    \caption{FFN-M}
    \label{fig:arxiv_FFN_M}
\end{subfigure}

\begin{subfigure}[t]{0.49\textwidth}\centering
    \includegraphics[width=0.49\linewidth]{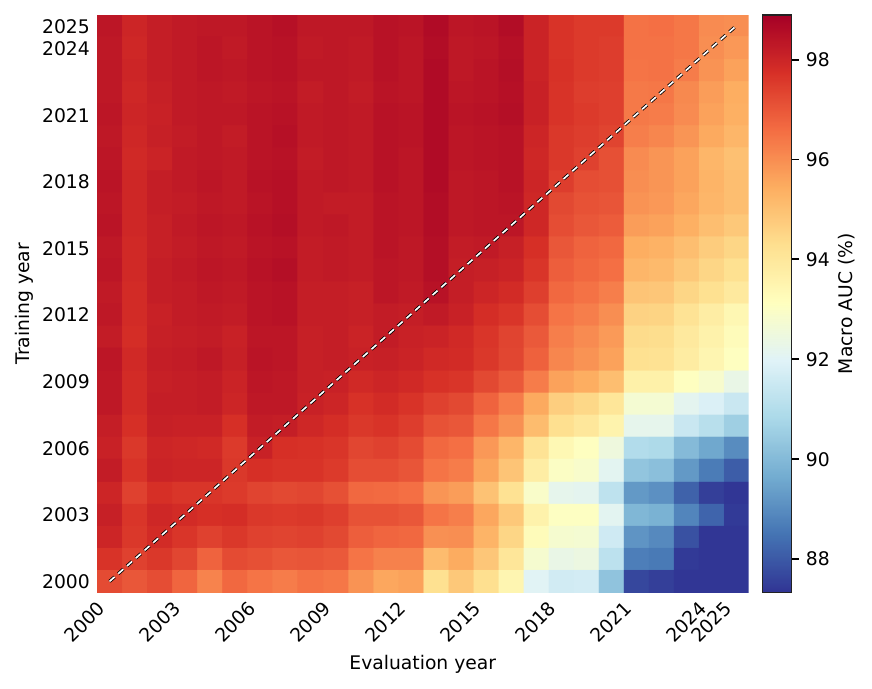}\hfill
    \includegraphics[width=0.49\linewidth]{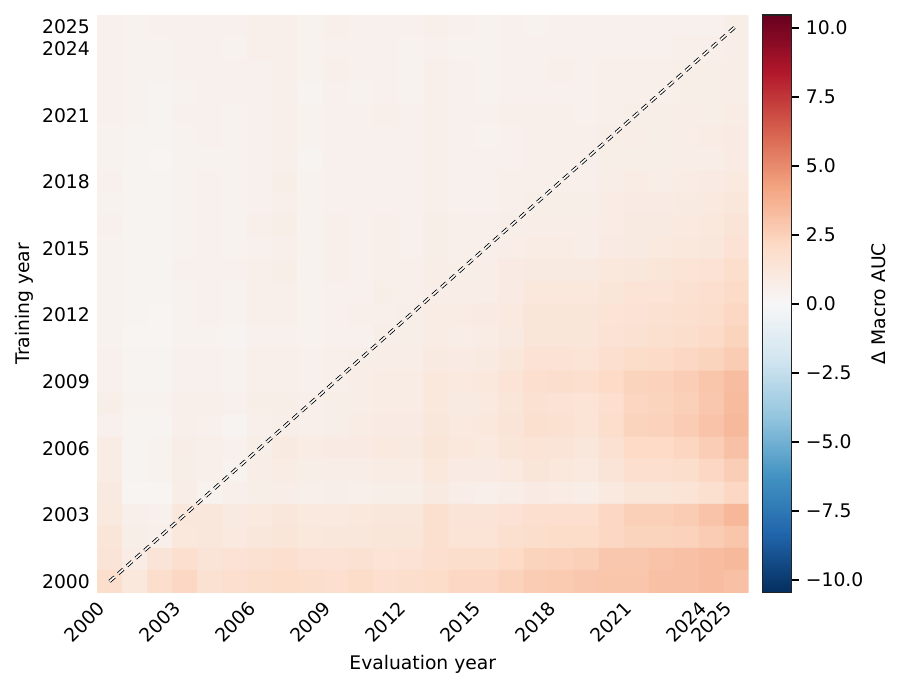}
    \caption{FFN-L}
    \label{fig:arxiv_FFN_L}
\end{subfigure}
\caption{FFN models: Macro AUC drift matrix $M^{(m)}$ and deviation from the cohort mean $\Delta^{(m)} = M^{(m)} - \bar{M}$ for each model, shown on a sequential and a zero-centred diverging scale, respectively.}
\label{fig:arxiv_family_text_ffn}
\end{figure}

\subsection{TextCNN}

\begin{figure}[H]
\centering
\begin{subfigure}[t]{0.49\textwidth}\centering
    \includegraphics[width=0.49\linewidth]{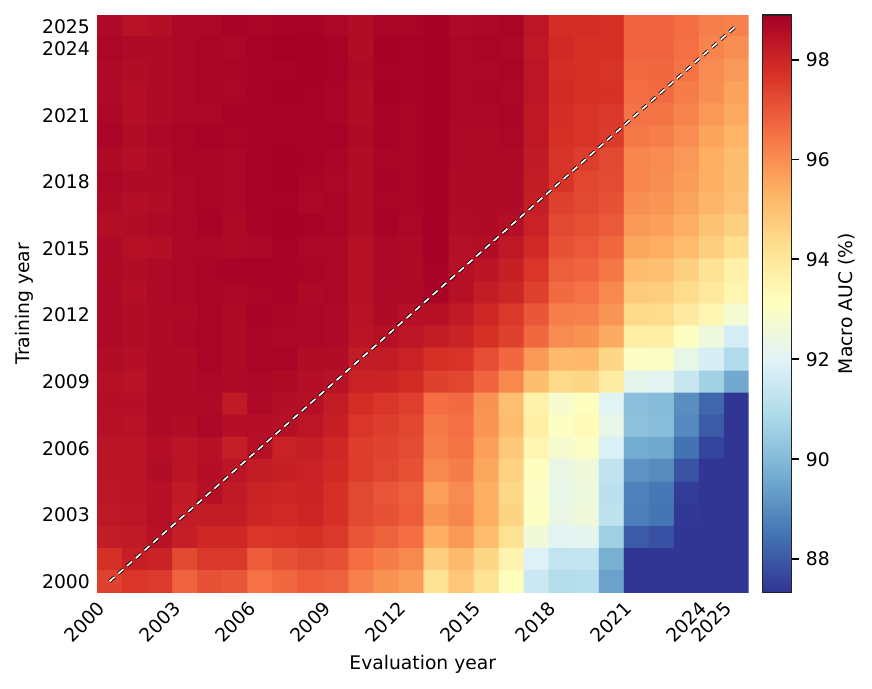}\hfill
    \includegraphics[width=0.49\linewidth]{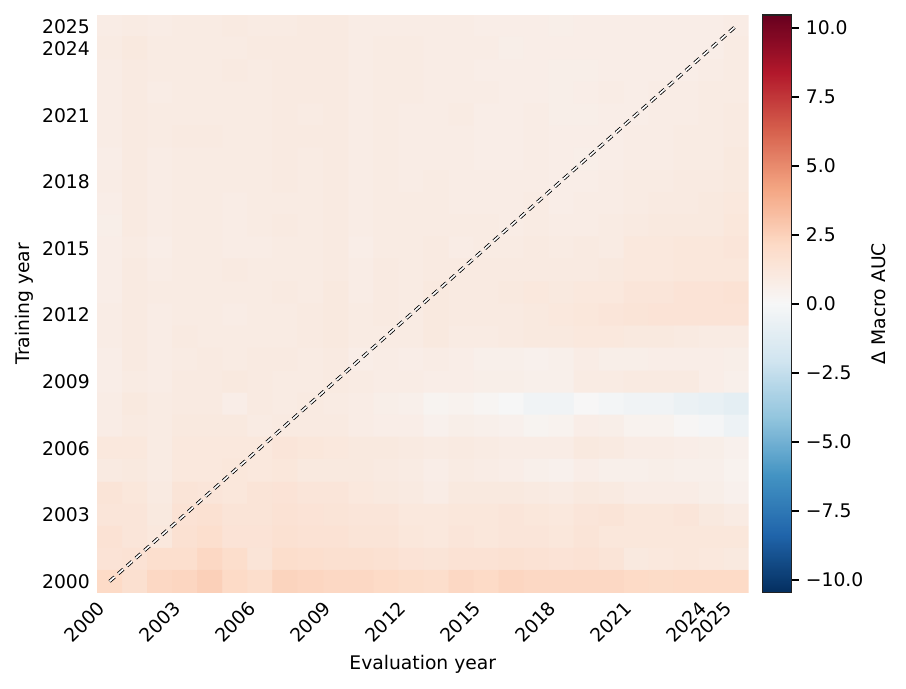}
    \caption{TextCNN-S}
    \label{fig:arxiv_TextCNN_S}
\end{subfigure}
\hfill
\begin{subfigure}[t]{0.49\textwidth}\centering
    \includegraphics[width=0.49\linewidth]{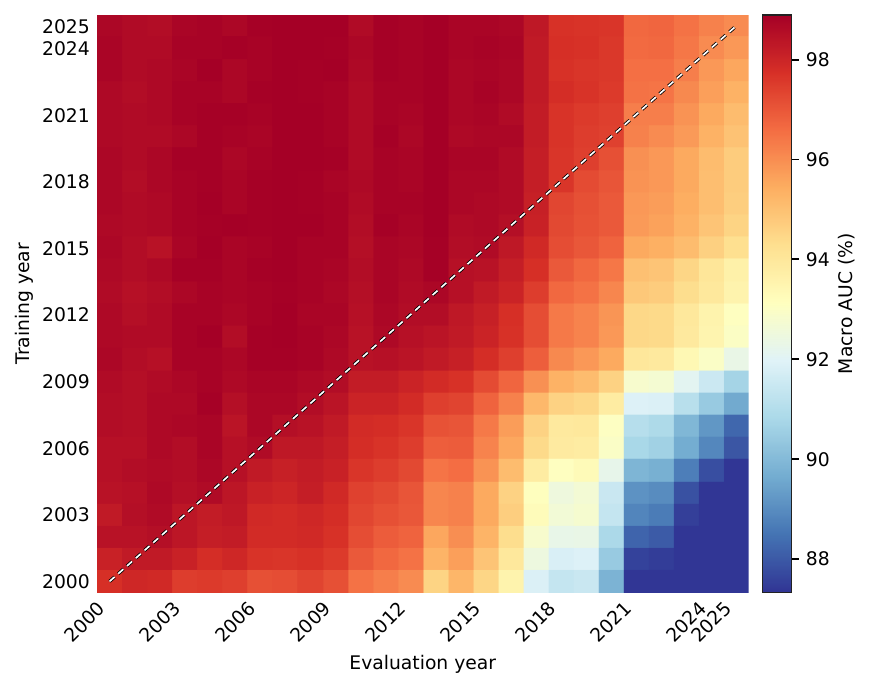}\hfill
    \includegraphics[width=0.49\linewidth]{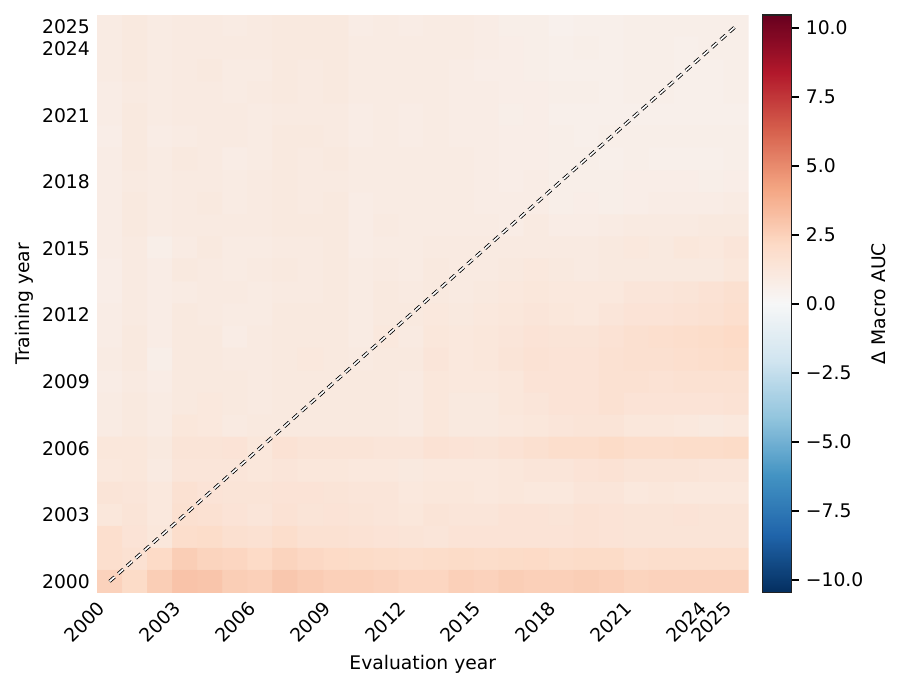}
    \caption{TextCNN-M}
    \label{fig:arxiv_TextCNN_M}
\end{subfigure}

\begin{subfigure}[t]{0.49\textwidth}\centering
    \includegraphics[width=0.49\linewidth]{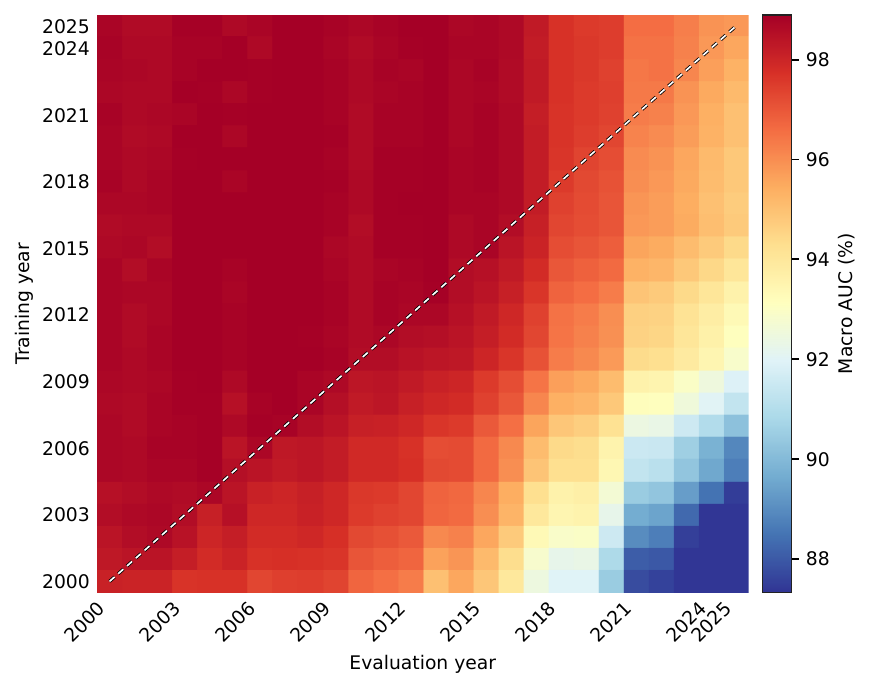}\hfill
    \includegraphics[width=0.49\linewidth]{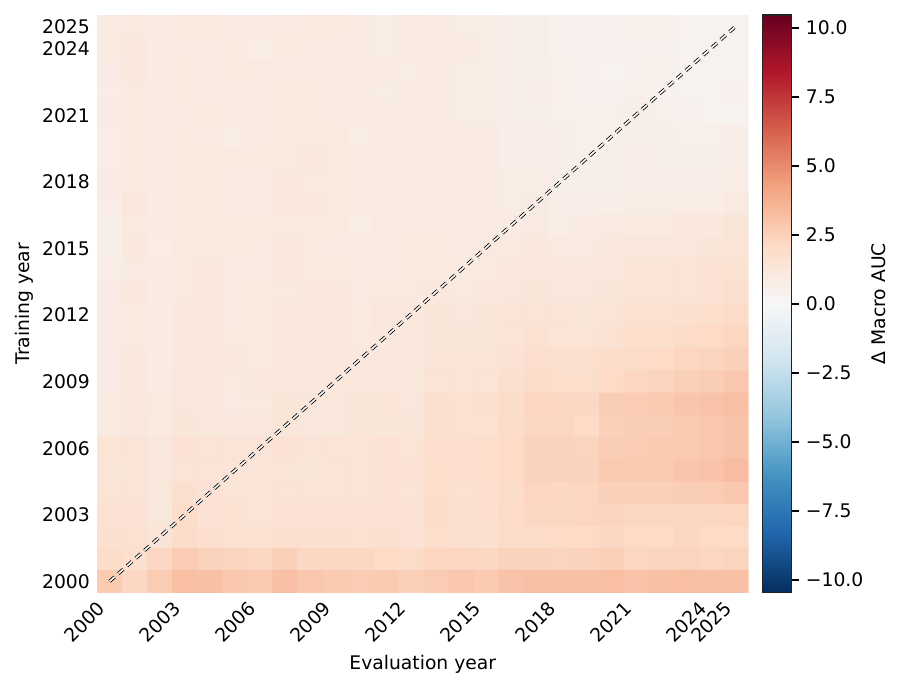}
    \caption{TextCNN-L}
    \label{fig:arxiv_TextCNN_L}
\end{subfigure}
\caption{TextCNN models: Macro AUC drift matrix $M^{(m)}$ and deviation from the cohort mean $\Delta^{(m)} = M^{(m)} - \bar{M}$ for each model, shown on a sequential and a zero-centred diverging scale, respectively.}
\label{fig:arxiv_family_text_textcnn}
\end{figure}

\subsection{Recurrent}

\begin{figure}[H]
\centering
\begin{subfigure}[t]{0.49\textwidth}\centering
    \includegraphics[width=0.49\linewidth]{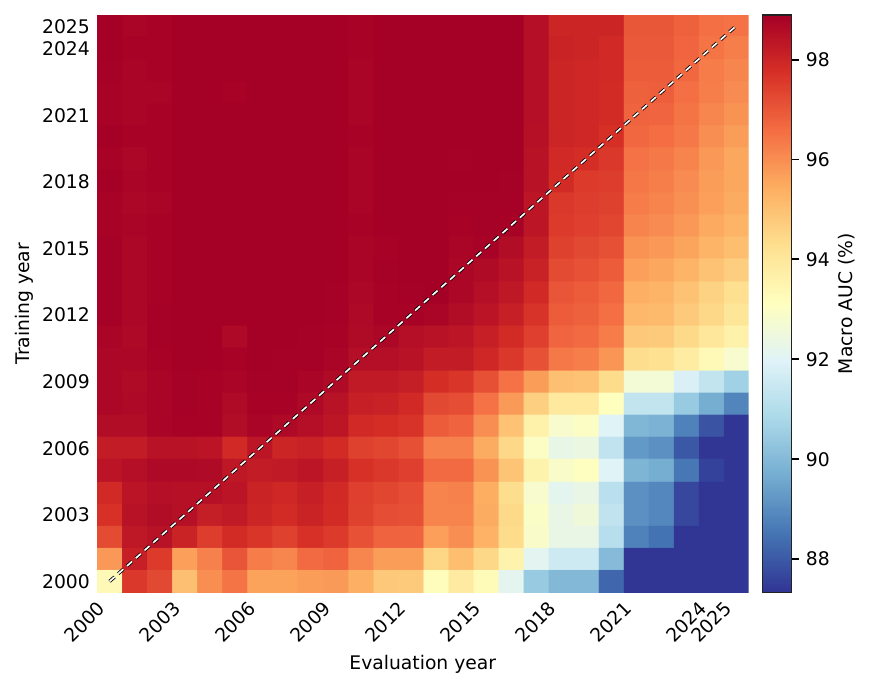}\hfill
    \includegraphics[width=0.49\linewidth]{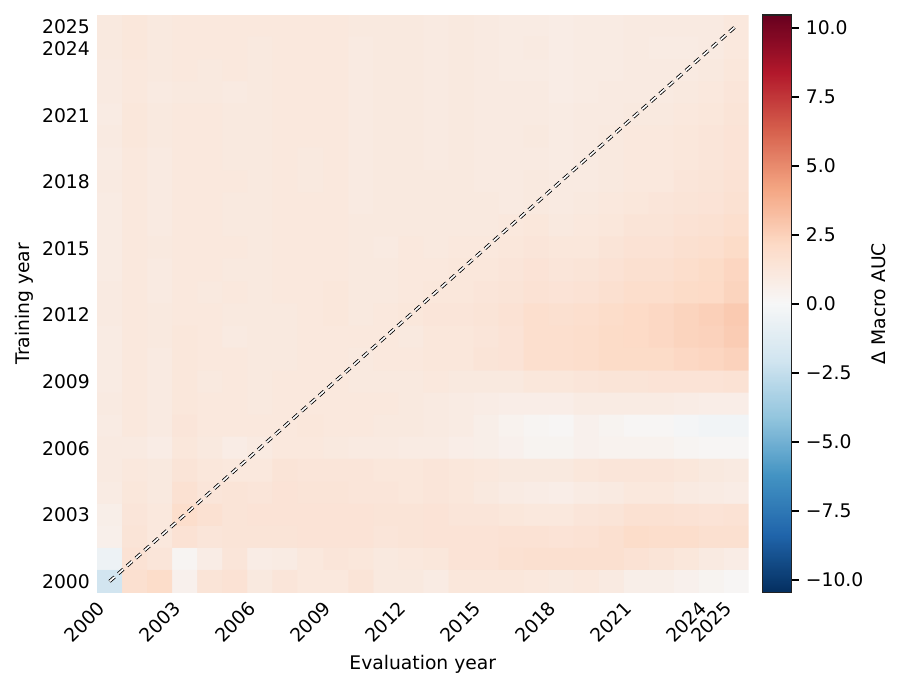}
    \caption{BiGRU-S}
    \label{fig:arxiv_BiGRU_S}
\end{subfigure}
\hfill
\begin{subfigure}[t]{0.49\textwidth}\centering
    \includegraphics[width=0.49\linewidth]{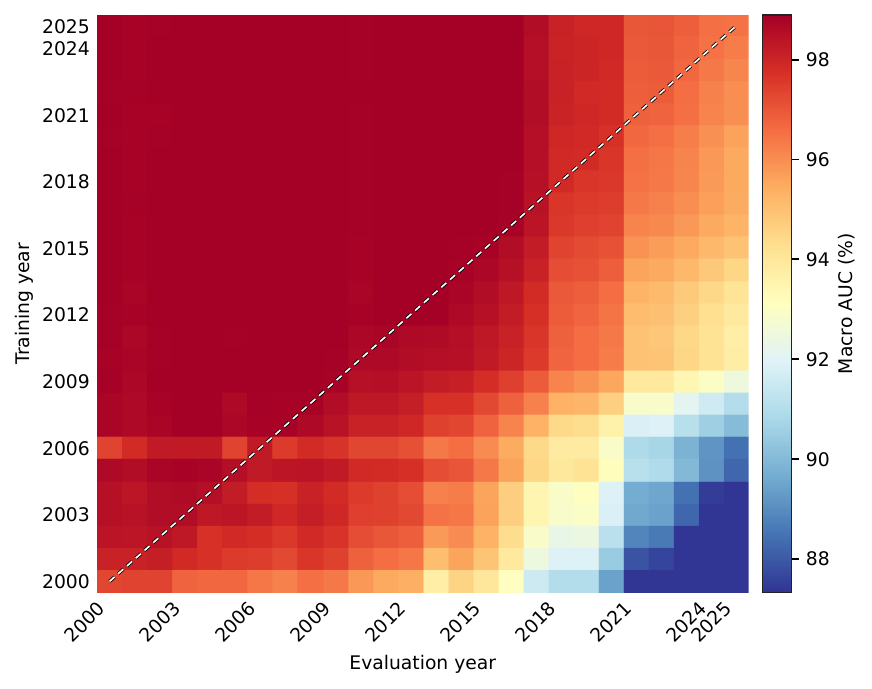}\hfill
    \includegraphics[width=0.49\linewidth]{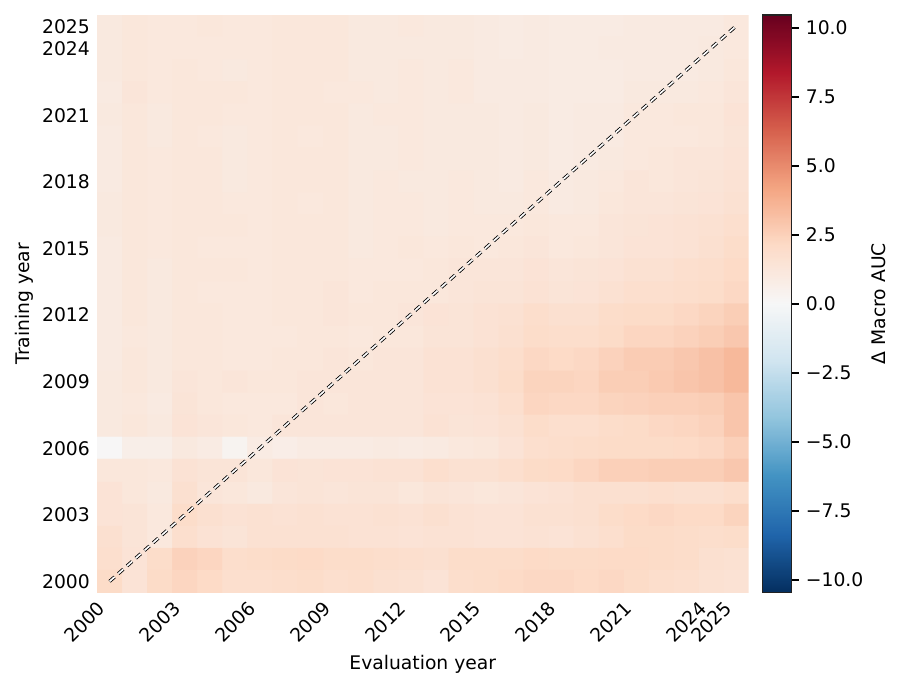}
    \caption{BiLSTM-M}
    \label{fig:arxiv_BiLSTM_M}
\end{subfigure}

\begin{subfigure}[t]{0.49\textwidth}\centering
    \includegraphics[width=0.49\linewidth]{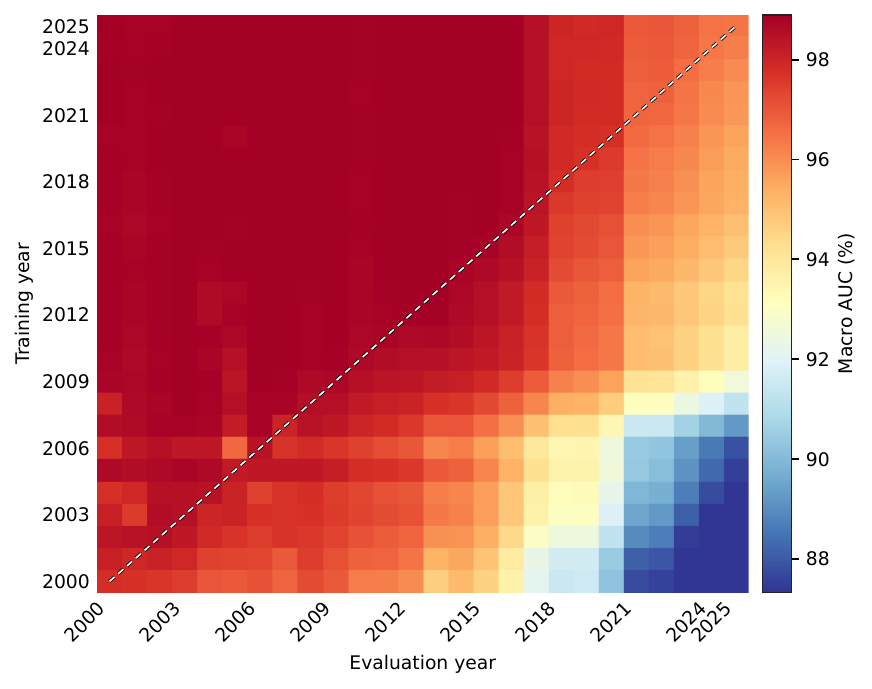}\hfill
    \includegraphics[width=0.49\linewidth]{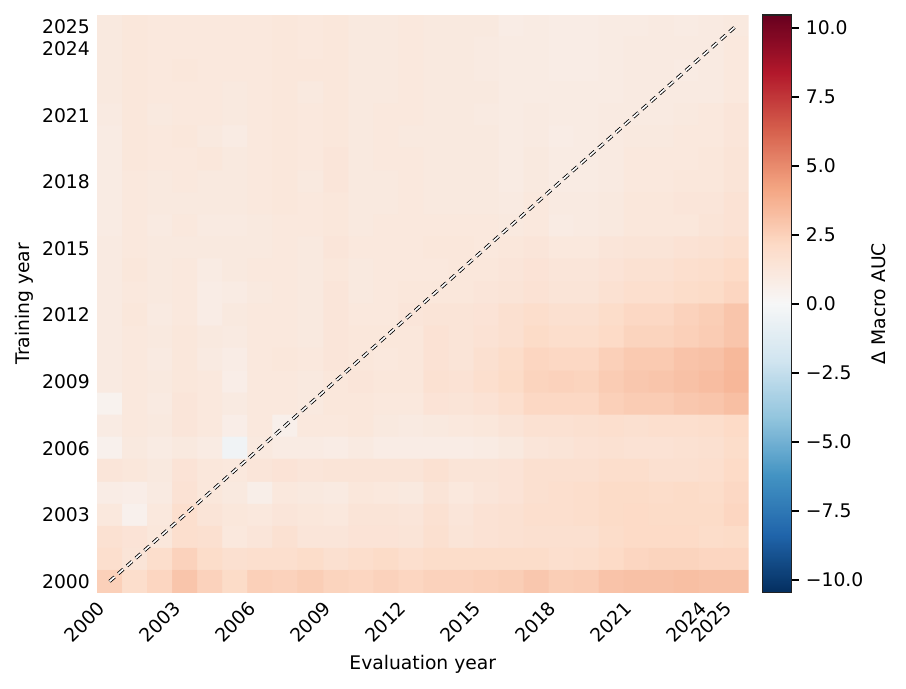}
    \caption{BiLSTM-Attn-L}
    \label{fig:arxiv_BiLSTM_Attn_L}
\end{subfigure}
\caption{Recurrent models: Macro AUC drift matrix $M^{(m)}$ and deviation from the cohort mean $\Delta^{(m)} = M^{(m)} - \bar{M}$ for each model, shown on a sequential and a zero-centred diverging scale, respectively.}
\label{fig:arxiv_family_text_rnn}
\end{figure}

\subsection{Transformer}

\begin{figure}[H]
\centering
\begin{subfigure}[t]{0.49\textwidth}\centering
    \includegraphics[width=0.49\linewidth]{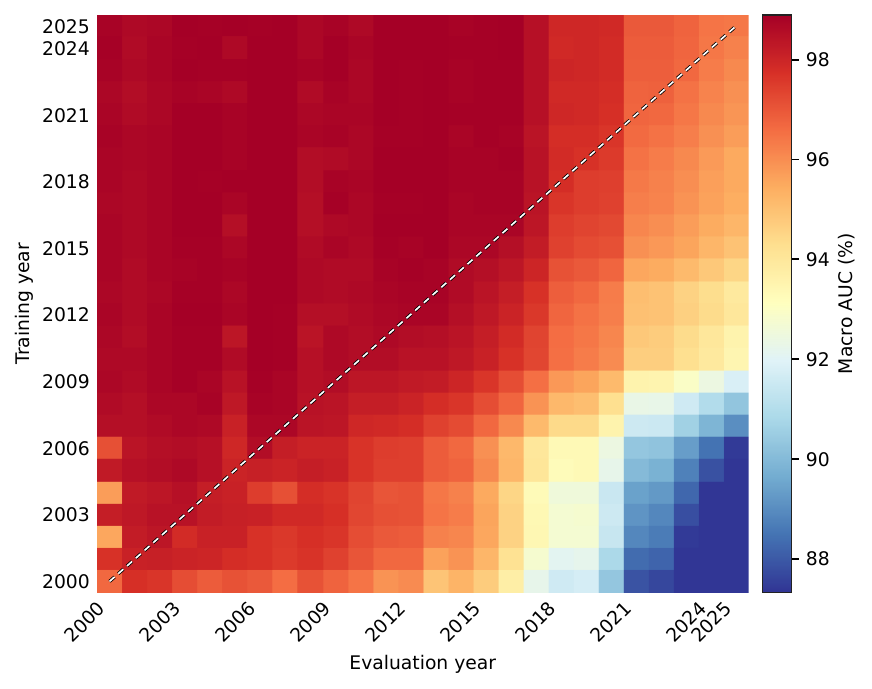}\hfill
    \includegraphics[width=0.49\linewidth]{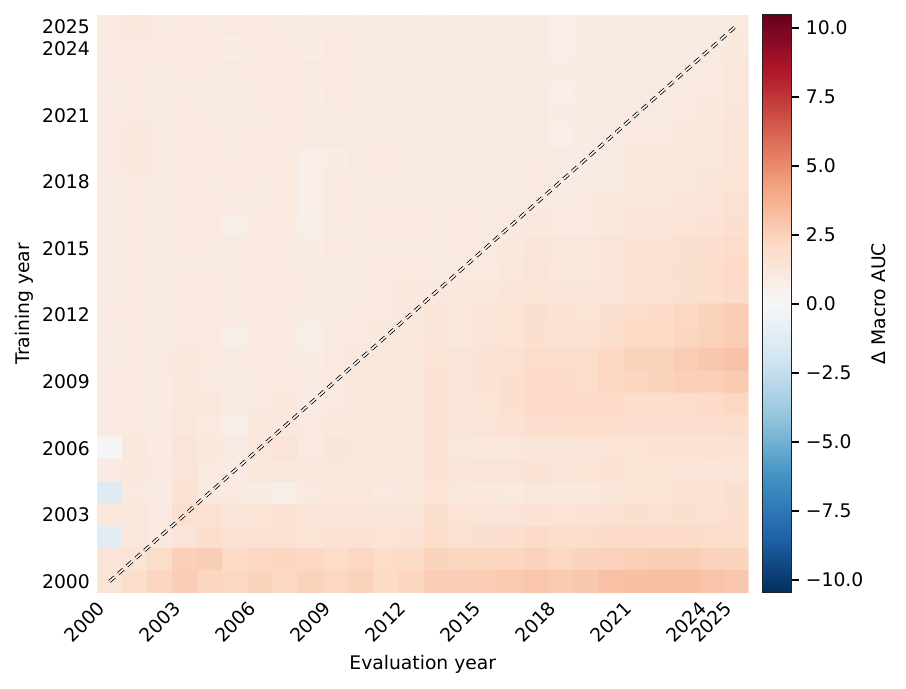}
    \caption{TX-S}
    \label{fig:arxiv_TX_S}
\end{subfigure}
\hfill
\begin{subfigure}[t]{0.49\textwidth}\centering
    \includegraphics[width=0.49\linewidth]{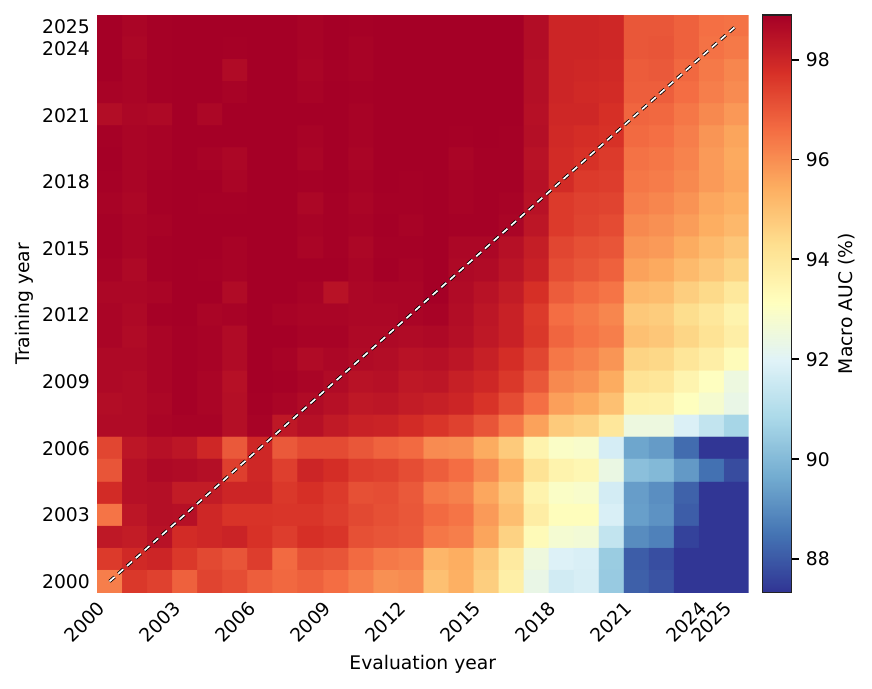}\hfill
    \includegraphics[width=0.49\linewidth]{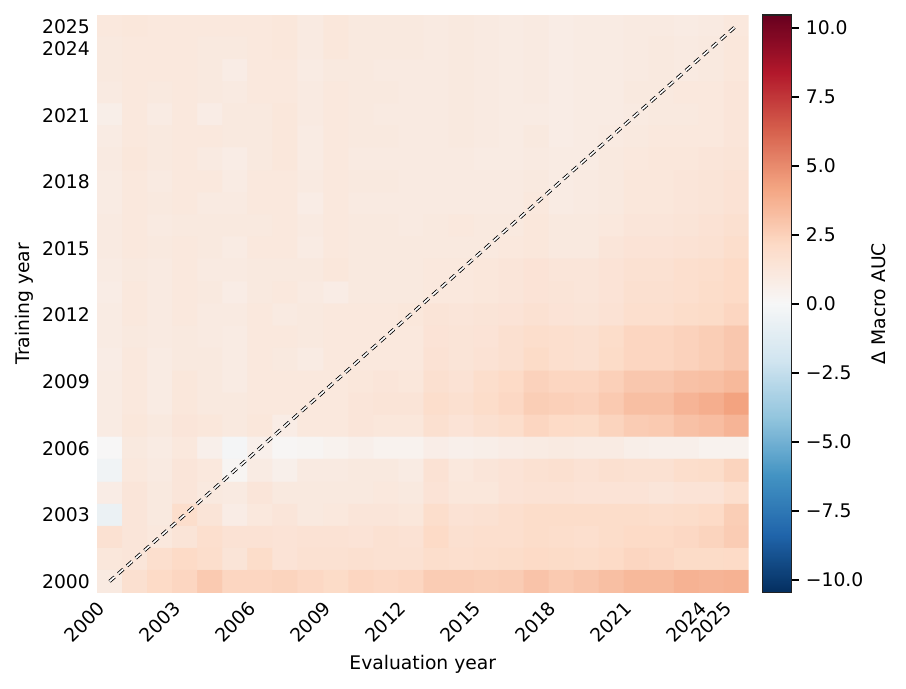}
    \caption{TX-M}
    \label{fig:arxiv_TX_M}
\end{subfigure}

\begin{subfigure}[t]{0.49\textwidth}\centering
    \includegraphics[width=0.49\linewidth]{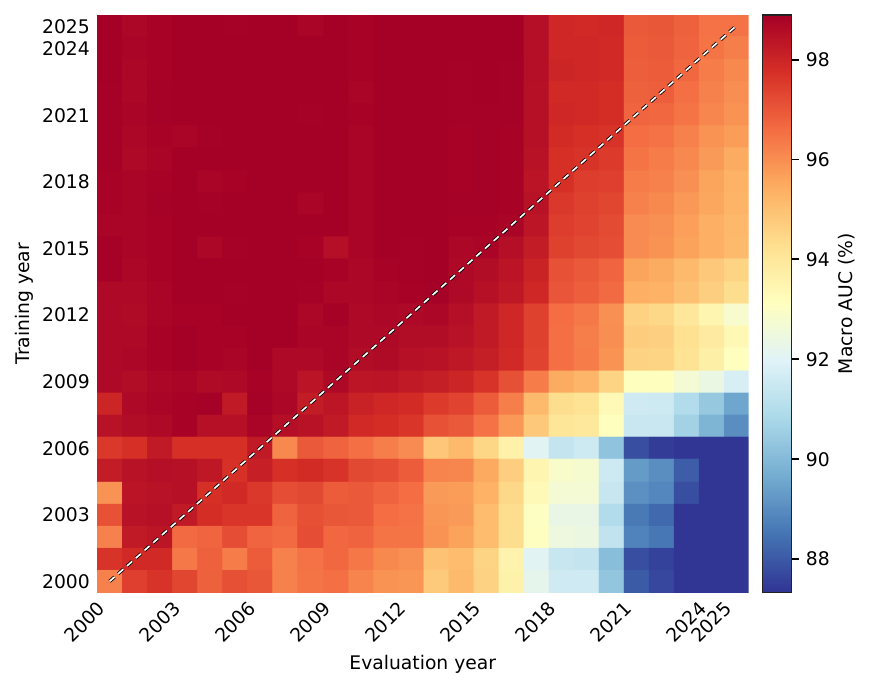}\hfill
    \includegraphics[width=0.49\linewidth]{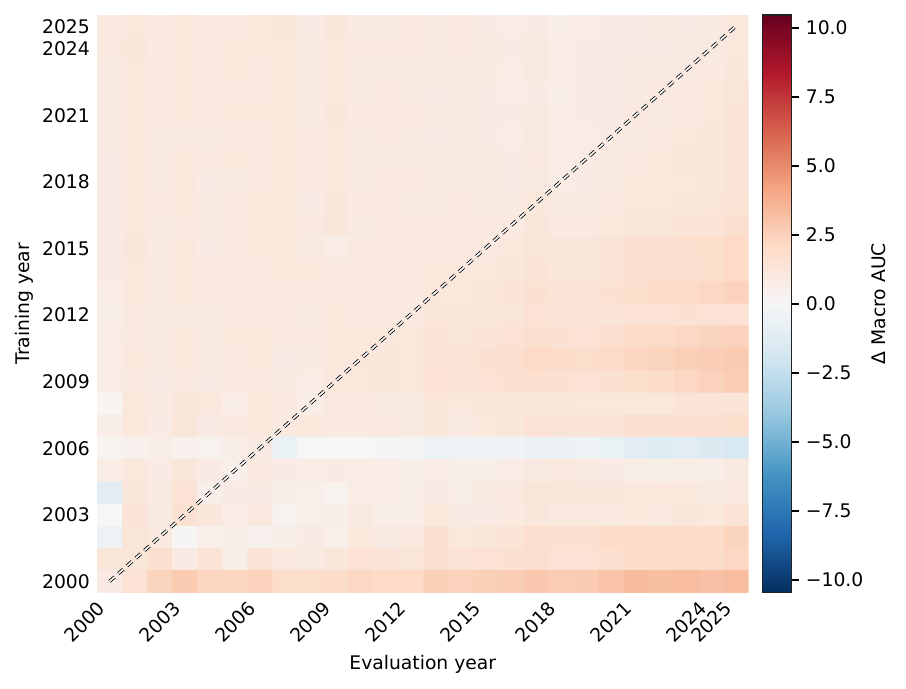}
    \caption{TX-L}
    \label{fig:arxiv_TX_L}
\end{subfigure}
\caption{Transformer models: Macro AUC drift matrix $M^{(m)}$ and deviation from the cohort mean $\Delta^{(m)} = M^{(m)} - \bar{M}$ for each model, shown on a sequential and a zero-centred diverging scale, respectively.}
\label{fig:arxiv_family_text_tx}
\end{figure}

\subsection{Frozen}

\begin{figure}[H]
\centering
\begin{subfigure}[t]{0.49\textwidth}\centering
    \includegraphics[width=0.49\linewidth]{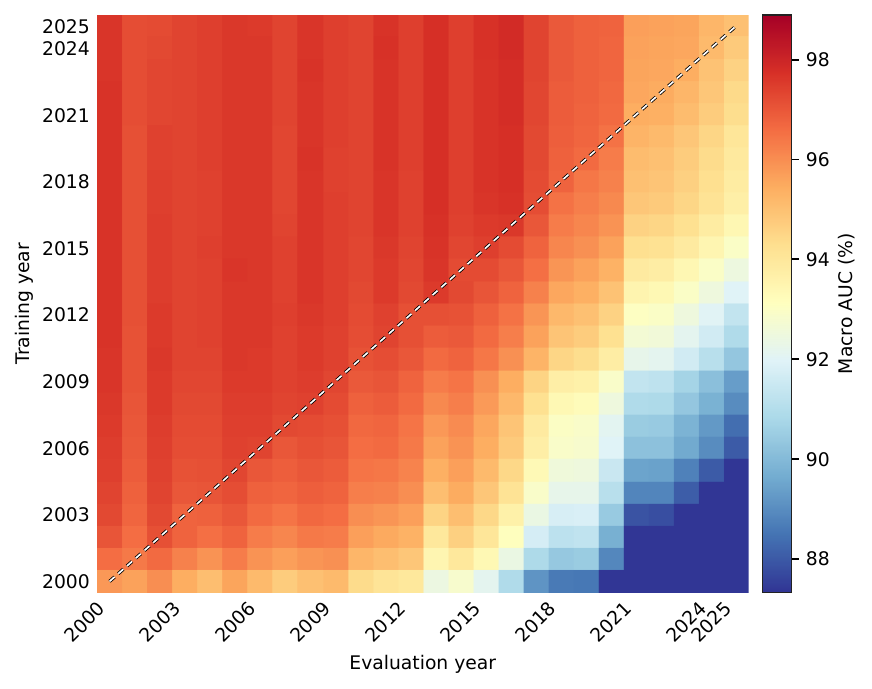}\hfill
    \includegraphics[width=0.49\linewidth]{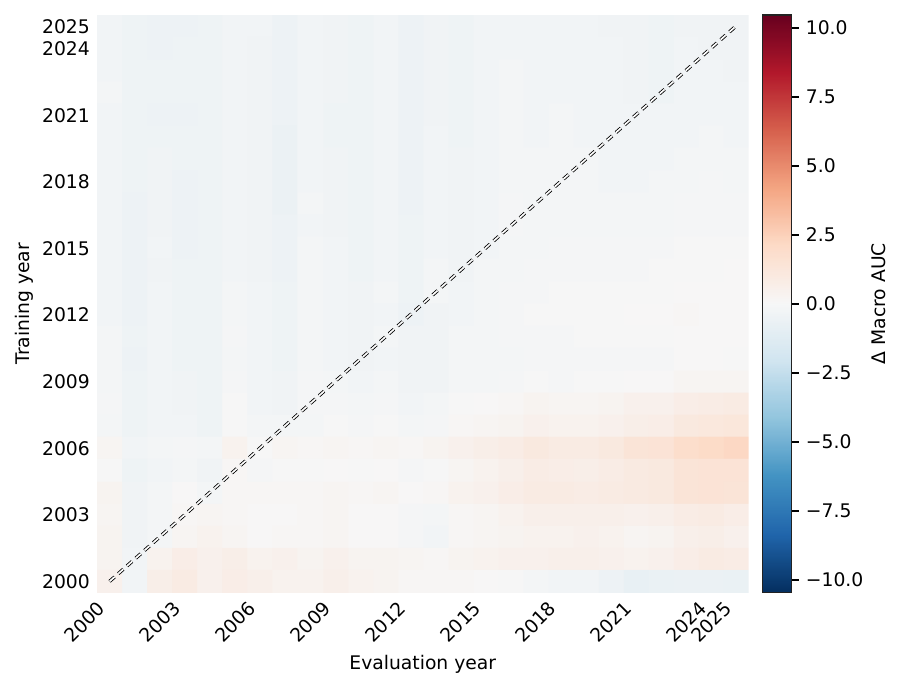}
    \caption{BERT}
    \label{fig:arxiv_BERT}
\end{subfigure}
\hfill
\begin{subfigure}[t]{0.49\textwidth}\centering
    \includegraphics[width=0.49\linewidth]{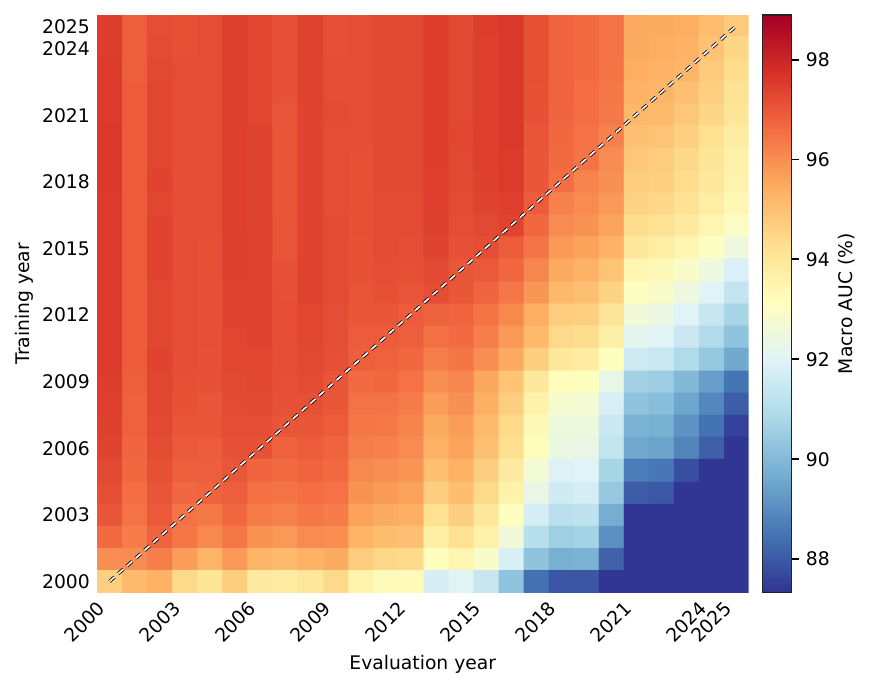}\hfill
    \includegraphics[width=0.49\linewidth]{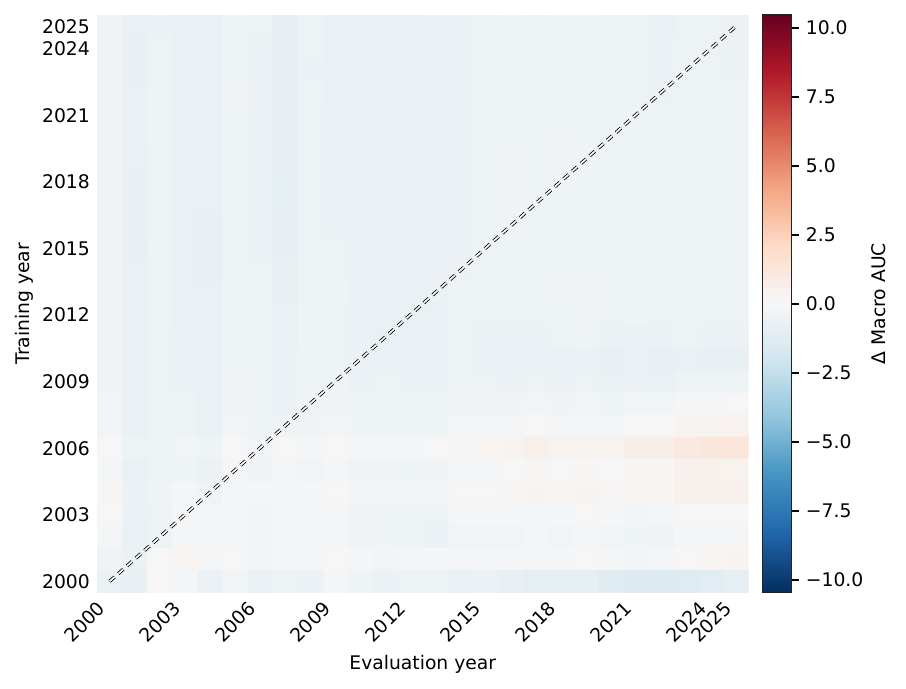}
    \caption{DistilBERT}
    \label{fig:arxiv_DistilBERT}
\end{subfigure}

\begin{subfigure}[t]{0.49\textwidth}\centering
    \includegraphics[width=0.49\linewidth]{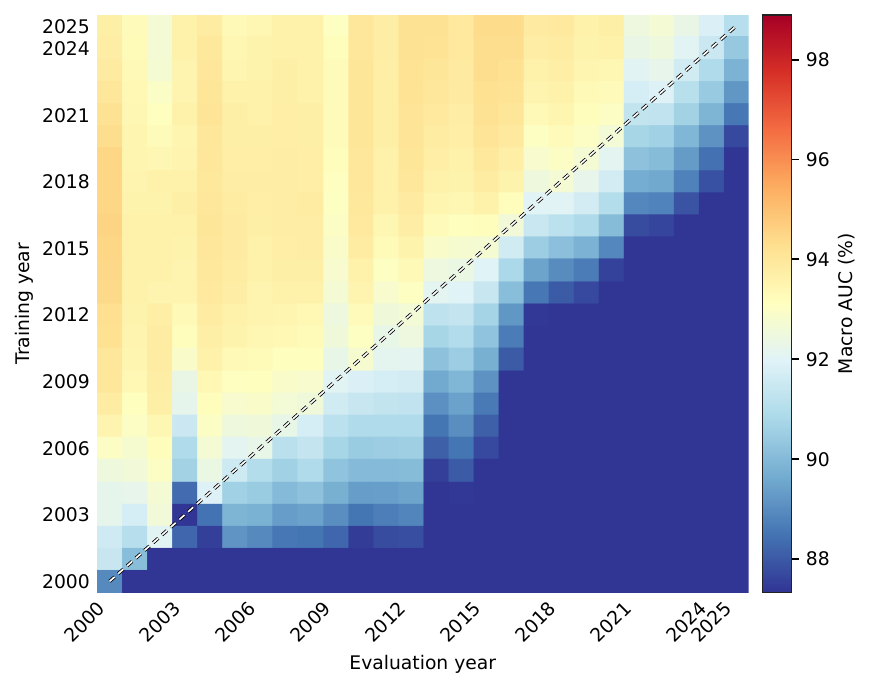}\hfill
    \includegraphics[width=0.49\linewidth]{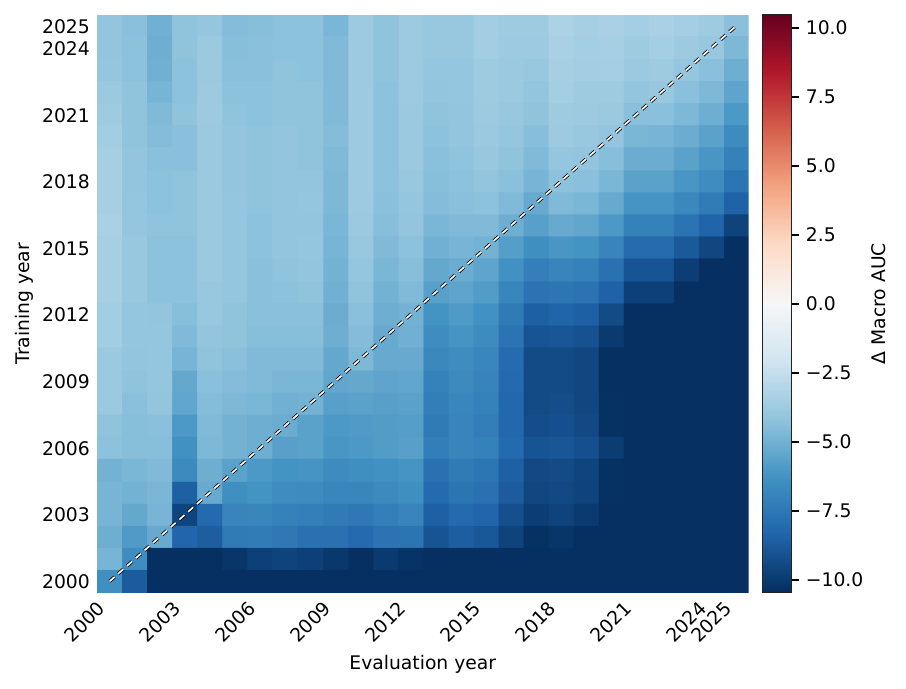}
    \caption{ELECTRA}
    \label{fig:arxiv_ELECTRA}
\end{subfigure}
\hfill
\begin{subfigure}[t]{0.49\textwidth}\centering
    \includegraphics[width=0.49\linewidth]{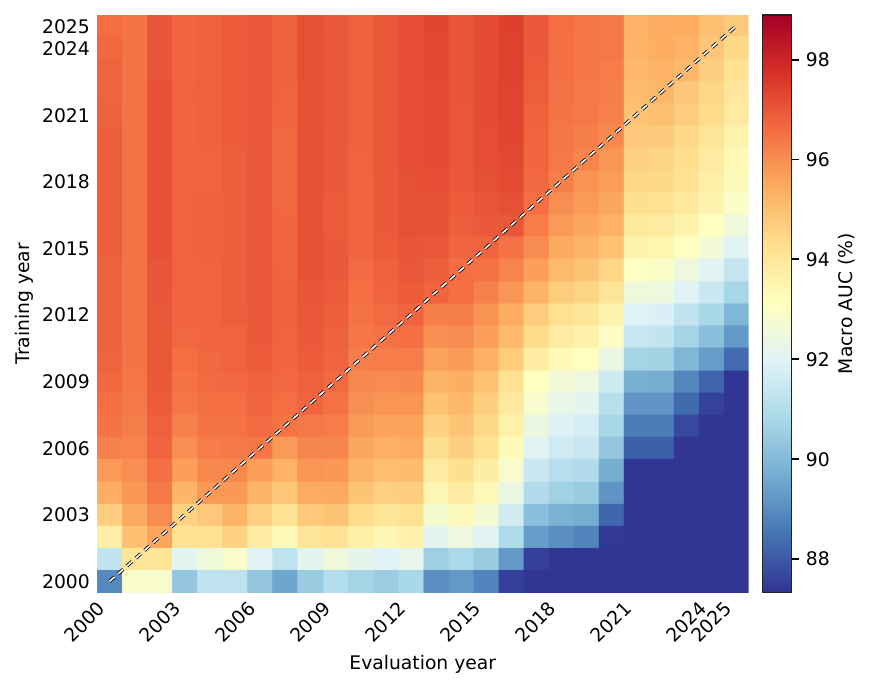}\hfill
    \includegraphics[width=0.49\linewidth]{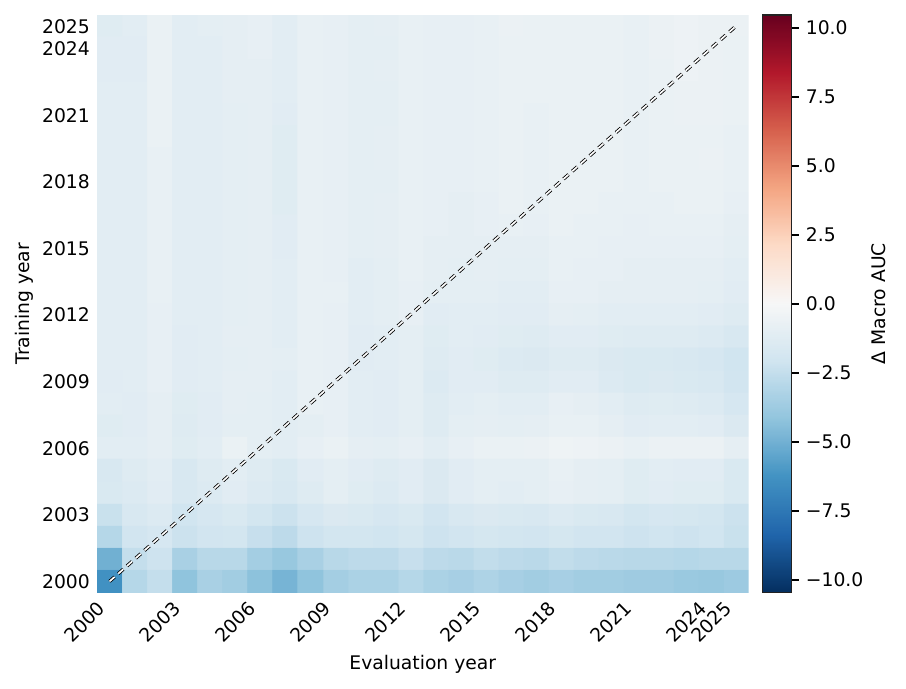}
    \caption{MPNet}
    \label{fig:arxiv_MPNet}
\end{subfigure}

\begin{subfigure}[t]{0.49\textwidth}\centering
    \includegraphics[width=0.49\linewidth]{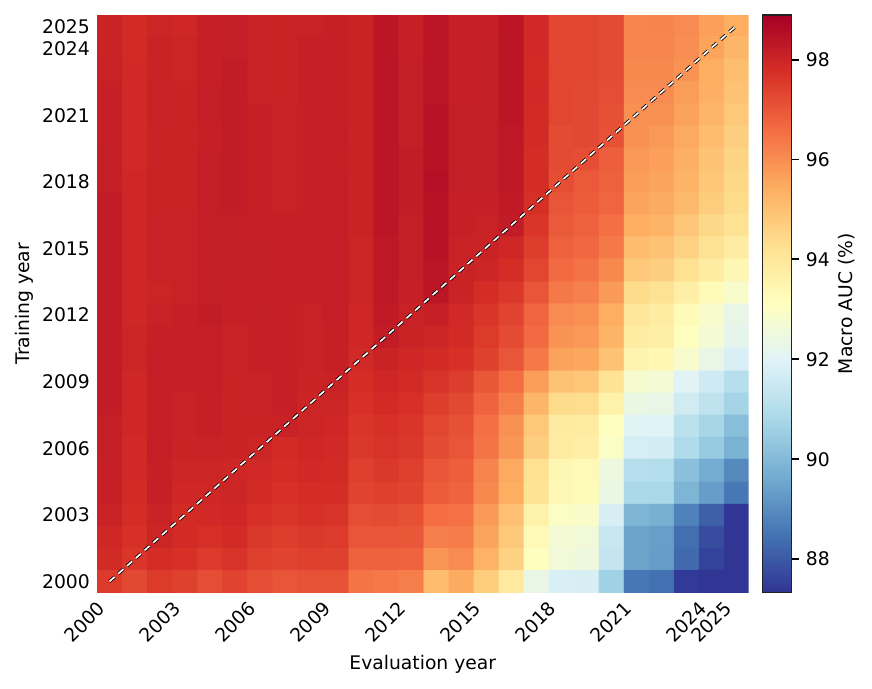}\hfill
    \includegraphics[width=0.49\linewidth]{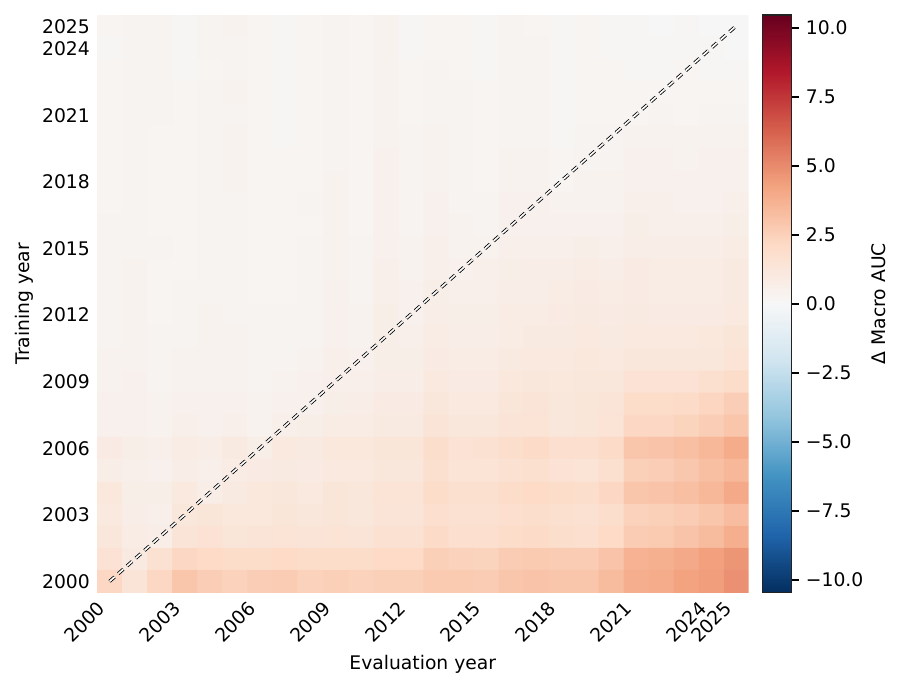}
    \caption{ModernBERT}
    \label{fig:arxiv_ModernBERT}
\end{subfigure}
\hfill
\begin{subfigure}[t]{0.49\textwidth}\centering
    \includegraphics[width=0.49\linewidth]{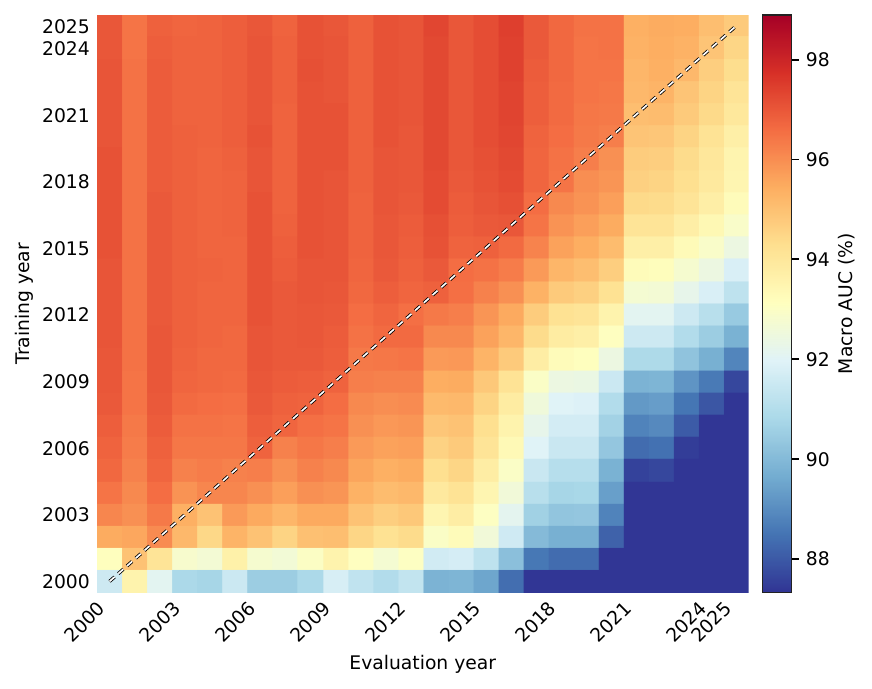}\hfill
    \includegraphics[width=0.49\linewidth]{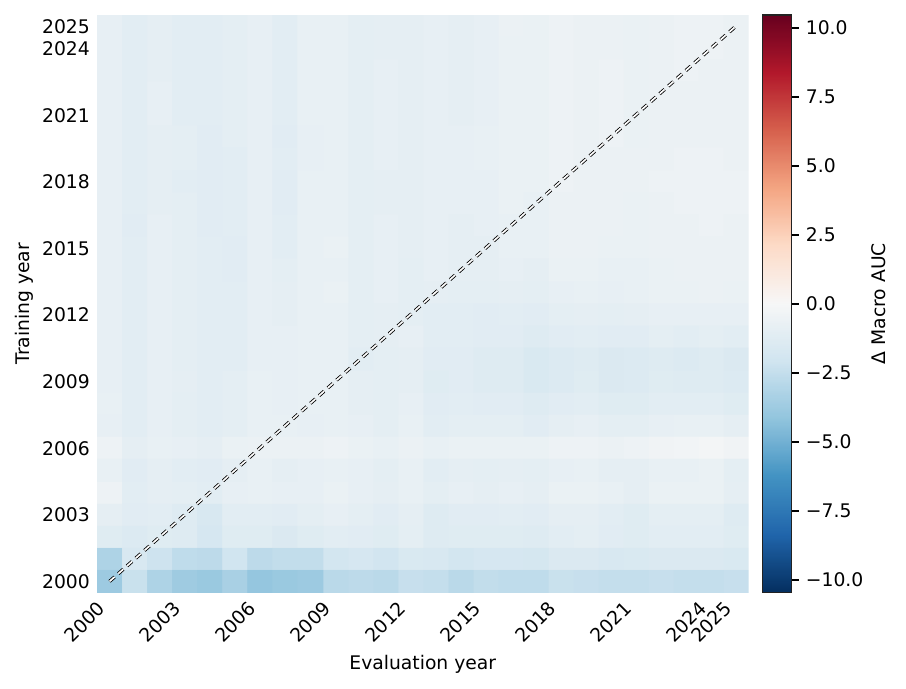}
    \caption{RoBERTa}
    \label{fig:arxiv_RoBERTa}
\end{subfigure}

\begin{subfigure}[t]{0.49\textwidth}\centering
    \includegraphics[width=0.49\linewidth]{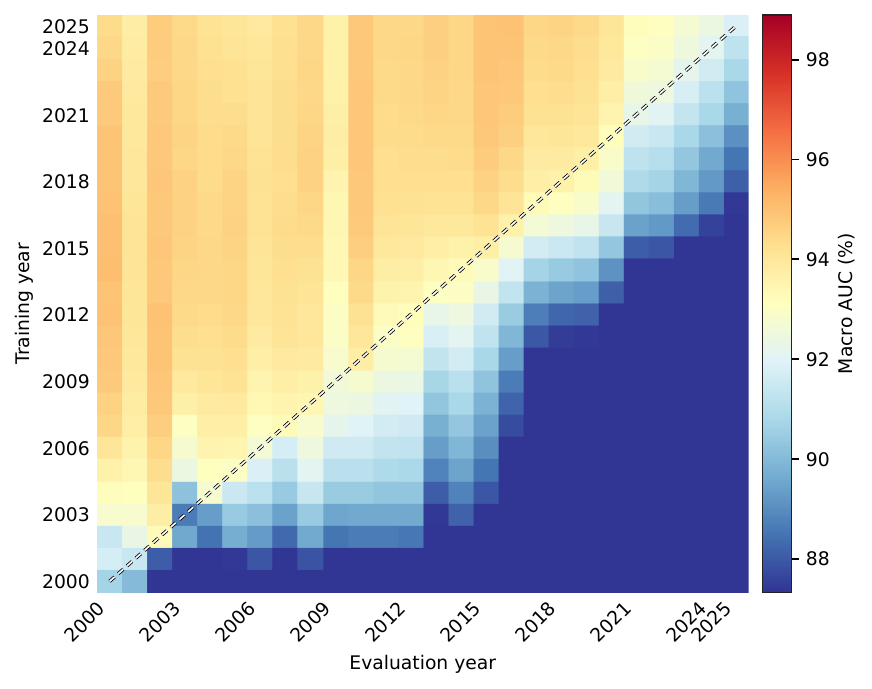}\hfill
    \includegraphics[width=0.49\linewidth]{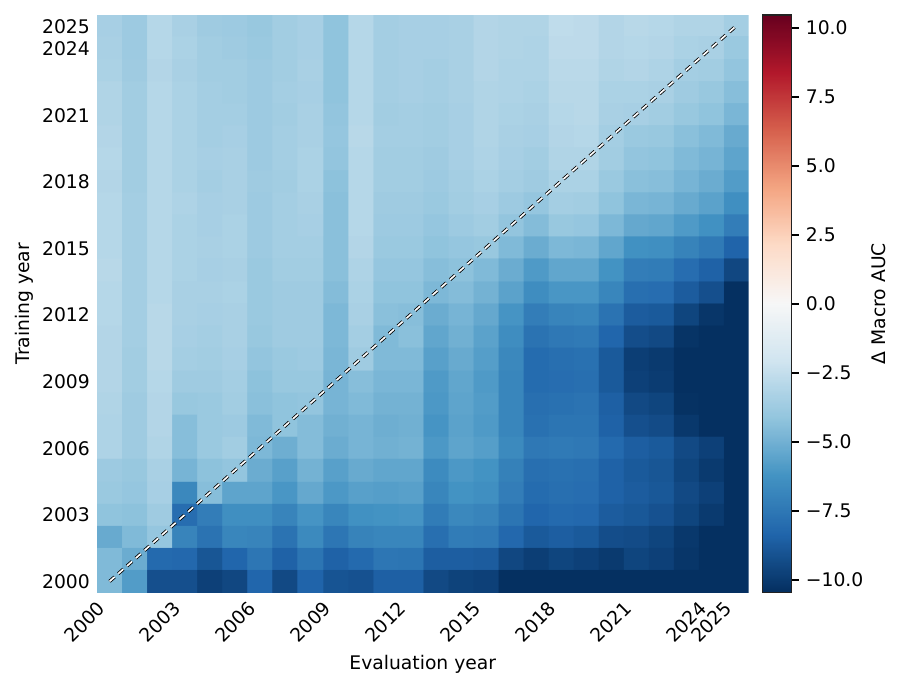}
    \caption{DeBERTa-v3}
    \label{fig:arxiv_DeBERTa_v3}
\end{subfigure}
\hfill
\begin{subfigure}[t]{0.49\textwidth}\centering
    \includegraphics[width=0.49\linewidth]{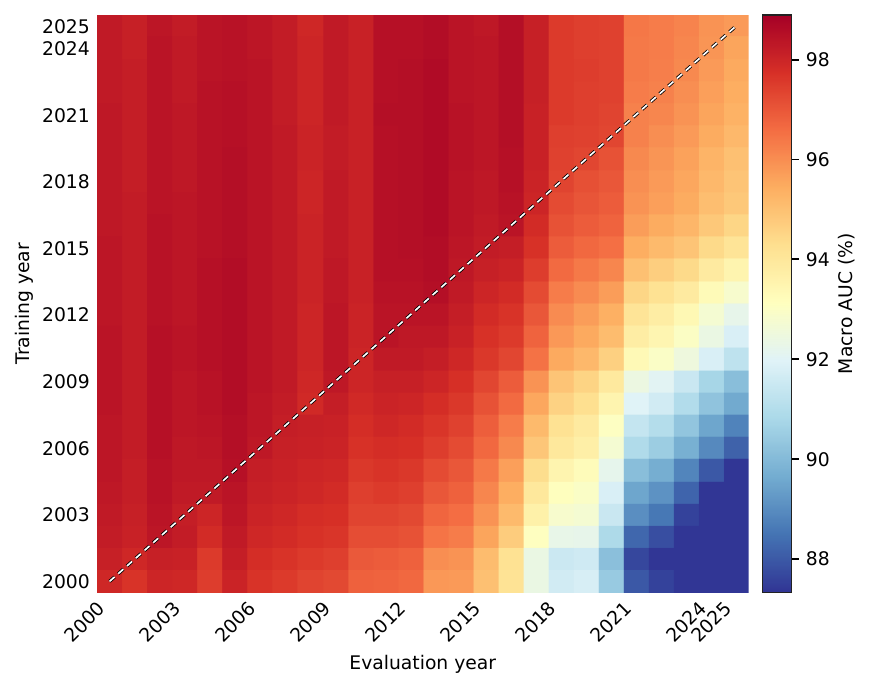}\hfill
    \includegraphics[width=0.49\linewidth]{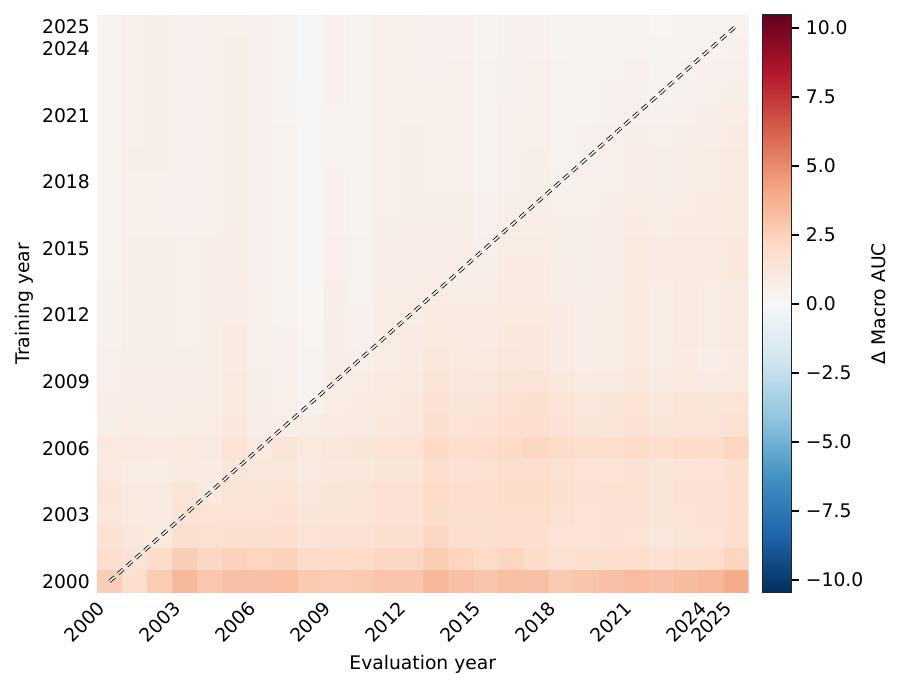}
    \caption{MiniLM-L6}
    \label{fig:arxiv_MiniLM_L6}
\end{subfigure}
\caption{Frozen models: Macro AUC drift matrix $M^{(m)}$ and deviation from the cohort mean $\Delta^{(m)} = M^{(m)} - \bar{M}$ for each model, shown on a sequential and a zero-centred diverging scale, respectively.}
\label{fig:arxiv_family_text_frozen_head}
\end{figure}

\subsection{Forgetting and Rankings}
\label{app:arxiv_forgetting}

\noindent To see how quickly each model forgets, we summarize its drift matrix as a forgetting curve. The curve plots the Macro AUC against the lag $\ell = j - i$, the number of slices between the training cutoff $i$ and the evaluation slice $j$. At each lag we average over all training cutoffs,
\[ F(\ell) = \operatorname{mean}_{i} M_{i,\,i+\ell}. \]
The result is the Macro AUC at a fixed temporal distance, independent of which period a model was trained on. This separates the effect of temporal distance from the difficulty of any single slice.

\begin{figure}[H]
\centering
\begin{subfigure}[t]{0.49\textwidth}\centering
    \includegraphics[width=\linewidth]{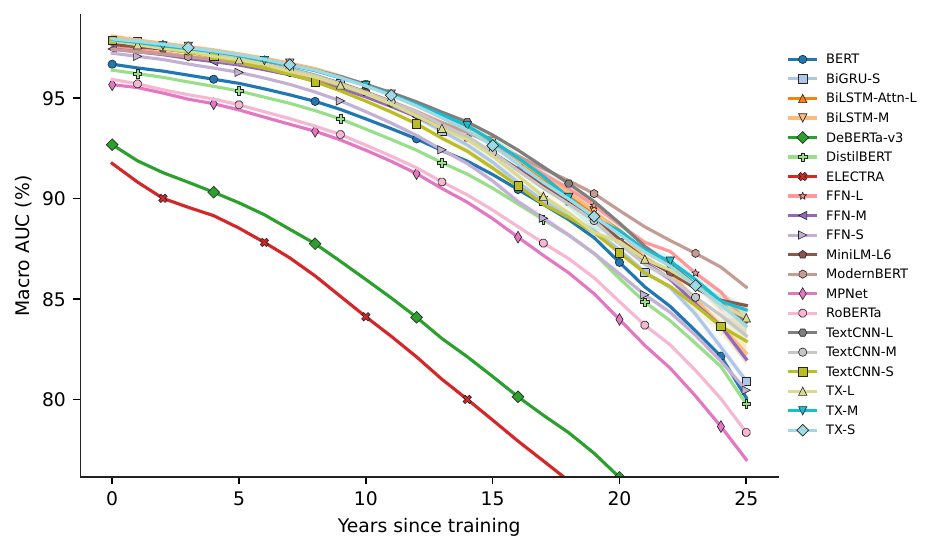}
    \caption{Per model}
    \label{fig:arxiv_forgetting}
\end{subfigure}\hfill
\begin{subfigure}[t]{0.49\textwidth}\centering
    \includegraphics[width=\linewidth]{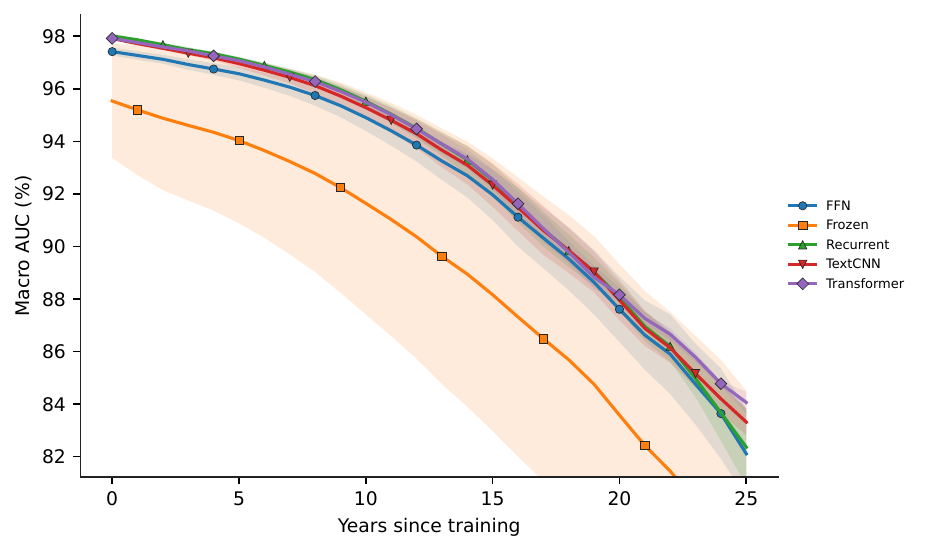}
    \caption{Per model family}
    \label{fig:arxiv_forgetting_family}
\end{subfigure}
\caption{Forgetting curves: each model (left) and averaged within each family (right).}
\label{fig:arxiv_forgetting_combined}
\end{figure}

\begin{figure}[H]
\centering
\begin{subfigure}[t]{0.49\textwidth}\centering
    \includegraphics[width=\linewidth]{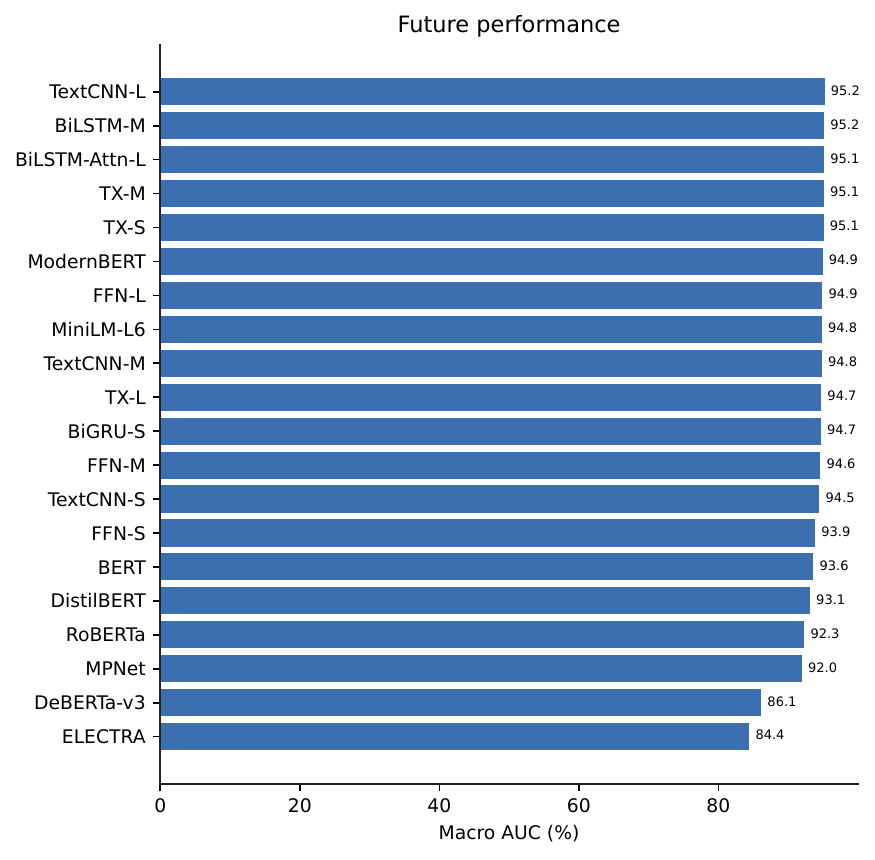}
    \caption{By future performance}
    \label{fig:arxiv_ranking_future}
\end{subfigure}\hfill
\begin{subfigure}[t]{0.49\textwidth}\centering
    \includegraphics[width=\linewidth]{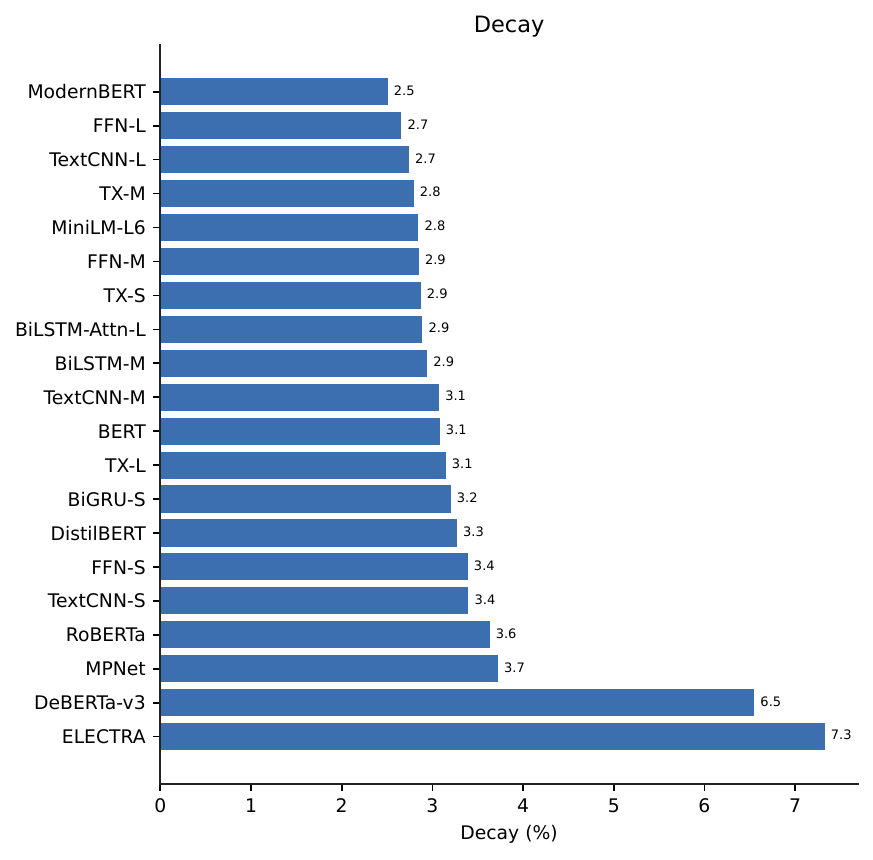}
    \caption{By decay}
    \label{fig:arxiv_ranking_decay}
\end{subfigure}
\caption{Models ranked by mean future performance and by temporal decay.}
\label{fig:arxiv_rankings}
\end{figure}

\subsection{Result Tables}

\begin{table}[H]
\centering
\footnotesize
\caption{Temporal robustness on arXiv.}
\label{tab:robustness_arxiv}
\begin{minipage}[t]{0.49\linewidth}
\centering
\resizebox{\ifdim\width>\linewidth \linewidth\else\width\fi}{!}{%
\begin{tabular}{l rr}
\toprule
Model & Future (\%) & Decay (\%) \\
\midrule
ModernBERT & 94.9 & 2.5 \\
FFN-L & 94.9 & 2.7 \\
TextCNN-L & 95.2 & 2.7 \\
TX-M & 95.1 & 2.8 \\
MiniLM-L6 & 94.8 & 2.8 \\
FFN-M & 94.6 & 2.9 \\
TX-S & 95.1 & 2.9 \\
BiLSTM-Attn-L & 95.1 & 2.9 \\
BiLSTM-M & 95.2 & 2.9 \\
TextCNN-M & 94.8 & 3.1 \\
\bottomrule
\end{tabular}}
\end{minipage}\hfill
\begin{minipage}[t]{0.49\linewidth}
\centering
\resizebox{\ifdim\width>\linewidth \linewidth\else\width\fi}{!}{%
\begin{tabular}{l rr}
\toprule
Model & Future (\%) & Decay (\%) \\
\midrule
BERT & 93.6 & 3.1 \\
TX-L & 94.7 & 3.1 \\
BiGRU-S & 94.7 & 3.2 \\
DistilBERT & 93.1 & 3.3 \\
FFN-S & 93.9 & 3.4 \\
TextCNN-S & 94.5 & 3.4 \\
RoBERTa & 92.3 & 3.6 \\
MPNet & 92.0 & 3.7 \\
DeBERTa-v3 & 86.1 & 6.5 \\
ELECTRA & 84.4 & 7.3 \\
\bottomrule
\end{tabular}}
\end{minipage}
\end{table}

\noindent
\begin{minipage}[t]{0.49\linewidth}
\centering
\scriptsize
\setlength{\tabcolsep}{4pt}
\captionof{table}{arXiv: models trained up to 2000, ordered by future performance.}
\label{tab:arxiv_cutoff_2000}
\resizebox{\ifdim\width>\linewidth \linewidth\else\width\fi}{!}{%
\begin{tabular}{c l rrr}
\toprule
Rank & Model & Macro AUC (\%) & Future (\%) & Decay (\%) \\
\midrule
1 & MiniLM-L6 & 97.9 & 93.9 & 4.0 \\
2 & ModernBERT & 97.5 & 93.9 & 3.7 \\
3 & TextCNN-L & 98.1 & 93.8 & 4.3 \\
4 & TX-M & 96.2 & 93.6 & 2.7 \\
5 & BiLSTM-Attn-L & 97.8 & 93.5 & 4.3 \\
6 & TX-S & 96.6 & 93.5 & 3.2 \\
7 & TX-L & 96.2 & 93.4 & 2.8 \\
8 & TextCNN-M & 97.7 & 93.4 & 4.3 \\
9 & FFN-L & 97.2 & 93.2 & 4.0 \\
10 & TextCNN-S & 97.4 & 93.0 & 4.4 \\
11 & BiLSTM-M & 97.3 & 92.8 & 4.5 \\
12 & FFN-M & 96.6 & 92.3 & 4.3 \\
13 & BiGRU-S & 93.3 & 91.9 & 1.4 \\
14 & FFN-S & 95.9 & 91.3 & 4.6 \\
15 & BERT & 95.8 & 90.9 & 4.9 \\
16 & DistilBERT & 94.6 & 90.2 & 4.4 \\
17 & RoBERTa & 91.5 & 88.0 & 3.5 \\
18 & MPNet & 88.9 & 87.3 & 1.6 \\
19 & DeBERTa-v3 & 90.7 & 81.2 & 9.5 \\
20 & ELECTRA & 88.9 & 77.3 & 11.7 \\
\bottomrule
\end{tabular}}
\end{minipage}
\hfill
\begin{minipage}[t]{0.49\linewidth}
\centering
\scriptsize
\setlength{\tabcolsep}{4pt}
\captionof{table}{arXiv: models trained up to 2008, ordered by future performance.}
\label{tab:arxiv_cutoff_2008}
\resizebox{\ifdim\width>\linewidth \linewidth\else\width\fi}{!}{%
\begin{tabular}{c l rrr}
\toprule
Rank & Model & Macro AUC (\%) & Future (\%) & Decay (\%) \\
\midrule
1 & TX-M & 98.6 & 96.0 & 2.6 \\
2 & TextCNN-L & 98.7 & 95.7 & 3.0 \\
3 & BiLSTM-Attn-L & 98.5 & 95.6 & 2.9 \\
4 & BiLSTM-M & 98.8 & 95.6 & 3.2 \\
5 & TX-S & 98.4 & 95.3 & 3.1 \\
6 & FFN-L & 98.0 & 95.2 & 2.8 \\
7 & ModernBERT & 98.0 & 95.0 & 3.0 \\
8 & MiniLM-L6 & 97.9 & 94.9 & 3.0 \\
9 & TextCNN-M & 98.6 & 94.9 & 3.7 \\
10 & FFN-M & 97.9 & 94.9 & 3.0 \\
11 & TX-L & 98.2 & 94.8 & 3.4 \\
12 & BiGRU-S & 98.6 & 94.5 & 4.2 \\
13 & BERT & 97.4 & 93.8 & 3.6 \\
14 & FFN-S & 97.6 & 93.7 & 4.0 \\
15 & TextCNN-S & 98.5 & 93.6 & 4.9 \\
16 & DistilBERT & 97.1 & 93.3 & 3.9 \\
17 & RoBERTa & 96.8 & 92.5 & 4.3 \\
18 & MPNet & 96.7 & 92.5 & 4.3 \\
19 & DeBERTa-v3 & 93.4 & 86.1 & 7.3 \\
20 & ELECTRA & 92.5 & 84.6 & 7.9 \\
\bottomrule
\end{tabular}}
\end{minipage}

\vspace{1.5ex}

\noindent
\begin{minipage}[t]{0.49\linewidth}
\centering
\scriptsize
\setlength{\tabcolsep}{4pt}
\captionof{table}{arXiv: models trained up to 2016, ordered by future performance.}
\label{tab:arxiv_cutoff_2016}
\resizebox{\ifdim\width>\linewidth \linewidth\else\width\fi}{!}{%
\begin{tabular}{c l rrr}
\toprule
Rank & Model & Macro AUC (\%) & Future (\%) & Decay (\%) \\
\midrule
1 & BiLSTM-M & 98.8 & 96.6 & 2.2 \\
2 & BiGRU-S & 98.8 & 96.6 & 2.2 \\
3 & TX-S & 98.7 & 96.5 & 2.2 \\
4 & TX-M & 98.7 & 96.5 & 2.2 \\
5 & TX-L & 98.7 & 96.5 & 2.2 \\
6 & BiLSTM-Attn-L & 98.7 & 96.4 & 2.3 \\
7 & TextCNN-L & 98.6 & 96.3 & 2.3 \\
8 & TextCNN-S & 98.5 & 96.2 & 2.3 \\
9 & TextCNN-M & 98.5 & 96.2 & 2.4 \\
10 & FFN-L & 98.3 & 96.2 & 2.2 \\
11 & FFN-M & 98.2 & 96.1 & 2.2 \\
12 & MiniLM-L6 & 98.4 & 96.0 & 2.4 \\
13 & FFN-S & 98.1 & 96.0 & 2.2 \\
14 & ModernBERT & 98.2 & 95.8 & 2.4 \\
15 & BERT & 97.6 & 95.1 & 2.5 \\
16 & DistilBERT & 97.3 & 94.8 & 2.5 \\
17 & RoBERTa & 97.0 & 94.6 & 2.4 \\
18 & MPNet & 97.0 & 94.4 & 2.5 \\
19 & DeBERTa-v3 & 93.6 & 90.0 & 3.7 \\
20 & ELECTRA & 92.6 & 88.3 & 4.3 \\
\bottomrule
\end{tabular}}
\end{minipage}
\hfill
\begin{minipage}[t]{0.49\linewidth}
\centering
\scriptsize
\setlength{\tabcolsep}{4pt}
\captionof{table}{arXiv: models trained up to 2024, ordered by future performance.}
\label{tab:arxiv_cutoff_2024}
\resizebox{\ifdim\width>\linewidth \linewidth\else\width\fi}{!}{%
\begin{tabular}{c l rrr}
\toprule
Rank & Model & Macro AUC (\%) & Future (\%) & Decay (\%) \\
\midrule
1 & TX-M & 96.5 & 96.3 & 0.1 \\
2 & BiLSTM-M & 96.5 & 96.3 & 0.2 \\
3 & TX-L & 96.4 & 96.3 & 0.2 \\
4 & BiLSTM-Attn-L & 96.4 & 96.3 & 0.2 \\
5 & BiGRU-S & 96.4 & 96.3 & 0.2 \\
6 & TX-S & 96.4 & 96.2 & 0.2 \\
7 & TextCNN-S & 96.2 & 96.0 & 0.2 \\
8 & FFN-L & 96.0 & 95.8 & 0.2 \\
9 & FFN-M & 96.0 & 95.8 & 0.2 \\
10 & TextCNN-M & 96.0 & 95.8 & 0.3 \\
11 & FFN-S & 95.9 & 95.7 & 0.2 \\
12 & MiniLM-L6 & 95.9 & 95.6 & 0.3 \\
13 & TextCNN-L & 95.8 & 95.5 & 0.3 \\
14 & ModernBERT & 95.6 & 95.2 & 0.4 \\
15 & BERT & 95.1 & 94.7 & 0.4 \\
16 & DistilBERT & 94.9 & 94.5 & 0.4 \\
17 & RoBERTa & 94.9 & 94.5 & 0.4 \\
18 & MPNet & 94.9 & 94.4 & 0.4 \\
19 & DeBERTa-v3 & 92.1 & 91.2 & 0.8 \\
20 & ELECTRA & 91.5 & 90.4 & 1.1 \\
\bottomrule
\end{tabular}}
\end{minipage}

\begin{table}[H]
\centering
\footnotesize
\caption{arXiv: future performance and decay by model family.}
\label{tab:arxiv_by_family}
\begin{tabular}{l rrrrrrrr}
\toprule
 & \multicolumn{2}{c}{2000} & \multicolumn{2}{c}{2008} & \multicolumn{2}{c}{2016} & \multicolumn{2}{c}{2024} \\
Family & Future & Decay & Future & Decay & Future & Decay & Future & Decay \\
\midrule
FFN & 92.2 & 4.3 & 94.6 & 3.3 & 96.1 & 2.2 & 95.8 & 0.2 \\
Frozen & 87.8 & 5.4 & 91.6 & 4.6 & 93.6 & 2.8 & 93.8 & 0.5 \\
Recurrent & 92.7 & 3.4 & 95.2 & 3.4 & 96.5 & 2.2 & 96.3 & 0.2 \\
TextCNN & 93.4 & 4.3 & 94.7 & 3.9 & 96.2 & 2.3 & 95.8 & 0.3 \\
Transformer & 93.5 & 2.9 & 95.4 & 3.0 & 96.5 & 2.2 & 96.3 & 0.2 \\
\bottomrule
\end{tabular}
\end{table}

\end{document}